\documentclass{article}

\newif\ifisarxiv
\isarxivtrue

\ifisarxiv
\usepackage{fullpage}
\newcommand{\citep}[1] {\cite{#1}}
\PassOptionsToPackage{numbers,compress,sort}{natbib}
\usepackage{natbib}
\else
\usepackage[final,nonatbib]{neurips_2021}
\PassOptionsToPackage{numbers,compress,sort}{natbib}
\usepackage{natbib}
\fi

\usepackage[utf8]{inputenc} 
\usepackage[T1]{fontenc}    
\usepackage{hyperref}       
\usepackage{url}            
\usepackage{booktabs}       
\usepackage{amsfonts}       
\usepackage{nicefrac}       
\usepackage{microtype}      
\usepackage{xcolor}         
\usepackage{enumitem}

\usepackage{cite}
\usepackage{graphicx}
\usepackage{float}
\usepackage{caption}
\usepackage{subcaption}
\usepackage{multirow}
\usepackage{amsthm}
\usepackage{amsmath}
\usepackage{pgffor}
\usepackage{comment}
\usepackage{array}
\theoremstyle{definition}
\newtheorem{mytheorem}{Theorem}

\newtheorem{remark}[mytheorem]{Remark}
\newtheorem{myDef}[mytheorem]{Definition}

\newcommand{\widesim}[2][1.5]{
  \mathrel{\overset{#2}{\scalebox{#1}[1]{$\sim$}}}
}

\newcolumntype{P}[1]{>{\centering\arraybackslash}p{#1}}
\newcolumntype{M}[1]{>{\centering\arraybackslash}m{#1}}

\newif\ifshowcomments

\showcommentstrue 

\ifshowcomments

\newcommand {\michael}[1]{{\color{red}\sf{[Michael: #1]}}}
\newcommand {\addressed}[1]{{\color{blue}\sf{[Michael: #1]}}}
\newcommand {\ryan}[1]{{\color{purple}\sf{[Ryan: #1]}}}
\newcommand {\lsh}[1]{{\color{green}\sf{[Liam: #1]}}}
\newcommand {\remove}[1]{{\color{yellow}{#1}}}
\newcommand{\yaoqing}[1]{\textcolor{orange}{\sf[Yaoqing:\ #1]}}

\else

\newcommand {\michael}[1]{}
\newcommand {\addressed}[1]{}
\newcommand {\ryan}[1]{}
\newcommand {\lsh}[1]{}
\newcommand {\remove}[1]{}
\newcommand{\yaoqing}[1]{}

\fi

\ifisarxiv
\title{Taxonomizing local versus global structure \\ in neural network loss landscapes}
\date{}

\author{%
  Yaoqing Yang$^{1}$, Liam Hodgkinson$^{1,2}$, Ryan Theisen$^{1}$, Joe Zou$^{1}$, \\Joseph E. Gonzalez$^{1}$, Kannan Ramchandran$^{1}$, Michael W. Mahoney$^{1,2}$ \\
  $^1$ University of California, Berkeley\\
  $^2$ International Computer Science Institute\\
  \texttt{  \{yqyang, liam.hodgkinson, theisen, joezou, jegonzal, kannanr,}\\
  \texttt{mahoneymw\}@berkeley.edu } \\
}
\else
\title{
Taxonomizing local versus global structure \\ in neural network loss landscapes
}
\author{%
  Yaoqing Yang$^{1}$, Liam Hodgkinson$^{1,2}$, Ryan Theisen$^{1}$, Joe Zou$^{1}$, \\
  {\bf Joseph E. Gonzalez$^{1}$, Kannan Ramchandran$^{1}$, Michael W. Mahoney$^{1,2}$ }\\
  $^1$ University of California, Berkeley\\
  $^2$ International Computer Science Institute\\
  \texttt{  \{yqyang, liam.hodgkinson, theisen, joezou, jegonzal, kannanr,}\\
  \texttt{mahoneymw\}@berkeley.edu } \\
}
\fi

\begin{document}

\maketitle

\begin{abstract}
    Viewing neural network models in terms of their loss landscapes has a long history in the statistical mechanics approach to learning, and in recent years it has received attention within machine learning proper.
    Among other things, local metrics (such as the smoothness of the loss landscape) have been shown to correlate with global properties of the model (such as good generalization performance). 
    Here, we perform a detailed empirical analysis of the loss landscape structure of thousands of neural network models, systematically varying learning tasks, model architectures, and/or quantity/quality of data.
    By considering a range of metrics that attempt to capture different aspects of the loss landscape,
    we demonstrate that the best test accuracy is obtained when: the loss landscape is globally well-connected; ensembles of trained models are more similar to each other; and models converge to locally smooth regions.
    We also show that globally poorly-connected landscapes can arise when models are small or when they are trained to lower quality data; and that, if the loss landscape is globally poorly-connected, then training to zero loss can actually lead to worse test accuracy.
    Our detailed empirical results shed light on phases of learning (and consequent double descent behavior), fundamental versus incidental determinants of good generalization, the role of load-like and temperature-like parameters in the learning process, different influences on the loss landscape from model and data, and the relationships between local and global metrics, all topics of recent interest.
\end{abstract}

\section{Introduction}

Among the many approaches to understanding the behavior of neural network (NN) models, the study of their loss landscapes \citep{li2018visualizing,ballard2017energy} has proven to be particularly fruitful. 
Indeed, analyzing loss landscapes has helped shed light on the workings of many popular techniques, including large-batch training \citep{keskar2016large,YGKM18_TRv1}, adversarial training \citep{yao2018hessian}, 
residual connections \citep{li2017convergence}, and BatchNorm \citep{santurkar2018does}. 
One particular concept of recent interest is the so-called \emph{sharpness} of local minima \citep{keskar2016large,dinh2017sharp,neyshabur2017pac,yao2018hessian,foret2020sharpness}.
While sharpness can be measured by first-order sensitivity measures, such as the Jacobian or Lipschitz constant, it is more appropriately measured by second-order sensitivity measures, typically via the Hessian spectrum \citep{yao2020pyhessian}. 
It has been observed that in some cases NNs generalize well when they converge to a relatively flat, i.e., non-sharp, local minimum \citep{keskar2016large}. 

While such local sharpness measures can provide insight, their focus on the local geometry of the loss landscape neglects the \emph{global} structure of the loss landscape (namely, precisely the sort of structure that statistical mechanics approaches to learning aim to quantify \citep{EB01_BOOK,martin2017rethinking}).
Indeed, it is well-known that existing sharpness-based metrics can be altered (trivially) by reparameterization tricks or (more interestingly) by taking algorithmic steps which have the effect of changing the local structures on the loss landscape \citep{dinh2017sharp,yao2018hessian,granziol2020flatness}. 
For example, 
\citep{yao2018hessian} shows that adversarial training can decrease the magnitude of Hessian eigenvalues and bias the model towards a locally smooth area, even though adversarial training can reduce clean test accuracy \citep{tsipras2018robustness}.
Similarly,
\citep{granziol2020flatness} shows that Hessian eigenvalues become smaller with reduced $\ell_2$ regularization, even though increased $\ell_2$ regularization is known to reduce overfitting and improve training, if used properly. 
More general considerations would suggest (and indeed our own empirical results, e.g., as reported in Figure~\ref{fig:Scaling_noisy_label}, demonstrate) that by training to data with noisy labels, one can find models that generalize poorly and yet simultaneously lie in very ``flat'' regions of the loss landscape, with small Hessian eigenvalues, and vice versa. 
These observations (and other observations we describe below) indicate that the previously-observed empirical correlation between very local metrics like sharpness and more global properties like generalization performance may be correlative and not causative, i.e., they may be due to the confounding factor that results in the published literature are on reasonably-good models trained to reasonably-good data, rather than due to some fundamental properties of deep NNs. They also raise the question of how to capture more global properties of the loss landscape.

\begin{figure}[]
    \centering
    \begin{tabular}{|p{2.1cm}|c|c|}
\hline
  \begin{tabular}{c}
    
  \end{tabular} &
  \begin{tabular}{c}
    Globally poorly-connected
  \end{tabular} &
  \begin{tabular}{c}
    Globally well-connected
  \end{tabular}
  \\
  \hline
  \hline
  \begin{tabular}{c}
    Locally sharp
  \end{tabular} &
  \begin{tabular}{c}
    Phase I \\
    \hline
    \includegraphics[width=0.2\textwidth]{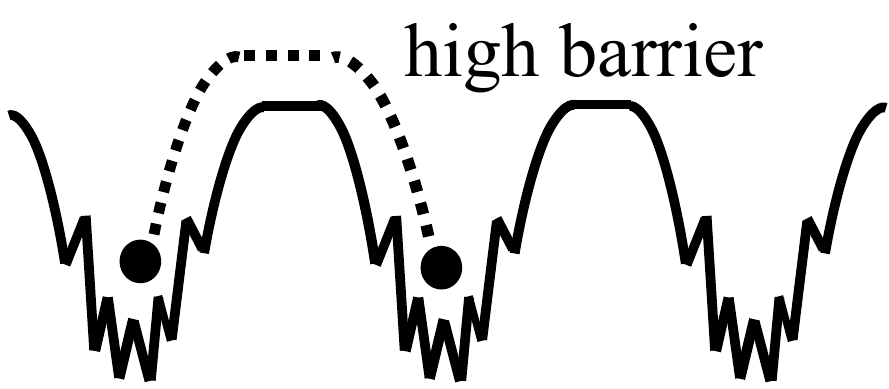}
  \end{tabular} &
  \begin{tabular}{c}
    Phase II \\
    \hline
    \includegraphics[width=0.2\textwidth]{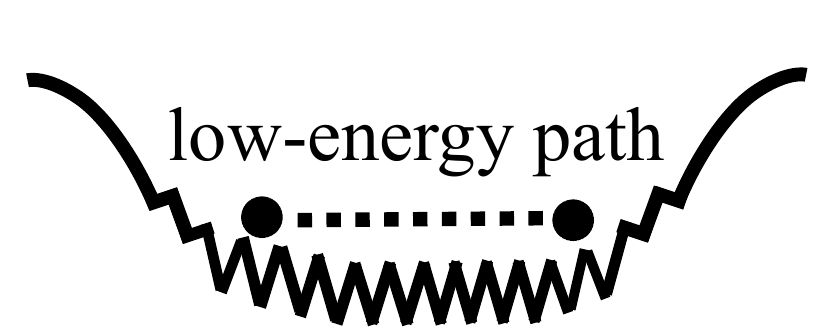}
  \end{tabular} \\
\hline
  \begin{tabular}{c}
    Locally flat
  \end{tabular} &
  \begin{tabular}{c}
    Phase III \\
    \hline
    \includegraphics[width=0.2\textwidth]{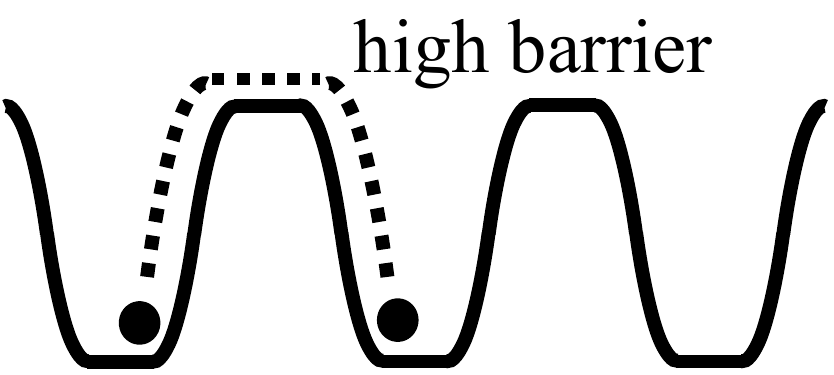}
  \end{tabular} &
  \begin{tabular}{c|c}
    Phase IV-A & Phase IV-B 
    \\
    \hline
    \begin{tabular}{c}
         \includegraphics[width=0.17\textwidth]{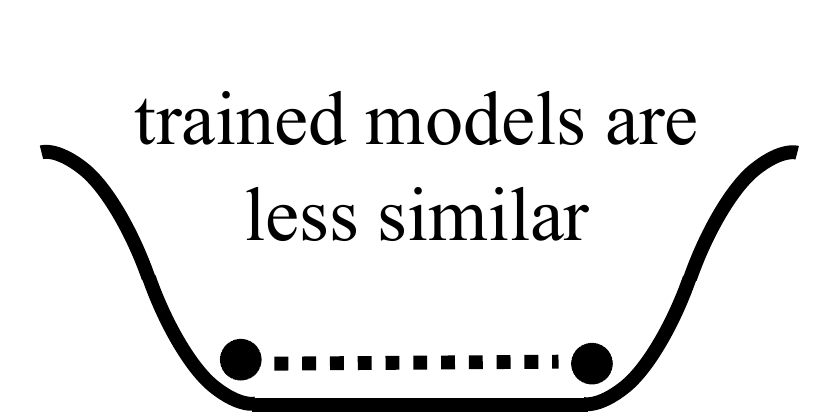}
    \end{tabular}
     & \begin{tabular}{c}
         \includegraphics[width=0.17\textwidth]{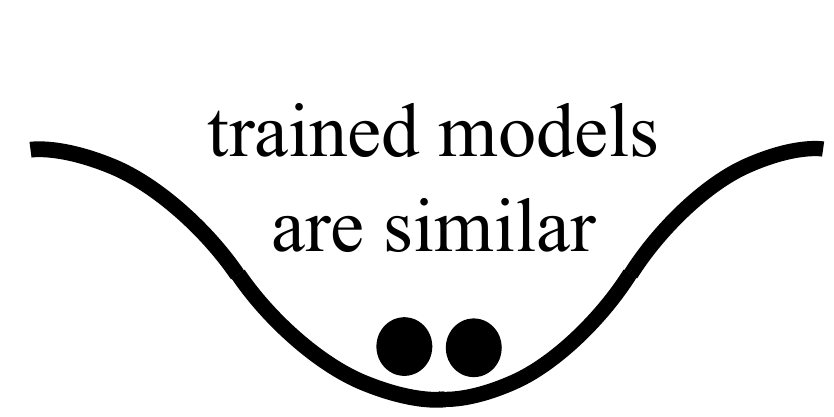}
    \end{tabular}
  \end{tabular} \\
 \hline
\end{tabular}
    \caption{{\bf (Caricature of different types of loss landscapes).} Globally well-connected versus globally poorly-connected loss landscapes; and locally sharp versus locally flat loss landscapes. 
    Globally well-connected loss landscapes can be interpreted in terms of a global ``rugged convexity''; and
    globally well-connected and locally flat loss landscapes can be further divided into two sub-cases, based on the similarity of trained~models.\vspace{-5mm}
    }
    \label{fig:two_by_two}
\end{figure}

Motivated by these considerations, we are interested in understanding 
local properties/structure versus global properties/structure of the loss landscape of realistic NN models.
While similar ideas underlie work that adopts a 
statistical mechanics perspective~\citep{EB01_BOOK,ballard2017energy,martin2017rethinking,martin2018implicit_JMLRversion}, here we are interested in adopting an 
operational machine learning (ML) perspective, where we employ metrics that have been used within ML as ``experimental probes'' to gain insight into local versus global properties.
To do so, we employ the following metrics. 
\begin{itemize}[noitemsep,leftmargin=*]
    \item 
    First, we consider \emph{Hessian-based metrics}, including the largest eigenvalue and the trace of the Hessian.
    These metrics try to capture \emph{local curvature properties of the loss landscape}~\citep{yao2020pyhessian}.
    \item
    Second, we use \emph{mode connectivity} \citep{garipov2018loss,draxler2018essentially}---in particular, the connectivity between trained models.
    This metric tries to capture \emph{how well-connected different local minima are} to each other.
    \item 
    Third, we use \emph{CKA similarity} \citep{kornblith2019similarity} to try to capture \emph{a correlation-like similarity} between the outputs of different trained models. 
    Averaging the CKA over several pairs of models can be thought of as an approximation to so-called overlap integrals frequently appearing in statistical mechanics~\citep{EB01_BOOK,Advani_2013,martin2017rethinking}.
\end{itemize}

We have considered many other metrics, but these three seem to be particularly useful for identifying global structure versus local structure in loss landscapes.
Informally, mode connectivity, as its name suggests, 
captures \emph{connectivity}, where 
well-connected models exhibit a single ``rugged basin'' with low-energy / low-loss, potentially non-linear, paths through the loss landscape, (i.e., continuous chains of models) all achieving a small loss value.
We expect this property to be important since the connectivity of local minima indicates efficiency of the training dynamics to explore the loss landscape, without becoming stuck at saddle points or in a ``bad'' local minimum.
Similarly, CKA captures \emph{similarity}, where an ensemble of good models will produce roughly similar outputs.
These two types of metrics are different and complementary; and both of them are very different than Hessian-based metrics, which clearly capture much more local information.

Here we briefly summarize our main contributions.
\begin{itemize}[noitemsep,leftmargin=*]
    \item  
    We design an experimental setup based on two control parameters, a \emph{temperature-like} parameter that correlates with the magnitude of SGD noise during training, e.g., batch size (in most figures), learning rate, or weight decay, and a \emph{load-like} parameter that measures the relationship between model size and data quantity and/or quality, e.g., the amount of data, size of intermediate layers, amount of exogenously-introduced label noise, etc. 
    By training thousands of models, under a variety of settings, and by measuring local and global metrics of the loss landscape, we identify four distinct phases in temperature-load space, with relatively sharp transitions between them.
    
    \item
    Using global connectivity (measured by mode connectivity) and local flatness (measured by the Hessian), we taxonomize loss landscapes into four categories, which are pictorially represented in Figure \ref{fig:two_by_two}, labelled Phase I through Phase IV. 
    For reasons observed in our empirical results in Section \ref{sec:experiments}, it is often convenient to further divide Phase IV into two subcategories, depending on whether the trained models produce similar representations (as measured by CKA similarity). 
    If the loss landscape satisfies the first property, we say it is \emph{globally well-connected}; and if the loss landscape \emph{also} satisfies the second property, we say it is \emph{globally nice}.
    \footnote{Our empirical results show that a loss landscape can generate dissimilar models while being globally well-connected; but we do not observe a loss landscape that generates similar models but is globally poorly-connected.}
    Depending on whether the Hessian eigenvalues are large or small, we say the loss landscape is \emph{locally sharp} or \emph{locally flat}.
    
    \item 
    Based on these results, as well as measured model quality, e.g., test accuracy, we empirically demonstrate that the global (but not necessarily local) structure of a loss landscape is well-correlated with good generalization performance, and that the best generalization occurs in the phase associated with a locally flat, globally nice loss landscape. 
    We demonstrate these results on a range of computer vision and natural language processing benchmarks (CIFAR-10, CIFAR-100, SVHN, and IWSLT 2016 De-En) and various models (ResNet, VGG, and Transformers). 
    We also vary the amount of data, the number of noisy labels, etc., to study both the effect of the quantity of data and the quality of data on changing the loss landscape.
    
    \item We observe the well-known double descent phenomenon \citep{belkin2019reconciling,belkin2020two} in our experiments, which exhibits itself as a ``bad fluctuation'' between the different phases (e.g., see the transition that separates Phase I and II from Phase III and IV in Figure \ref{fig:Noisy_label_accuracy}).
    Our empirical observations on double descent corroborates recent theoretical analysis 
    \citep{liao2020random,derezinski2019exact}, which views the phenomenon as a consequence of a transition between qualitatively different phases of learning~\citep{martin2017rethinking}.
\end{itemize}

Computing connectivity and similarity requires comparing multiple distinct models. 
Compared to Hessian computations, this could be computationally expensive, especially if model training is expensive.
For many reasonably-sized models, however, the metrics we consider are sufficiently tractable so as to be useful, e.g.,
\ifisarxiv
during model training (although a full analysis of that is outside the scope of this paper). 
Moreover, both connectivity and similarity can be computed only from the training data and trained networks, without access to any testing data, thus providing non-trivial predictors of generalization performance. 
We should also note that the use of connectivity and similarity for studying the global structure of NN loss landscapes has motivations in classical spin-glass theory (which has been widely applied in studying NNs \citep{amit1985storing,EB01_BOOK,choromanska2015loss,agliari2014walk}, as well as neuroscience \citep{fuhs2006spin,hudetz2014spin,recio2016emergence}); and was inspired by \citep{martin2017rethinking}, whose results suggest a \emph{rugged convexity} in the NN loss landscape, as well as the concept of a \emph{folding funnel} in the (statistical mechanics of) protein folding literature \citep{bryngelson1987spin,garstecki1999energy,klemm2007funnels,brooks2001taking}. 
Also motivating our approach is a large body of work related to energy landscapes~\citep{wales_book,stillinger_book}. We discuss related work further in Section~\ref{sec:related_work}.
\fi

\section{Setup}

In the sequel, we consider training a NN $f_\theta : \mathbb{R}^{d_{\text{in}}}\to\mathbb{R}^{d_{\text{out}}}$, with trainable parameters $\theta$, to a dataset consisting of $n$ datapoint/label pairs $S_{\text{train}} = \{(\boldsymbol{x}_1,y_1),\dots,(\boldsymbol{x}_n,y_n)\}$. 
Our nominal training objective is to minimize a loss function of the form
\begin{align}\label{eqn:training_loss}
    \mathcal{L}(\theta) = \frac{1}{n}\sum_{(\boldsymbol{x},y)\in S_{\text{train}}} \ell(f_\theta(\boldsymbol{x}),y) + \lambda \|\theta\|_2^2.
\end{align}
Here $\ell$ is a loss function, typically chosen to be the cross entropy loss. 
The parameter $\lambda$ is the weight decay parameter, which controls the level of $\ell_2$ regularization. We consider optimizing NN models using standard minibatch SGD, with iterates of the form
\begin{align}
\label{eq:sgd-def}
    \theta \leftarrow \theta - \eta \tilde{\boldsymbol{g}}(\theta),\hspace{5mm} \tilde{\boldsymbol{g}}(\theta) = \frac{1}{B}\sum_{j=1}^B \nabla_\theta \ell(f_\theta(\boldsymbol{x}_{i_j}),y_{i_j}),
\end{align}
where $\eta$ is the learning rate, $0<B\leq n$ is the batch size, and the indices $i_1,\dots, i_B$ of each minibatch are sampled without replacement from $\{1,\dots, n\}$. For classification tasks, we consider also the training/testing accuracy, which is simply the fraction of correctly classified points, $\text{acc}_{\text{train}}(\theta) = \frac1n\sum_{(\boldsymbol{x},y)\in S_{\text{train}}} \mathbf{1}(f_\theta(\boldsymbol{x}) = y)$,
and similarly for $\text{acc}_{\text{test}}(\theta)$ on a given test set $S_{\text{test}}$. 

We now briefly introduce the main metrics and control parameters which we will consider.
\ifisarxiv
Due to space constraints, we defer further details to Appendix~\ref{sec:definitions}.
\else
\fi

\textbf{Temperature and load.}
In the sequel, a \emph{load-like parameter} of a loss landscape refers to some quantity related to the amount and/or quality of data, relative to the size of the model. 
Specifically, we vary either i) model size (e.g., width, which captures the size of an internal representation of the data), for fixed training set size $n$, ii) training set size $n$, for fixed model size, or iii) the ``quality'' of training data, which is varied by randomizing a fraction $\alpha$ of the training labels.
Each of these control parameters \emph{directly} induces a different loss landscape by changing the data $S_\text{train}$ and/or architecture $f_\theta$ for which the loss $\mathcal{L}(\theta)$ is being computed. 
For example, we expect that increasing width will result in a smoother loss landscape \citep{neal2018modern}; we shall see this effect with CKA similarity in the transition from Phase IV-A to IV-B.

The second control parameter we vary in our experiments is a \emph{temperature-like parameter}, representing the amount of noise introduced in the SGD iterates (\ref{eq:sgd-def}). 
Most commonly, we take this to be the batch size $B$, although we will also use the learning rate $\eta$ and the weight decay parameter $\lambda$. 
Increasing temperature corresponds to smaller batch size, and large learning rate or weight decay. 
Varying the temperature does not directly define a different loss function $\mathcal{L}(\theta)$, but rather it \emph{indirectly} induces a different \emph{effective loss function}. 
This is because, at different temperatures, the iterates of SGD concentrate on different regions of the loss landscape. Due to the noise in the stochastic optimization, the training dynamics may not be able to ``see'' certain features of the loss landscape.

\textbf{CKA similarity.}
To measure the similarity of two NN representations, we use the \textit{centered kernel alignment} (CKA) metric, proposed in \citep{kornblith2019similarity}. 
For a NN $f_\theta$, let $F_\theta = \begin{bmatrix}f_\theta(\boldsymbol{x}_1)&\cdots &f_\theta(\boldsymbol{x}_m)\end{bmatrix}^\top \in \mathbb{R}^{m\times d_{\text{out}}}$ denote the concatenation of the outputs%
\footnote{For this work, we focus on the similarity of representations at the output layer, i.e., after the softmax is applied, although the CKA similarity can be used to compare the representations at any layer.} 
of the network over a set of $m$ randomly sampled datapoints.  
Then the (linear) CKA similarity between two parameter configurations $\theta,\theta'$ is given by
\begin{align}\label{eq:CKA}
    s(\theta,\theta') = \frac{\text{Cov}(F_\theta, F_{\theta'})}{\sqrt{\text{Cov}(F_\theta, F_{\theta})\text{Cov}(F_{\theta'}, F_{\theta'})}},
\end{align}
where for $X,Y\in \mathbb{R}^{m\times d}$, we define $\text{Cov}(X,Y) = (m-1)^{-2}\text{tr}(XX^\top H_m YY^\top H_m)$, and $H_m = I_m - m^{-1}\mathbf{1}\mathbf{1}^\top$ is the centering matrix. 
The CKA similarity is known to be an effective way to compare the overall representations learned by two different trained NNs \citep{kornblith2019similarity}.  
Rather than computing the similarity directly on the original training points, we measure CKA on a perturbed training set comprised of Mixup samples \citep{zhang2017mixup}; this can reduce trivial similarity that occurs when the models are trained to exactly or approximately zero training error.
\ifisarxiv
See also Appendix~\ref{sec:CKA_ablation} for the ablation study on different perturbed training sets.
\fi

\textbf{Mode connectivity.} 
For two parameter configurations $\theta,\theta'$, computing mode connectivity involves finding a \textit{low-energy curve} $\gamma(t)$, $t\in [0,1]$, for which $\gamma(0) = \theta, \gamma(1) = \theta'$, such that $\int\mathcal{L}(\gamma(t))dt$ is minimized \citep{draxler2018essentially, garipov2018loss}. 
A number of techniques have been proposed to find such curves $\gamma$.
In this work, we use the technique proposed in \citep{garipov2018loss}, which parameterizes the Bezier curve with $k+1$ bends, given by
$\gamma_\phi(t) = \sum_{j=0}^{k} \binom{k}{j}(1-t)^{k-j}t^{j}\theta_j$ for $t\in[0,1],$ 
where $\theta_0 = \theta, \theta_k = \theta'$, and $\phi = \{\theta_1,\dots,\theta_{k-1}\}$ are trainable parameters of additional models, defining ``bends'' on the curve $\gamma_\phi(t)$. 
We use Bezier curves with three bends ($k=2$). Given the curve $\gamma_\phi(t)$, we define the mode connectivity of the models $\theta,\theta'$ to be
\begin{align}\label{eq:MD}
    \textsf{mc}(\theta,\theta') = \frac{1}{2}(\mathcal{L}(\theta)+\mathcal{L}(\theta')) - \mathcal{L}(\gamma_\phi(t^\ast)),
\end{align}
where $t^\ast$ maximizes the deviation $t \mapsto | \frac12 (\mathcal{L}(\theta) + \mathcal{L}(\theta')) - \mathcal{L}(\gamma_\phi(t))|$.
There are three possibilities for mode connectivity. 
If $\textsf{mc}(\theta,\theta') < 0$, then $\frac{1}{2}(\mathcal{L}(\theta)+\mathcal{L}(\theta')) < \mathcal{L}(\gamma_\phi(t^\ast))$, which means there is a ``barrier'' of high loss between $\theta,\theta'$; in this case, we will say that the loss landscape is poorly-connected or simply say that mode connectivity is \emph{poor}.
If $\textsf{mc}(\theta,\theta') > 0$, then this implies a curve of low loss connecting $\theta,\theta'$, but it also implies that the training failed to locate a reasonable optimum, i.e., $\mathcal{L}(\theta)$ and $\mathcal{L}(\theta')$ are large.
If $\textsf{mc}(\theta,\theta') \approx 0$, then we will say that the loss landscape is well-connected or simply say that the mode connectivity is \emph{good}.
Note that for all the experiments except neural machine translation, we use the training error (0-1 loss) when computing mode connectivity, so that mode connectivity is always normalized to the range of [$-100$, $100$]. 
We provide additional details on this procedure, as well as an ablation study on different mode connectivity hyperparameters, \ifisarxiv
in Appendix~\ref{sec:MD_lr}.
\fi

\textbf{Hessian.}
The Hessian at a given point $\theta_0$ in parameter space is represented by the matrix $\nabla_\theta^2 \mathcal{L}(\theta_0)$. 
To summarize the Hessian in a single scalar value, we report the dominant eigenvalue $\lambda_{\text{max}}(\nabla_\theta^2 \mathcal{L}(\theta_0))$ and/or the trace $\text{tr}(\nabla_\theta^2 \mathcal{L}(\theta_0))$, calculated using the \texttt{PyHessian} software~\citep{yao2020pyhessian}.

\textbf{$\ell_2$ distance.} 
We will also occasionally report the $\ell_2$ distance between two parameter configurations $\|\theta-\theta'\|_2$ as a measure of similarity between models, although we typically find that the CKA similarity is a more informative measure.

\section{Empirical results on taxonomizing local versus global structure}\label{sec:experiments}
\vspace{-1mm}

In this section, we present our main empirical results. 
Among other things, our results will highlight the presence of globally nice, globally well-connected/poorly-connected, and locally flat/sharp loss landscapes, and the phase transitions which separate them.
In addition to test accuracy, results on six other metrics are presented, including training loss, leading Hessian eigenvalue, trace of Hessian, CKA similarity, mode connectivity, and $\ell_2$ distance measured between model weights. 
For each metric, the results are presented in a 2D diagram, in which the horizontal dimension is the load (with increasing load to the right), and the vertical dimension is the temperature (with increasing temperature to the top).

We will illustrate our main results in a simple setting, and then consider several variants of this setting to illustrate how these results do or do not change when various parameters and design decisions are modified.
To start, we will consider ResNets \citep{he2016deep} trained on CIFAR-10 \citep{krizhevsky2009learning} as the standard setting to demonstrate different loss landscapes. 
We will scale the network width to change the size of the network. 
For ResNet18 which contains four major blocks with channel width $\{k, 2k, 4k, 8k\}$, we select different values of $k$ to obtain ResNet models with different widths. 
In the standard setting, batch size, learning rate, and weight decay are kept constant throughout training to study interactions between temperature-like parameters, load-like parameters, and the loss landscape.
Below, we will apply learning rate decay and consider other variations of this standard setting, in separate experiments.
More details on the experimental setup can be found in
\ifisarxiv
Appendix~\ref{sec:implementation}.
\fi

\vspace{-1mm}
\subsection{Types of loss landscapes and phase transitions}\label{sec:two_by_two}

\def \figname {ResNet18}
\begin{figure}
  \begin{tabular}[c]{cccc}
  \hspace{-5mm}
    \multirow{2}{*}{
    \begin{subfigure}{0.30\textwidth}
        \vspace{-5mm}
      \includegraphics[width=\textwidth]{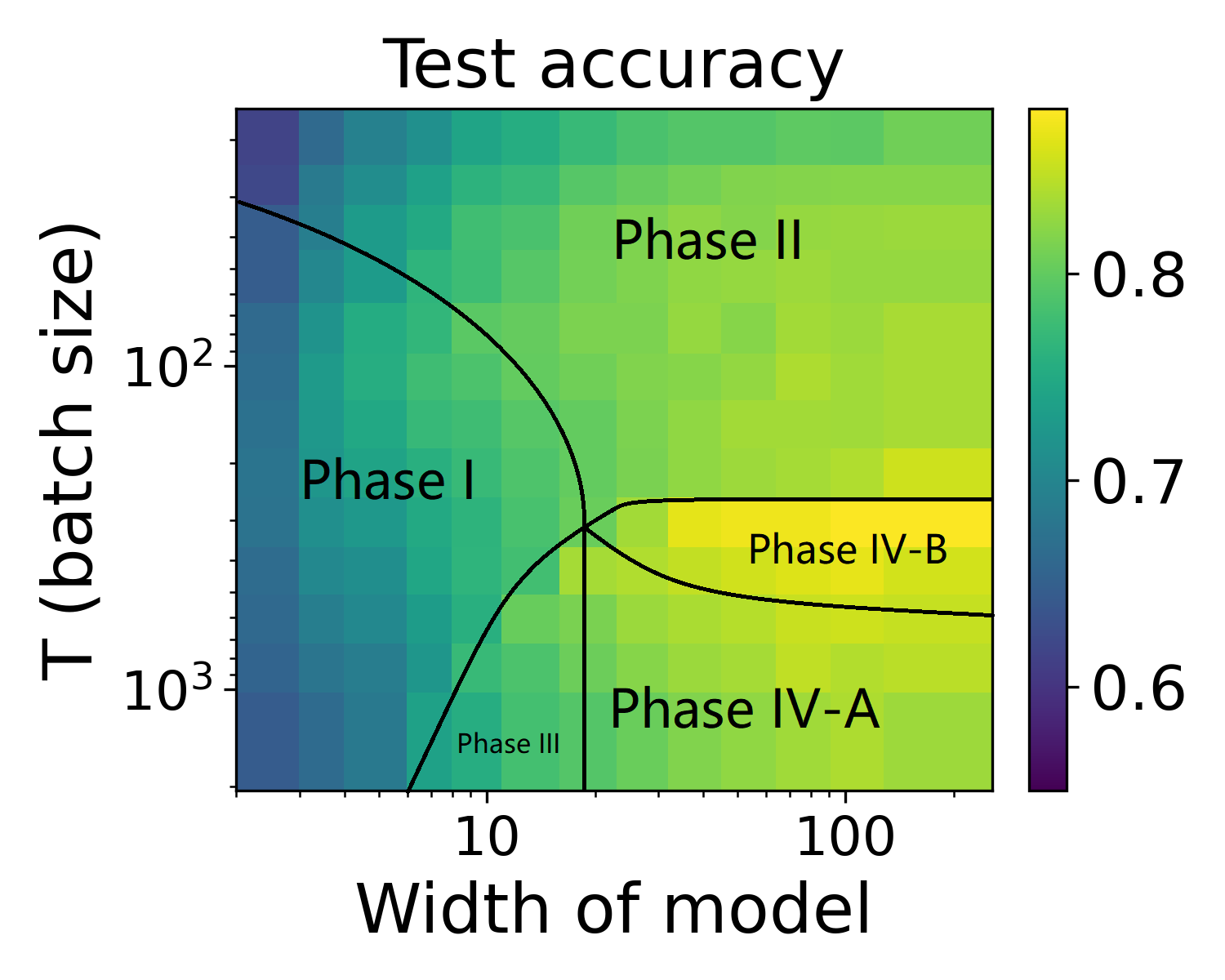}
      \caption{Test accuracy\label{fig:\figname_accuracy}}
    \end{subfigure}
} 
& \begin{subfigure}[c]{0.2\textwidth}
      \includegraphics[width=\textwidth]{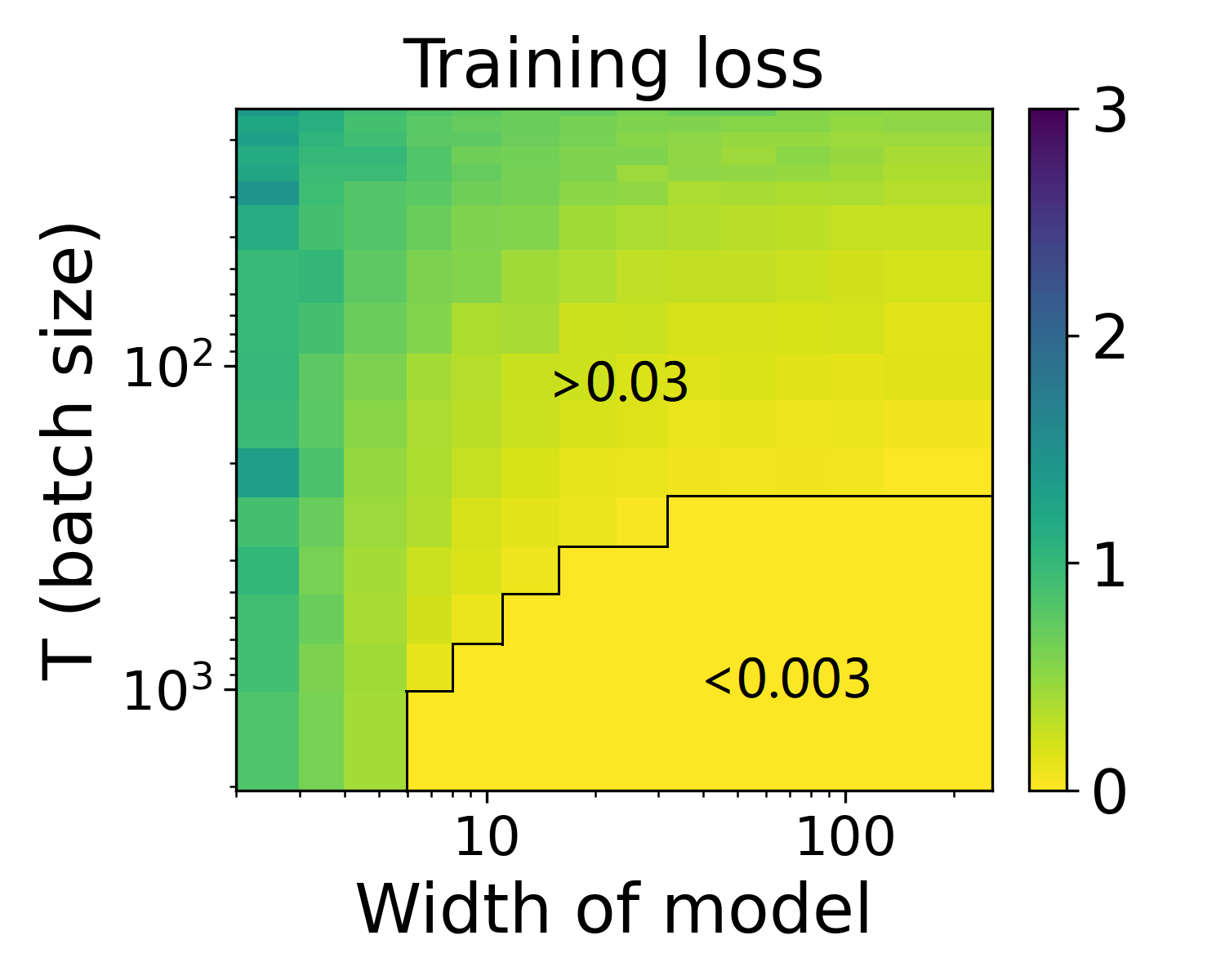}
      \vspace{-6mm}
      \caption{Training loss\label{fig:\figname_loss}}
    \end{subfigure}&
    \begin{subfigure}[c]{0.2\textwidth}
      \includegraphics[width=\textwidth]{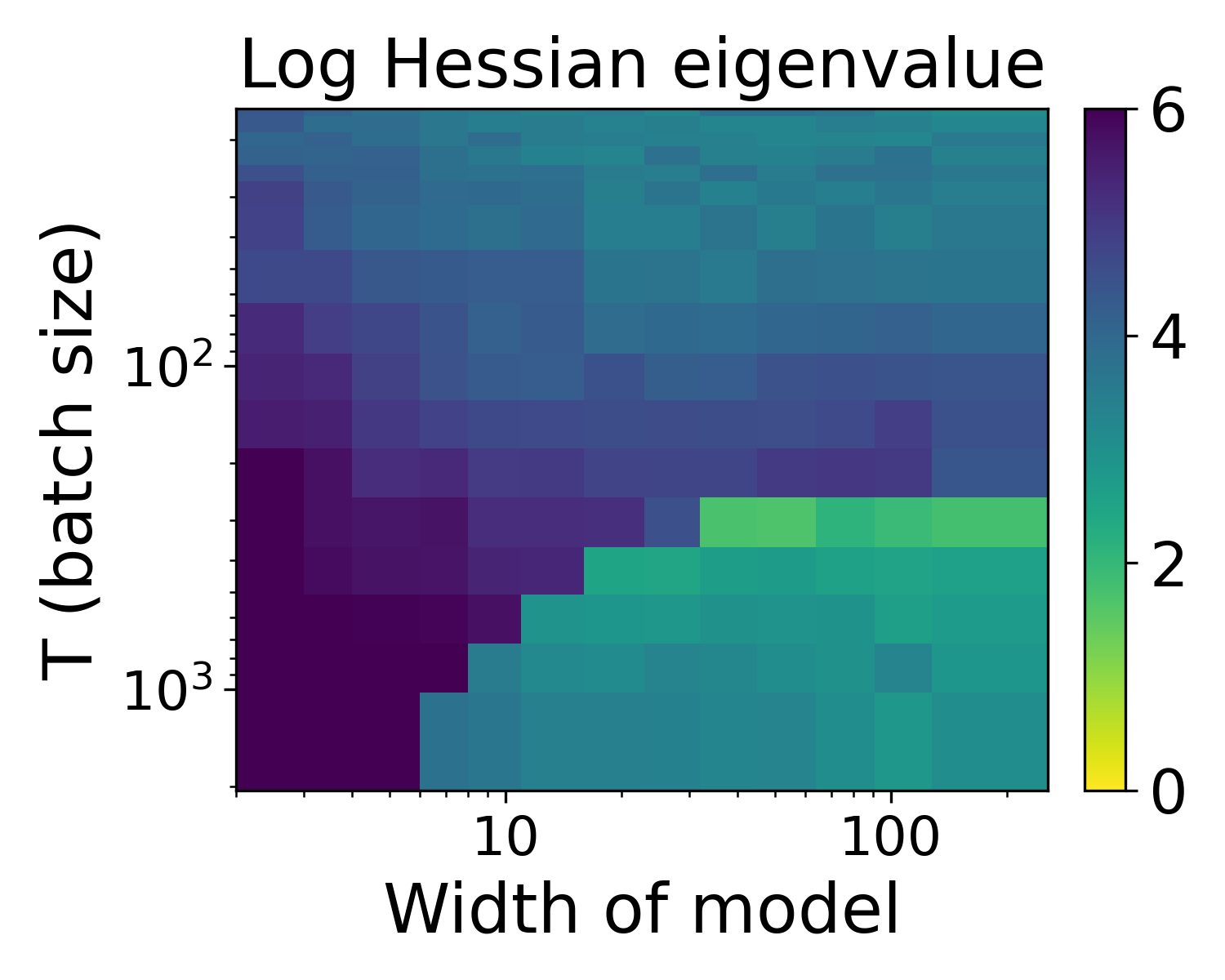}
      \vspace{-6mm}
      \caption{Hessian eigenvalue\label{fig:\figname_hessian_e}}
    \end{subfigure}&
    \begin{subfigure}[c]{0.2\textwidth}
      \includegraphics[width=\textwidth]{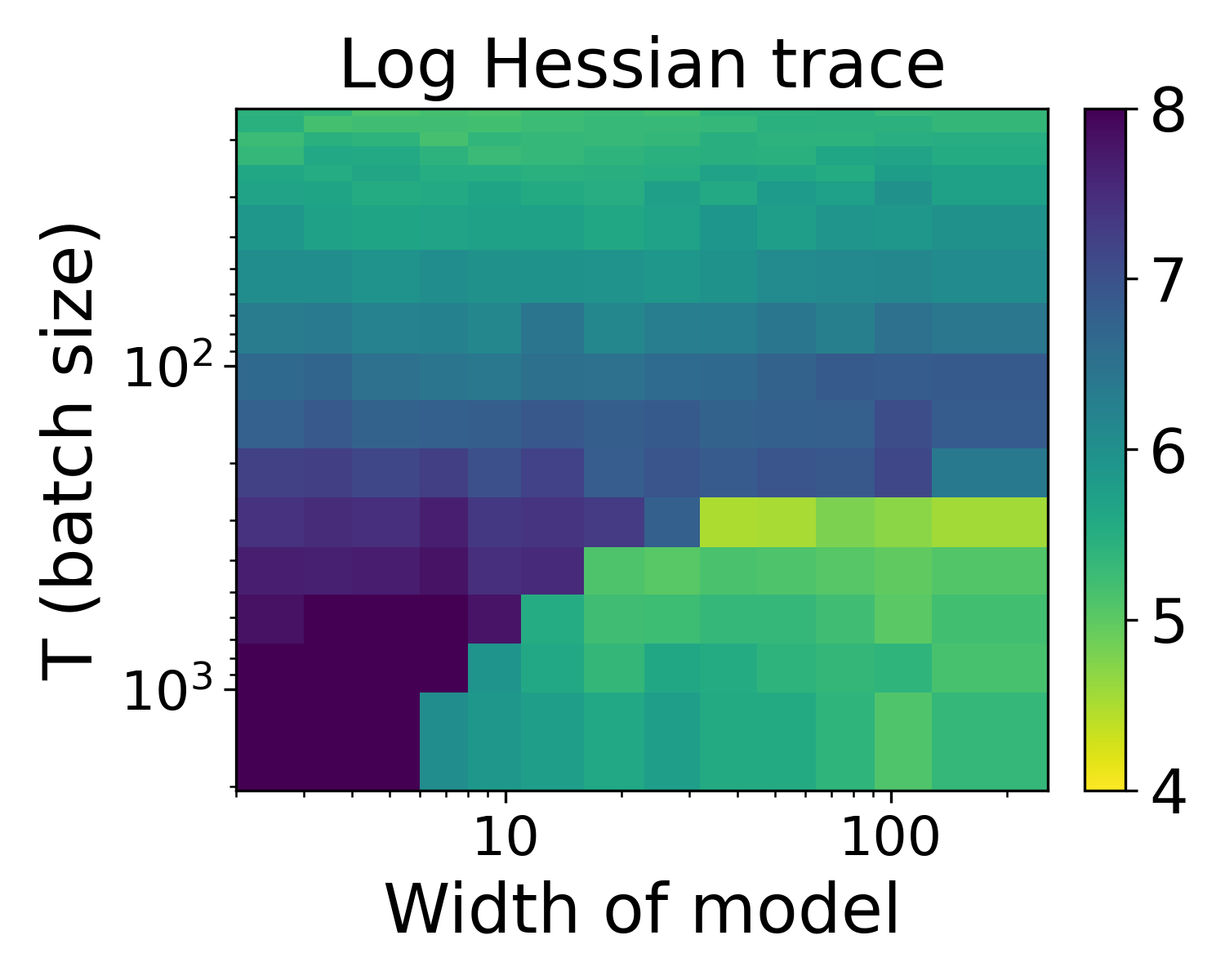}
      \vspace{-6mm}
      \caption{Hessian trace\label{fig:\figname_hessian_t}}
    \end{subfigure}\\
& \begin{subfigure}[c]{0.2\textwidth}
      \includegraphics[width=\textwidth]{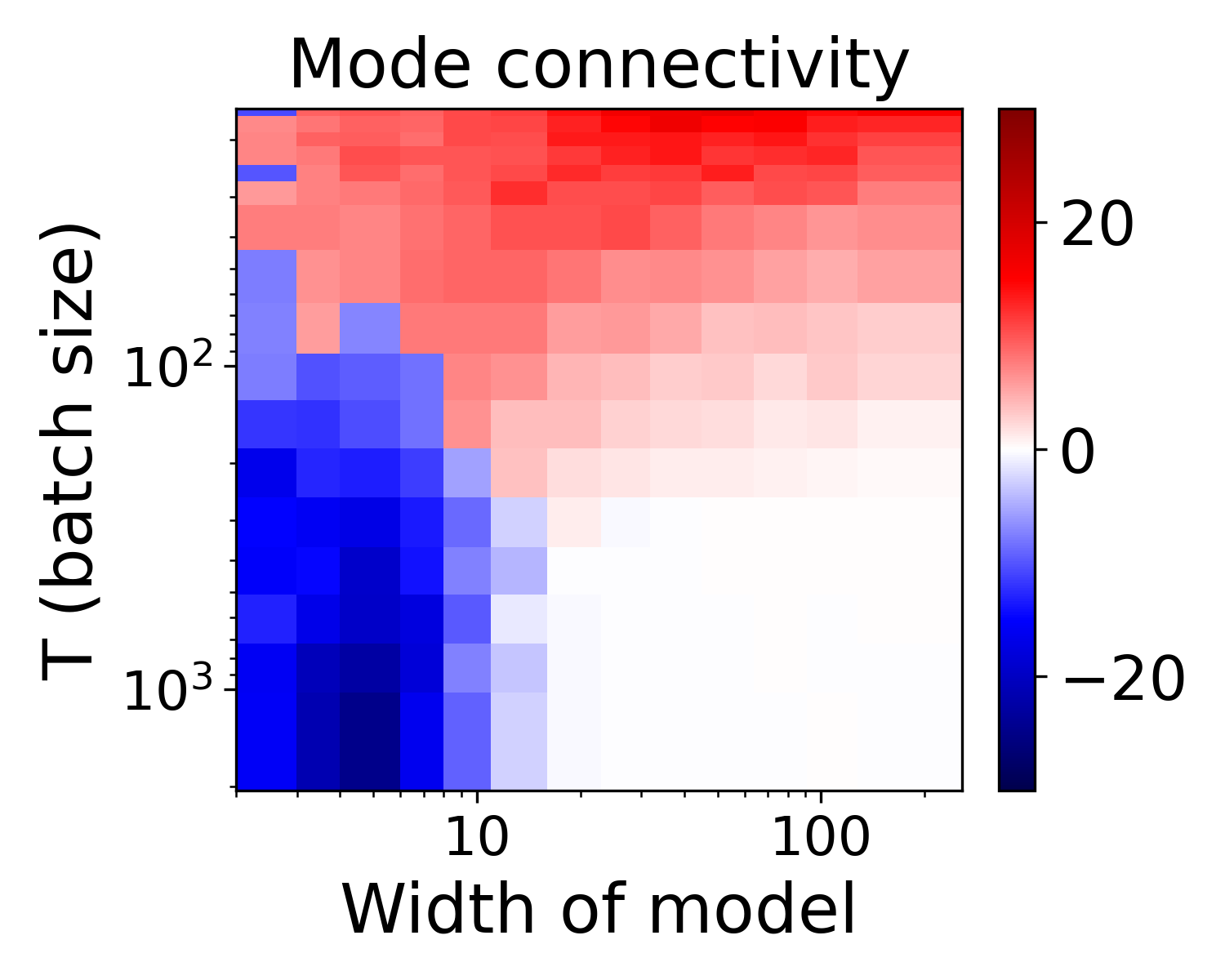}
      \vspace{-6mm}
      \caption{Mode connectivity\label{fig:\figname_curve}}
    \end{subfigure}&
    \begin{subfigure}[c]{0.2\textwidth}
      \includegraphics[width=\textwidth]{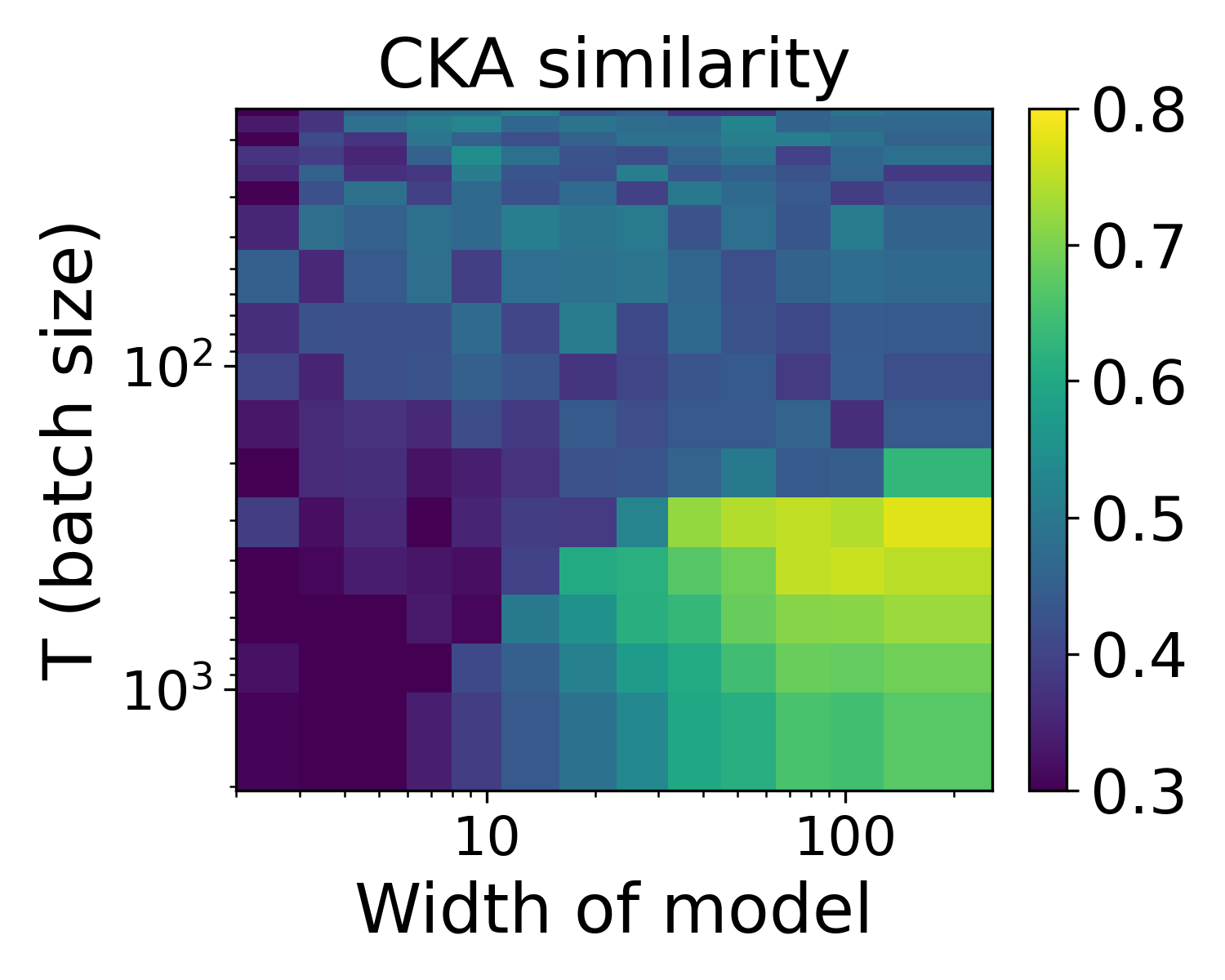}
      \vspace{-6mm}
      \caption{CKA similarity\label{fig:\figname_CKA}}
    \end{subfigure}&
    \begin{subfigure}[c]{0.2\textwidth}
      \includegraphics[width=\textwidth]{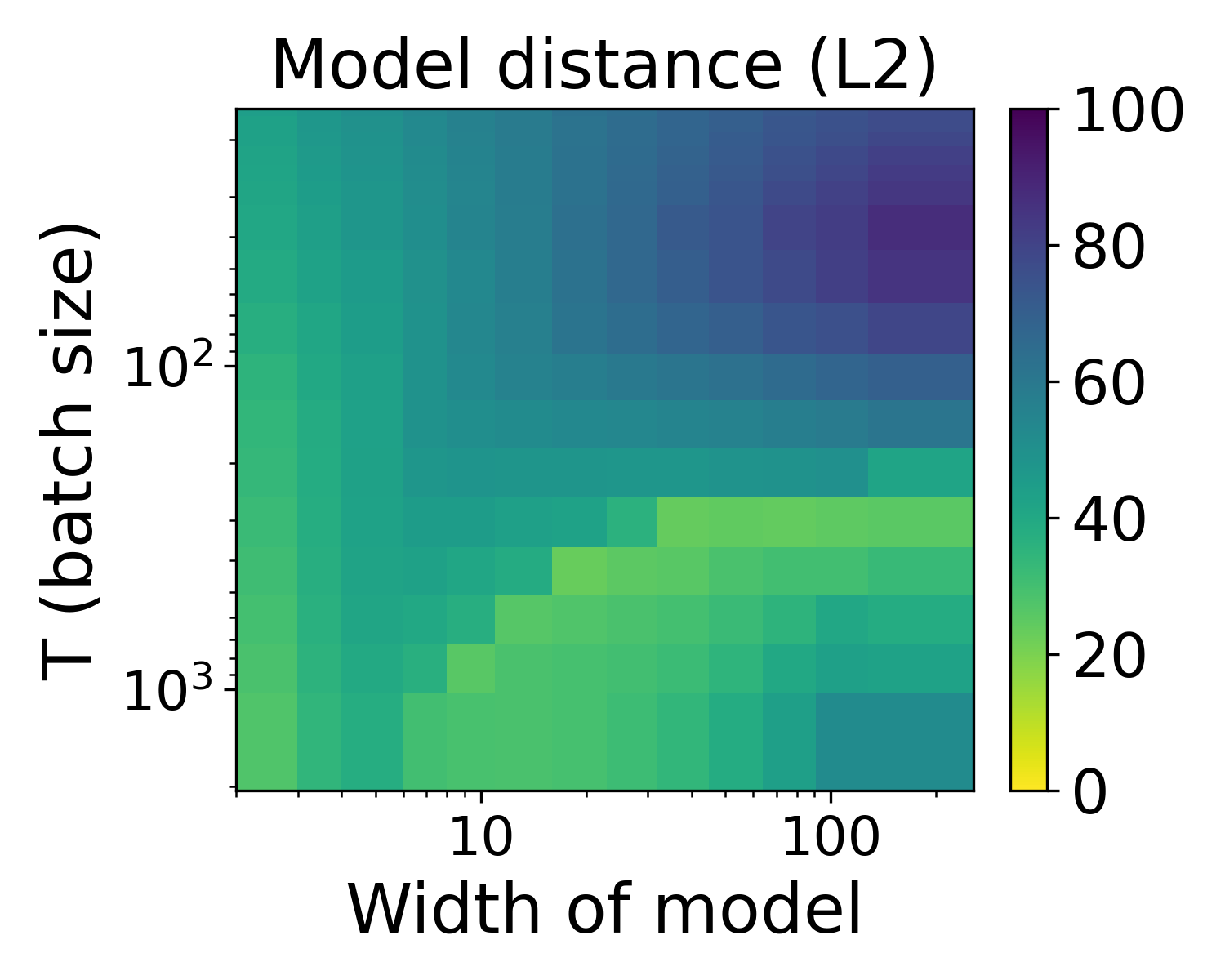}
      \vspace{-6mm}
      \caption{$\ell_2$ distance\label{fig:\figname_dist}}
    \end{subfigure}\\
  \end{tabular}    
  \caption{{\bf (Standard setting).} Partitioning the 2D load-like---temperature-like diagram into different phases of learning, using batch size as the temperature and varying model width to change load.  Models are trained with ResNet18 on CIFAR-10. All plots are on the same set of axes. We note that batch size is inverse temperature, and thus it has smaller values at the top of the y-axis and larger values at the bottom.\vspace{-5mm}
  }
  \label{fig:ResNet18}
\end{figure}

In this subsection, we discuss our standard setting, in which we vary model width as the load-like parameter and batch size as the temperature-like parameter.
A summary of the results is displayed in Figure \ref{fig:ResNet18}.
Each pixel represents a specific training configuration tuple (width, batch size), averaged over five independent~runs. Observe that there are two phase transitions (identified by different metrics) that separate each plot into four primary regions (corresponding to those shown in Figure \ref{fig:two_by_two}).

\begin{itemize}[leftmargin=*,noitemsep]
\item
{\bf Hessian distinguishes locally sharp versus locally flat loss landscapes.} 
The first phase transition is displayed in Figure \ref{fig:ResNet18_hessian_e} and \ref{fig:ResNet18_hessian_t}, separating Phase I/II from Phase III/IV.
A larger Hessian eigenvalue or Hessian trace (darker color) represents a sharper local loss landscape  \citep{yao2020pyhessian,yao2018hessian}.
In Figure \ref{fig:ResNet18_loss}, we find this transition coincides with a significant decrease in the training loss. Indeed, the training loss experiences a more than tenfold reduction when transitioning from the upper side to the lower side on the right of the figure.
Comparing Figures \ref{fig:ResNet18_accuracy} and \ref{fig:ResNet18_hessian_e}-\ref{fig:ResNet18_hessian_t}, categorizing loss landscapes based solely on the Hessian (or, from other results, other local flatness metrics) is insufficient to predict test accuracy, e.g., the test accuracy in Phase III is lower than Phase IV-A but the Hessian eigenvalues are almost the same.

\item
{\bf Mode connectivity distinguishes globally well-connected versus globally poorly-connected loss landscapes.} 
The second phase transition is shown in Figure \ref{fig:ResNet18_curve}.
The white region represents near-zero mode connectivity which, according to our definition, implies a flat curve in the loss landscape between freshly-trained weights; the blue region represents negative mode connectivity which implies a high barrier between weights; and the red region represents positive mode connectivity which implies a low-loss curve between weights, although the weights are not trained to a reasonable optimum.  
The loss along individual mode connectivity curves can be found in
\ifisarxiv
Appendix~\ref{sec:individual_curves}.
\fi
In contrast to training loss, test accuracy only appears to show significant improvements after this transition.
In particular, for well-connected loss landscapes, one can improve the test accuracy with suitable choice of temperature.
This phase transition forms a curve separating Phase I from II, and separates Phase III from IV.
\end{itemize}

\noindent
Based on the two transitions, we now classify the loss landscapes into the following phases.
\begin{itemize}[leftmargin=*,noitemsep]
    \item 
    {\bf Phase I: Globally poorly-connected and locally sharp}:
    Training loss is high; Hessian eigenvalue and trace are large; and mode connectivity is poor.
    \item 
    {\bf Phase II: Globally well-connected and locally sharp}:
    Training loss is high; Hessian eigenvalue and trace are large; and mode connectivity is poor because the trained weights fail to locate a reasonable minimum.
    \item 
    {\bf Phase III: Globally poorly-connected and locally flat}:
    Training loss is small; Hessian eigenvalue and trace are small; yet mode connectivity still remains poor.
    \item 
    {\bf Phase IV: Globally well-connected and locally flat}:
    Training loss is small; Hessian eigenvalue and trace are small; and mode connectivity is good (near-zero).
\end{itemize}

\noindent
We remark that in Figure~\ref{fig:ResNet18} (and subsequent figures below) the load-like and temperature-like parameters are on the X and Y axes, respectively, and we have, to the extent possible, kept other control parameters (in particular, those which are also load-like and temperature-like) fixed, so as to isolate the effect of load-like and temperature-like behavior on trained models.
One might wonder (or even criticize our experimental setup, if one were not to realize that we are trying to isolate the effects of load-like and temperature-like parameters) 
what would be the effect of varying learning rate (which is another temperature-like parameter) during the training process.
Thus, we include the setting with decaying learning rate during training in Section~\ref{sec:corroborating_results}.

Here are two additional observations we can make from Figure \ref{fig:two_by_two}.
\begin{itemize}[leftmargin=*,noitemsep]
\item
{\bf CKA further distinguishes two subcategories in Phase IV.} From Figure \ref{fig:ResNet18_CKA}, CKA can be used to further divide Phase IV into Phase IV-A and Phase IV-B, with the latter exhibiting larger CKA similarity. We remark that the transition from Phase IV-A to Phase IV-B is more like a smooth crossover than a sharp transition. Thus, we name both of them Phase IV.

\item
{\bf Simple $\ell_2$ distance is not enough.} 
A challenge in measuring similarity between models is that the same model can be realized using different weights \citep{dinh2017sharp}.
To reconcile this effect, the distance between two models is commonly defined in terms of their predictions instead of weights. Indeed, the representation-based CKA similarity is seen to be preferable to the weights-based $\ell_2$ distance.
For example, from Figure \ref{fig:ResNet18_dist}, the $\ell_2$ distance provides some limited information, but it is not as informative as CKA similarity.
\end{itemize}

\noindent
Based on these results, we assert the following central claim of this work: 
{\bf optimal test accuracy is obtained when the loss landscape is globally nice and the trained model converges to a locally flat region; and we can diagnose these different phases in the load-like--temperature-like phase diagram with Hessian, mode connectivity, and CKA metrics.} 
Importantly, both similarity and connectivity metrics are required for a globally nice loss landscape. 
Phase IV-B is precisely the region with \emph{globally nice} landscapes, exhibiting the highest test accuracies.

\subsection{Corroborating results}\label{sec:corroborating_results}

In this subsection, we consider initial corroborating results, modifying the setup of Section~\ref{sec:two_by_two} to train with learning rate decay, or to data with exogenously-introduced noisy labels, etc.
Still more results can be found in Section~\ref{sec:ablation_study} and in the Appendix.

{\bf Training with learning rate decay.}
Next, we consider a similar experimental setup and the same phase diagram, except with the same learning rate decay schedule applied in the middle of training rather than with a fixed learning rate throughout. We still vary batch size to change temperature.
The results are presented in Figure \ref{fig:Lr_decay}. 
Comparing Figure \ref{fig:Lr_decay} with Figure \ref{fig:ResNet18}, we see that the four phases are still present, and the test accuracy is maximized when the loss landscape is globally nice and locally flat. Therefore, our central claim is unaffected by the learning rate decay schedule. 
In Figures \ref{fig:Lr_decay_hessian_e} and \ref{fig:Lr_decay_hessian_t}, smaller temperatures (or larger batch size) in Phase IV-A appear to increase the size of the Hessian.
This is a well-known issue with large-batch training \citep{keskar2016large}.
Finally, note that the optimal test accuracy achieved improves in the presence of learning rate decay.
\ifisarxiv
See Appendix~\ref{sec:lr_decay_helps} for further discussions on learning rate decay.
\else
\fi

\def \figname {Lr_decay}
\begin{figure}
  \begin{tabular}[c]{cccc}
  \hspace{-5mm}
    \multirow{2}{*}{
    \begin{subfigure}{0.30\textwidth}
        \vspace{-5mm}
      \includegraphics[width=\textwidth]{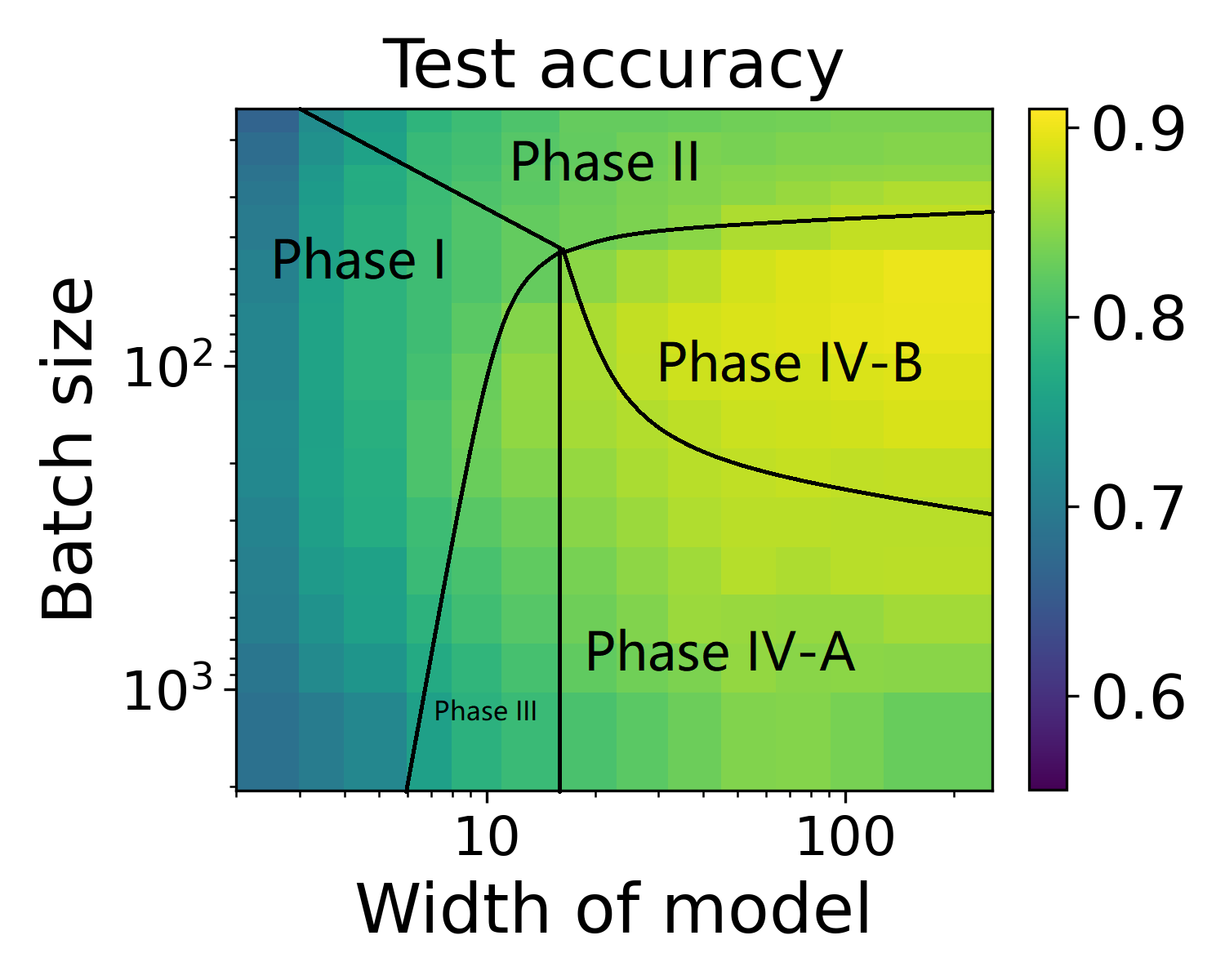}
      \caption{Test accuracy\label{fig:\figname_accuracy}}
    \end{subfigure}
} 
& \begin{subfigure}[c]{0.2\textwidth}
      \includegraphics[width=\textwidth]{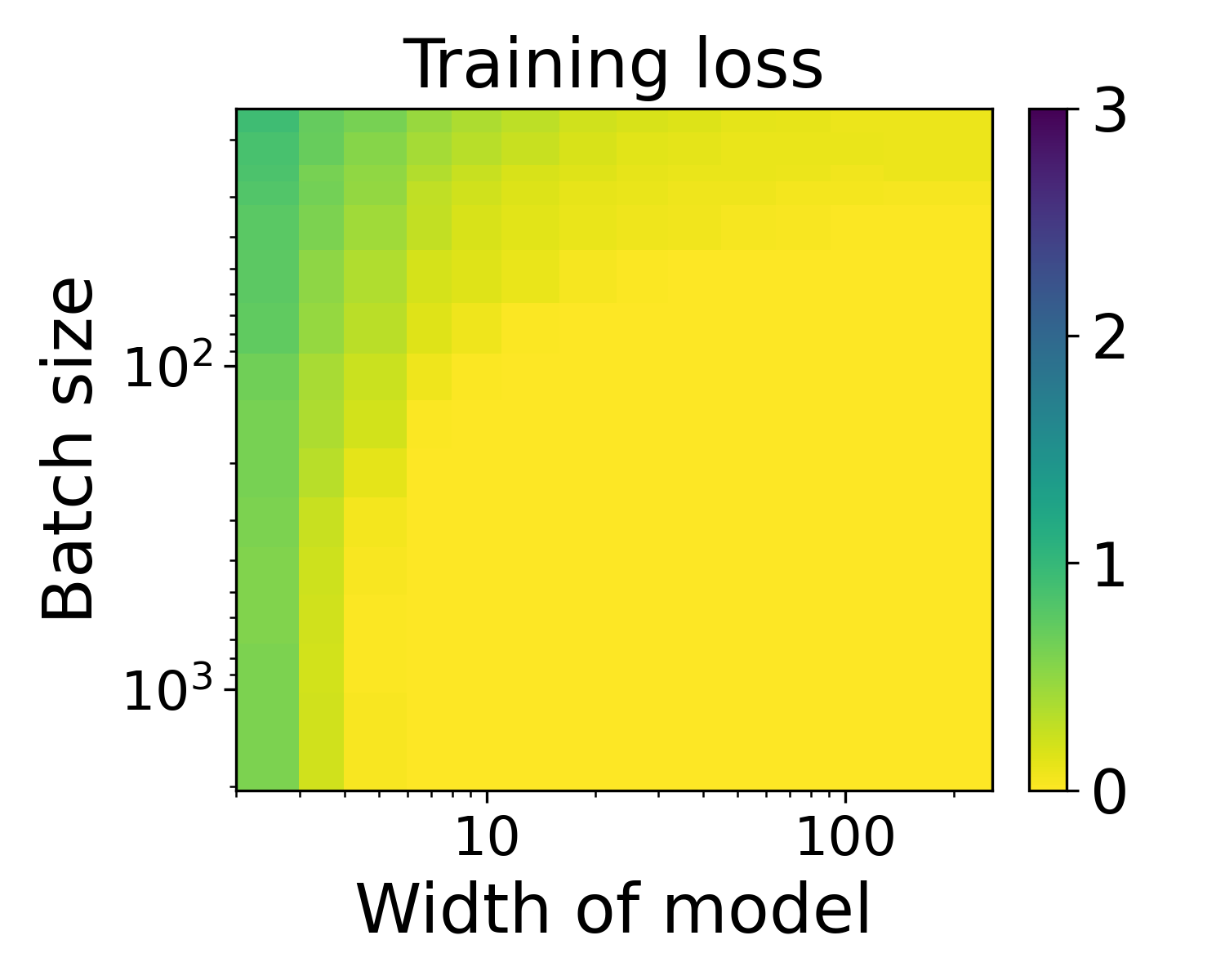}
      \vspace{-6mm}
      \caption{Training loss\label{fig:\figname_loss}}
    \end{subfigure}&
    \begin{subfigure}[c]{0.2\textwidth}
      \includegraphics[width=\textwidth]{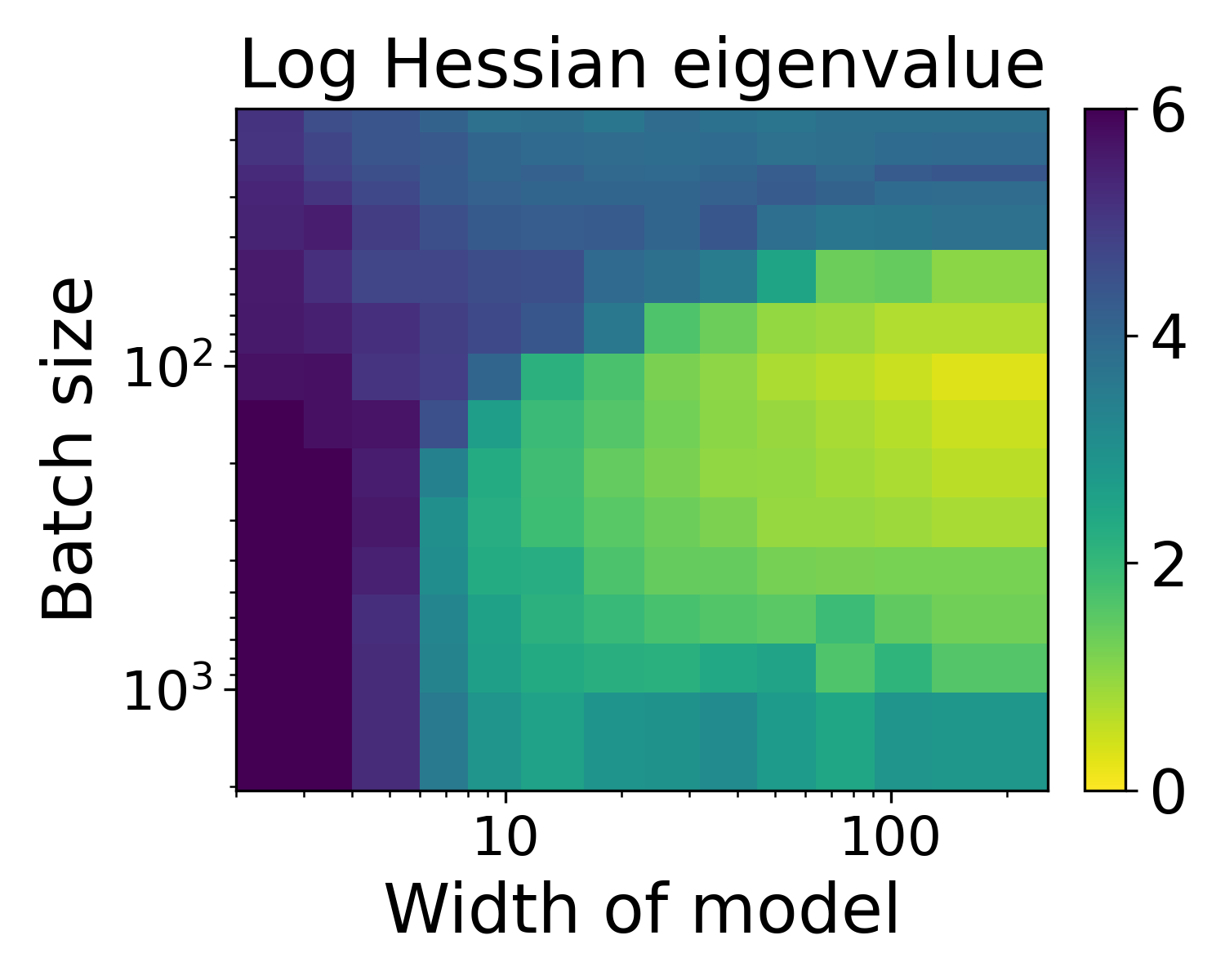}
      \vspace{-6mm}
      \caption{Hessian eigenvalue\label{fig:\figname_hessian_e}}
    \end{subfigure}&
    \begin{subfigure}[c]{0.2\textwidth}
      \includegraphics[width=\textwidth]{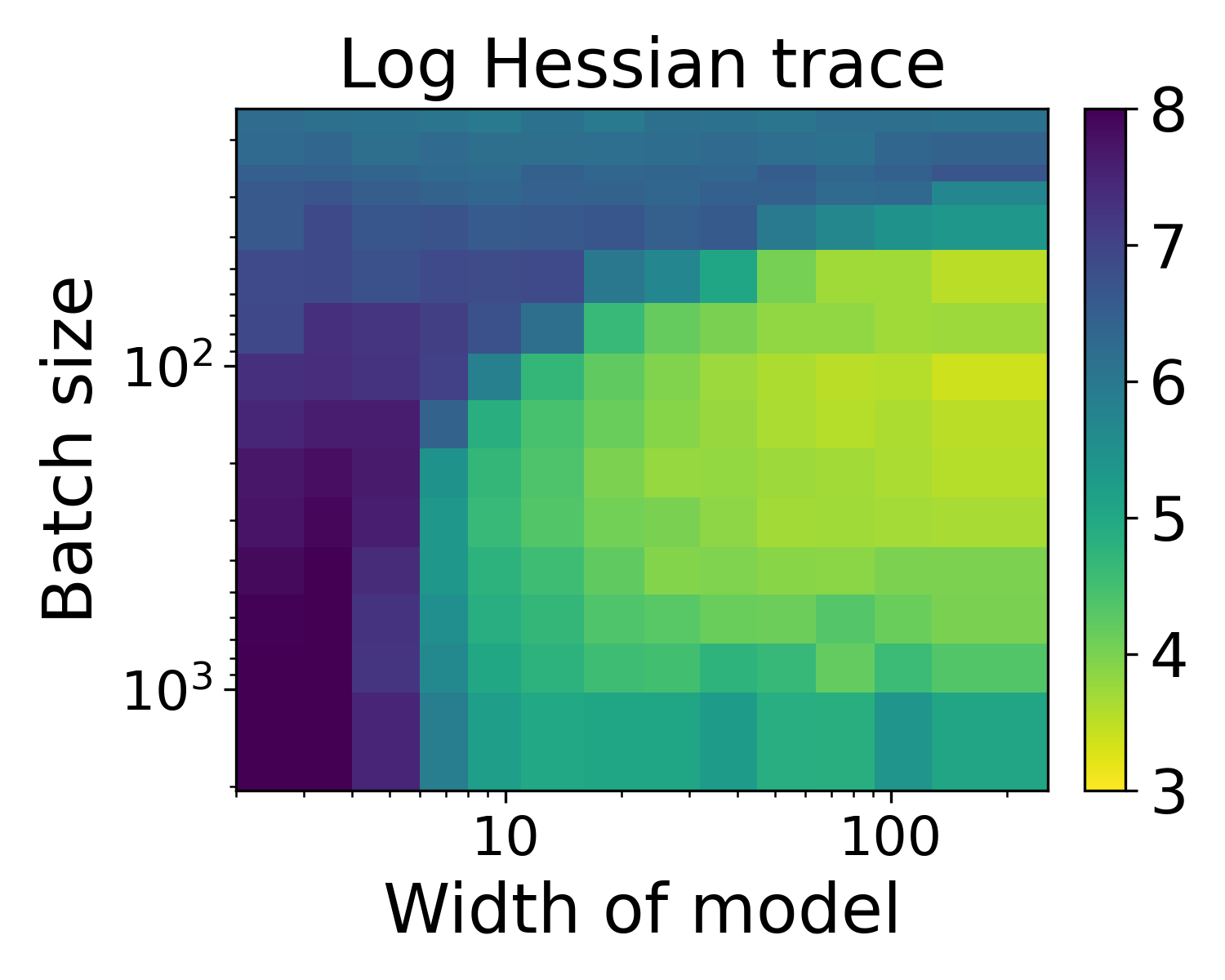}
      \vspace{-6mm}
      \caption{Hessian trace\label{fig:\figname_hessian_t}}
    \end{subfigure}\\
& \begin{subfigure}[c]{0.2\textwidth}
      \includegraphics[width=\textwidth]{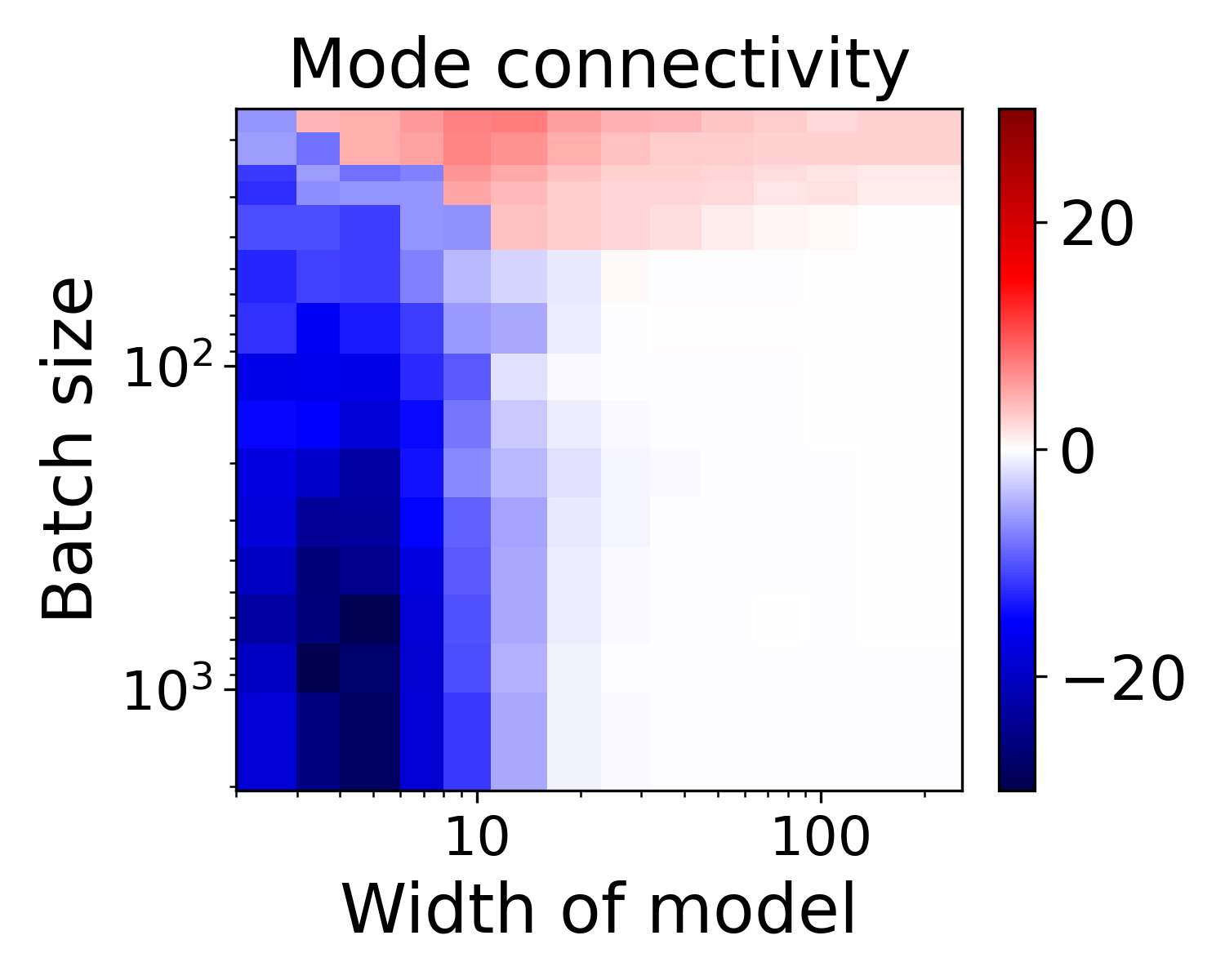}
      \vspace{-6mm}
      \caption{Mode connectivity\label{fig:\figname_curve}}
    \end{subfigure}&
    \begin{subfigure}[c]{0.2\textwidth}
      \includegraphics[width=\textwidth]{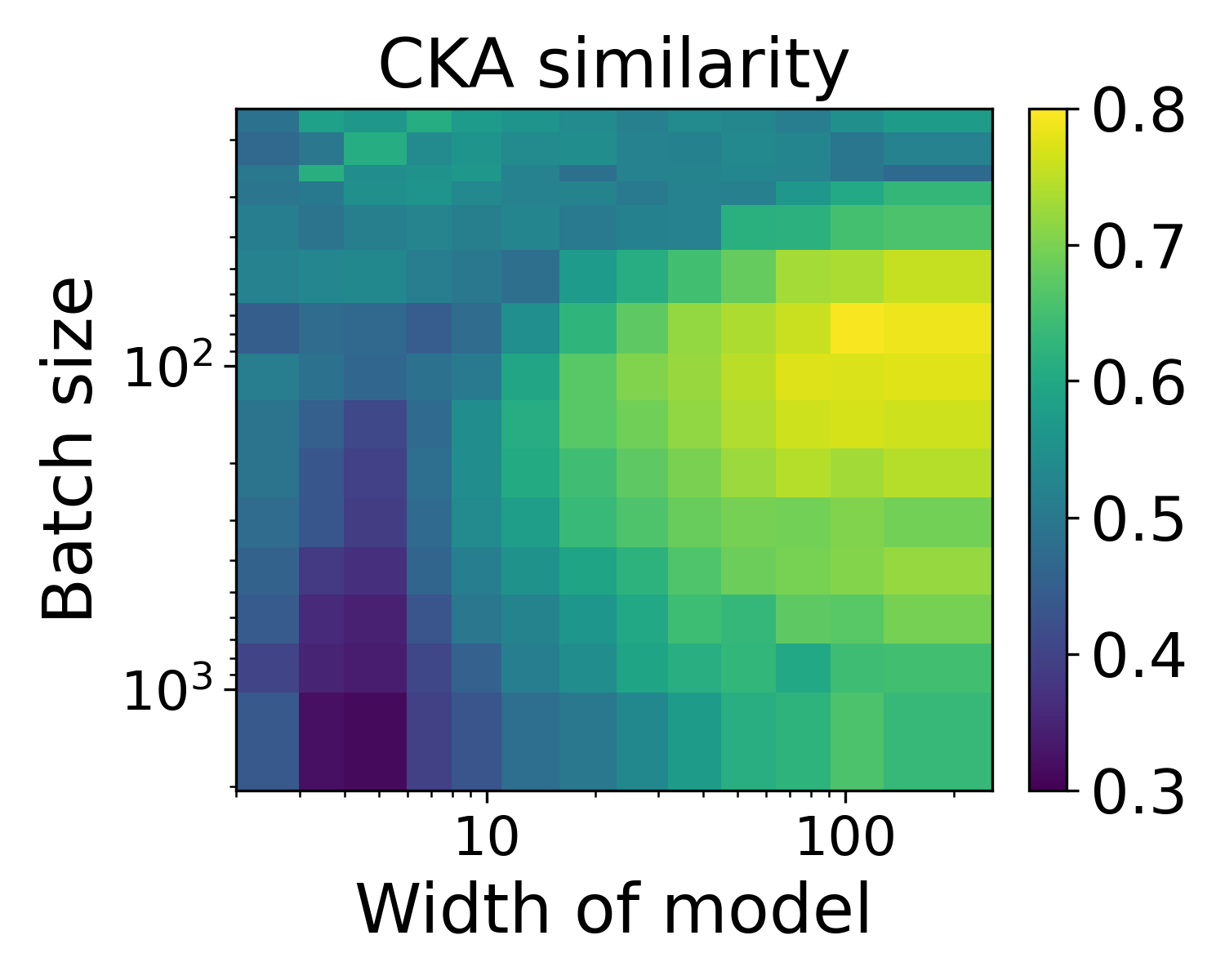}
      \vspace{-6mm}
      \caption{CKA similarity\label{fig:\figname_CKA}}
    \end{subfigure}&
    \begin{subfigure}[c]{0.2\textwidth}
      \includegraphics[width=\textwidth]{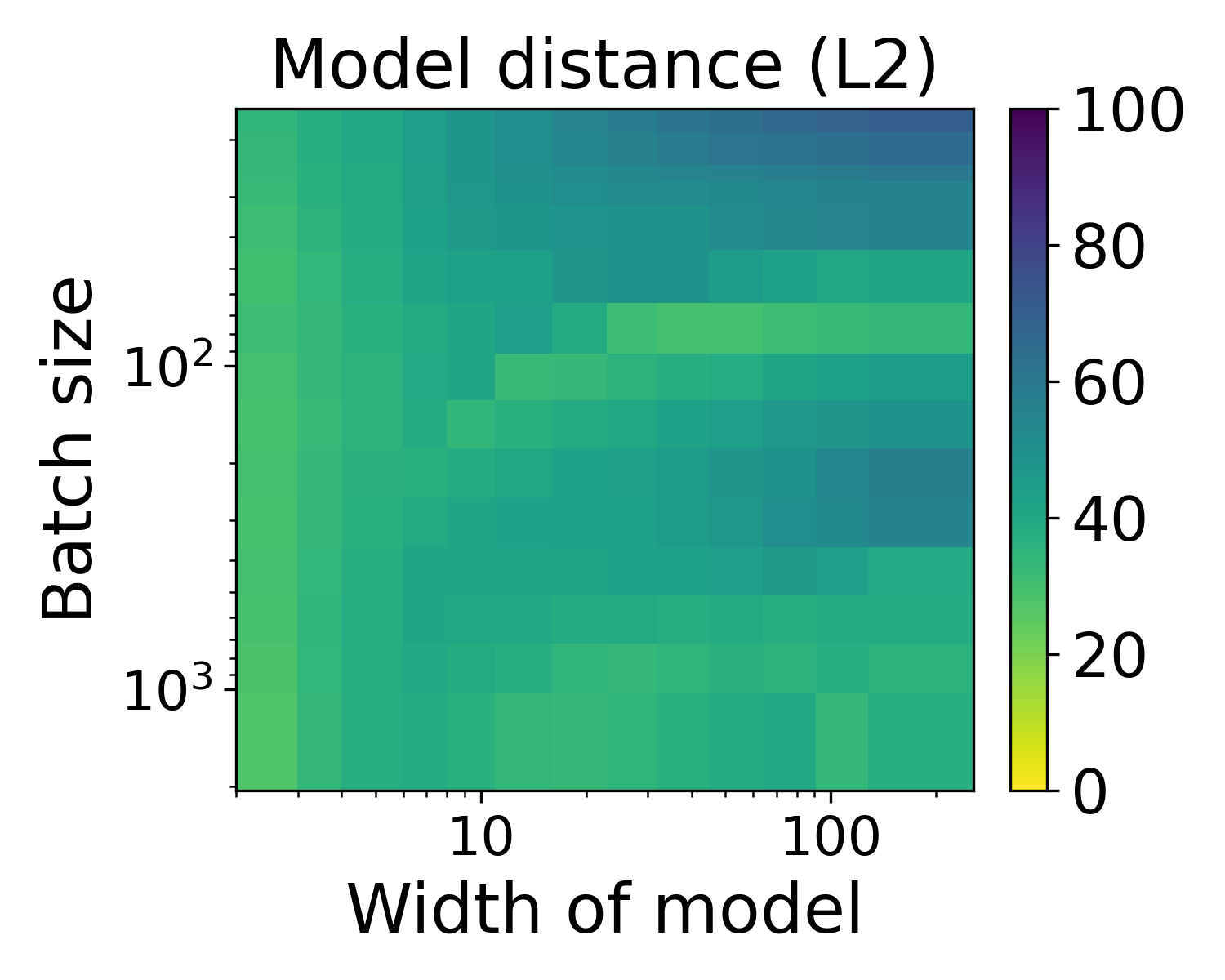}
      \vspace{-6mm}
      \caption{$\ell_2$ distance\label{fig:\figname_dist}}
    \end{subfigure}\\
  \end{tabular}    
  \caption{{\bf (Learning rate decay).} Partitioning the 2D load-like---temperature-like diagram into different phases of learning, varying batch size to change temperature and varying model width to change load. Learning rate decay is applied during training. Models are trained with ResNet18 on CIFAR-10. All plots are on the same set of axes.
  \ifisarxiv
  We note that batch size is inverse temperature, and thus it has smaller values at the top of the y-axis and larger values at the bottom.
  \fi
  \vspace{-5mm}
  }
  \label{fig:Lr_decay}
\end{figure}

{\bf Training to noisy labels and double descent.}
Next, we consider a similar experimental setup and the same phase diagram, except that we randomize 10\% of the training labels (similar to \citep{nakkiran2019deep}).
The results are presented in Figure~\ref{fig:Noisy_label}.
Comparing with Figure~\ref{fig:ResNet18}, we see that our main conclusion still holds, i.e., the loss landscape which is both globally nice and locally flat achieves the best test accuracy, shown in Phase IV-B.
However, an additional observation can be made: if we compare Figure \ref{fig:Noisy_label_accuracy} with Figure \ref{fig:ResNet18_accuracy}, a ``dark band'' arises between different learning phases.
In particular, from Figure \ref{fig:Noisy_label_accuracy}, we see that the test accuracy exhibits both width-wise and temperature-wise double descent \citep{belkin2019reconciling,belkin2020two,nakkiran2019deep,derezinski2019exact,liao2020random}, for certain parameter choices.
In particular, the shape of the dark band matches that of the transitions shown in Figure \ref{fig:Noisy_label_hessian_e} and \ref{fig:Noisy_label_hessian_t}.

{\bf Double descent and phases of learning.}
The significance of this ``dark band'' is the following.
A central prediction when viewing different phases of optimization landscapes from a statistical mechanics perspective~\citep{EB01_BOOK,martin2017rethinking} is that there should be ``bad fluctuations'' between qualitatively different phases of learning (e.g., see the transition that separates Phase I and II from Phase III and IV in Figure \ref{fig:Noisy_label_accuracy}). 
The connection between phases and fluctuations in the popular double descent \citep{belkin2019reconciling,belkin2020two} was made precise in analyzable settings \citep{liao2020random,derezinski2019exact}.
Here, we complement \citep{liao2020random,derezinski2019exact} by exhibiting the same type of transitions empirically between different phases in our taxonomy, and demonstrating that empirical double descent is a consequence of qualitatively different phases of learning.

{\bf Training to zero loss.}
Next, we use Figure~\ref{fig:Noisy_label} to discuss whether to train to (approximately) zero loss, which is popular in recent work. 
From Figure~\ref{fig:Noisy_label_loss}, we observe that Phase III and Phase IV achieve almost exactly zero loss, while Phase I and Phase II do not.
Once again, the loss experiences a more than tenfold decay when transitioning from Phase I/II to Phase III/IV.
However, if we restrict to globally poorly-connected regions and we restrict to a particular width value, i.e., selecting one column slice in the diagram that cuts through Phase III, such as the red block shown in Figure \ref{fig:Noisy_label_accuracy}, we see that the best test accuracy is obtained in Phase I/II, instead of Phase III.
Note that Phase I/II not only does not achieve zero loss, but it also has locally sharp minima (observed from Figure \ref{fig:Noisy_label_hessian_e} and \ref{fig:Noisy_label_hessian_t}).
This means that {\bf for globally poorly-connected loss landscapes, it is possible that converging to a locally flat region achieves lower accuracy than a locally sharp region.}
\ifisarxiv
More interestingly, this locally sharp region does not even converge to close-to-zero training loss. 

We should note that this observation is obtained only for a constant learning rate, without learning rate decay, which has been studied in Figure \ref{fig:Lr_decay}.
In other words, for Phase III, training is done with a low temperature throughout.
This can restrict the ability of SGD to ``explore'' the loss landscape to find a better minima.
Thus, even though Phase III achieves even lower accuracy than non-converged training configurations in Phase I/II, we attribute it to insufficient exploration.
Also, we note that Phase III has both poor mode connectivity and small CKA similarity, as shown in Figure~\ref{fig:Noisy_label_curve} and Figure~\ref{fig:Noisy_label_CKA}, respectively. 
However, one will wrongly predict that Phase III outperforms Phase I/II if one only looks at local sharpness, e.g., the Hessian plots shown in Figure~\ref{fig:Noisy_label_hessian_e}-\ref{fig:Noisy_label_hessian_t}, because both the Hessian eigenvalue and the Hessian trace are smaller in Phase III than in Phase I/II.
\fi

\def \figname {Noisy_label}
\begin{figure}
  \begin{tabular}[c]{cccc}
  \hspace{-5mm}
    \multirow{2}{*}{
    \begin{subfigure}{0.30\textwidth}
        \vspace{-5mm}
      \includegraphics[width=\textwidth]{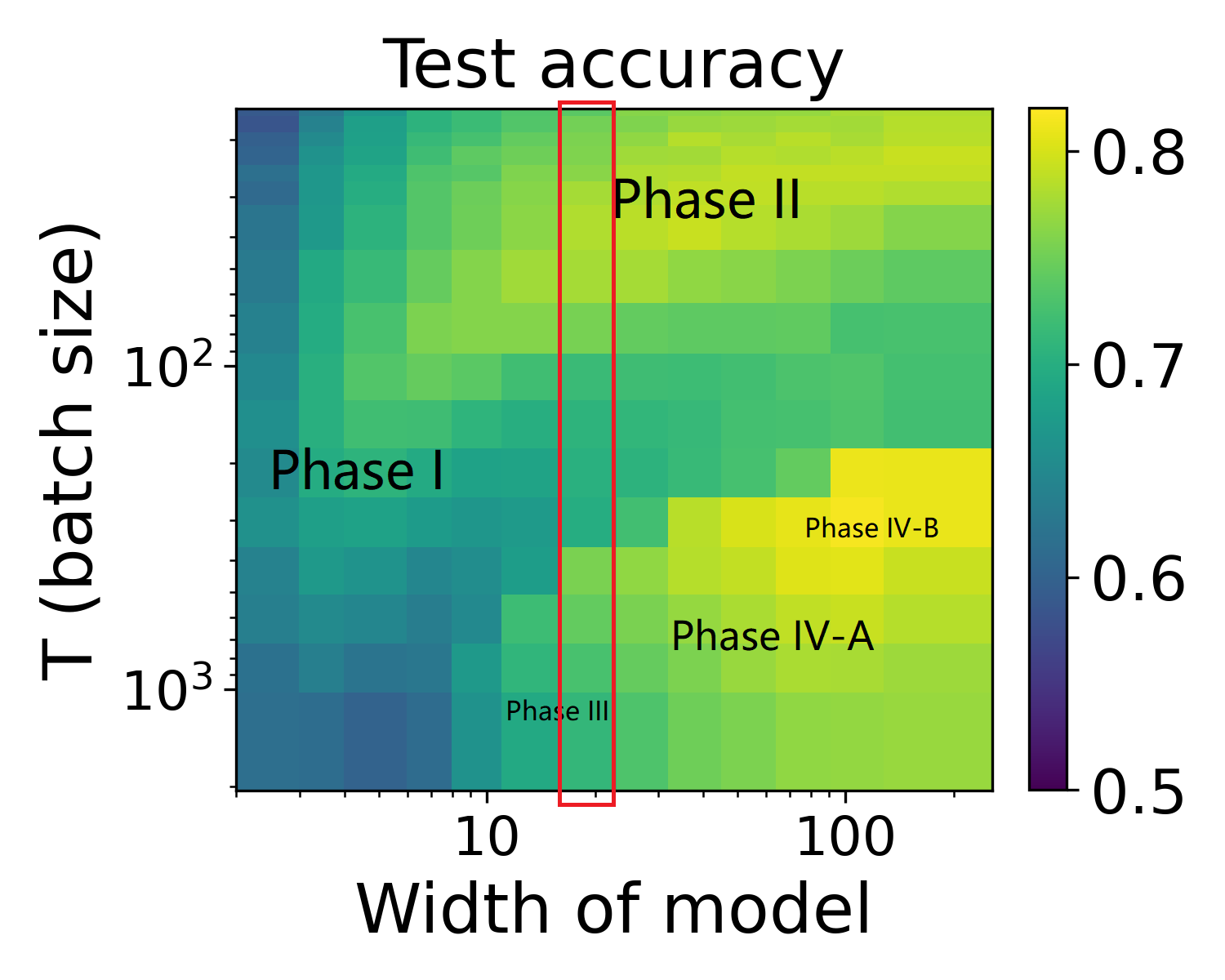}
      \caption{Test accuracy\label{fig:\figname_accuracy}}
    \end{subfigure}
} 
& \begin{subfigure}[c]{0.21\textwidth}
      \includegraphics[width=\textwidth]{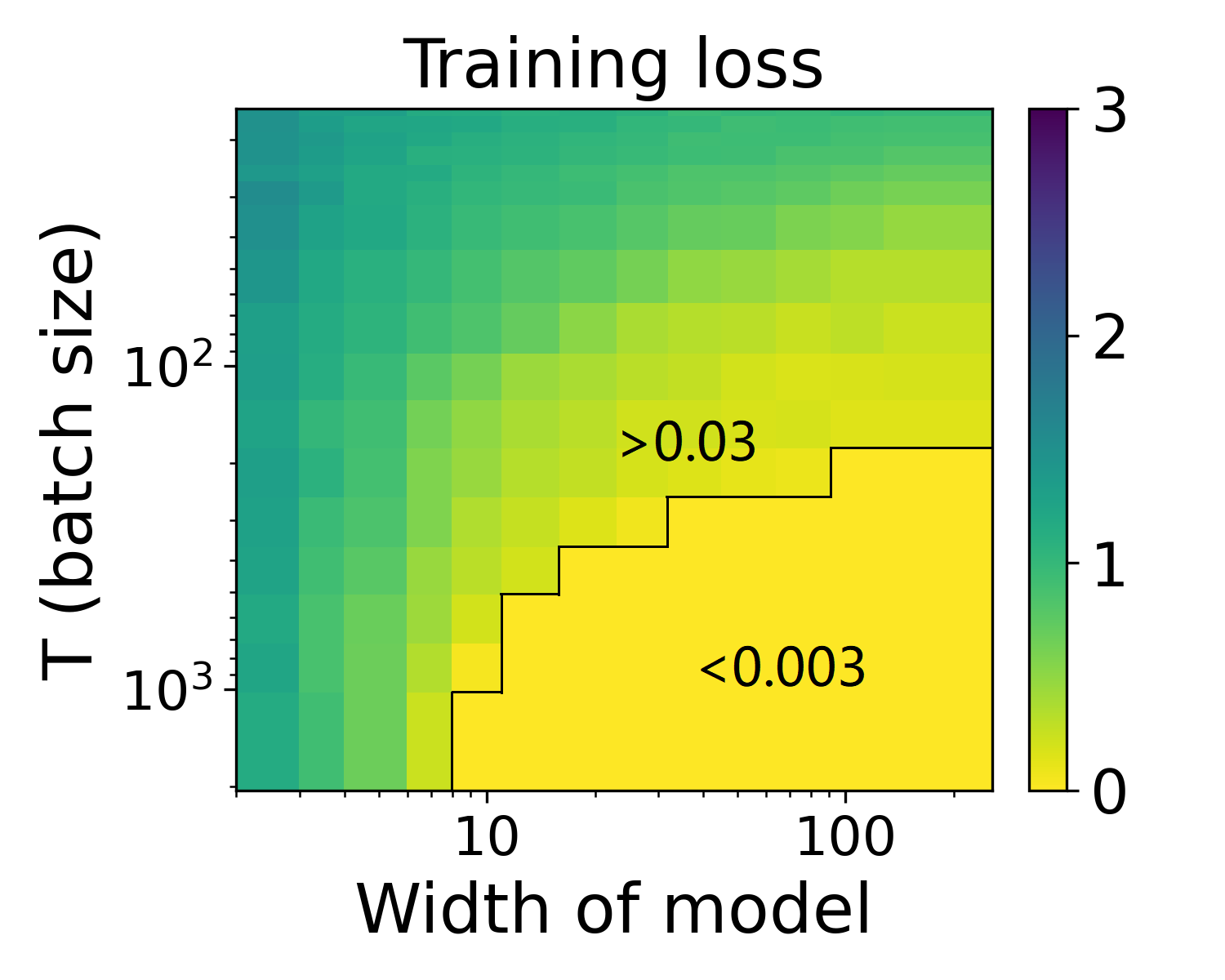}
      \caption{Training loss\label{fig:\figname_loss}}
    \end{subfigure}&
    \begin{subfigure}[c]{0.21\textwidth}
      \includegraphics[width=\textwidth]{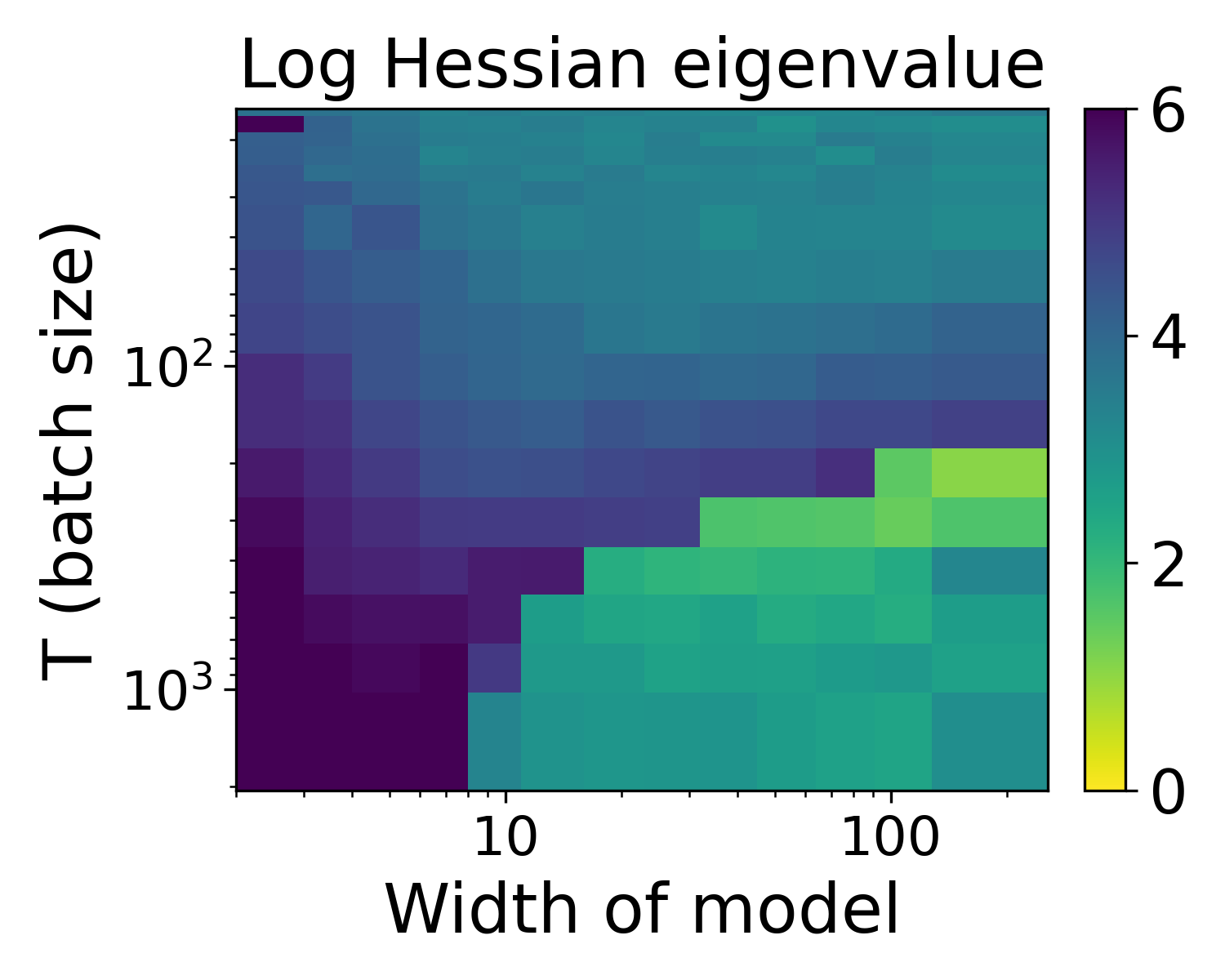}
      \caption{Hessian eigenvalue\label{fig:\figname_hessian_e}}
    \end{subfigure}&
    \begin{subfigure}[c]{0.21\textwidth}
      \includegraphics[width=\textwidth]{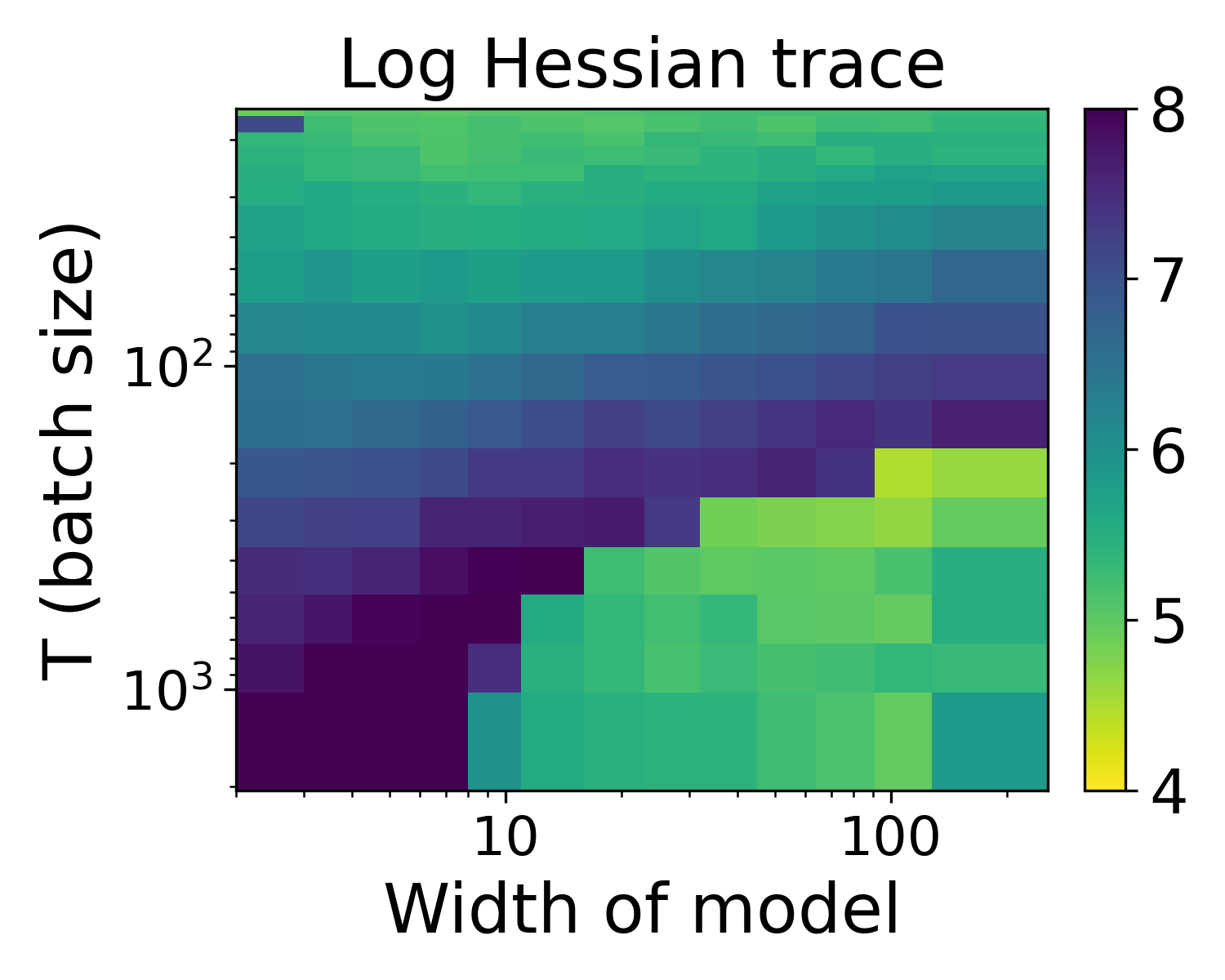}
      \caption{Hessian trace\label{fig:\figname_hessian_t}}
    \end{subfigure}\\
& \begin{subfigure}[c]{0.21\textwidth}
      \includegraphics[width=\textwidth]{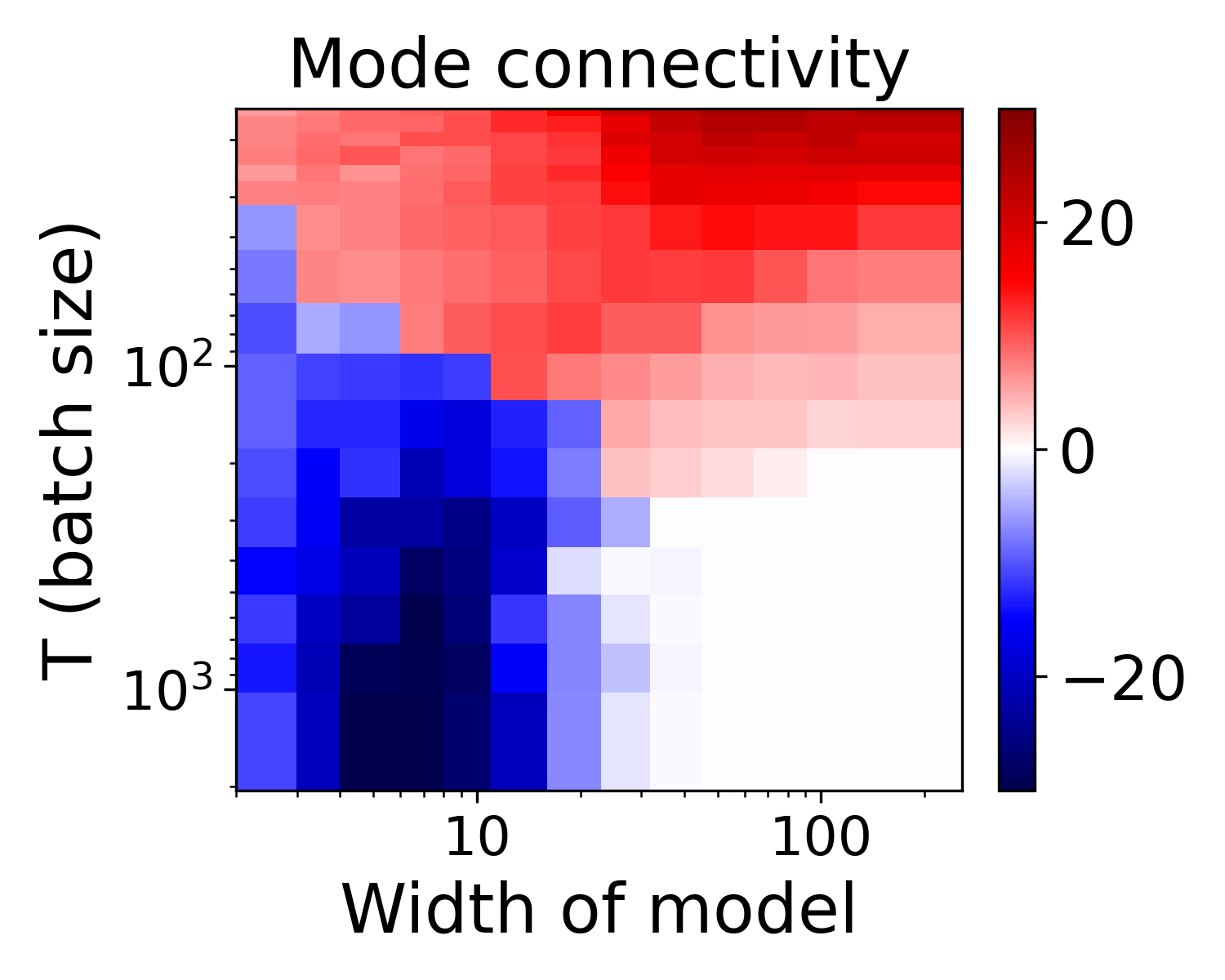}
      \caption{Mode connectivity\label{fig:\figname_curve}}
    \end{subfigure}&
    \begin{subfigure}[c]{0.21\textwidth}
      \includegraphics[width=\textwidth]{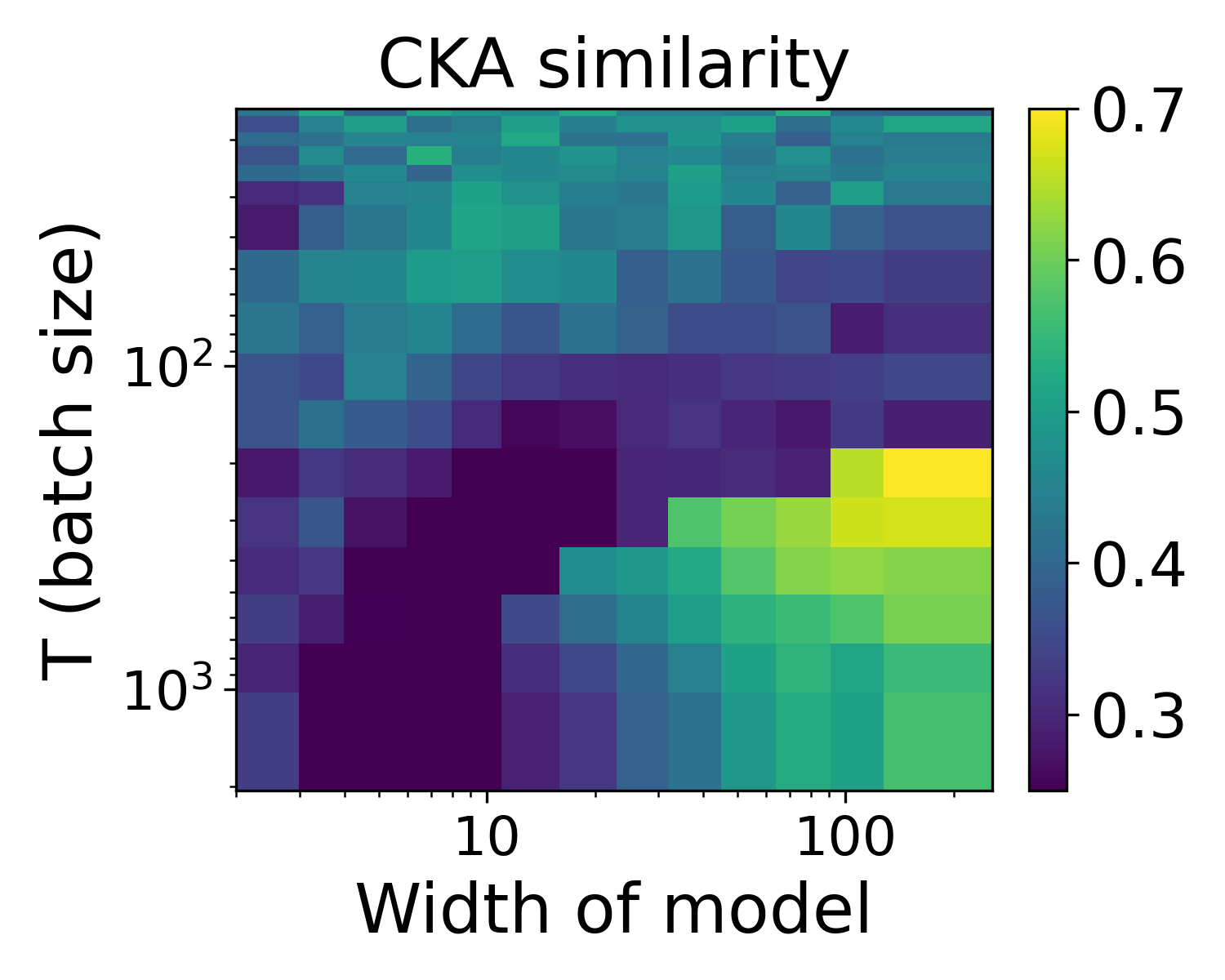}
      \caption{CKA similarity\label{fig:\figname_CKA}}
    \end{subfigure}&
    \begin{subfigure}[c]{0.21\textwidth}
      \includegraphics[width=\textwidth]{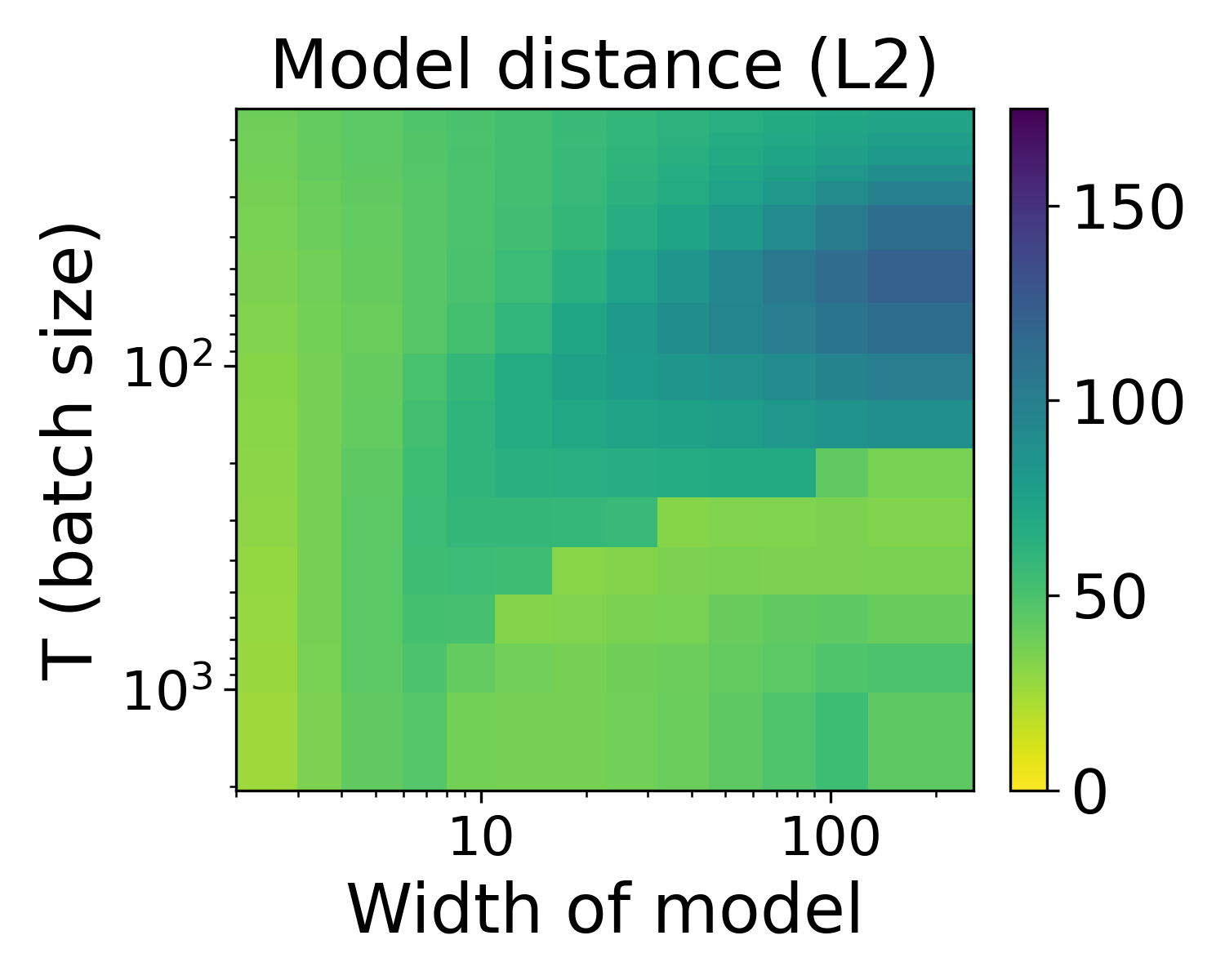}
      \caption{$\ell_2$ distance\label{fig:\figname_dist}}
    \end{subfigure}\\
  \end{tabular}    
  \caption{{\bf (Training to noisy labels and double descent).} Partitioning the 2D load-like---temperature-like diagram into different phases of learning, using batch size as the temperature and varying model width to change load.
  10\% of labels are randomized, and double descent is observed between different phases.
  For an arbitrary column slice that cuts through Phase III (e.g., the red block), optimal accuracy is achieved in Phase I/II with locally sharp minima.
  Models are trained with ResNet18 on CIFAR-10. All plots are on the same set of axes. \vspace{-2mm}
  }
  \label{fig:Noisy_label}
\end{figure}

\subsection{Ablation study}
\label{sec:ablation_study}

{\bf Different temperature parameters.}
First, we study weight decay as an alternative temperature parameter, in addition to batch size. 
We change the temperature parameter from batch size used in Figure~\ref{fig:ResNet18} to weight decay, and we report the results in Figure~\ref{fig:WD}.
The results shown in Figure \ref{fig:WD} are similar to those seen in Figure \ref{fig:ResNet18}. 
One observation is that, once again, the best test accuracy is obtained when the loss landscape is both globally nice and locally flat.
Another observation with Figure \ref{fig:WD_hessian_t} is that when training a wide model with small weight decay (which is shown on the bottom of the figure), the Hessian trace becomes extremely small. 
This matches observations in \citep{granziol2020flatness} that decreasing weight decay reduces the size of the Hessian. Since weight decay is known to improve generalization, this also demonstrates that local metrics alone are insufficient to predict test performance.

\def \figname {WD}
\begin{figure}
  \centering
    \begin{subfigure}{0.23\textwidth}
      \includegraphics[width=\textwidth]{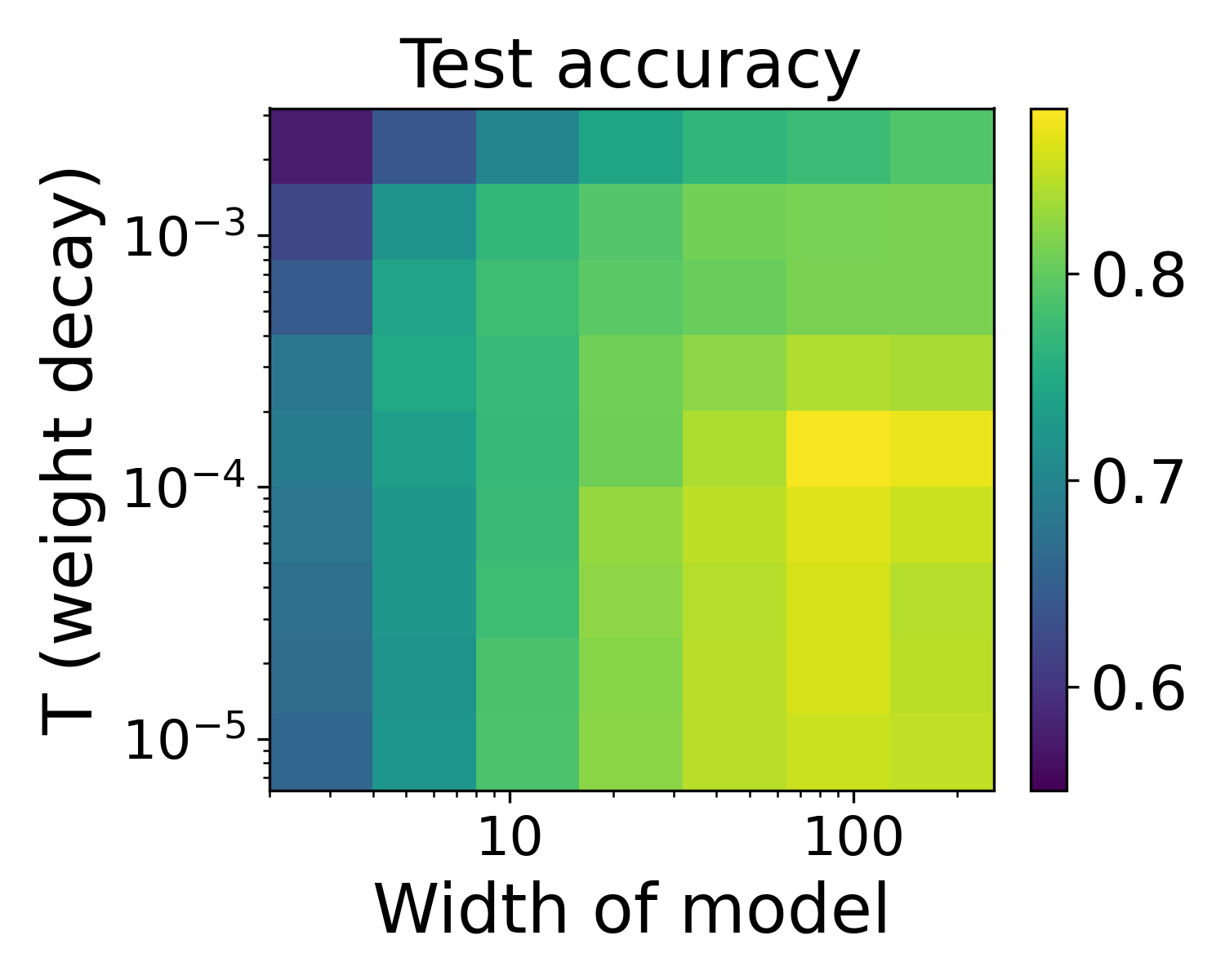}
      \vspace{-6mm}
      \caption{Test accuracy\label{fig:\figname_accuracy}}
    \end{subfigure}
    \begin{subfigure}[c]{0.23\textwidth}
      \includegraphics[width=\textwidth]{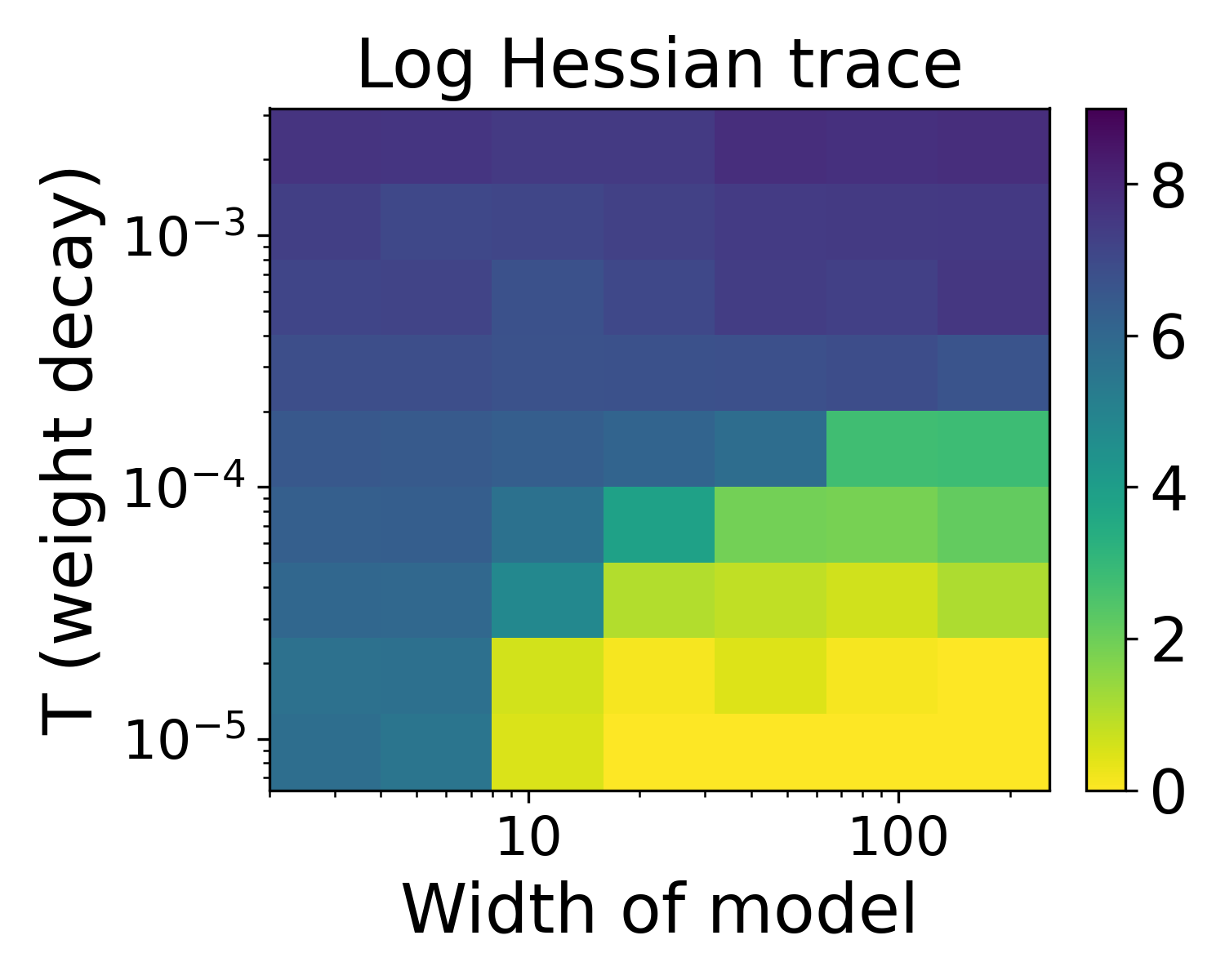}
      \vspace{-6mm}
      \caption{Hessian trace\label{fig:\figname_hessian_t}}
    \end{subfigure}
  \begin{subfigure}[c]{0.23\textwidth}
      \includegraphics[width=\textwidth]{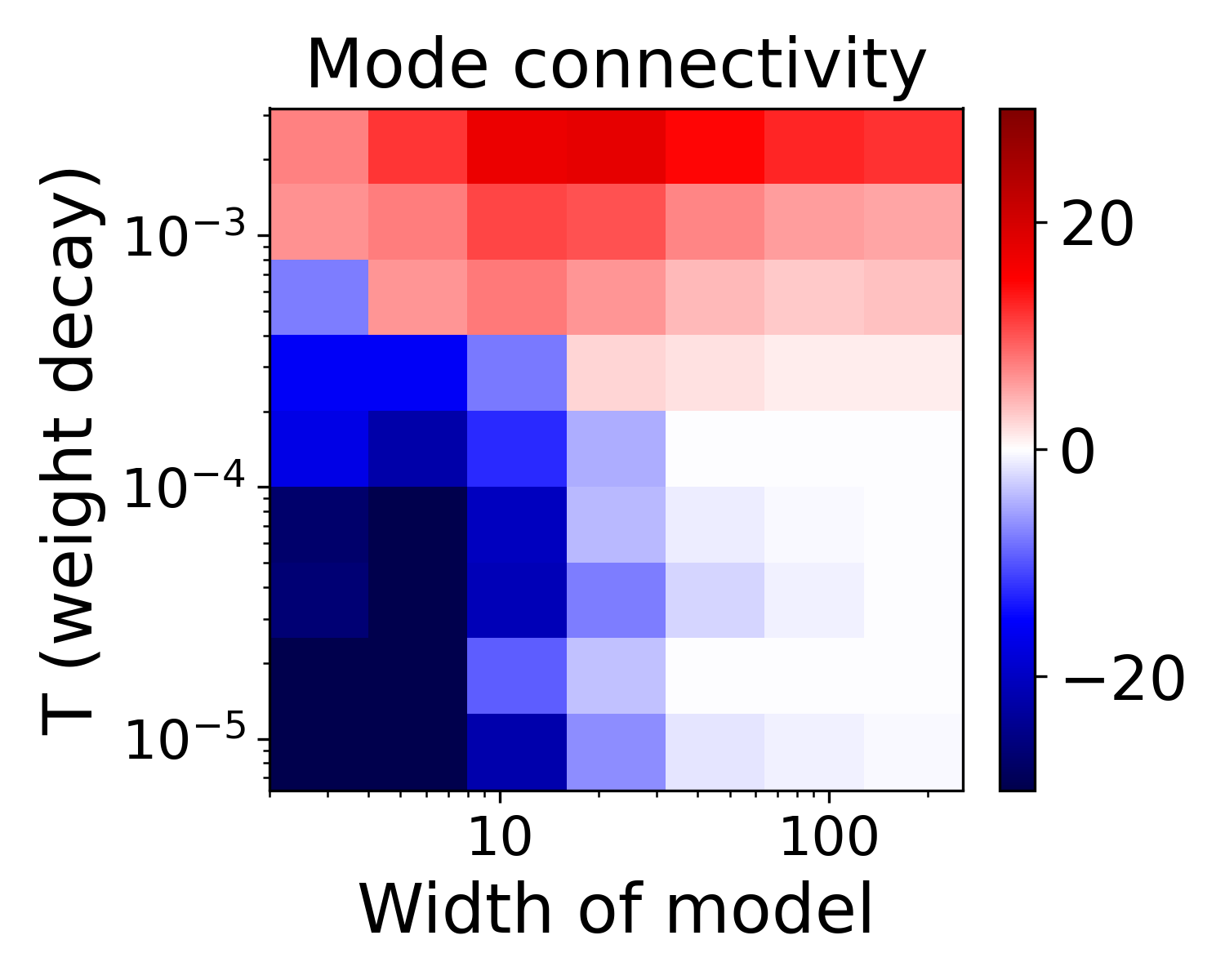}
      \vspace{-6mm}
      \caption{Mode connectivity\label{fig:\figname_curve}}
    \end{subfigure}
    \begin{subfigure}[c]{0.23\textwidth}
      \includegraphics[width=\textwidth]{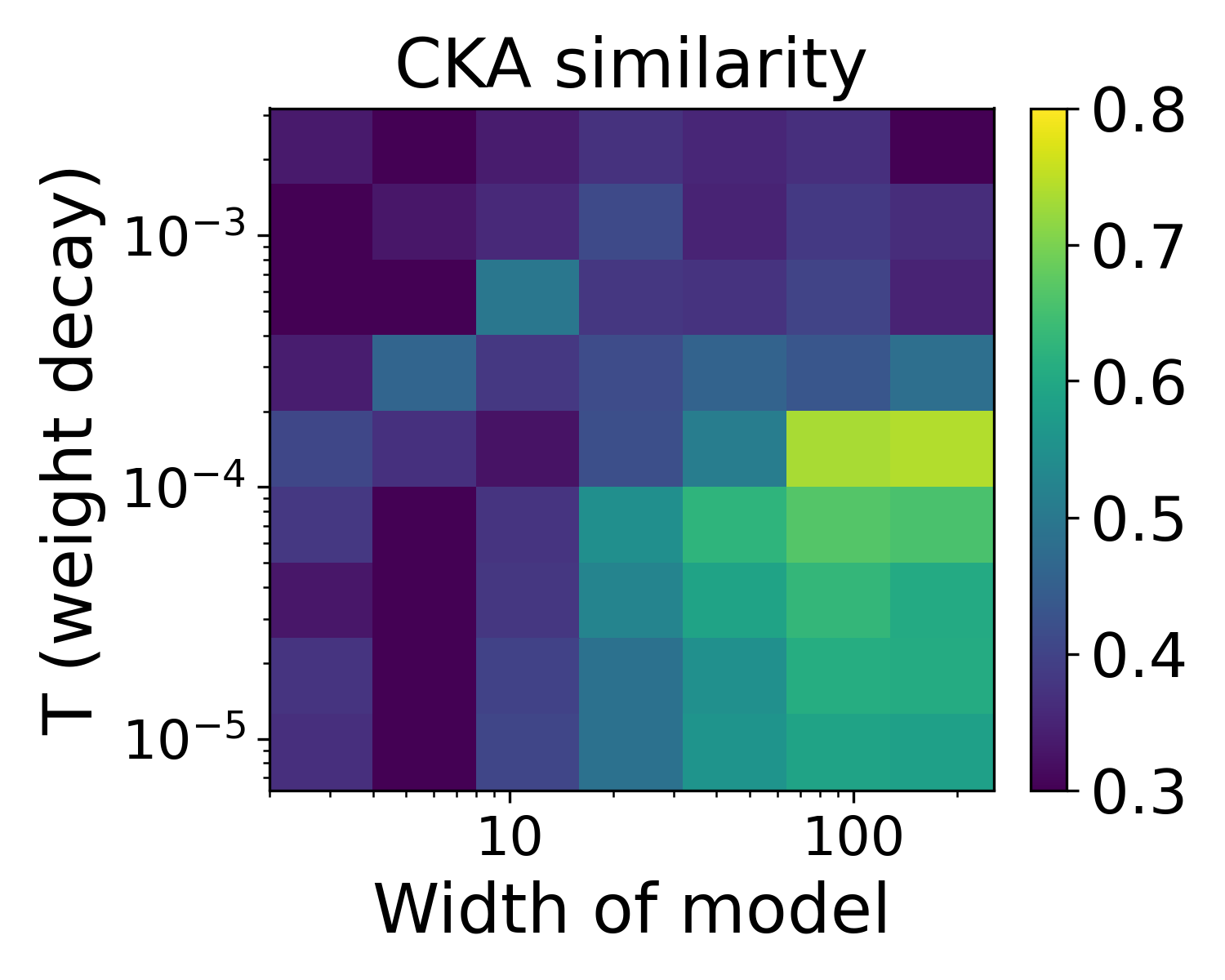}
      \vspace{-6mm}
      \caption{CKA similarity\label{fig:\figname_CKA}}
    \end{subfigure}
  \caption{{\bf (Weight decay as temperature).}
  Partitioning the 2D load-like---temperature-like diagram into different phases of learning, using weight decay as the temperature and varying model width to change load.
  Models are trained with ResNet18 on CIFAR-10. All plots are on the same set of axes.  \label{fig:WD}}
\end{figure}

{\bf Different amount of training data.}
Next, we vary the amount of training data (as another way of changing load) and see how that affects our results.
We vary the number of training samples in CIFAR-10 by a factor of ten.
Results are shown in Figure \ref{fig:different_data}. 
Again, the optimal test accuracy is achieved when the Hessian eigenvalue and trace are small, mode connectivity is near-zero, and CKA similarity is large. 
Perhaps unsurprisingly, better test accuracy is achieved with more data.
Here, CKA provides useful complementary information to the Hessian and mode connectivity for explaining the utility of larger data. 
The Hessian alone cannot predict the correct trend, as it \emph{increases} in magnitude with data. 
Mode connectivity alone cannot predict the correct trend either, becoming increasingly poor with larger data (see the shrinking white region). 
Indeed, it appears that larger models are required to keep the loss landscape well-connected with increasing data. 
In contrast, CKA precisely captures the relationship of increasing test accuracy with additional data.

\ifisarxiv
To make these trends more visible, we rearrange the plots in Figure~\ref{fig:different_data} by replacing the X-axis of each figure with the amount of data, while having separate plots for different model width. 
We also include results on both training with or without noisy labels. 
See the rearranged results in Figure~\ref{fig:different_data_flip}. 
Now, we can clearly see that CKA is the only metric that precisely captures the relationship of increasing test accuracy with additional data. 
Interestingly, we observe double descent in the test-accuracy plots, for both training with and without noisy labels, consistent with our previous results.
\fi

These observations also imply that the utilities of extra data and larger models are different: 
larger models can increase connectivity in the loss landscape (e.g., Figure~\ref{fig:ResNet18_curve}); while increasing data boosts signal in the landscape, enabling trained models to become more similar to each other.
Clearly, researchers have been increasing both the size of data and the size of models in recent years; our methodology suggests obvious directions for doing this in more principled ways.

\def \width {0.13}
\def \firstwidth {0.11}
\begin{figure}
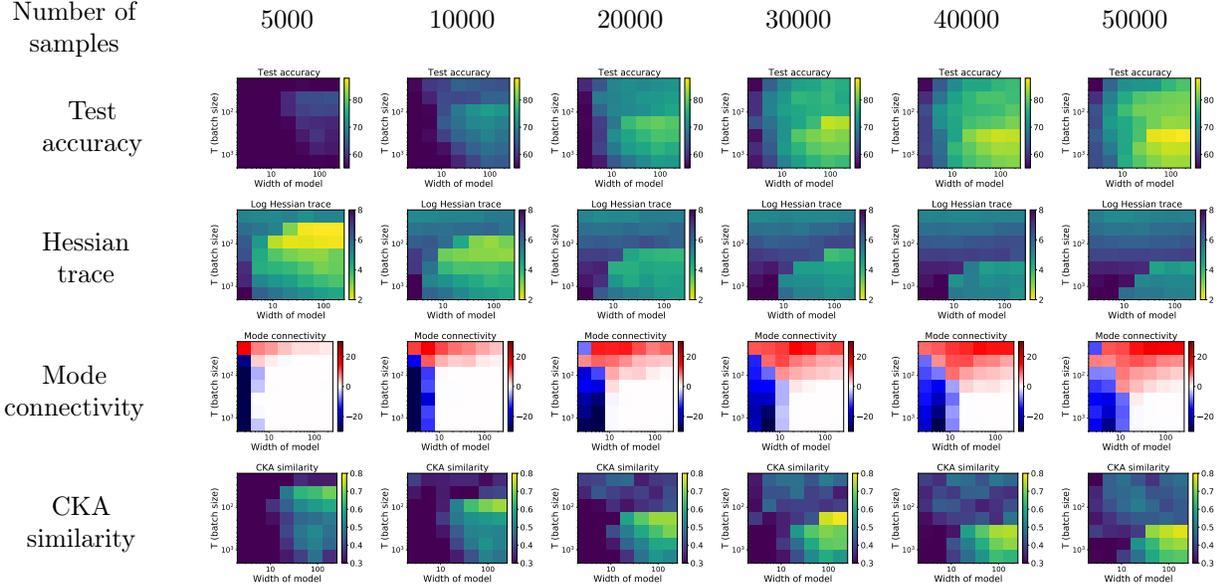

    \begin{tabular}{m{0.15\textwidth}m{\firstwidth\textwidth}m{\firstwidth\textwidth}m{\firstwidth\textwidth}m{\firstwidth\textwidth}m{\firstwidth\textwidth}m{\firstwidth\textwidth}}
    \begin{tabular}{c}\hspace{-4mm}Number of\\\hspace{-4mm}samples\end{tabular}&\begin{tabular}{c}5000\end{tabular}&\begin{tabular}{c}10000\end{tabular}&\begin{tabular}{c}20000\end{tabular}&\begin{tabular}{c}30000\end{tabular}&\begin{tabular}{c}40000\end{tabular}&\begin{tabular}{c}50000\end{tabular}
    \end{tabular}
    \foreach \l / \s in
    {\begin{tabular}{c}\hspace{-5mm}Test\\ \hspace{-5mm}accuracy\end{tabular}/accuracy,\begin{tabular}{c}\hspace{-5mm}Hessian\\ \hspace{-5mm}trace\end{tabular}/hessian_t,\begin{tabular}{c}\hspace{-3mm}Mode\\ \hspace{-3mm}connectivity\end{tabular}/curve,\begin{tabular}{c}CKA\\ similarity\end{tabular}/CKA}{
    \begin{minipage}[c]{0.15\textwidth}\quad\quad\l
    \end{minipage}
    \begin{minipage}[c]{\width\textwidth}
    \includegraphics[width=\textwidth]{figs/Subset_01_\s.pdf}
    \end{minipage}
    \begin{minipage}[c]{\width\textwidth}
    \includegraphics[width=\textwidth]{figs/Subset_02_\s.pdf}
    \end{minipage}
    \begin{minipage}[c]{\width\textwidth}
    \includegraphics[width=\textwidth]{figs/Subset_04_\s.pdf}
    \end{minipage}
    \begin{minipage}[c]{\width\textwidth}
    \includegraphics[width=\textwidth]{figs/Subset_06_\s.pdf}
    \end{minipage}
    \begin{minipage}[c]{\width\textwidth}
    \includegraphics[width=\textwidth]{figs/Subset_08_\s.pdf}
    \end{minipage}
    \begin{minipage}[c]{\width\textwidth}
    \includegraphics[width=\textwidth]{figs/Subset_10_\s.pdf}
    \end{minipage}
    \\
    }
    \caption{{\bf (Varying amount of training data).}
    Partitioning the 2D load-like---temperature-like diagram into different phases of learning, using batch size as the temperature and varying model width to change load.
    We vary quantities of training data from CIFAR-10 in different columns. All plots are on the same set of~axes.
    \label{fig:different_data}
    \vspace{-5mm}
    }
\end{figure}

\ifisarxiv
\def \metricwidth {0.14}
\def \firstwidth {0.12}
\begin{figure}
    \begin{tabular}{M{0.10\textwidth}M{\firstwidth\textwidth}M{\firstwidth\textwidth}M{0.14\textwidth}|M{\firstwidth\textwidth}M{\firstwidth\textwidth}M{\firstwidth\textwidth}}
    \begin{tabular}{c}\hspace{0mm}Model \\\hspace{0mm}width\end{tabular}&\begin{tabular}{c}4\end{tabular}&\begin{tabular}{c}8\end{tabular}&\begin{tabular}{c}16\end{tabular}&\begin{tabular}{c}8\end{tabular}&\begin{tabular}{c}16\end{tabular}&\begin{tabular}{c}32\end{tabular}\\
    \hline
    \begin{tabular}{c}\hspace{0mm}Noisy \\\hspace{0mm}label\end{tabular}&\begin{tabular}{c}No\end{tabular}&\begin{tabular}{c}No\end{tabular}&\begin{tabular}{c}No\end{tabular}&\begin{tabular}{c}Yes\end{tabular}&\begin{tabular}{c}Yes\end{tabular}&\begin{tabular}{c}Yes\end{tabular}\\
    \hline
    \begin{tabular}{c}\hspace{0mm}Test \\\hspace{0mm}accuracy\end{tabular}
    &
    \includegraphics[width=\metricwidth\textwidth]{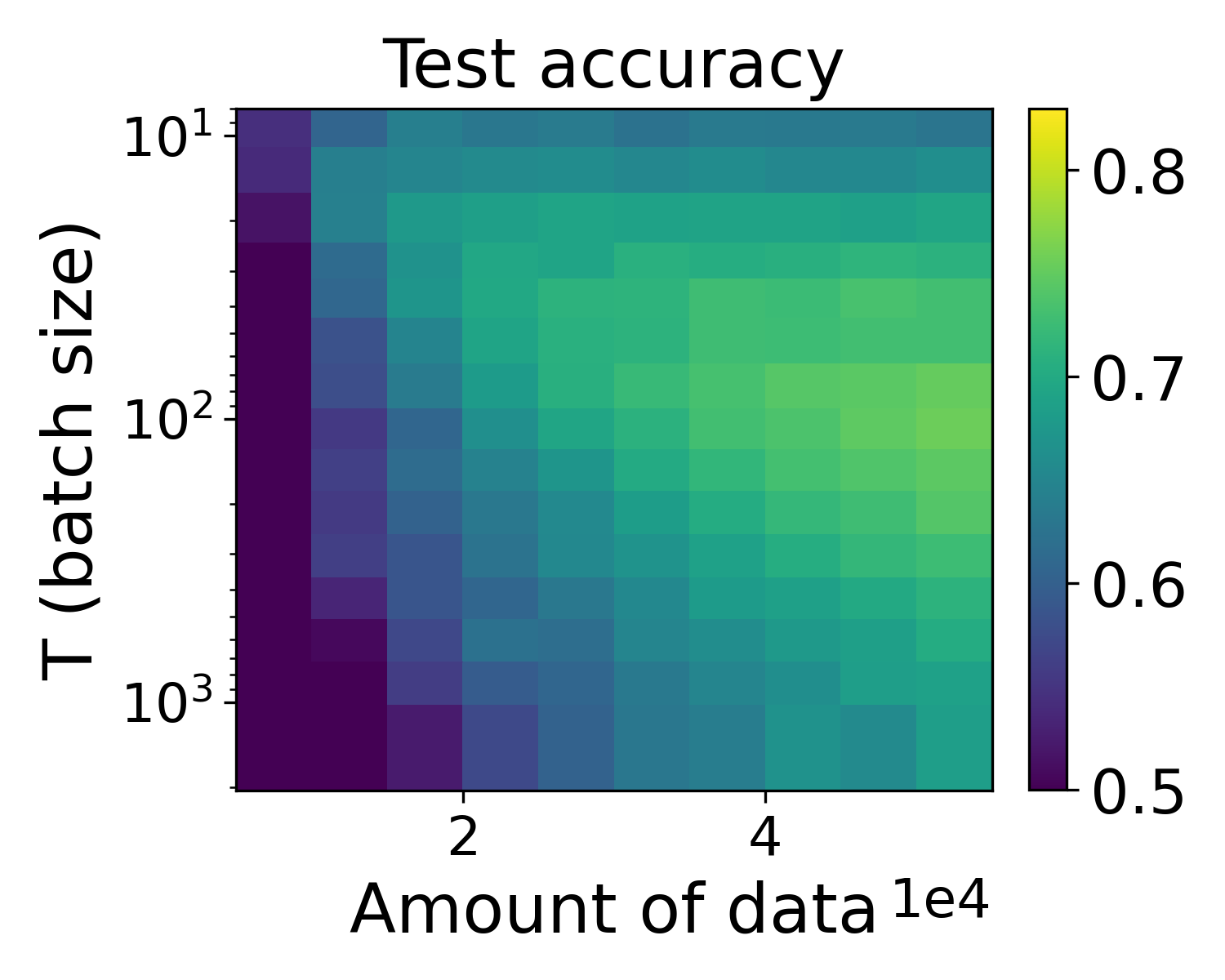}
    &
    \includegraphics[width=\metricwidth\textwidth]{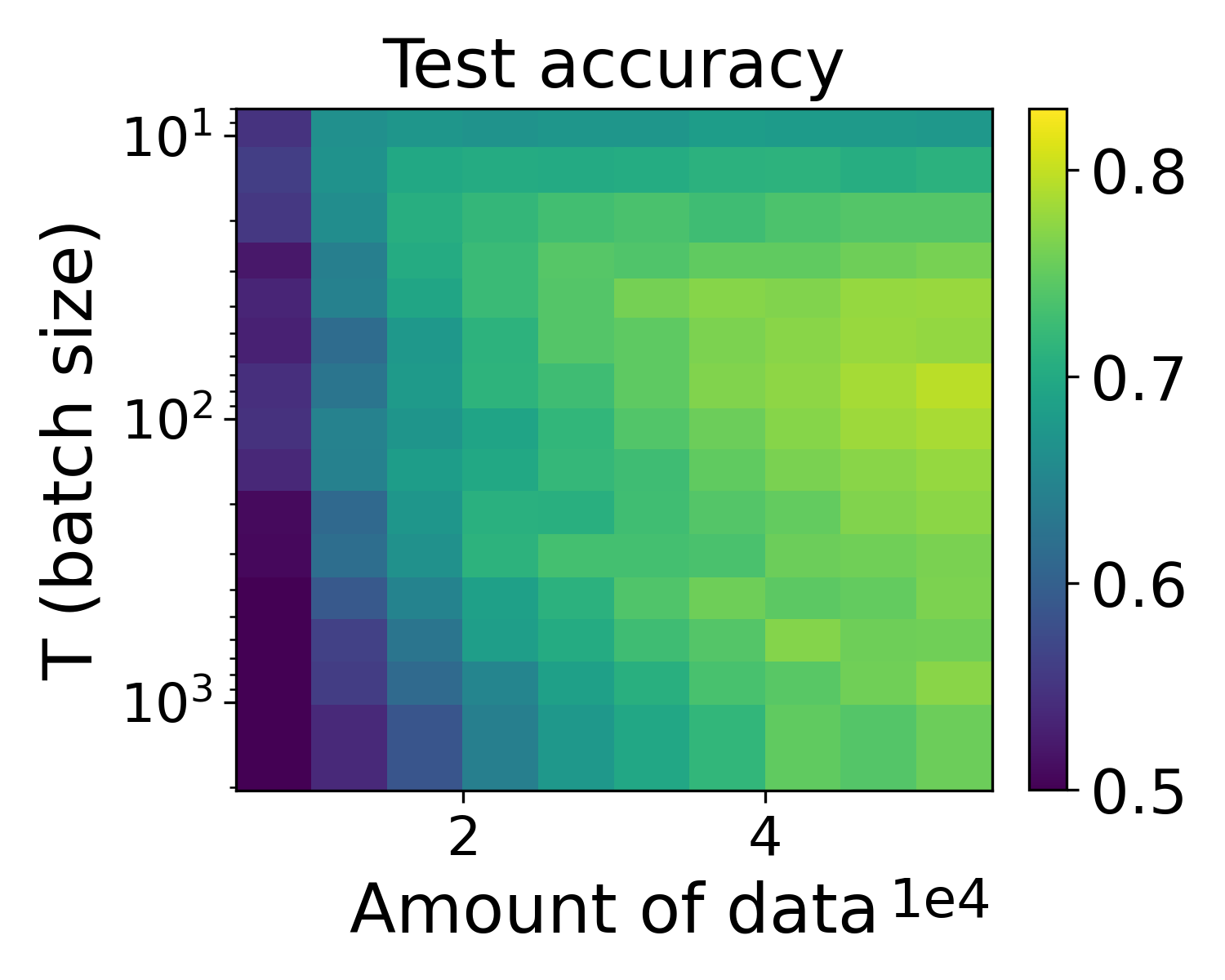}
    &
    \includegraphics[width=\metricwidth\textwidth]{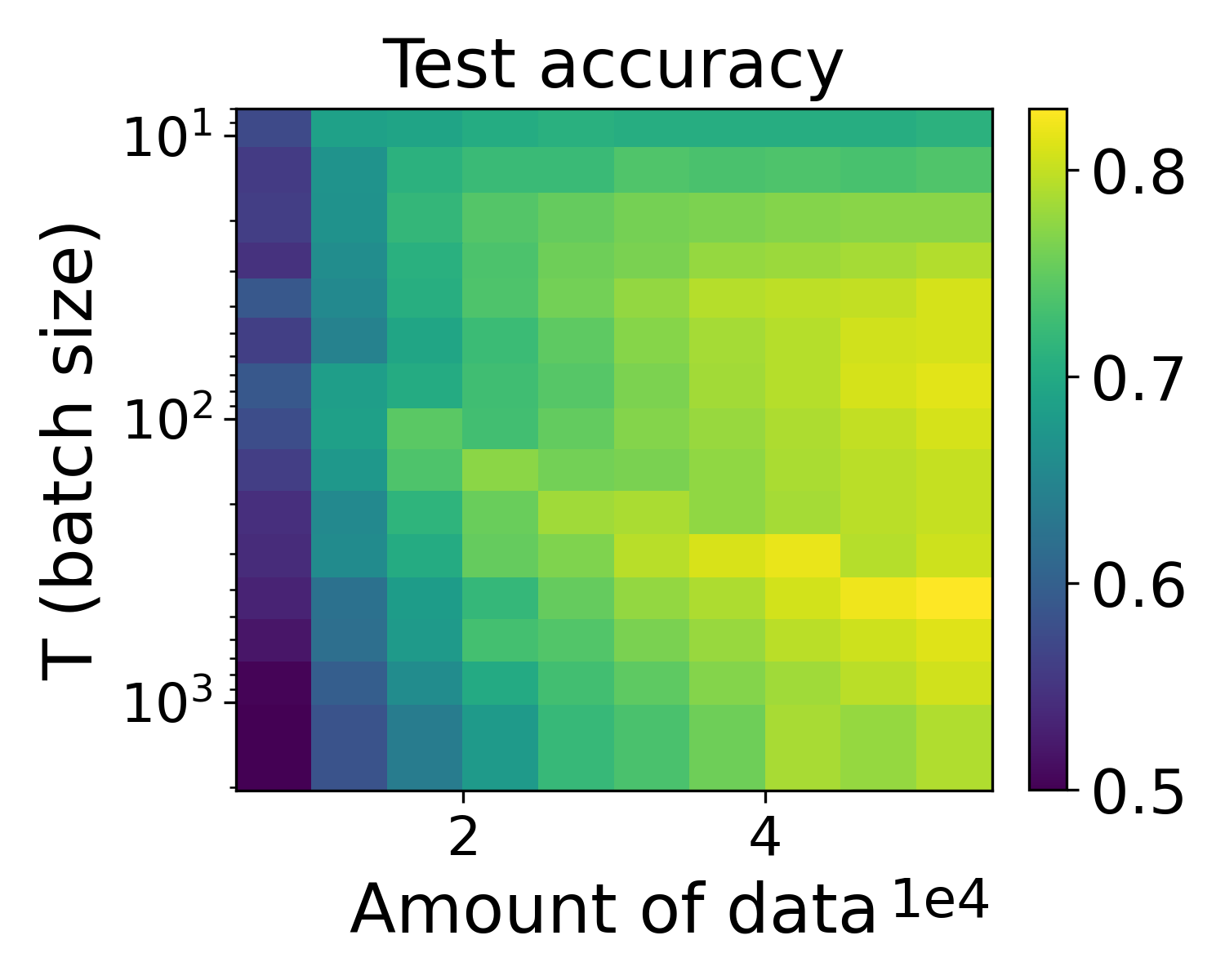}
    &
    \includegraphics[width=\metricwidth\textwidth]{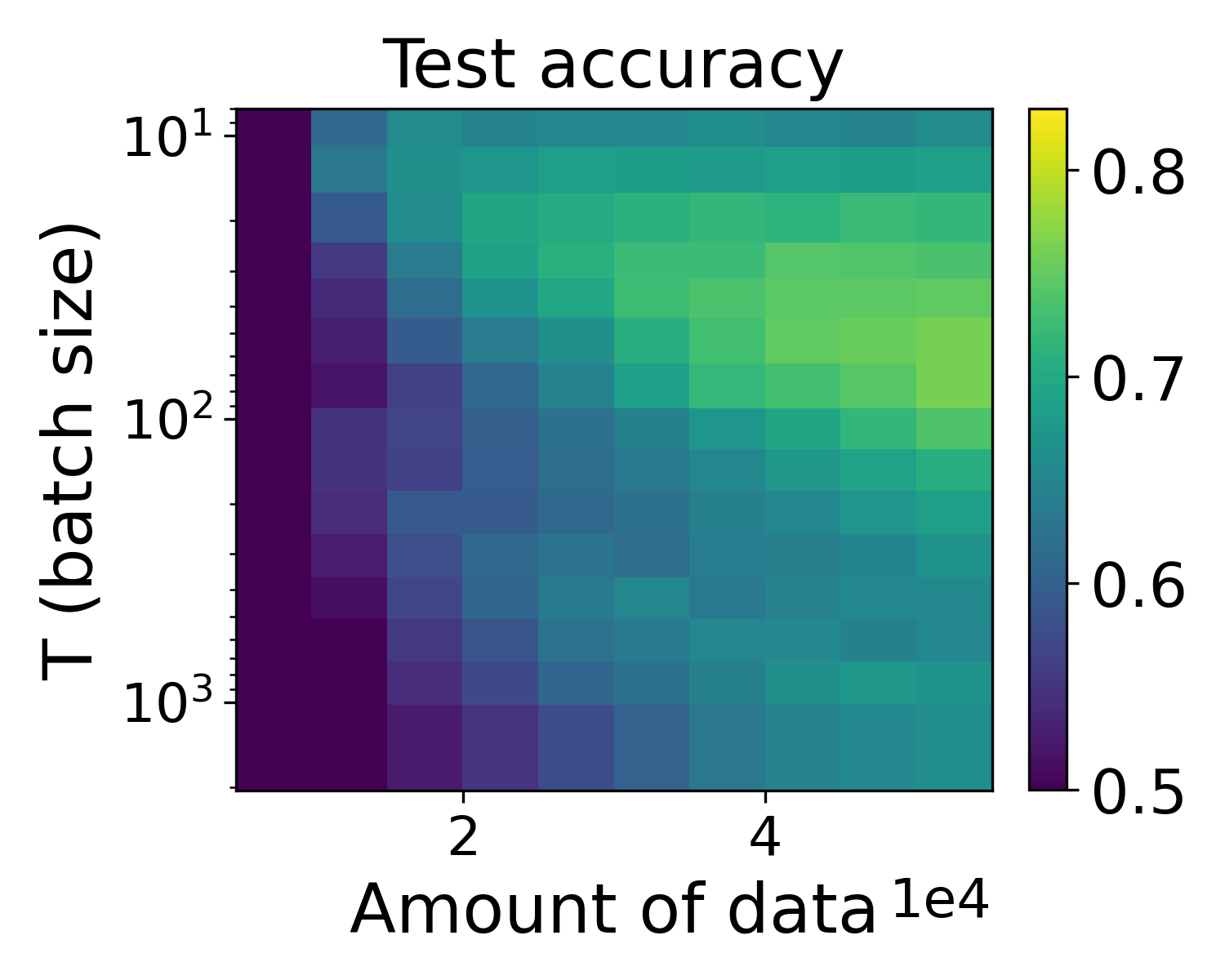}
    &
    \includegraphics[width=\metricwidth\textwidth]{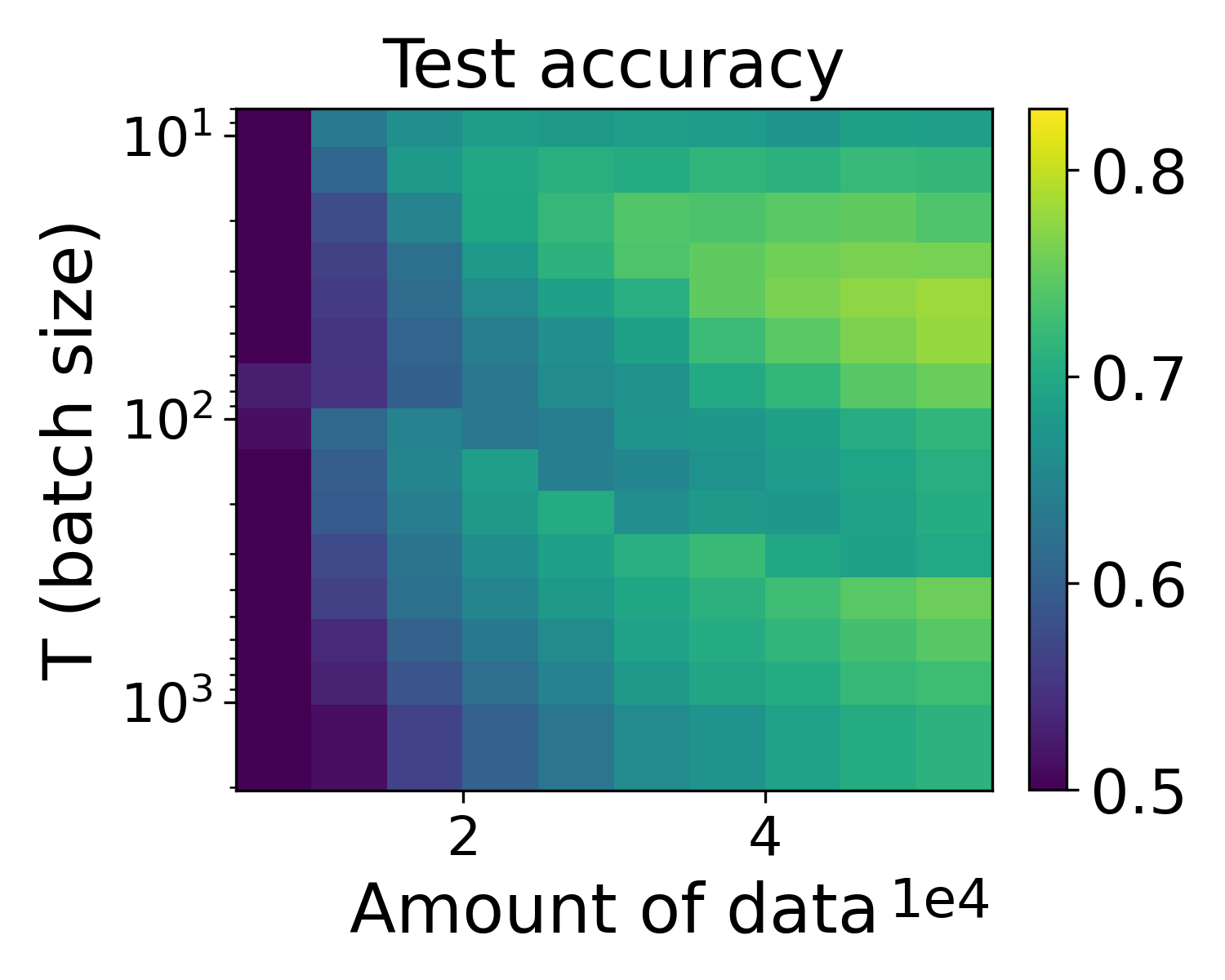}
    &
    \includegraphics[width=\metricwidth\textwidth]{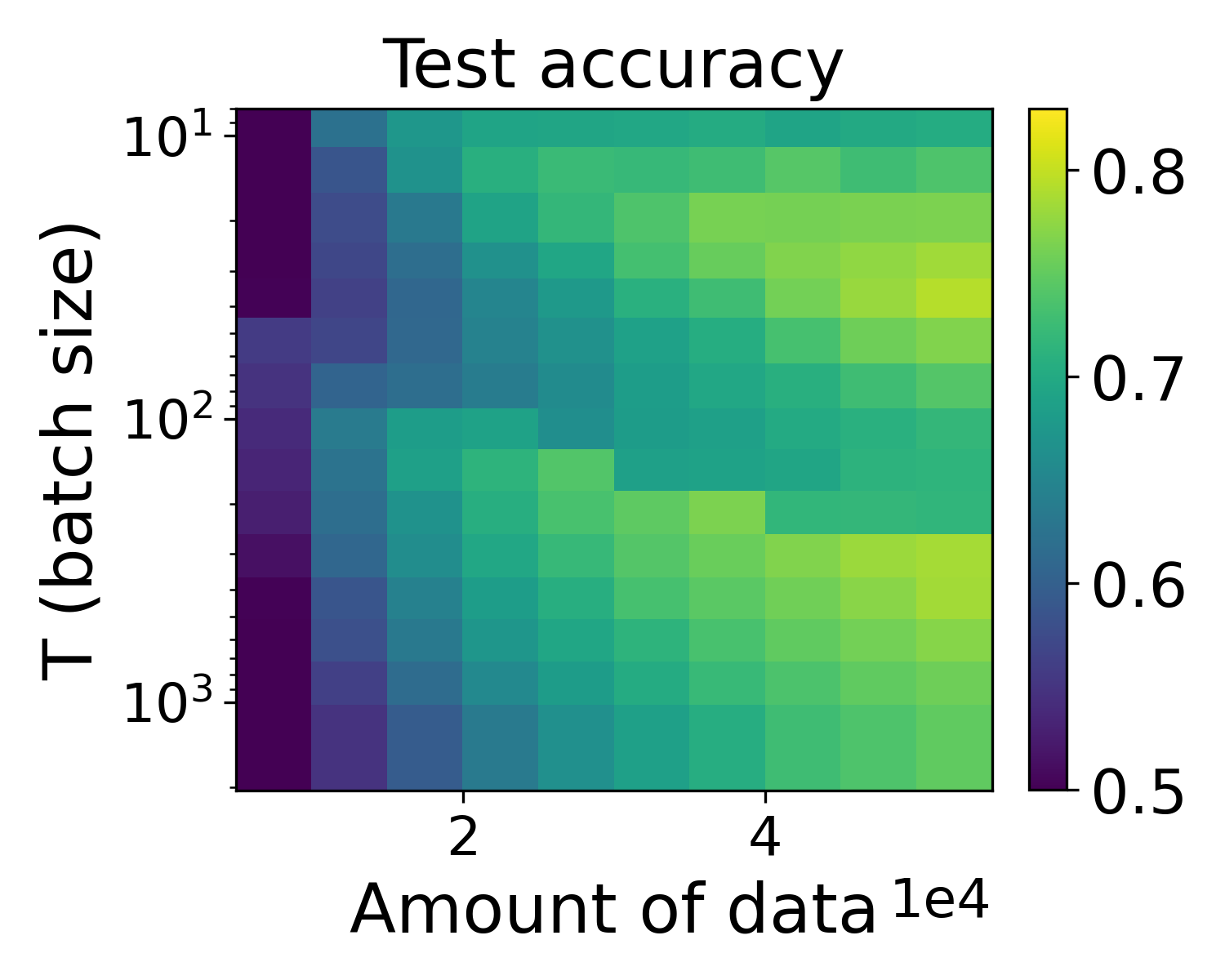}
    \\
    \begin{tabular}{c}\hspace{0mm}Hessian \\\hspace{0mm}trace\end{tabular}
    &
    \includegraphics[width=\metricwidth\textwidth]{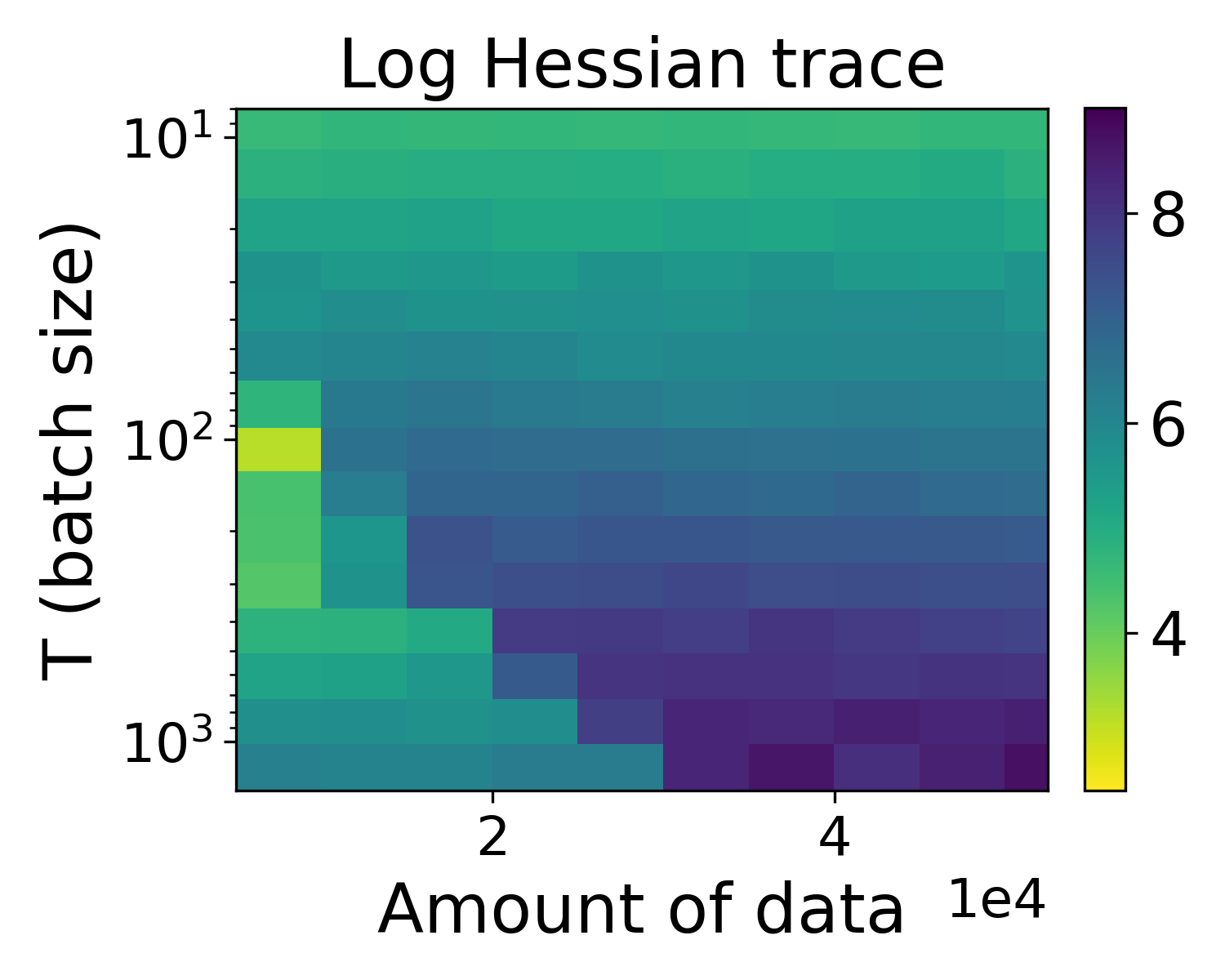}
    &
    \includegraphics[width=\metricwidth\textwidth]{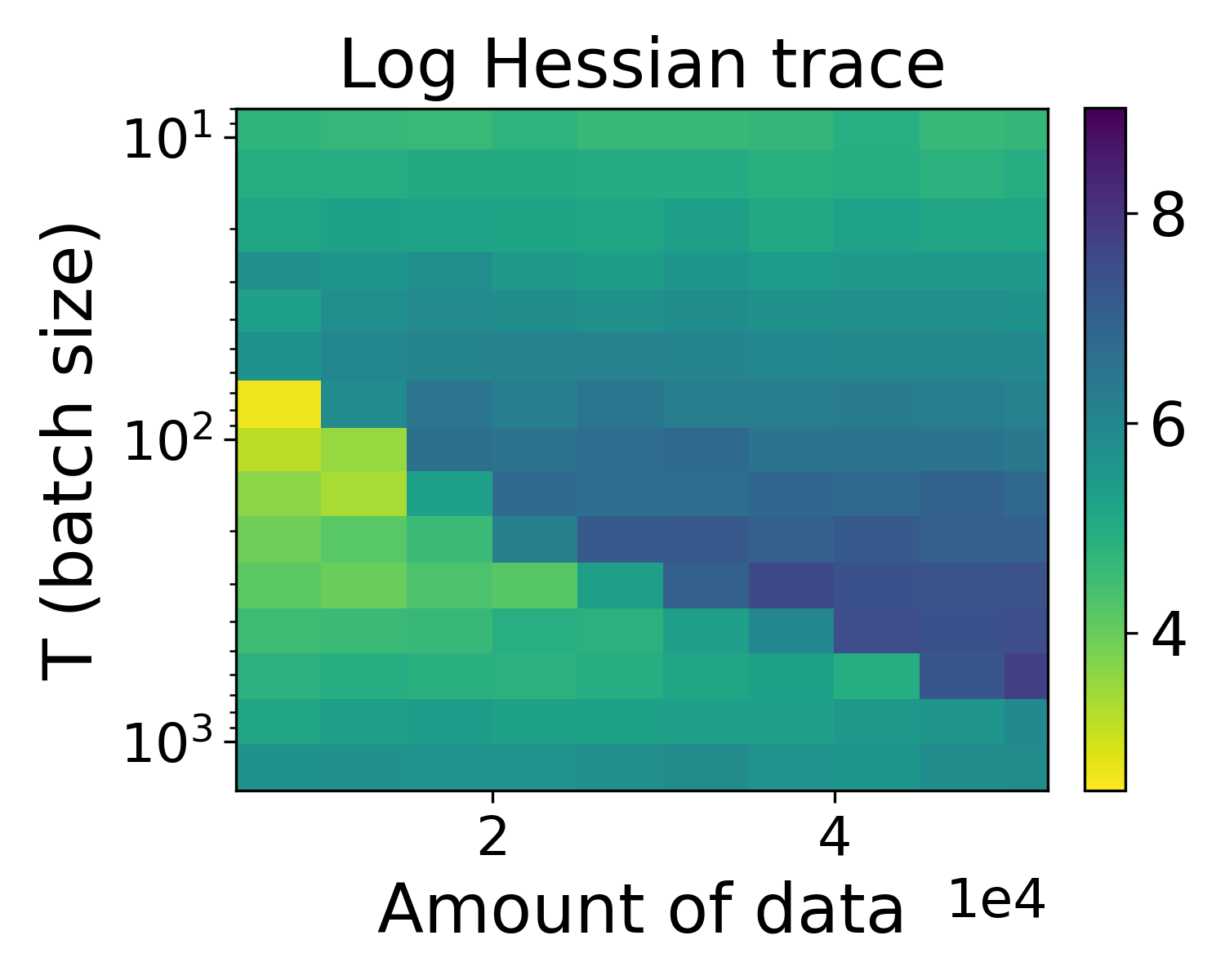}
    &
    \includegraphics[width=\metricwidth\textwidth]{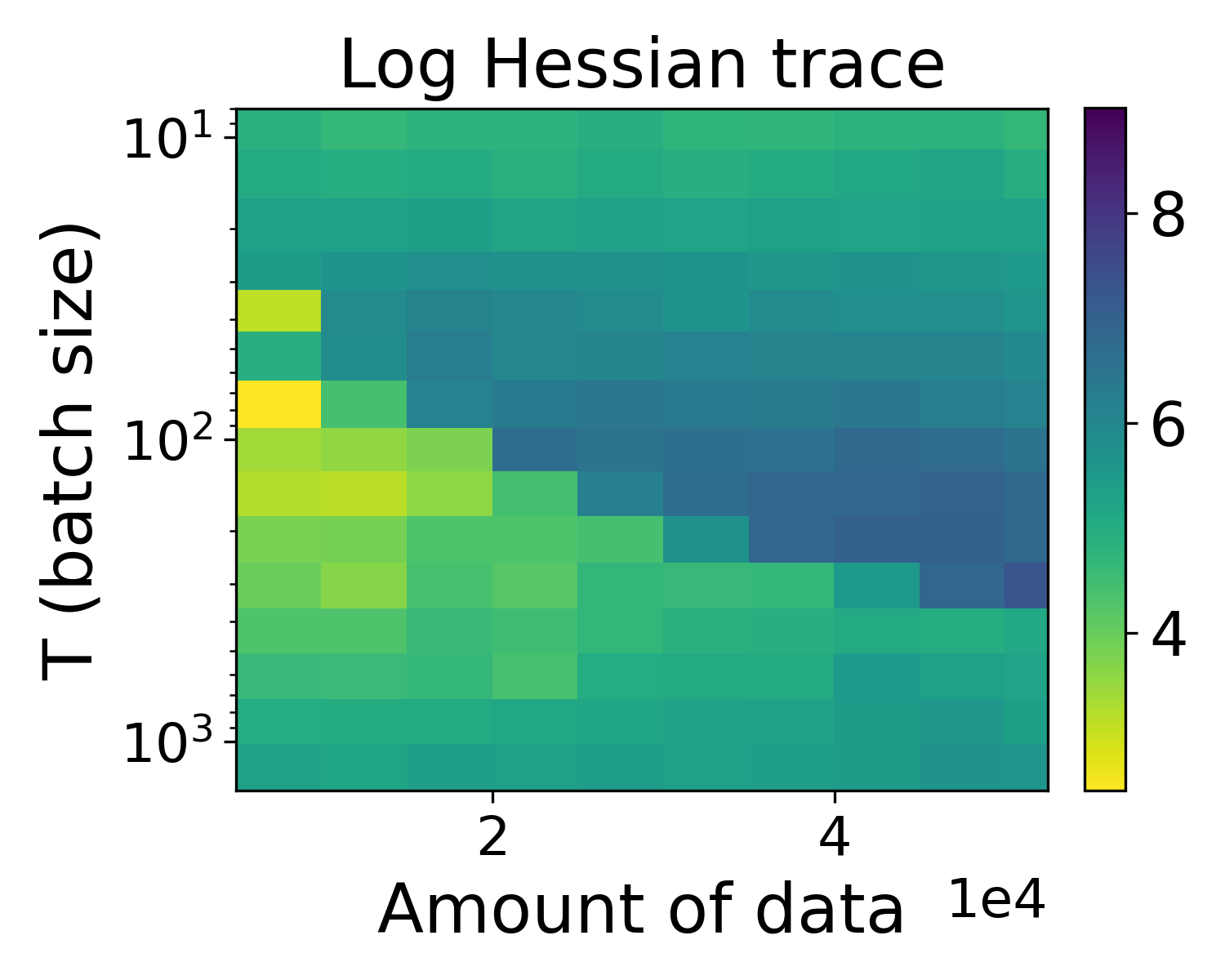}
    &
    \includegraphics[width=\metricwidth\textwidth]{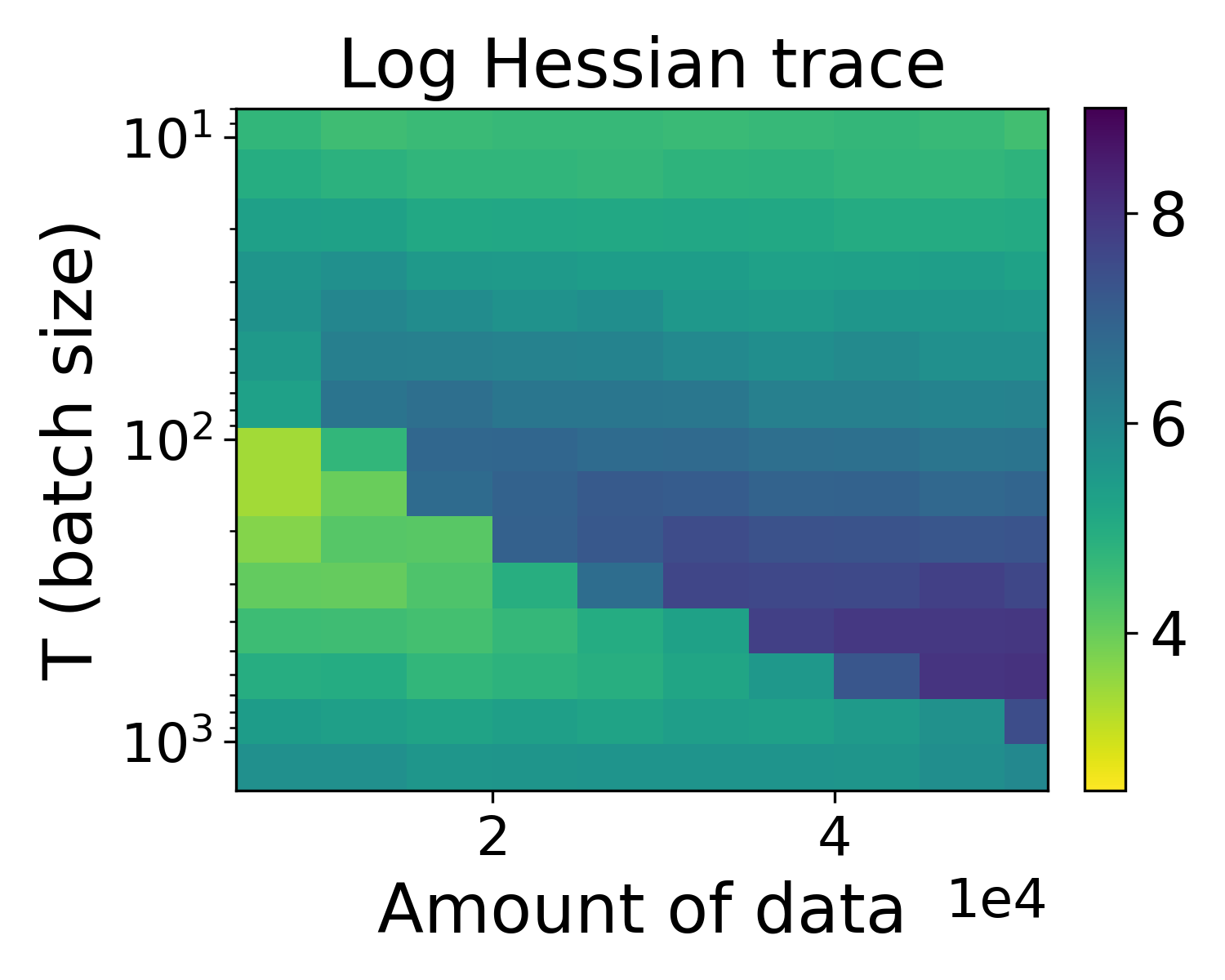}
    &
    \includegraphics[width=\metricwidth\textwidth]{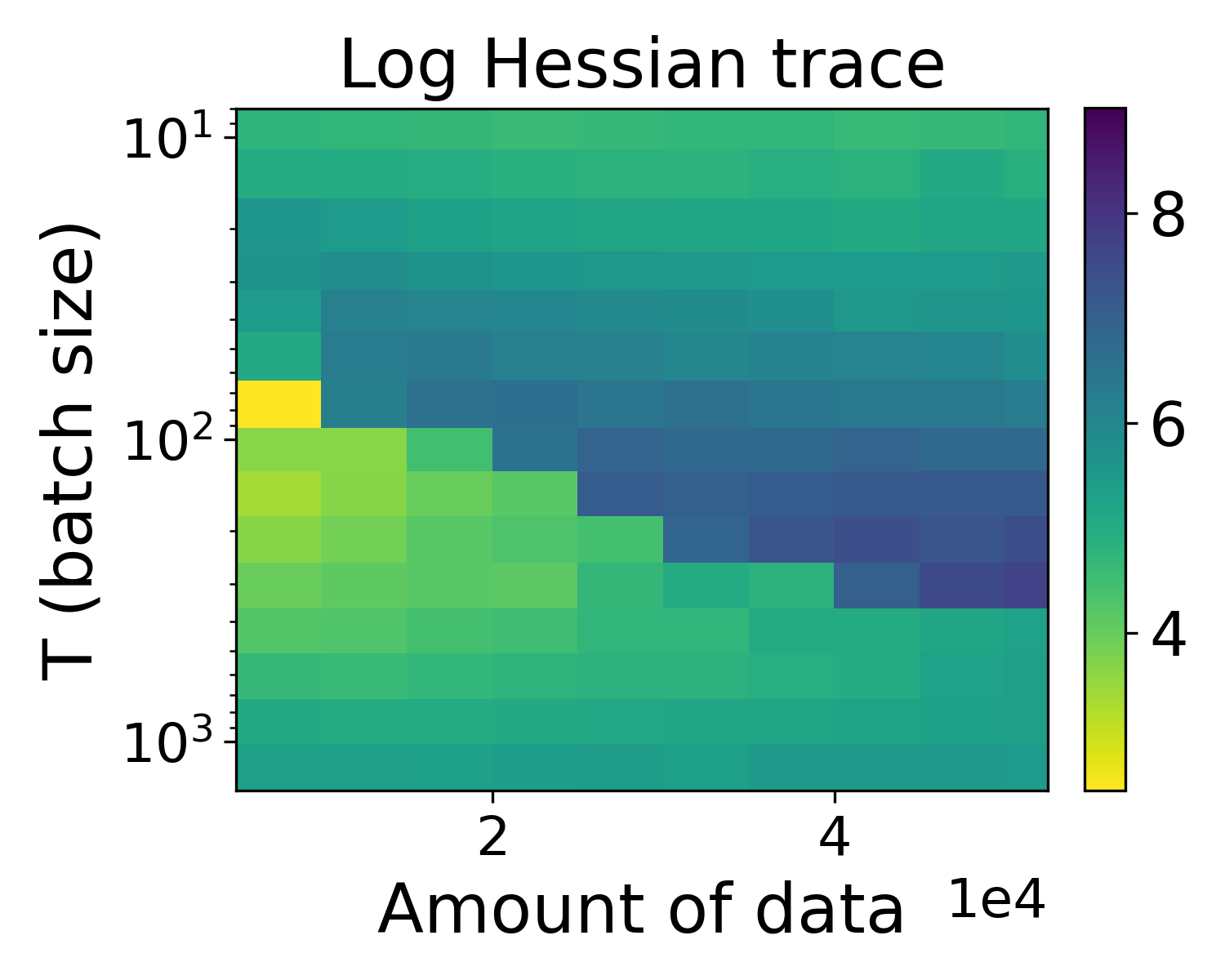}
    &
    \includegraphics[width=\metricwidth\textwidth]{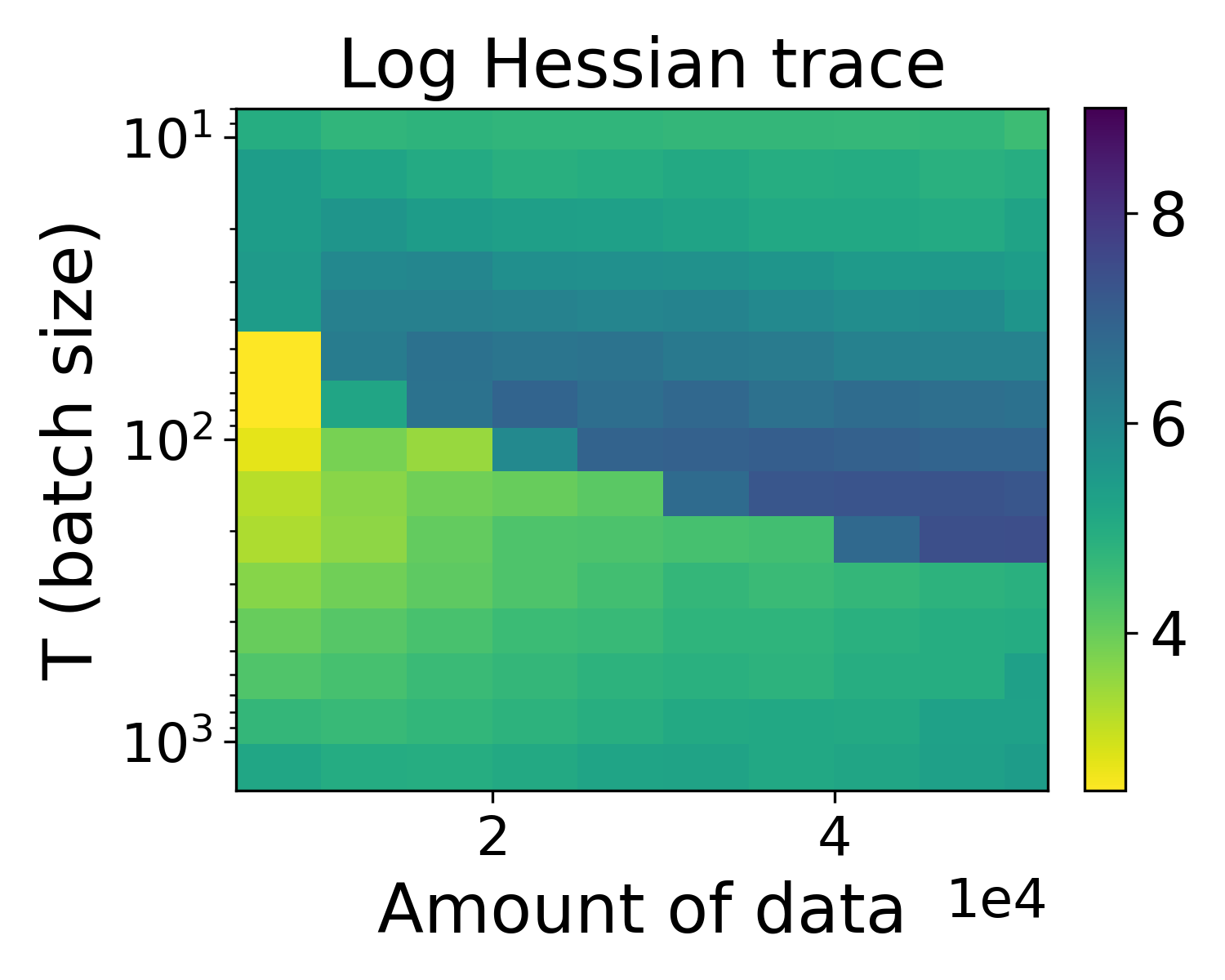}
    \\
    \begin{tabular}{c}\hspace{-4mm}Mode \\\hspace{-4mm}connectivity\end{tabular}
    &
    \includegraphics[width=\metricwidth\textwidth]{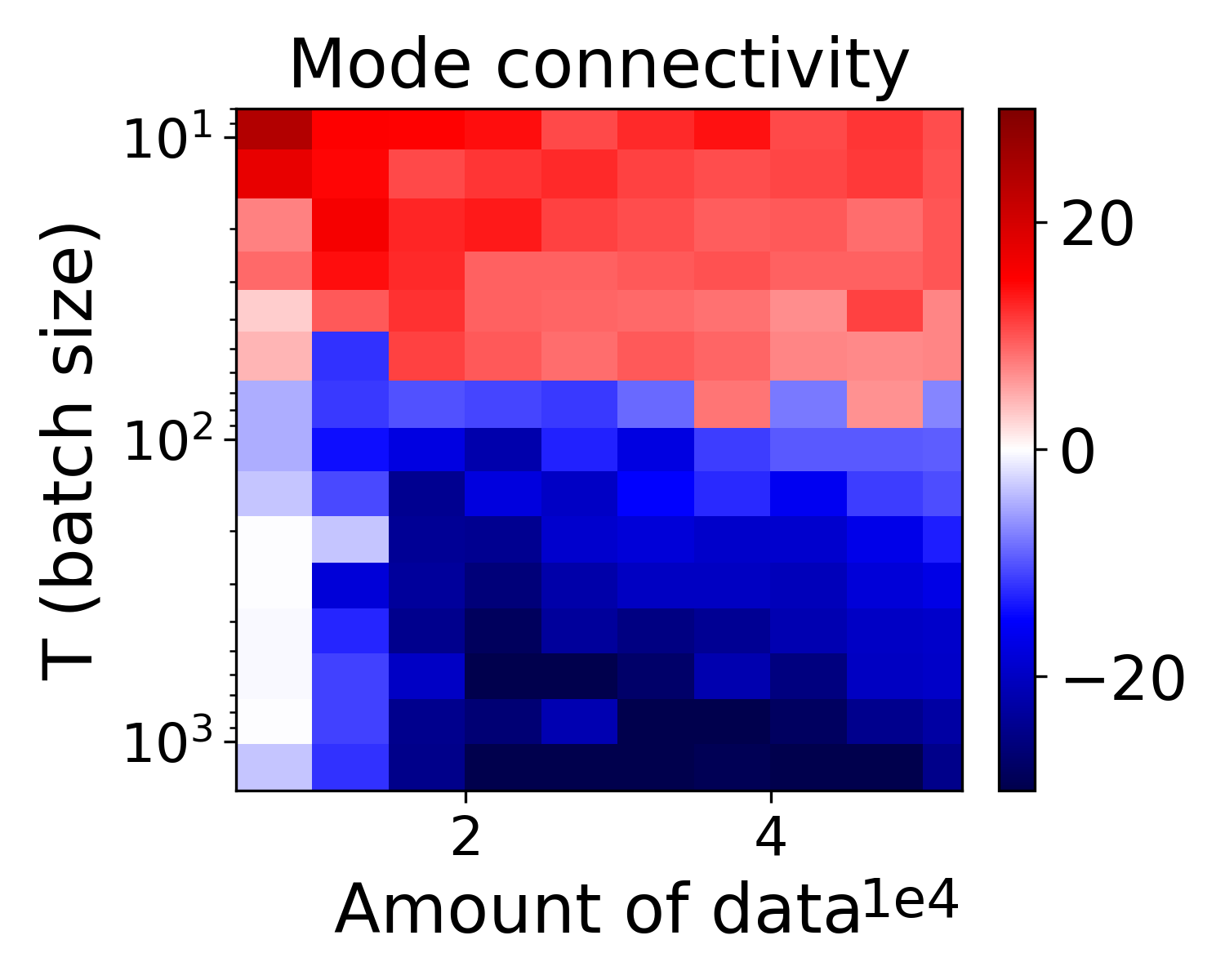}
    &
    \includegraphics[width=\metricwidth\textwidth]{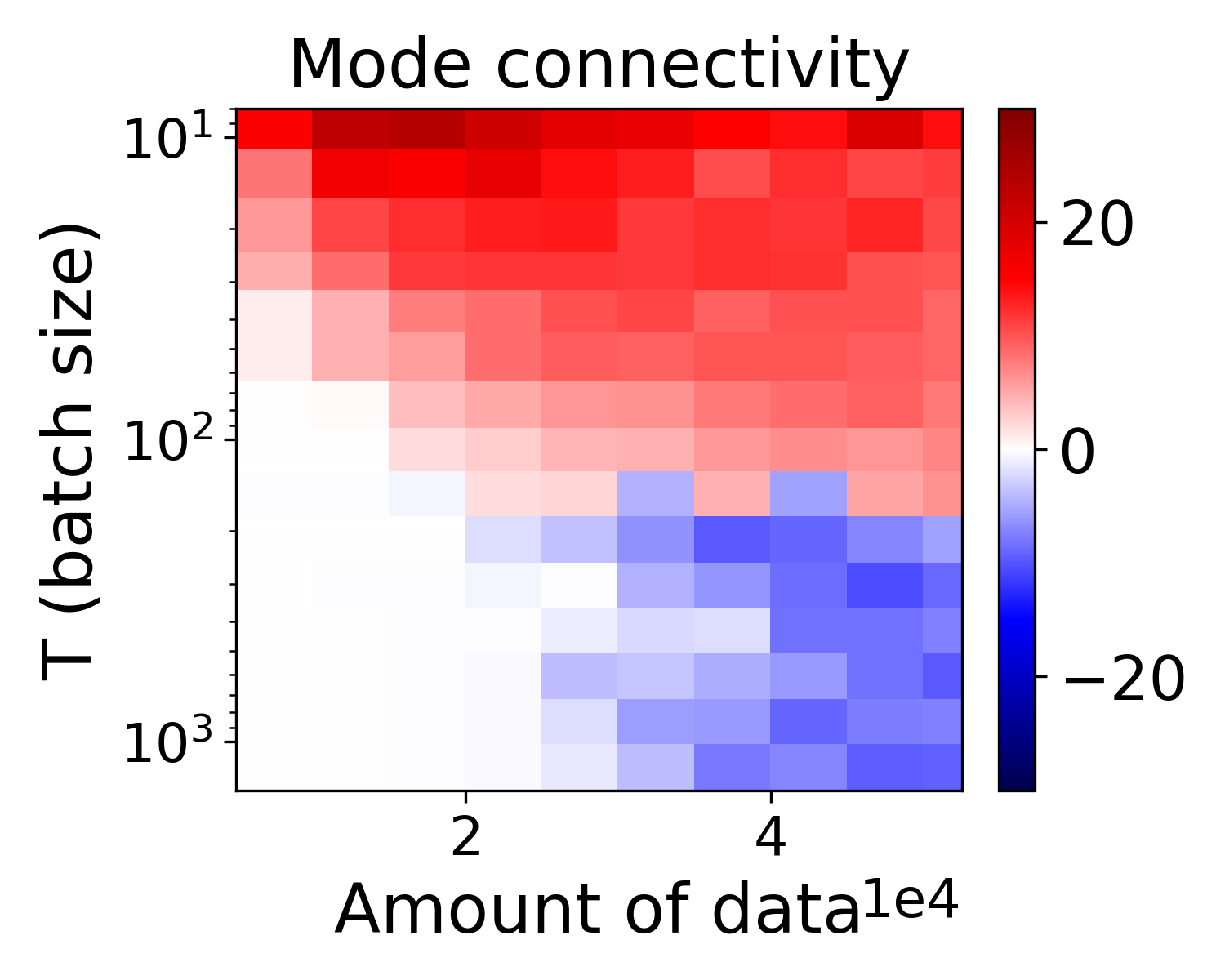}
    &
    \includegraphics[width=\metricwidth\textwidth]{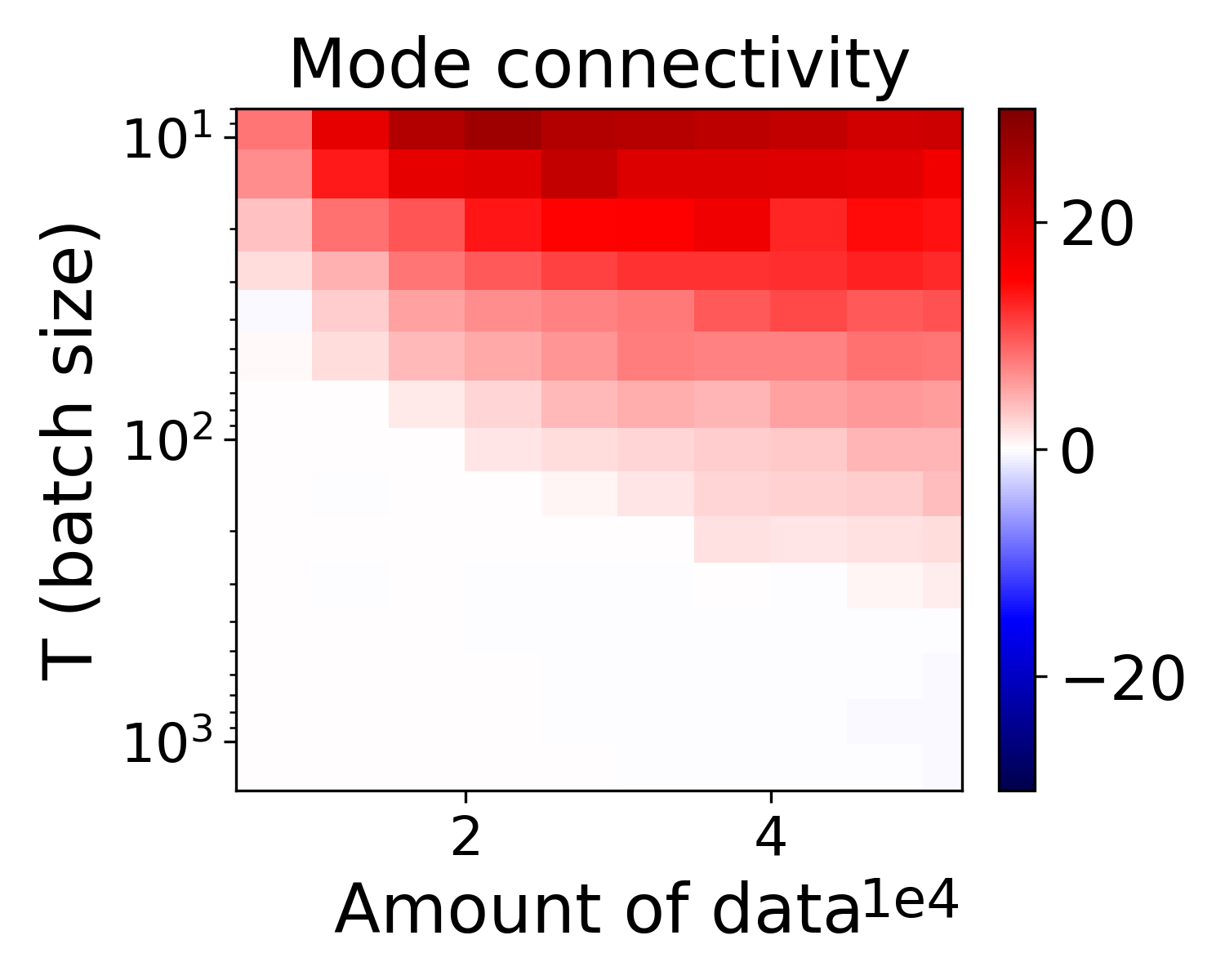}
    &
    \includegraphics[width=\metricwidth\textwidth]{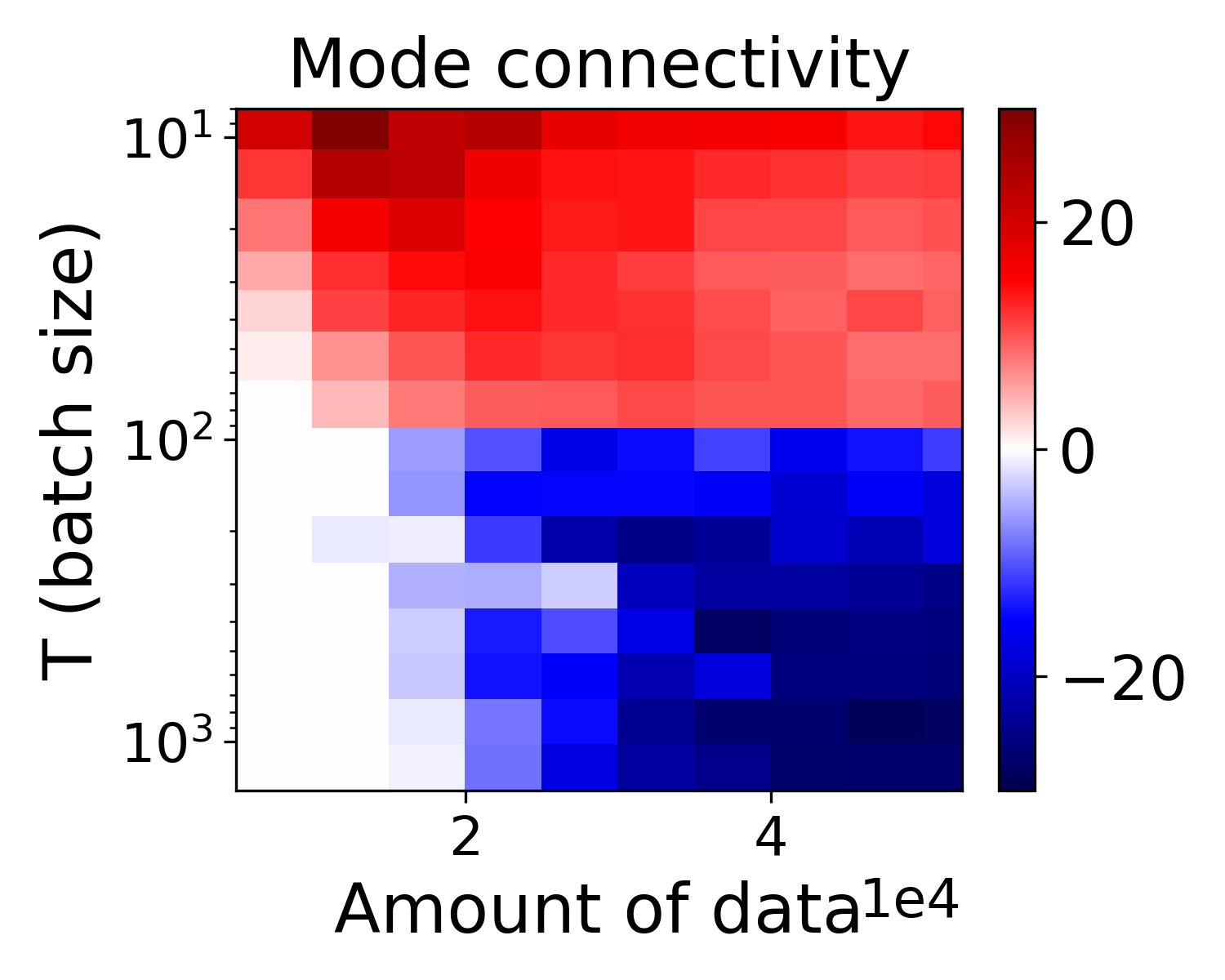}
    &
    \includegraphics[width=\metricwidth\textwidth]{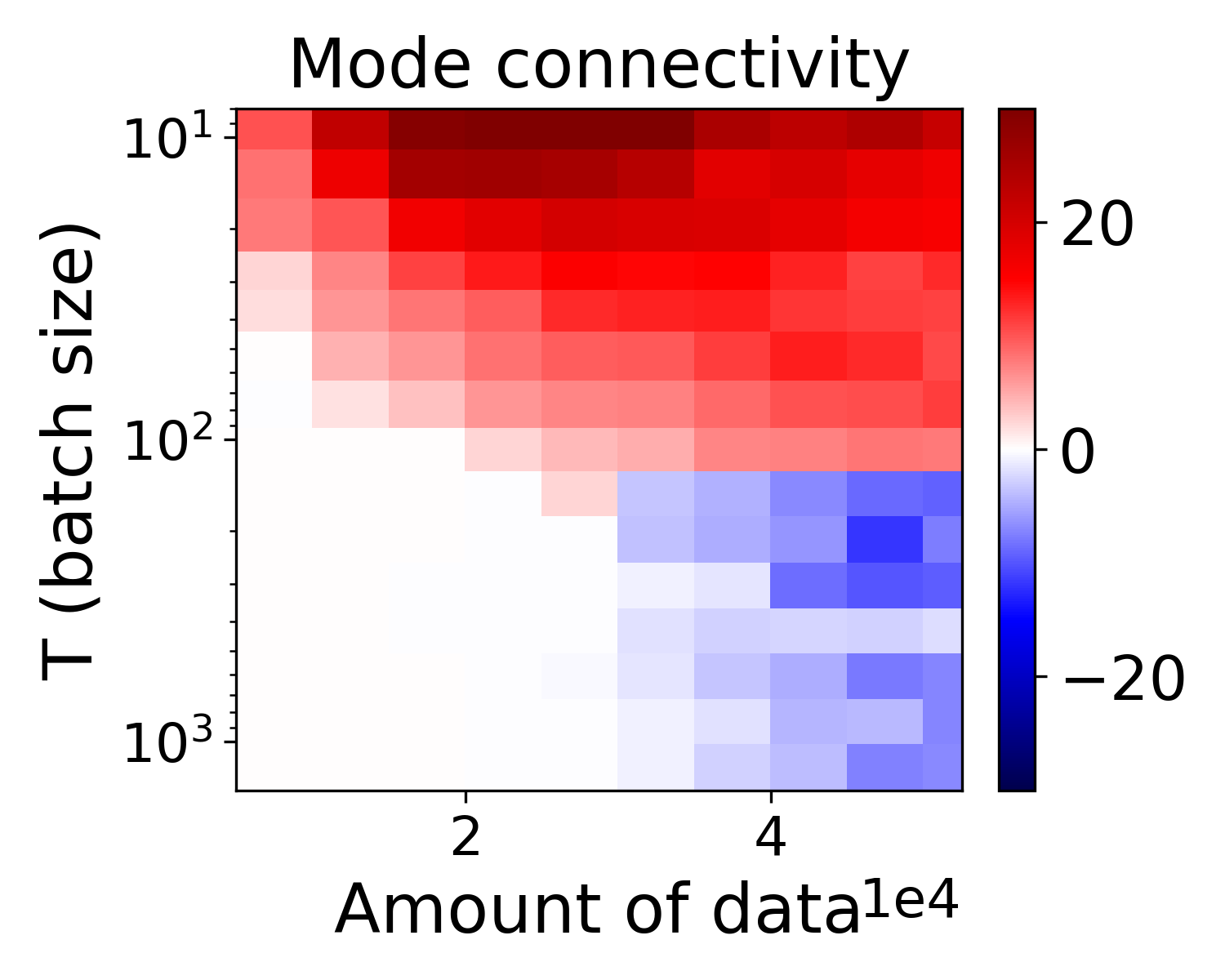}
    &
    \includegraphics[width=\metricwidth\textwidth]{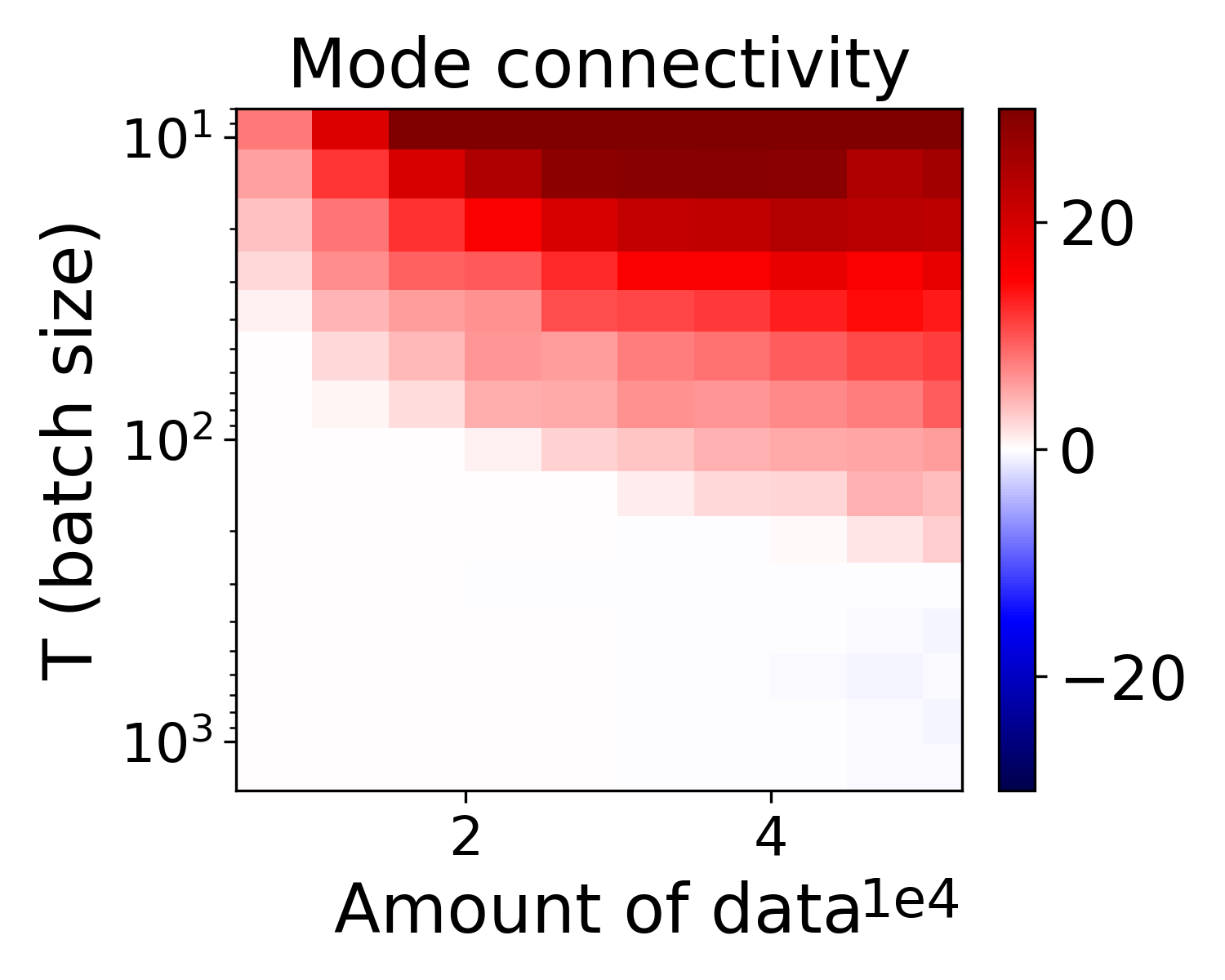}
    \\
    \begin{tabular}{c}\hspace{0mm}CKA \\\hspace{0mm}similarity\end{tabular}
    &
    \includegraphics[width=\metricwidth\textwidth]{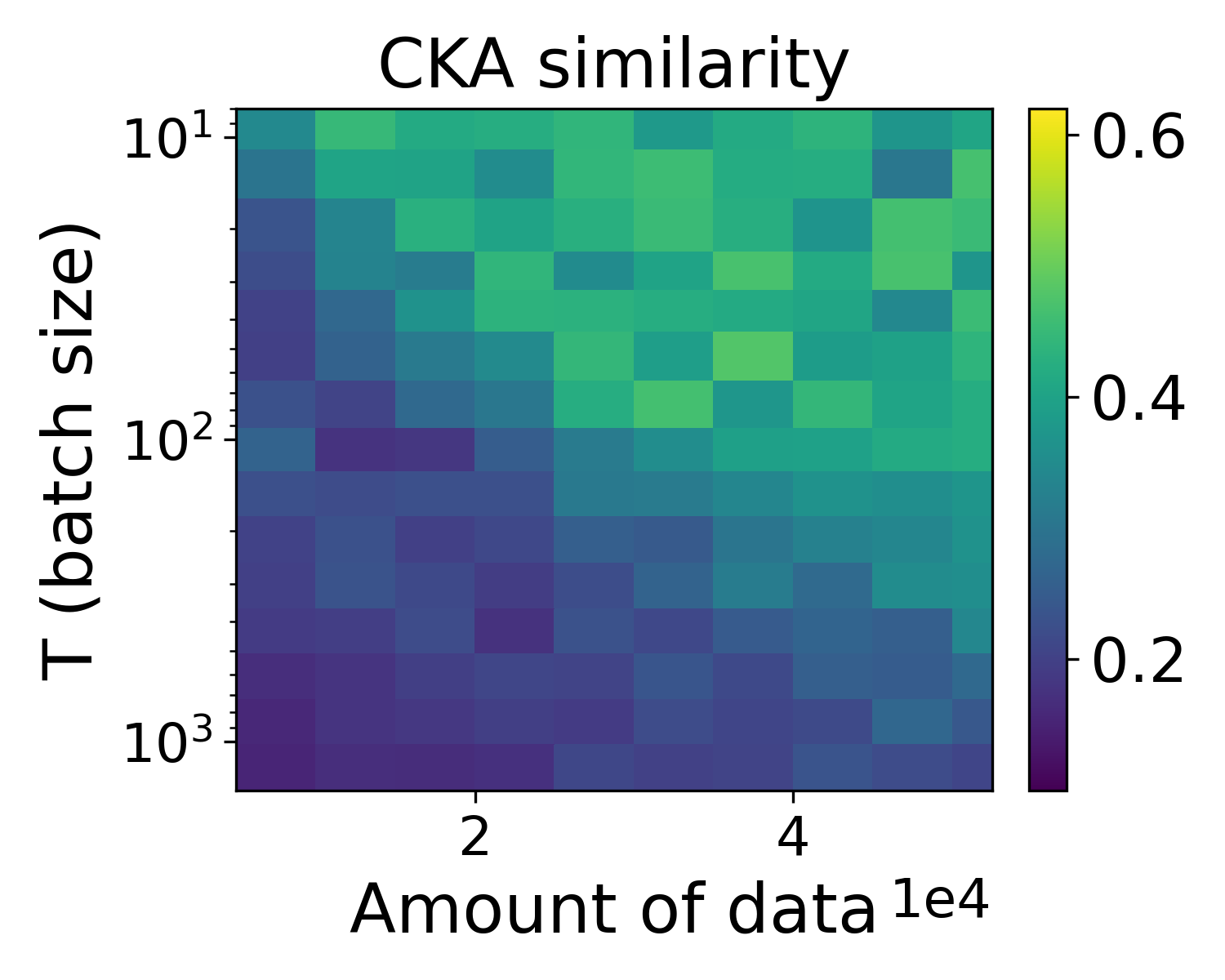}
    &
    \includegraphics[width=\metricwidth\textwidth]{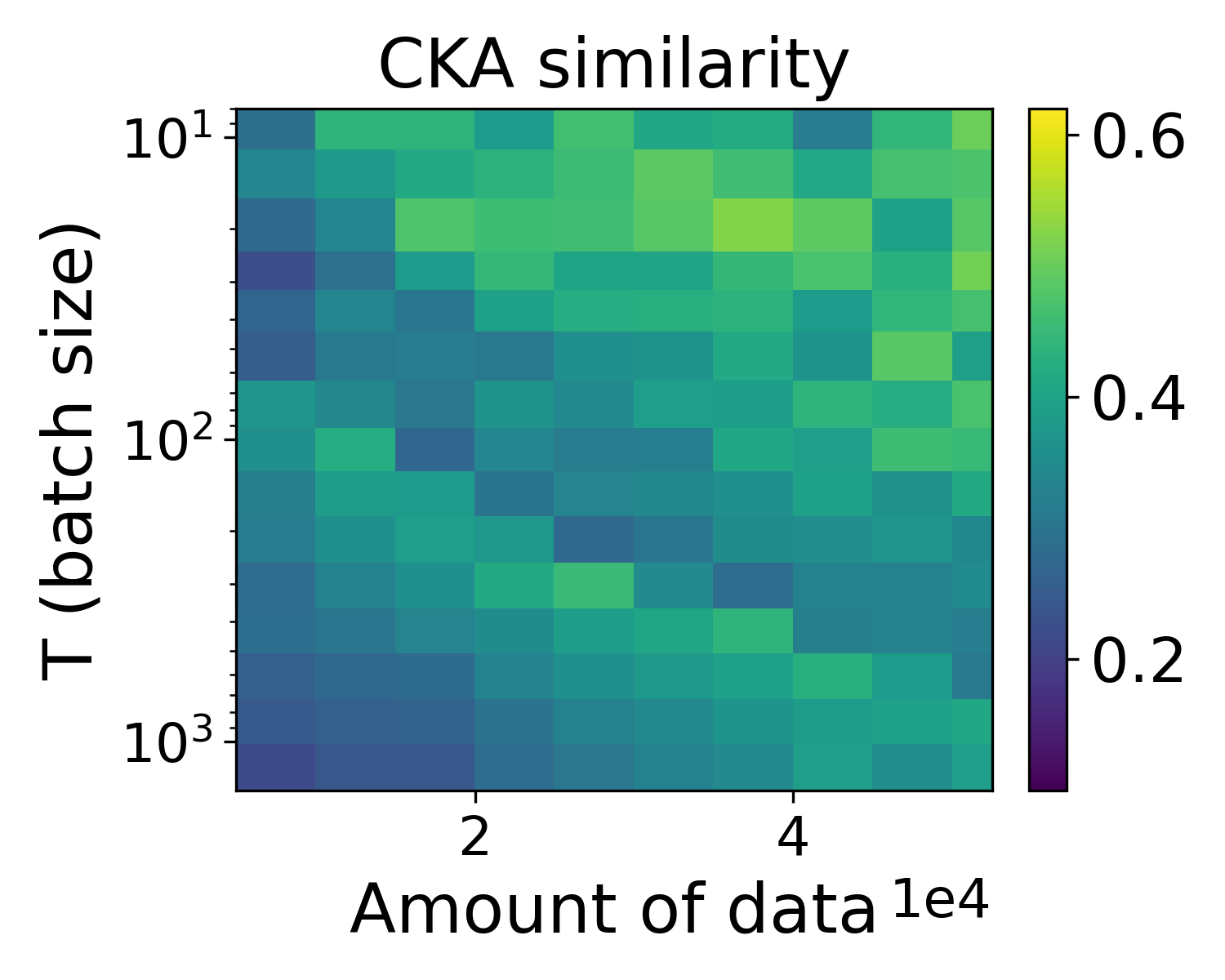}
    &
    \includegraphics[width=\metricwidth\textwidth]{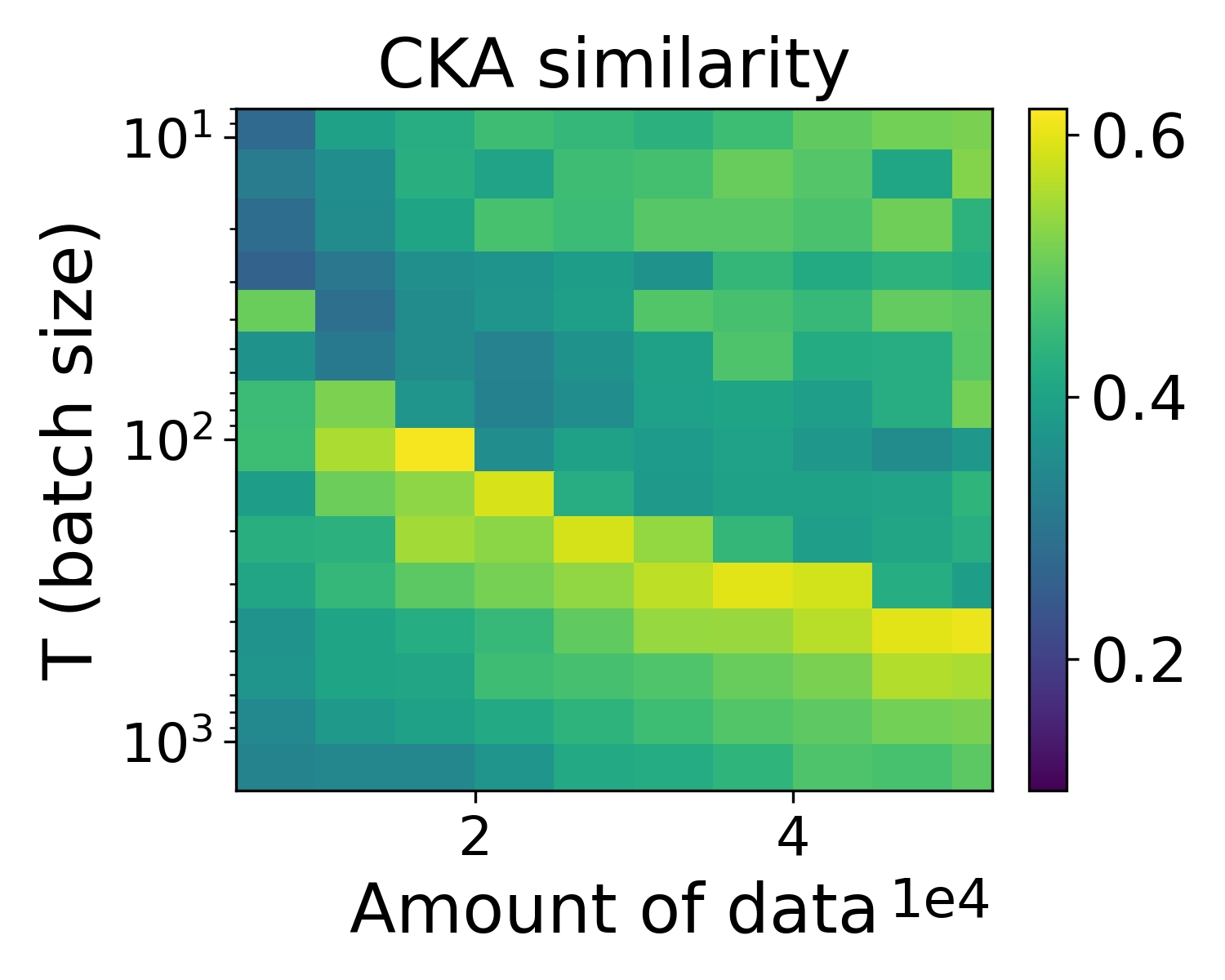}
    &
    \includegraphics[width=\metricwidth\textwidth]{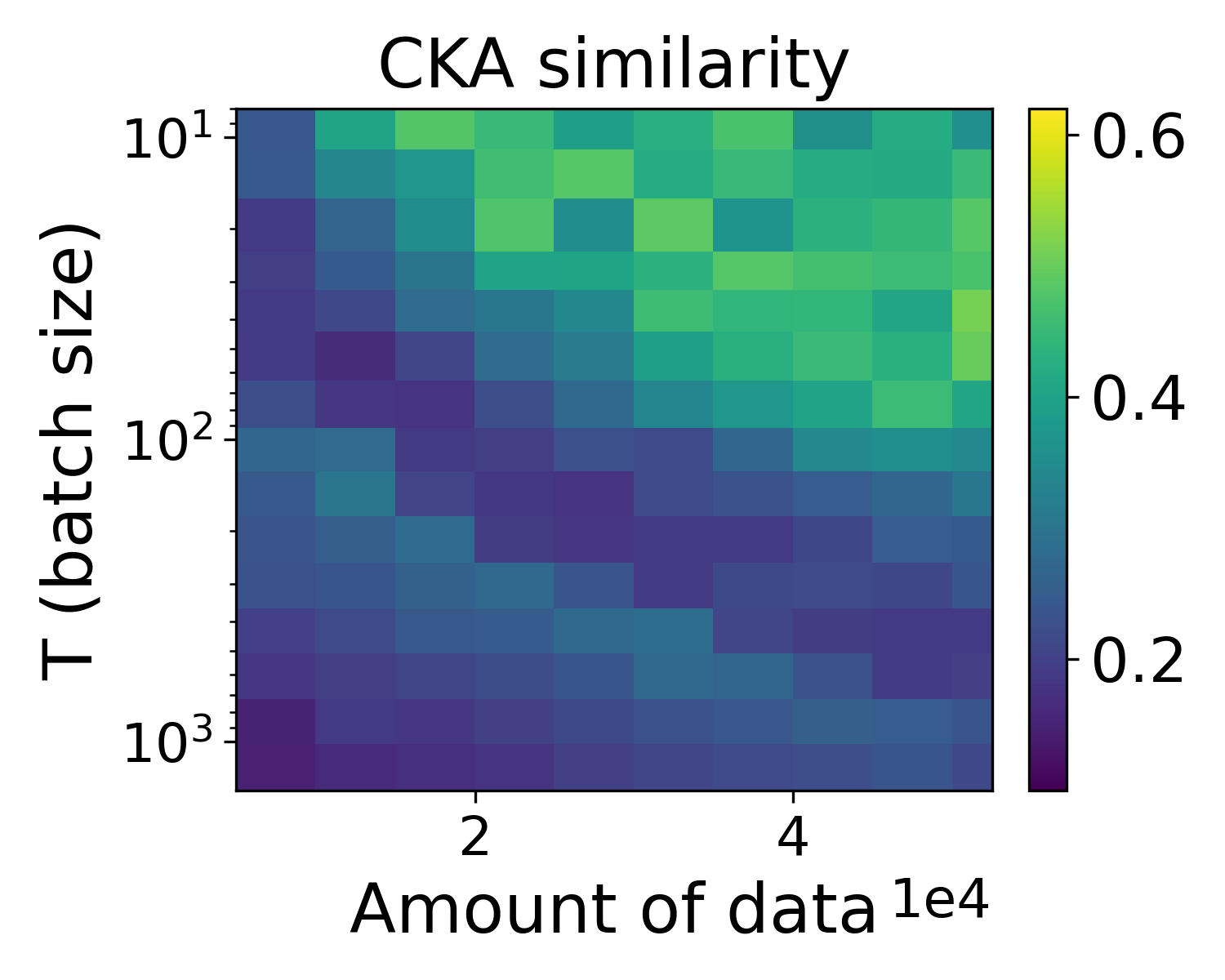}
    &
    \includegraphics[width=\metricwidth\textwidth]{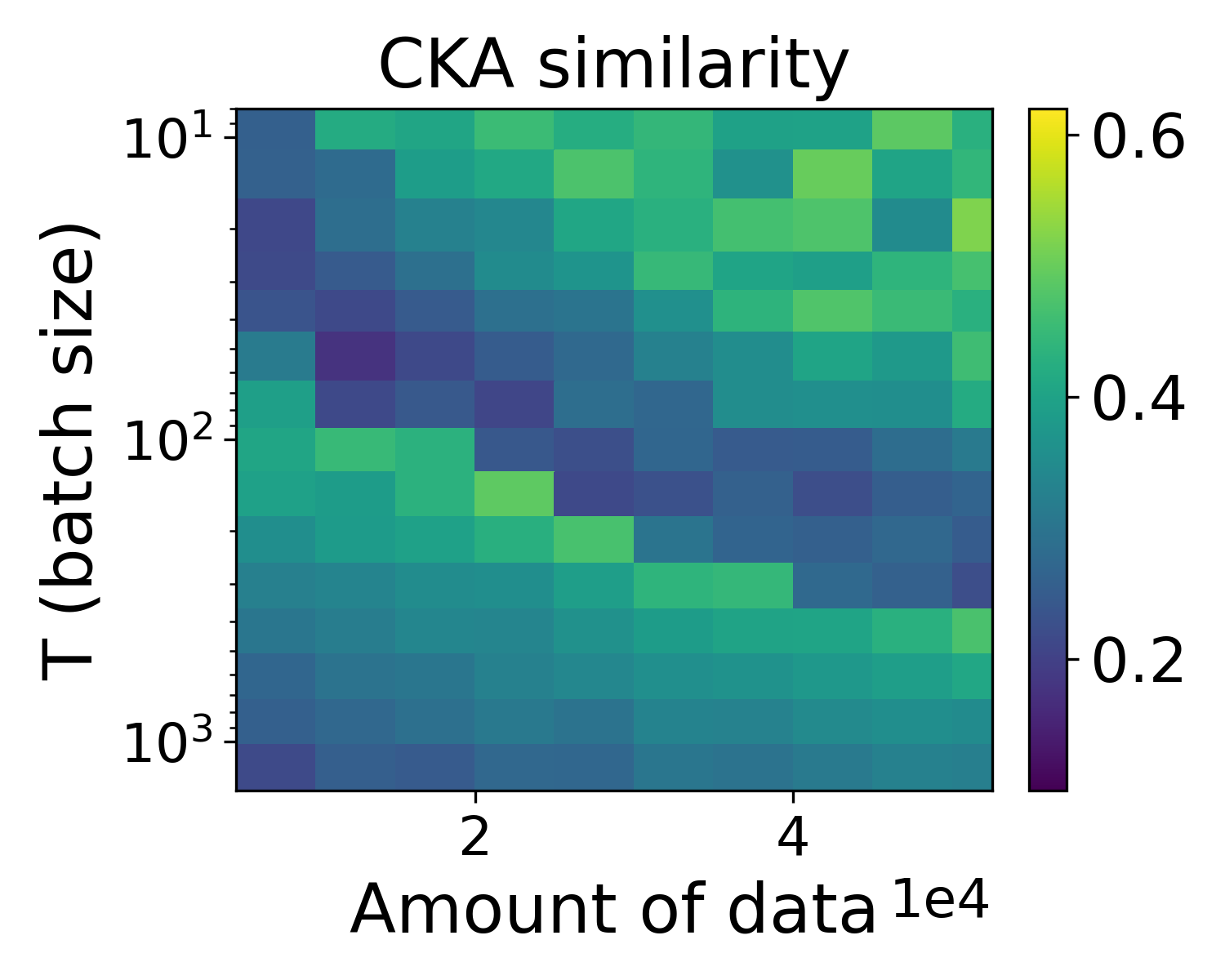}
    &
    \includegraphics[width=\metricwidth\textwidth]{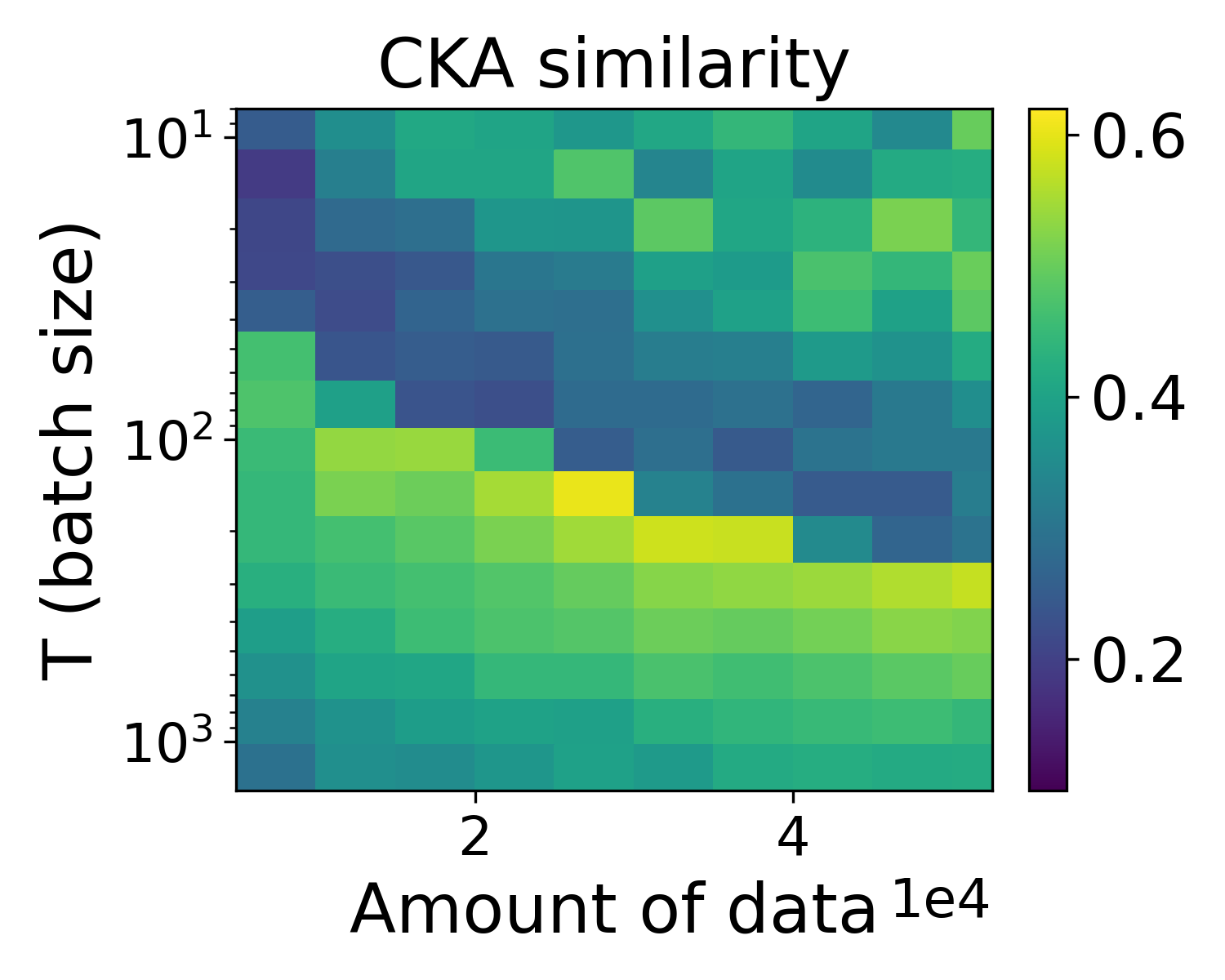}
    \end{tabular}
    \caption{{\bf (Amount of training data as load).} Partitioning the 2D load-like---temperature-like diagram into different phases of learning, using batch size as the temperature and varying the amount of training data to change load. 
    All plots are on the same set of axes.
    {\bf (Left three columns).} Original CIFAR-10 data. Models are ResNet18 with different width. {\bf (Right three columns).} Randomizing 10\% of training labels in CIFAR-10, and still training with ResNet18 of different width.
    }
    \label{fig:different_data_flip}
\end{figure}
\fi

{\bf Different quality of data by changing the amount of noisy labels.}
Next, we vary the proportion of randomized labels to simulate the change in the \emph{quality} of data, as another way to change load.
To generate randomized labels, a percentage of the training data is randomly selected and altered to an incorrect target class. 
Results are shown in Figure \ref{fig:Scaling_noisy_label}.
Once again, local information alone fails to measure the quality of training data. 
We can see that training with a large amount of noise does not significantly affect the Hessian --- see Figure \ref{fig:Scaling_noisy_label_hessian_t}.
In particular, as the temperature decreases (down to the bottom of Figure \ref{fig:Scaling_noisy_label_hessian_t}), the Hessian becomes smaller, independent of the quantity of noisy labels.
This is especially evident in Figure \ref{fig:Scaling_noisy_label_hessian_explicit}, where we plot Hessian trace against batch size.
However, looking instead at mode connectivity in Figure \ref{fig:Scaling_noisy_label_curve} and CKA in Figure \ref{fig:Scaling_noisy_label_CKA}, one can easily deduce that training with more noisy labels leads to more poorly-connected loss landscapes.

\def \figname {Scaling_noisy_label}
\begin{figure}
  \centering
    \begin{subfigure}{0.19\textwidth}
      \includegraphics[width=\textwidth]{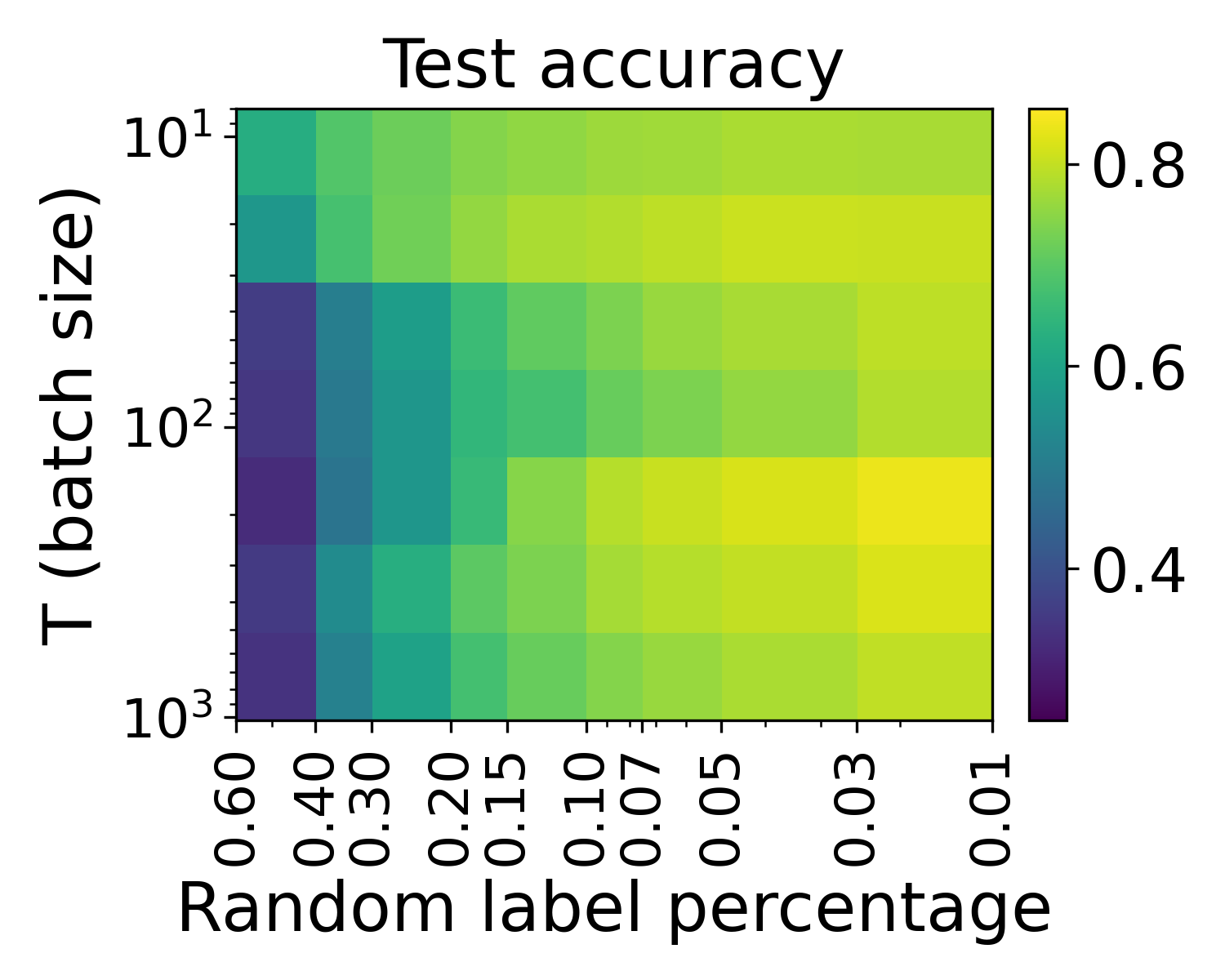}
      \vspace{-6mm}
      \caption{Test accuracy\label{fig:\figname_accuracy}}
    \end{subfigure}
    \begin{subfigure}[c]{0.19\textwidth}
      \includegraphics[width=\textwidth]{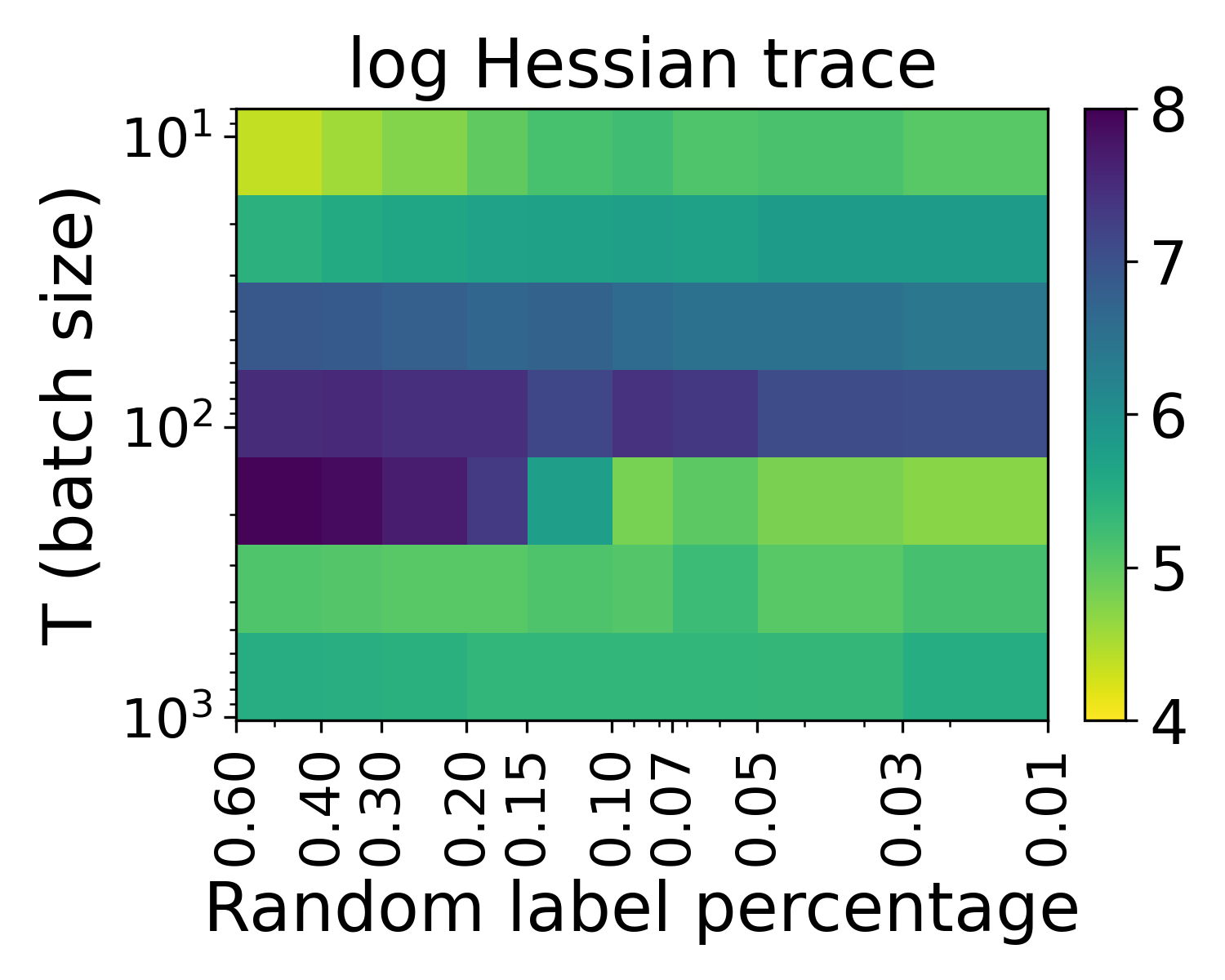}
      \vspace{-6mm}
      \caption{Hessian trace\label{fig:\figname_hessian_t}}
    \end{subfigure}
  \begin{subfigure}[c]{0.19\textwidth}
      \includegraphics[width=\textwidth]{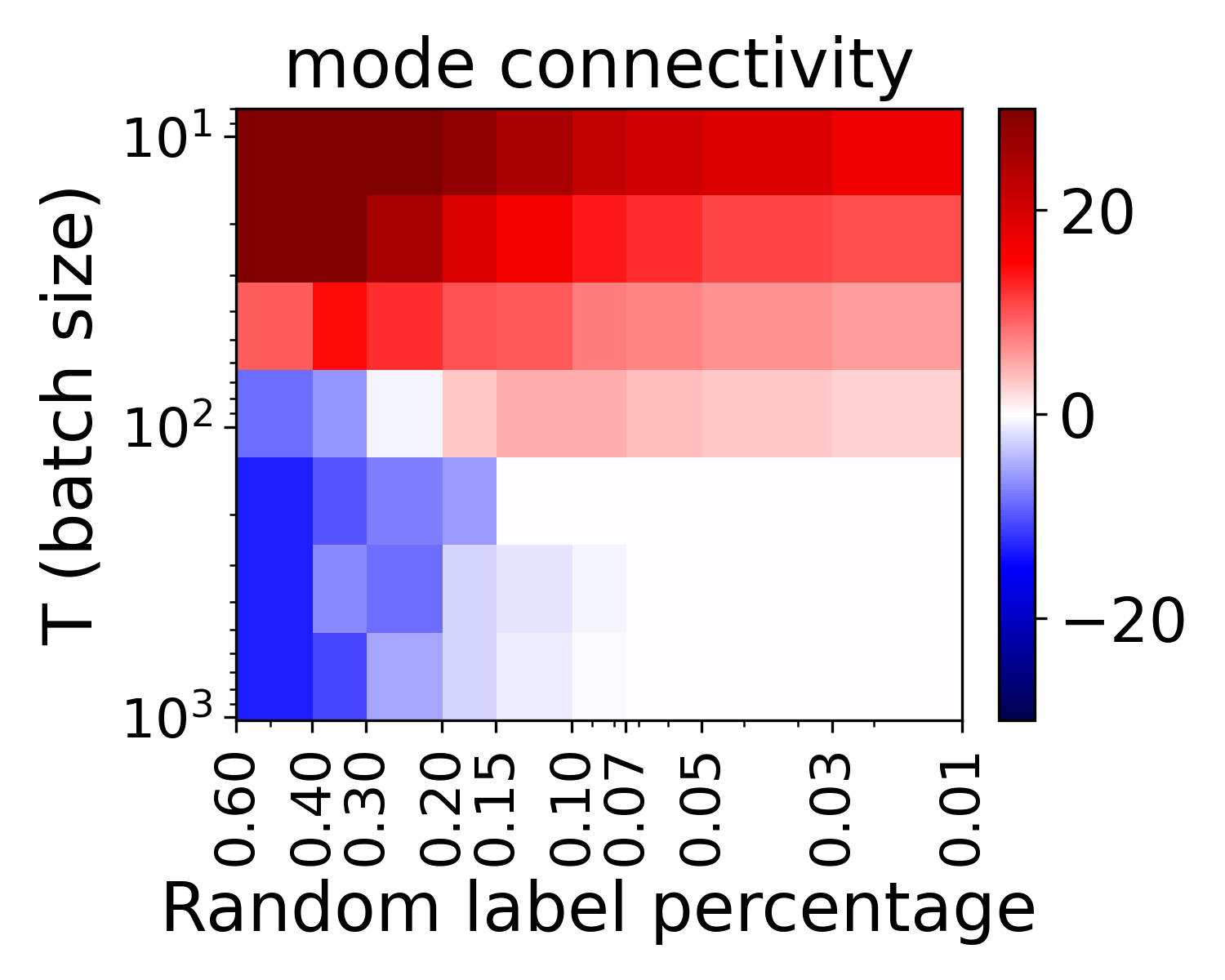}
      \vspace{-6mm}
      \caption{Mode connectivity\label{fig:\figname_curve}}
    \end{subfigure}
    \begin{subfigure}[c]{0.19\textwidth}
      \includegraphics[width=\textwidth]{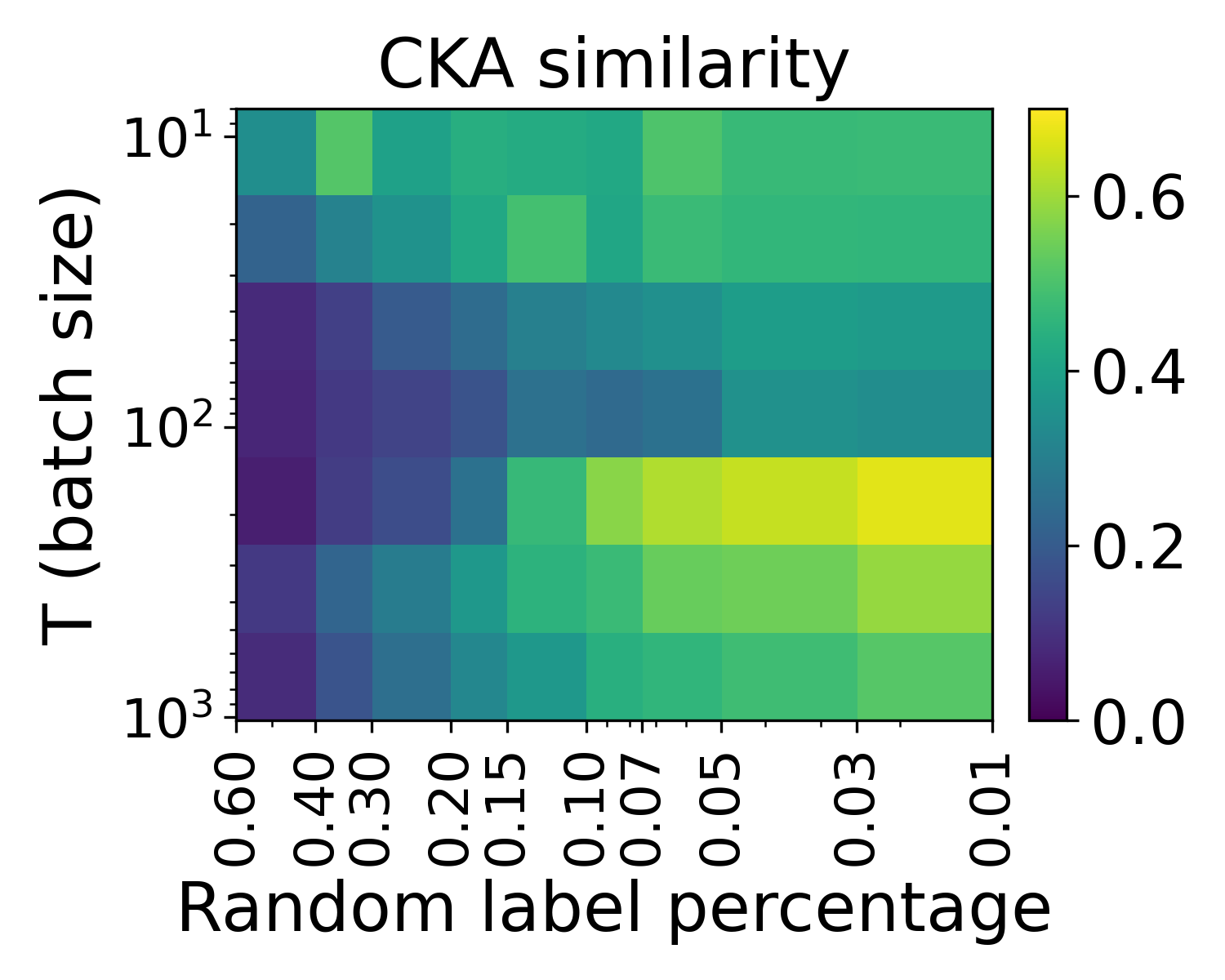}
      \vspace{-6mm}
      \caption{CKA similarity\label{fig:\figname_CKA}}
    \end{subfigure}
    \begin{subfigure}[c]{0.19\textwidth}
      \includegraphics[width=\textwidth]{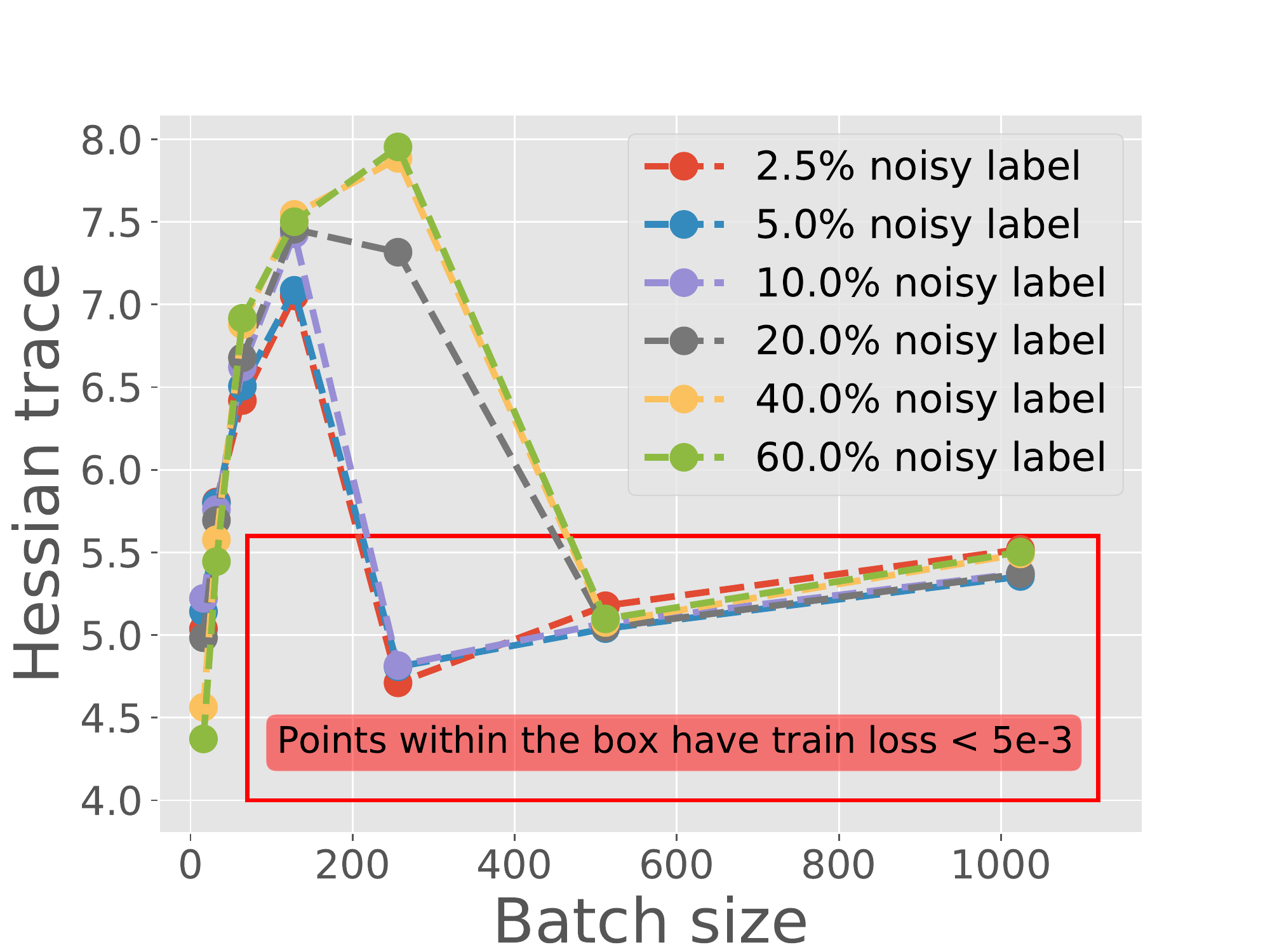}
      \vspace{-6mm}
      \caption{Hessian trace versus batch size\label{fig:\figname_hessian_explicit}}
    \end{subfigure}
  \caption{{\bf (Proportion of randomized labels as load).}
  Partitioning the 2D load-like---temperature-like diagram into different phases of learning, using batch size as the temperature and varying proportion of randomized training labels to change load.
  Models are trained with ResNet18 on CIFAR-10. All plots are on the same set of axes. 
  (e) shows that the Hessian trace changes slowly with the proportion of noisy labels when training loss is small.\vspace{-5mm}
  }
  \label{fig:Scaling_noisy_label}
\end{figure}

{\bf Different datasets, architectures, load/temperature parameters, and training schemes.}
We have performed a wide range of other experiments, only a subset of which we report here.
\ifisarxiv
In Appendix~\ref{sec:additional_results},
\fi
we cover additional datasets, including SVHN, CIFAR-100, and IWSLT 2016 German to English (De-En) (a machine translation dataset), as well as additional NN architectures, including VGG11 and Transformers. 
While there are many subtleties in such a detailed analysis (several of which point to future research directions), all experiments support our main conclusions.
Here, we briefly summarize these results.

\ifisarxiv
In Appendix~\ref{sec:double_descent},
\fi
we study an analogous plot to Figure~\ref{fig:Noisy_label}, training with 10\% noisy labels but replacing the temperature-like parameter from batch size to learning rate. Again, we observe the double descent phenomenon. 
Using this experiment, we infer that the decision to train to zero loss (traditionally a rule-of-thumb in computer vision tasks, although note that recent work has highlighted how the difference between exactly zero versus approximately zero can matter~\citep{MM21a_simpsons_TR}) should depend on the global connectivity of the loss landscape. 
Indeed, for small models with poor connectivity, we find that training to zero loss can harm test accuracy.
This suggests that the common wisdom to fit training data to zero loss is derived from experiments involving relatively high-quality data and models, and is not a principle of learning more~generally.

\ifisarxiv
In Appendix~\ref{sec:NLP},
\fi
we show that in the setting of machine translation, the loss landscape remains poorly-connected (i.e., the mode connectivity remains negative) even for a reasonably large embedding dimension up to 512
\ifisarxiv
(see Figure~\ref{fig:NMT_constant_lr}).
\fi
In this case, generalization can be quite poor when training to zero loss. 
This conclusion matches (with hindsight) the observations in practice, e.g., dropout and early stopping can improve test loss \citep{barone2017regularization,gal2016theoretically}.
It also suggests that an embedding size of dimension 512 (for six-layer Transformers with eight attention heads used in our experiments) is still not large enough for baseline machine translation, and that certain (different) training schemes should be designed to improve the optimization on these loss~landscapes.

\ifisarxiv
In Appendix~\ref{sec:temperature_LR},
\fi
we study learning rate as an alternative temperature parameter, which produces analogous results to Figure \ref{fig:ResNet18}.
\ifisarxiv
In Appendix~\ref{sec:large_batch_size},
\fi
we study large-batch training and show that it increases local sharpness.
Note that for most experiments, we intentionally keep a constant learning rate when varying the batch size to study the change in the landscape with a changing temperature; thus, in
\ifisarxiv
Appendix~\ref{sec:LR_scaling},
\fi
we provide additional results on tuning learning rate with changing batch size, including the commonly used ``linear scaling rule''~\citep{goyal2017accurate}.

\section{Related work}\label{sec:related_work}

In this section, we review prior work related to the optimization of loss landscapes, including both local and global approaches, as well as the double descent phenomenon.

\textbf{Loss landscape.}
Loss landscapes and the connections to SGD training have been important topics in ML research for years. 
Many of these ideas have roots in statistical mechanics and chemical physics~\citep{stillinger_book,wales_book,ballard2017energy,brooks2001taking}.
More recently, within ML, 
\citep{choromanska2015loss} shows that for large-size NNs, the local minima of the loss function often stay within a band close to the global minima; 
\citep{xing2018walk} observes that the loss interpolating between trained weights before and after each training iteration is convex, thus deducing that SGD moves in valley-like regions on the loss landscape; and 
\citep{kleinberg2018alternative} observes that SGD training can be viewed as a way to smooth the loss function, providing a way to rigorously analyze the effect of stochastic noise.
More recently, \citep{martin2018implicit_JMLRversion,martin2019traditional,martin2020heavy,martin2020predicting_NatComm,MM21a_simpsons_TR} use ideas from heavy-tailed random matrix theory to measure the energy landscape through the empirical spectral densities of learned weight matrices, without even using the training data.
They show that the heavy-tail distributions in these spectral densities can predict generalization.
Subsequently, 
\citep{simsekli2019tail} questions the common Brownian motion-based analyses and shows that a heavy-tailed random variable can better capture the ``jump'' phenomena in SGD exploration; and
\citep{hodgkinson2020multiplicative} shows that the multiplicative noise that commonly arises in SGD can improve the exploration of SGD dynamics on the non-convex loss landscape through hopping between basins.
Fort \emph{et al.} \citep{fort2019deep} analyzes deep ensembles from the perspective of loss landscapes and shows that random initializations can explore more diverse modes than functions obtained along an optimization trajectory.
Another paper that is closely related to ours is \citep{fort2020deep}, which also studies both local and global geometry around trained networks. 
It uses a method of parent child spawning to study training dynamics, and it shows that the transition from chaotic to stabilized training happens during the first few epochs.
For more discussions on the global properties of loss landscapes, see \citep{sun2020global} for a survey of recent theoretical and empirical results.
There are also a large number of papers that theoretically analyze the convergence properties of gradient-descent-based methods \citep{dauphin2014identifying,ge2015escaping,du2017gradient,jin2017escape,safran2018spurious,soltanolkotabi2018theoretical,du2019gradient}. 
Another line of related papers considers the properties of NNs measured in the input space, such as the sensitivity measure characterized by the input-output Jacobian norm \citep{novak2018sensitivity}, and the similarity of gradients among different samples, which is called generalized signal-to-noise ratio~\citep{liu2020understanding}.

\textbf{Sharpness and Hessian spectrum.}
Sharpness-based analysis is an important building block of current research on loss landscapes.
\citep{keskar2016large} proposes a sharpness-based metric and shows that large-batch training can bias trained NNs to sharp local minima, thus making generalization worse.
Several papers \citep{mcallester1999pac,neyshabur2017pac,dziugaite2017computing} use a PAC-Bayesian approach to bound generalization, which can also derive sharpness-based bounds \citep{neyshabur2017pac}.
Although researchers have found counter arguments to the belief that flat minima generalize better \citep{dinh2017sharp,granziol2020flatness}, it has been shown recently \citep{jiang2019fantastic} that sharpness-based metrics perform well relative to other complexity metrics that aim to predict generalization; see also \citep{MM21a_simpsons_TR} for a discussion of the connection between sharpness and weight analysis.
One way to measure sharpness is to look at the Hessian spectrum using (randomized) numerical linear algebra approaches.
\citep{yao2018hessian} measures the Hessian spectrum and shows that large-batch training leads to larger Hessian eigenvalues. 
The paper also shows that robust training can allow convergence to flat regions.
\citep{geiger2019jamming} shows that there is a connection between overparameterization of deep NNs and the ``jamming transition'' of repulsive ellipses.
Moreover, when the training loss function has exactly zero value, the Hessian spectrum of both systems can have a sharp phase transition.
Several papers \citep{sagun2016eigenvalues,gur2018gradient,yao2018hessian,papyan2020traces,fort2019emergent} show that the Hessian spectrum is sparse and contains outliers that are dictated by a class and ``cross-class'' structure.
This is connected to the perhaps surprising observation that the training dynamics seems to happen in a low-dimensional structure in the weight space.
Beyond sharp and flat minima, \citep{he2019asymmetric} proposes the concept of ``asymmetric valley'' which contains asymmetric directions on the loss landscape, along which the loss can change sharply on one side and flatly on the other, and the paper shows that solutions biased towards the flat side generalize better.
Sharpness-based and spectrum-based analysis also leads to many practical ways to improve the training and use of NNs, such as in \citep{chaudhari2019entropy,izmailov2018averaging,foret2020sharpness,dong2019hawq,shen2020q}.

\textbf{Connectivity of loss landscape.} 
\citep{goodfellow2014qualitatively} shows that, unlike the common belief regarding the difficulties of non-convex optimization, the linear path connecting the initialization and the minima found by SGD often shows smooth and monotonic loss changes.
\citep{draxler2018essentially} and \citep{garipov2018loss} propose the concept of \emph{mode connectivity} to explicitly construct nonlinear curves connecting trained solutions in the weight space, on which the loss remains small.
This finding is essential in that it might suggest the whole concept of ``local minima'' is flawed because there might (effectively) only exists one large complex connected minima (akin to the rugged convexity we discuss).
This is consistent with our results suggesting a global rugged convexity for a well-trained NN loss landscape.
Another practical motivation for studying mode connectivity is to find better optima on the curve or through some ensemble technique.
On the theory side, \citep{cooper2018loss} proves that the locus of global minima of an overparameterized NN is a ``connected submanifold''.
Another paper \citep{freeman2016topology} studies a more general property on the connectivity of ``sublevel sets'' for deep linear NNs and one-hidden-layer ReLU networks.
Further, \citep{nguyen2019connected} proves that the sublevel sets of deep NNs are connected if one of the hidden layers has more neurons than the number of training samples.
In \citep{kuditipudi2019explaining}, mode connectivity of multilayer ReLU networks is proved by assuming properties such as dropout stability.
The paper also constructs an interesting two-layer network for which overparameterization does not lead to connection between all local minima.
Then, \citep{shevchenko2020landscape} shows that in the mean field regime, dropout stability holds for deep and wide NNs.
Using the dropout stability, it justifies the empirical observation in \citep{draxler2018essentially} that mode connectivity improves with the size of the NN.
\citep{frankle2020linear} shows that linear low-loss paths can be found between two networks if they originate from shared trained initialization.
\citep{fort2019large} extends the one-dimensional curves studied in previous literature to high-dimensional ``tunnels'' between a set of optima, and it shows that many regularization hyperparameters can increase the ``angular width'' of the tunnels that connect different local~minima.

\textbf{Double descent.}
It has recently been observed that the classical U-shape generalization curve can be extended to a double descent curve   \citep{belkin2019reconciling,belkin2020two,nakkiran2019deep,muthukumar2020harmless,mei2019generalization,bartlett2020benign}. 
This curve reconciles the fact that large models which completely overfit to training data can still generalize \citep{zhang2016understanding}.
\citep{chen2020multiple,adlam2020neural,d2020triple} show that even double descent might not be the complete picture, and more complex non-monotonicity in the overparameterized regime may exist.
See also \citep{DKM20_TR} for similar results.
\citep{yang2020rethinking} derives the bias-variance tradeoff and shows that the peak seen in double descent arises from increased variance.
That the fluctuational properties around phase transitions would lead to a double descent phenomenon in values of volume/entropy measures (such as generalization) has long been known \citep{EB01_BOOK,martin2017rethinking}; and 
the most relevant results to our paper are those which explicitly show that double descent can indeed arise from transitions between different learning phases \citep{liao2020random,derezinski2019exact} (which is a natural prediction from the existence of these learning phases \citep{martin2017rethinking}).

\section{Conclusions}

Motivated by recent work in the statistical mechanics of learning, we have performed a detailed empirical analysis of the loss landscape of realistic models, with particular attention to how properties vary as load-like and temperature-like control parameters are varied.
In particular, local properties (such as those based on Hessian) are relatively easy to measure; and while more global properties of a loss landscape are more challenging to measure, we have found success with a combination of similarity metrics and connectivity metrics.
This complements recent work that uses tools from statistical mechanics and heavy-tailed random matrix theory, as we can perform large-scale empirical evaluations using metrics (CKA, mode connectivity, Hessian eigenvalues, etc.) that are more familiar to the ML community.
We interpreted these metrics in terms of connectivity and similarity, and we used them to obtain insight into the local versus global properties of NN loss landscapes.

Here, we summarize a few observations from our connectivity and similarity plots (that we expect will be increasingly relevant as larger data sets and models are considered).
\ifisarxiv
\begin{itemize}[leftmargin=*]
    \item {\bf A larger width improves connectivity:} From Figure~\ref{fig:ResNet18_curve}, \ref{fig:Lr_decay_curve}, \ref{fig:Noisy_label_curve}, \ref{fig:WD_curve}, \ref{fig:SVHN_curve}, \ref{fig:VGG_curve}, and \ref{fig:LR_curve} (see the Appendices), we see that increasing model width improves connectivity.
    \item {\bf More data improves similarity:} From the CKA-similarity row in Figure \ref{fig:different_data}, we see that increasing the quantity of data can improve similarity.
    \item {\bf Better data quality improves connectivity:} From Figure \ref{fig:Scaling_noisy_label_curve}, we see that increasing the quality of data by reducing the amount of randomized labels can improve connectivity.
    \item {\bf A larger width and a higher temperature in Phase IV improves similarity:} From Figure~\ref{fig:ResNet18_CKA}, \ref{fig:Lr_decay_CKA}, \ref{fig:Noisy_label_CKA}, \ref{fig:WD_CKA}, \ref{fig:SVHN_CKA}, \ref{fig:VGG_CKA}, and \ref{fig:LR_CKA}, we see that i) a larger width increases similarity of trained models; and ii) for Phase IV (globally well-connected and locally flat minima), using a relatively large temperature can improve similarity. 
\end{itemize}
\fi
These observations can be restated in a way to guide the operations in training:
\begin{itemize}[leftmargin=*,noitemsep]
    \item A negative mode connectivity suggests that the data quality is low or the model size is small.
    \item A large positive mode connectivity or a large Hessian leading eigenvalue/trace indicates that the training fails to converge to the bottom of a local minimum.
    \item A small CKA similarity suggests that generalization is not good, which can be caused by various factors. However, if, in addition, the mode connectivity is close to zero, and the Hessian is small, a large CKA similarity indicates the ``lack of signal'' from the data. In other words, one should get more high-quality data for training.
\end{itemize}

In future work, we aim to provide a more detailed study on using the metrics for improved training, and we will look at phase diagrams outside of the load/temperature form, especially in the low-connectivity regime, which is most challenging according to our taxonomy.

\vspace{10mm}
\textbf{Acknowledgements.}
We want to thank Charles Martin, Rajiv Khanna, Zhewei Yao, and Amir Gholami for helpful discussions. 
Michael W. Mahoney would like to acknowledge the UC Berkeley CLTC, ARO, IARPA (contract W911NF20C0035), NSF, and ONR for providing partial support of this work. 
Kannan Ramchandran would like to acknowledge support from NSF CIF-2007669, CIF-1703678, and  CIF-2002821.
Joseph E. Gonzalez would like to acknowledge supports from NSF CISE Expeditions Award CCF-1730628 and gifts from Alibaba Group, Amazon Web Services, Ant Group, CapitalOne, Ericsson, Facebook, Futurewei, Google, Intel, Microsoft, Nvidia, Scotiabank, Splunk and VMware. 
Our conclusions do not necessarily reflect the position or the policy of our sponsors, and no official endorsement should be~inferred.

\bibliographystyle{unsrtnat}
\bibliography{papers}

\begin{thebibliography}{97}
\providecommand{\natexlab}[1]{#1}
\providecommand{\url}[1]{\texttt{#1}}
\expandafter\ifx\csname urlstyle\endcsname\relax
  \providecommand{\doi}[1]{doi: #1}\else
  \providecommand{\doi}{doi: \begingroup \urlstyle{rm}\Url}\fi

\bibitem[Li et~al.(2018)Li, Xu, Taylor, Studer, and
  Goldstein]{li2018visualizing}
Hao Li, Zheng Xu, Gavin Taylor, Christoph Studer, and Tom Goldstein.
\newblock Visualizing the loss landscape of neural nets.
\newblock In \emph{Conference on Neural Information Processing Systems}, pages
  6389--6399, 2018.

\bibitem[Ballard et~al.(2017)Ballard, Das, Martiniani, Mehta, Sagun, Stevenson,
  and Wales]{ballard2017energy}
Andrew~J Ballard, Ritankar Das, Stefano Martiniani, Dhagash Mehta, Levent
  Sagun, Jacob~D Stevenson, and David~J Wales.
\newblock Energy landscapes for machine learning.
\newblock \emph{Physical Chemistry Chemical Physics}, 19\penalty0
  (20):\penalty0 12585--12603, 2017.

\bibitem[Keskar et~al.(2017)Keskar, Nocedal, Tang, Mudigere, and
  Smelyanskiy]{keskar2016large}
Nitish~Shirish Keskar, Jorge Nocedal, Ping Tak~Peter Tang, Dheevatsa Mudigere,
  and Mikhail Smelyanskiy.
\newblock On large-batch training for deep learning: Generalization gap and
  sharp minima.
\newblock In \emph{International Conference on Learning Representations}, 2017.

\bibitem[Yao et~al.(2018{\natexlab{a}})Yao, Gholami, Keutzer, and
  Mahoney]{YGKM18_TRv1}
Zhewei Yao, Amir Gholami, Kurt Keutzer, and Michael~W Mahoney.
\newblock Large batch size training of neural networks with adversarial
  training and second-order information.
\newblock Technical Report Preprint: arXiv:1810.01021, 2018{\natexlab{a}}.

\bibitem[Yao et~al.(2018{\natexlab{b}})Yao, Gholami, Lei, Keutzer, and
  Mahoney]{yao2018hessian}
Zhewei Yao, Amir Gholami, Qi~Lei, Kurt Keutzer, and Michael~W Mahoney.
\newblock Hessian-based analysis of large batch training and robustness to
  adversaries.
\newblock In \emph{Conference on Neural Information Processing Systems},
  volume~31, pages 4949--4959, 2018{\natexlab{b}}.

\bibitem[Li and Yuan(2017)]{li2017convergence}
Yuanzhi Li and Yang Yuan.
\newblock Convergence analysis of two-layer neural networks with {ReLU}
  activation.
\newblock In \emph{Conference on Neural Information Processing Systems}, pages
  597--607, 2017.

\bibitem[Santurkar et~al.(2018)Santurkar, Tsipras, Ilyas, and
  M{\k{a}}dry]{santurkar2018does}
Shibani Santurkar, Dimitris Tsipras, Andrew Ilyas, and Aleksander M{\k{a}}dry.
\newblock How does batch normalization help optimization?
\newblock In \emph{Neural Information Processing Systems}, pages 2488--2498,
  2018.

\bibitem[Dinh et~al.(2017)Dinh, Pascanu, Bengio, and Bengio]{dinh2017sharp}
Laurent Dinh, Razvan Pascanu, Samy Bengio, and Yoshua Bengio.
\newblock Sharp minima can generalize for deep nets.
\newblock In \emph{International Conference on Machine Learning}, pages
  1019--1028, 2017.

\bibitem[Neyshabur et~al.(2018)Neyshabur, Bhojanapalli, and
  Srebro]{neyshabur2017pac}
Behnam Neyshabur, Srinadh Bhojanapalli, and Nathan Srebro.
\newblock A {PAC}-{B}ayesian approach to spectrally-normalized margin bounds
  for neural networks.
\newblock In \emph{International Conference on Learning Representations}, 2018.

\bibitem[Foret et~al.(2020)Foret, Kleiner, Mobahi, and
  Neyshabur]{foret2020sharpness}
Pierre Foret, Ariel Kleiner, Hossein Mobahi, and Behnam Neyshabur.
\newblock Sharpness-aware minimization for efficiently improving
  generalization.
\newblock \emph{arXiv preprint arXiv:2010.01412}, 2020.

\bibitem[Yao et~al.(2020)Yao, Gholami, Keutzer, and Mahoney]{yao2020pyhessian}
Zhewei Yao, Amir Gholami, Kurt Keutzer, and Michael~W Mahoney.
\newblock Py{H}essian: Neural networks through the lens of the hessian.
\newblock In \emph{IEEE International Conference on Big Data (Big Data)}, pages
  581--590, 2020.

\bibitem[Engel and den Broeck(2001)]{EB01_BOOK}
Andreas Engel and Christian P. L.~Van den Broeck.
\newblock \emph{Statistical mechanics of learning}.
\newblock Cambridge University Press, New York, NY, USA, 2001.

\bibitem[Martin and Mahoney(2017)]{martin2017rethinking}
Charles~H Martin and Michael~W Mahoney.
\newblock Rethinking generalization requires revisiting old ideas: statistical
  mechanics approaches and complex learning behavior.
\newblock Technical Report Preprint: arXiv:1710.09553, 2017.

\bibitem[Granziol(2020)]{granziol2020flatness}
Diego Granziol.
\newblock Flatness is a false friend.
\newblock Technical Report Preprint: arXiv:2006.09091, 2020.

\bibitem[Tsipras et~al.(2019)Tsipras, Santurkar, Engstrom, Turner, and
  Madry]{tsipras2018robustness}
Dimitris Tsipras, Shibani Santurkar, Logan Engstrom, Alexander Turner, and
  Aleksander Madry.
\newblock Robustness may be at odds with accuracy.
\newblock In \emph{International Conference on Learning Representations}, 2019.

\bibitem[Martin and
  Mahoney(2021{\natexlab{a}})]{martin2018implicit_JMLRversion}
Charles~H Martin and Michael~W Mahoney.
\newblock Implicit self-regularization in deep neural networks: Evidence from
  random matrix theory and implications for learning.
\newblock \emph{Journal of Machine Learning Research}, 22\penalty0
  (165):\penalty0 1--73, 2021{\natexlab{a}}.

\bibitem[Garipov et~al.(2018)Garipov, Izmailov, Podoprikhin, Vetrov, and
  Wilson]{garipov2018loss}
Timur Garipov, Pavel Izmailov, Dmitrii Podoprikhin, Dmitry Vetrov, and
  Andrew~Gordon Wilson.
\newblock Loss surfaces, mode connectivity, and fast ensembling of dnns.
\newblock In \emph{Conference on Neural Information Processing Systems}, pages
  8803--8812, 2018.

\bibitem[Draxler et~al.(2018)Draxler, Veschgini, Salmhofer, and
  Hamprecht]{draxler2018essentially}
Felix Draxler, Kambis Veschgini, Manfred Salmhofer, and Fred Hamprecht.
\newblock Essentially no barriers in neural network energy landscape.
\newblock In \emph{International conference on machine learning}, pages
  1309--1318, 2018.

\bibitem[Kornblith et~al.(2019)Kornblith, Norouzi, Lee, and
  Hinton]{kornblith2019similarity}
Simon Kornblith, Mohammad Norouzi, Honglak Lee, and Geoffrey Hinton.
\newblock Similarity of neural network representations revisited.
\newblock In \emph{International Conference on Machine Learning}, pages
  3519--3529, 2019.

\bibitem[Advani et~al.(2013)Advani, Lahiri, and Ganguli]{Advani_2013}
Madhu Advani, Subhaneil Lahiri, and Surya Ganguli.
\newblock Statistical mechanics of complex neural systems and high dimensional
  data.
\newblock \emph{Journal of Statistical Mechanics: Theory and Experiment},
  2013\penalty0 (03):\penalty0 P03014, mar 2013.
\newblock \doi{10.1088/1742-5468/2013/03/p03014}.
\newblock URL \url{https://doi.org/10.1088/1742-5468/2013/03/p03014}.

\bibitem[Belkin et~al.(2019)Belkin, Hsu, Ma, and Mandal]{belkin2019reconciling}
Mikhail Belkin, Daniel Hsu, Siyuan Ma, and Soumik Mandal.
\newblock Reconciling modern machine-learning practice and the classical
  bias--variance trade-off.
\newblock \emph{Proceedings of the National Academy of Sciences}, 116\penalty0
  (32):\penalty0 15849--15854, 2019.

\bibitem[Belkin et~al.(2020)Belkin, Hsu, and Xu]{belkin2020two}
Mikhail Belkin, Daniel Hsu, and Ji~Xu.
\newblock Two models of double descent for weak features.
\newblock \emph{SIAM Journal on Mathematics of Data Science}, 2\penalty0
  (4):\penalty0 1167--1180, 2020.

\bibitem[Liao et~al.(2020)Liao, Couillet, and Mahoney]{liao2020random}
Zhenyu Liao, Romain Couillet, and Michael~W Mahoney.
\newblock A random matrix analysis of random fourier features: beyond the
  {G}aussian kernel, a precise phase transition, and the corresponding double
  descent.
\newblock In \emph{Conference on Neural Information Processing Systems}, 2020.

\bibitem[Derezi{\'n}ski et~al.(2020{\natexlab{a}})Derezi{\'n}ski, Liang, and
  Mahoney]{derezinski2019exact}
Micha{\l} Derezi{\'n}ski, Feynman Liang, and Michael~W Mahoney.
\newblock Exact expressions for double descent and implicit regularization via
  surrogate random design.
\newblock In \emph{Conference on Neural Information Processing Systems},
  volume~33, 2020{\natexlab{a}}.

\bibitem[Amit et~al.(1985)Amit, Gutfreund, and Sompolinsky]{amit1985storing}
Daniel~J Amit, Hanoch Gutfreund, and Haim Sompolinsky.
\newblock Storing infinite numbers of patterns in a spin-glass model of neural
  networks.
\newblock \emph{Physical Review Letters}, 55\penalty0 (14):\penalty0 1530,
  1985.

\bibitem[Choromanska et~al.(2015)Choromanska, Henaff, Mathieu, Arous, and
  LeCun]{choromanska2015loss}
Anna Choromanska, Mikael Henaff, Michael Mathieu, G{\'e}rard~Ben Arous, and
  Yann LeCun.
\newblock The loss surfaces of multilayer networks.
\newblock In \emph{Artificial intelligence and statistics}, pages 192--204,
  2015.

\bibitem[Agliari et~al.(2014)Agliari, Barra, Galluzzi, Tantari, and
  Tavani]{agliari2014walk}
Elena Agliari, Adriano Barra, Andrea Galluzzi, Daniele Tantari, and Flavia
  Tavani.
\newblock A walk in the statistical mechanical formulation of neural networks.
\newblock In \emph{International Joint Conference on Computational
  Intelligence}, pages 210--217, 2014.

\bibitem[Fuhs and Touretzky(2006)]{fuhs2006spin}
Mark~C Fuhs and David~S Touretzky.
\newblock A spin glass model of path integration in rat medial entorhinal
  cortex.
\newblock \emph{The Journal of neuroscience: the official journal of the
  Society for Neuroscience}, 26\penalty0 (16):\penalty0 4266--4276, 2006.

\bibitem[Hudetz et~al.(2014)Hudetz, Humphries, and Binder]{hudetz2014spin}
Anthony~G Hudetz, Colin~J Humphries, and Jeffrey~R Binder.
\newblock Spin-glass model predicts metastable brain states that diminish in
  anesthesia.
\newblock \emph{Frontiers in systems neuroscience}, 8:\penalty0 234, 2014.

\bibitem[Recio and Torres(2016)]{recio2016emergence}
Ibon Recio and Joaqu{\'\i}n~J Torres.
\newblock Emergence of low noise frustrated states in {E}/{I} balanced neural
  networks.
\newblock \emph{Neural Networks}, 84:\penalty0 91--101, 2016.

\bibitem[Bryngelson and Wolynes(1987)]{bryngelson1987spin}
Joseph~D Bryngelson and Peter~G Wolynes.
\newblock Spin glasses and the statistical mechanics of protein folding.
\newblock \emph{Proceedings of the National Academy of sciences}, 84\penalty0
  (21):\penalty0 7524--7528, 1987.

\bibitem[Garstecki et~al.(1999)Garstecki, Hoang, and
  Cieplak]{garstecki1999energy}
Piotr Garstecki, Trinh~Xuan Hoang, and Marek Cieplak.
\newblock Energy landscapes, supergraphs, and “folding funnels” in spin
  systems.
\newblock \emph{Physical Review E}, 60\penalty0 (3):\penalty0 3219, 1999.

\bibitem[Klemm et~al.(2008)Klemm, Flamm, and Stadler]{klemm2007funnels}
Konstantin Klemm, Christoph Flamm, and Peter~F Stadler.
\newblock Funnels in energy landscapes.
\newblock \emph{The European Physical Journal B}, 63\penalty0 (3):\penalty0
  387--391, 2008.

\bibitem[Brooks et~al.(2001)Brooks, Onuchic, and Wales]{brooks2001taking}
Charles~L Brooks, Jos{\'e}~N Onuchic, and David~J Wales.
\newblock Taking a walk on a landscape.
\newblock \emph{Science}, 293\penalty0 (5530):\penalty0 612--613, 2001.

\bibitem[Wales(2003)]{wales_book}
D.~J. Wales.
\newblock \emph{Energy Landscapes: Applications to Clusters, Biomolecules and
  Glasses}.
\newblock Cambridge University Press, 2003.

\bibitem[Stillinger(2016)]{stillinger_book}
F.~H. Stillinger.
\newblock \emph{Energy Landscapes, Inherent Structures, and Condensed-Matter
  Phenomena}.
\newblock Princeton University Press, 2016.

\bibitem[Neal et~al.(2018)Neal, Mittal, Baratin, Tantia, Scicluna,
  Lacoste-Julien, and Mitliagkas]{neal2018modern}
Brady Neal, Sarthak Mittal, Aristide Baratin, Vinayak Tantia, Matthew Scicluna,
  Simon Lacoste-Julien, and Ioannis Mitliagkas.
\newblock A modern take on the bias-variance tradeoff in neural networks.
\newblock Technical Report Preprint: arXiv:1810.08591, 2018.

\bibitem[Zhang et~al.(2018)Zhang, Cisse, Dauphin, and
  Lopez-Paz]{zhang2017mixup}
Hongyi Zhang, Moustapha Cisse, Yann~N Dauphin, and David Lopez-Paz.
\newblock mixup: Beyond empirical risk minimization.
\newblock In \emph{International Conference on Learning Representations}, 2018.

\bibitem[He et~al.(2016)He, Zhang, Ren, and Sun]{he2016deep}
Kaiming He, Xiangyu Zhang, Shaoqing Ren, and Jian Sun.
\newblock Deep residual learning for image recognition.
\newblock In \emph{IEEE conference on computer vision and pattern recognition},
  pages 770--778, 2016.

\bibitem[Krizhevsky et~al.(2009)Krizhevsky, Hinton,
  et~al.]{krizhevsky2009learning}
Alex Krizhevsky, Geoffrey Hinton, et~al.
\newblock Learning multiple layers of features from tiny images.
\newblock 2009.

\bibitem[Nakkiran et~al.(2019)Nakkiran, Kaplun, Bansal, Yang, Barak, and
  Sutskever]{nakkiran2019deep}
Preetum Nakkiran, Gal Kaplun, Yamini Bansal, Tristan Yang, Boaz Barak, and Ilya
  Sutskever.
\newblock Deep double descent: Where bigger models and more data hurt.
\newblock In \emph{International Conference on Learning Representations}, 2019.

\bibitem[Martin and Mahoney(2021{\natexlab{b}})]{MM21a_simpsons_TR}
Charles~H Martin and Michael~W Mahoney.
\newblock Post-mortem on a deep learning contest: a {S}impson's paradox and the
  complementary roles of scale metrics versus shape metrics.
\newblock Technical Report Preprint: arXiv:2106.00734, 2021{\natexlab{b}}.

\bibitem[Miceli-Barone et~al.(2017)Miceli-Barone, Haddow, Germann, and
  Sennrich]{barone2017regularization}
Antonio~Valerio Miceli-Barone, Barry Haddow, Ulrich Germann, and Rico Sennrich.
\newblock Regularization techniques for fine-tuning in neural machine
  translation.
\newblock In \emph{Conference on Empirical Methods in Natural Language
  Processing}, pages 1489--1494, 2017.

\bibitem[Gal and Ghahramani(2016)]{gal2016theoretically}
Yarin Gal and Zoubin Ghahramani.
\newblock A theoretically grounded application of dropout in recurrent neural
  networks.
\newblock In \emph{Conference on Neural Information Processing Systems},
  volume~29, pages 1019--1027, 2016.

\bibitem[Goyal et~al.(2017)Goyal, Doll{\'a}r, Girshick, Noordhuis, Wesolowski,
  Kyrola, Tulloch, Jia, and He]{goyal2017accurate}
Priya Goyal, Piotr Doll{\'a}r, Ross Girshick, Pieter Noordhuis, Lukasz
  Wesolowski, Aapo Kyrola, Andrew Tulloch, Yangqing Jia, and Kaiming He.
\newblock Accurate, large minibatch {SGD}: Training {I}mage{N}et in 1 hour.
\newblock Technical Report Preprint: arXiv:1706.02677, 2017.

\bibitem[Xing et~al.(2018)Xing, Arpit, Tsirigotis, and Bengio]{xing2018walk}
Chen Xing, Devansh Arpit, Christos Tsirigotis, and Yoshua Bengio.
\newblock A walk with {SGD}.
\newblock Technical Report Preprint: arXiv:1802.08770, 2018.

\bibitem[Kleinberg et~al.(2018)Kleinberg, Li, and
  Yuan]{kleinberg2018alternative}
Bobby Kleinberg, Yuanzhi Li, and Yang Yuan.
\newblock An alternative view: When does sgd escape local minima?
\newblock In \emph{International Conference on Machine Learning}, pages
  2698--2707, 2018.

\bibitem[Mahoney and Martin(2019)]{martin2019traditional}
Michael Mahoney and Charles Martin.
\newblock Traditional and heavy tailed self regularization in neural network
  models.
\newblock In \emph{International Conference on Machine Learning}, pages
  4284--4293, 2019.

\bibitem[Martin and Mahoney(2020)]{martin2020heavy}
Charles~H Martin and Michael~W Mahoney.
\newblock Heavy-tailed universality predicts trends in test accuracies for very
  large pre-trained deep neural networks.
\newblock In \emph{SIAM International Conference on Data Mining}, pages
  505--513. SIAM, 2020.

\bibitem[Martin et~al.(2021)Martin, Peng, and
  Mahoney]{martin2020predicting_NatComm}
Charles~H Martin, Tongsu~Serena Peng, and Michael~W Mahoney.
\newblock Predicting trends in the quality of state-of-the-art neural networks
  without access to training or testing data.
\newblock \emph{Nature Communications}, 12\penalty0 (1):\penalty0 1--13, 2021.

\bibitem[Simsekli et~al.(2019)Simsekli, Sagun, and
  Gurbuzbalaban]{simsekli2019tail}
Umut Simsekli, Levent Sagun, and Mert Gurbuzbalaban.
\newblock A tail-index analysis of stochastic gradient noise in deep neural
  networks.
\newblock In \emph{International Conference on Machine Learning}, pages
  5827--5837, 2019.

\bibitem[Hodgkinson and Mahoney(2017)]{hodgkinson2020multiplicative}
Liam Hodgkinson and Michael~W Mahoney.
\newblock Multiplicative noise and heavy tails in stochastic optimization.
\newblock In \emph{International Conference on Machine Learning}, pages
  1019--1028, 2017.

\bibitem[Fort et~al.(2019)Fort, Hu, and Lakshminarayanan]{fort2019deep}
Stanislav Fort, Huiyi Hu, and Balaji Lakshminarayanan.
\newblock Deep ensembles: A loss landscape perspective.
\newblock \emph{arXiv preprint arXiv:1912.02757}, 2019.

\bibitem[Fort et~al.(2020)Fort, Dziugaite, Paul, Kharaghani, Roy, and
  Ganguli]{fort2020deep}
Stanislav Fort, Gintare~Karolina Dziugaite, Mansheej Paul, Sepideh Kharaghani,
  Daniel~M Roy, and Surya Ganguli.
\newblock Deep learning versus kernel learning: an empirical study of loss
  landscape geometry and the time evolution of the neural tangent kernel.
\newblock In \emph{Conference on Neural Information Processing Systems}, 2020.

\bibitem[Sun et~al.(2020)Sun, Li, Liang, Ding, and Srikant]{sun2020global}
Ruoyu Sun, Dawei Li, Shiyu Liang, Tian Ding, and Rayadurgam Srikant.
\newblock The global landscape of neural networks: An overview.
\newblock \emph{IEEE Signal Processing Magazine}, 37\penalty0 (5):\penalty0
  95--108, 2020.

\bibitem[Dauphin et~al.(2014)Dauphin, Pascanu, Gulcehre, Cho, Ganguli, and
  Bengio]{dauphin2014identifying}
Yann~N Dauphin, Razvan Pascanu, Caglar Gulcehre, Kyunghyun Cho, Surya Ganguli,
  and Yoshua Bengio.
\newblock Identifying and attacking the saddle point problem in
  high-dimensional non-convex optimization.
\newblock In \emph{Conference on Neural Information Processing Systems}, pages
  2933--2941, 2014.

\bibitem[Ge et~al.(2015)Ge, Huang, Jin, and Yuan]{ge2015escaping}
Rong Ge, Furong Huang, Chi Jin, and Yang Yuan.
\newblock Escaping from saddle points—online stochastic gradient for tensor
  decomposition.
\newblock In \emph{Conference on learning theory}, pages 797--842, 2015.

\bibitem[Du et~al.(2017)Du, Jin, Jordan, P{\'o}czos, Singh, and
  Lee]{du2017gradient}
SS~Du, C~Jin, MI~Jordan, B~P{\'o}czos, A~Singh, and JD~Lee.
\newblock Gradient descent can take exponential time to escape saddle points.
\newblock In \emph{Conference on Neural Information Processing Systems}, pages
  1068--1078, 2017.

\bibitem[Jin et~al.(2017)Jin, Ge, Netrapalli, Kakade, and
  Jordan]{jin2017escape}
Chi Jin, Rong Ge, Praneeth Netrapalli, Sham~M Kakade, and Michael~I Jordan.
\newblock How to escape saddle points efficiently.
\newblock In \emph{International Conference on Machine Learning}, pages
  1724--1732, 2017.

\bibitem[Safran and Shamir(2018)]{safran2018spurious}
Itay Safran and Ohad Shamir.
\newblock Spurious local minima are common in two-layer relu neural networks.
\newblock In \emph{International Conference on Machine Learning}, pages
  4433--4441, 2018.

\bibitem[Soltanolkotabi et~al.(2018)Soltanolkotabi, Javanmard, and
  Lee]{soltanolkotabi2018theoretical}
Mahdi Soltanolkotabi, Adel Javanmard, and Jason~D Lee.
\newblock Theoretical insights into the optimization landscape of
  over-parameterized shallow neural networks.
\newblock \emph{IEEE Transactions on Information Theory}, 65\penalty0
  (2):\penalty0 742--769, 2018.

\bibitem[Du et~al.(2019)Du, Lee, Li, Wang, and Zhai]{du2019gradient}
Simon Du, Jason Lee, Haochuan Li, Liwei Wang, and Xiyu Zhai.
\newblock Gradient descent finds global minima of deep neural networks.
\newblock In \emph{International Conference on Machine Learning}, pages
  1675--1685, 2019.

\bibitem[Novak et~al.(2018)Novak, Bahri, Abolafia, Pennington, and
  Sohl-Dickstein]{novak2018sensitivity}
Roman Novak, Yasaman Bahri, Daniel~A Abolafia, Jeffrey Pennington, and Jascha
  Sohl-Dickstein.
\newblock Sensitivity and generalization in neural networks: an empirical
  study.
\newblock In \emph{International Conference on Learning Representations}, 2018.

\bibitem[Liu et~al.(2019)Liu, Bai, Jiang, Chen, and Wang]{liu2020understanding}
Jinlong Liu, Yunzhi Bai, Guoqing Jiang, Ting Chen, and Huayan Wang.
\newblock Understanding why neural networks generalize well through {GSNR} of
  parameters.
\newblock In \emph{International Conference on Learning Representations}, 2019.

\bibitem[McAllester(1999)]{mcallester1999pac}
David~A McAllester.
\newblock {PAC}-{B}ayesian model averaging.
\newblock In \emph{Annual Conference on Computational Learning Theory}, pages
  164--170, 1999.

\bibitem[Dziugaite and Roy(2017)]{dziugaite2017computing}
Gintare~Karolina Dziugaite and Daniel~M Roy.
\newblock Computing nonvacuous generalization bounds for deep (stochastic)
  neural networks with many more parameters than training data.
\newblock In \emph{Annual Conference on Uncertainty in Artificial Intelligence
  (UAI)}, 2017.

\bibitem[Jiang et~al.(2019)Jiang, Neyshabur, Mobahi, Krishnan, and
  Bengio]{jiang2019fantastic}
Yiding Jiang, Behnam Neyshabur, Hossein Mobahi, Dilip Krishnan, and Samy
  Bengio.
\newblock Fantastic generalization measures and where to find them.
\newblock In \emph{International Conference on Learning Representations}, 2019.

\bibitem[Geiger et~al.(2019)Geiger, Spigler, d'Ascoli, Sagun, Baity-Jesi,
  Biroli, and Wyart]{geiger2019jamming}
Mario Geiger, Stefano Spigler, St{\'e}phane d'Ascoli, Levent Sagun, Marco
  Baity-Jesi, Giulio Biroli, and Matthieu Wyart.
\newblock Jamming transition as a paradigm to understand the loss landscape of
  deep neural networks.
\newblock \emph{Physical Review E}, 100\penalty0 (1):\penalty0 012115, 2019.

\bibitem[Sagun et~al.(2016)Sagun, Bottou, and LeCun]{sagun2016eigenvalues}
Levent Sagun, Leon Bottou, and Yann LeCun.
\newblock Eigenvalues of the {H}essian in deep learning: Singularity and
  beyond.
\newblock Technical Report Preprint: arXiv:1611.07476, 2016.

\bibitem[Gur-Ari et~al.(2018)Gur-Ari, Roberts, and Dyer]{gur2018gradient}
Guy Gur-Ari, Daniel~A Roberts, and Ethan Dyer.
\newblock Gradient descent happens in a tiny subspace.
\newblock Technical Report Preprint: arXiv:1812.04754, 2018.

\bibitem[Papyan(2020)]{papyan2020traces}
Vardan Papyan.
\newblock Traces of class/cross-class structure pervade deep learning spectra.
\newblock \emph{Journal of Machine Learning Research}, 21\penalty0
  (252):\penalty0 1--64, 2020.

\bibitem[Fort and Ganguli(2019)]{fort2019emergent}
Stanislav Fort and Surya Ganguli.
\newblock Emergent properties of the local geometry of neural loss landscapes.
\newblock Technical Report Preprint: arXiv:1910.05929, 2019.

\bibitem[He et~al.(2019)He, Huang, and Yuan]{he2019asymmetric}
Haowei He, Gao Huang, and Yang Yuan.
\newblock Asymmetric valleys: Beyond sharp and flat local minima.
\newblock In \emph{Conference on Neural Information Processing Systems}, pages
  2553--2564, 2019.

\bibitem[Chaudhari et~al.(2019)Chaudhari, Choromanska, Soatto, LeCun, Baldassi,
  Borgs, Chayes, Sagun, and Zecchina]{chaudhari2019entropy}
Pratik Chaudhari, Anna Choromanska, Stefano Soatto, Yann LeCun, Carlo Baldassi,
  Christian Borgs, Jennifer Chayes, Levent Sagun, and Riccardo Zecchina.
\newblock Entropy-{SGD}: biasing gradient descent into wide valleys.
\newblock \emph{Journal of Statistical Mechanics: Theory and Experiment},
  12\penalty0 (12):\penalty0 124018, 2019.

\bibitem[Izmailov et~al.(2018)Izmailov, Wilson, Podoprikhin, Vetrov, and
  Garipov]{izmailov2018averaging}
P~Izmailov, AG~Wilson, D~Podoprikhin, D~Vetrov, and T~Garipov.
\newblock Averaging weights leads to wider optima and better generalization.
\newblock In \emph{Conference on Uncertainty in Artificial Intelligence}, pages
  876--885, 2018.

\bibitem[Dong et~al.(2019)Dong, Yao, Gholami, Mahoney, and
  Keutzer]{dong2019hawq}
Zhen Dong, Zhewei Yao, Amir Gholami, Michael~W Mahoney, and Kurt Keutzer.
\newblock {HAWQ}: Hessian aware quantization of neural networks with
  mixed-precision.
\newblock In \emph{IEEE/CVF International Conference on Computer Vision}, pages
  293--302, 2019.

\bibitem[Shen et~al.(2020)Shen, Dong, Ye, Ma, Yao, Gholami, Mahoney, and
  Keutzer]{shen2020q}
Sheng Shen, Zhen Dong, Jiayu Ye, Linjian Ma, Zhewei Yao, Amir Gholami,
  Michael~W Mahoney, and Kurt Keutzer.
\newblock Q-{BERT}: Hessian based ultra low precision quantization of bert.
\newblock In \emph{AAAI Conference on Artificial Intelligence}, volume~34,
  pages 8815--8821, 2020.

\bibitem[Goodfellow et~al.(2014)Goodfellow, Vinyals, and
  Saxe]{goodfellow2014qualitatively}
Ian~J Goodfellow, Oriol Vinyals, and Andrew~M Saxe.
\newblock Qualitatively characterizing neural network optimization problems.
\newblock Technical Report Preprint: arXiv:1412.6544, 2014.

\bibitem[Cooper(2018)]{cooper2018loss}
Yaim Cooper.
\newblock The loss landscape of overparameterized neural networks.
\newblock Technical Report Preprint: arXiv:1804.10200, 2018.

\bibitem[Freeman and Bruna(2017)]{freeman2016topology}
C~Daniel Freeman and Joan Bruna.
\newblock Topology and geometry of half-rectified network optimization.
\newblock In \emph{International Conference on Learning Representations}, 2017.

\bibitem[Nguyen(2019)]{nguyen2019connected}
Quynh Nguyen.
\newblock On connected sublevel sets in deep learning.
\newblock In \emph{International Conference on Machine Learning}, pages
  4790--4799, 2019.

\bibitem[Kuditipudi et~al.(2019)Kuditipudi, Wang, Lee, Zhang, Li, Hu, Ge, and
  Arora]{kuditipudi2019explaining}
Rohith Kuditipudi, Xiang Wang, Holden Lee, Yi~Zhang, Zhiyuan Li, Wei Hu, Rong
  Ge, and Sanjeev Arora.
\newblock Explaining landscape connectivity of low-cost solutions for
  multilayer nets.
\newblock \emph{Advances in Neural Information Processing Systems},
  32:\penalty0 14601--14610, 2019.

\bibitem[Shevchenko and Mondelli(2020)]{shevchenko2020landscape}
Alexander Shevchenko and Marco Mondelli.
\newblock Landscape connectivity and dropout stability of {SGD} solutions for
  over-parameterized neural networks.
\newblock In \emph{International Conference on Machine Learning}, pages
  8773--8784, 2020.

\bibitem[Frankle et~al.(2020)Frankle, Dziugaite, Roy, and
  Carbin]{frankle2020linear}
Jonathan Frankle, Gintare~Karolina Dziugaite, Daniel Roy, and Michael Carbin.
\newblock Linear mode connectivity and the lottery ticket hypothesis.
\newblock In \emph{International Conference on Machine Learning}, pages
  3259--3269, 2020.

\bibitem[Fort and Jastrzebski(2019)]{fort2019large}
Stanislav Fort and Stanislaw Jastrzebski.
\newblock Large scale structure of neural network loss landscapes.
\newblock \emph{Advances in Neural Information Processing Systems},
  32:\penalty0 6709--6717, 2019.

\bibitem[Muthukumar et~al.(2020)Muthukumar, Vodrahalli, Subramanian, and
  Sahai]{muthukumar2020harmless}
Vidya Muthukumar, Kailas Vodrahalli, Vignesh Subramanian, and Anant Sahai.
\newblock Harmless interpolation of noisy data in regression.
\newblock \emph{IEEE Journal on Selected Areas in Information Theory},
  1\penalty0 (1):\penalty0 67--83, 2020.

\bibitem[Mei and Montanari()]{mei2019generalization}
Song Mei and Andrea Montanari.
\newblock The generalization error of random features regression: Precise
  asymptotics and the double descent curve.
\newblock \emph{Communications on Pure and Applied Mathematics}.

\bibitem[Bartlett et~al.(2020)Bartlett, Long, Lugosi, and
  Tsigler]{bartlett2020benign}
Peter~L Bartlett, Philip~M Long, G{\'a}bor Lugosi, and Alexander Tsigler.
\newblock Benign overfitting in linear regression.
\newblock \emph{Proceedings of the National Academy of Sciences}, 117\penalty0
  (48):\penalty0 30063--30070, 2020.

\bibitem[Zhang et~al.(2017)Zhang, Bengio, Hardt, Recht, and
  Vinyals]{zhang2016understanding}
Chiyuan Zhang, Samy Bengio, Moritz Hardt, Benjamin Recht, and Oriol Vinyals.
\newblock Understanding deep learning requires rethinking generalization.
\newblock In \emph{International Conference on Learning Representations}, 2017.

\bibitem[Chen et~al.(2020)Chen, Min, Belkin, and Karbasi]{chen2020multiple}
Lin Chen, Yifei Min, Mikhail Belkin, and Amin Karbasi.
\newblock Multiple descent: Design your own generalization curve.
\newblock Technical Report Preprint: arXiv:2008.01036, 2020.

\bibitem[Adlam and Pennington(2020)]{adlam2020neural}
Ben Adlam and Jeffrey Pennington.
\newblock The neural tangent kernel in high dimensions: Triple descent and a
  multi-scale theory of generalization.
\newblock In \emph{International Conference on Machine Learning}, pages 74--84,
  2020.

\bibitem[d'Ascoli et~al.(2020)d'Ascoli, Sagun, and Biroli]{d2020triple}
St{\'e}phane d'Ascoli, Levent Sagun, and Giulio Biroli.
\newblock Triple descent and the two kinds of overfitting: Where \& why do they
  appear?
\newblock In \emph{Conference on Neural Information Processing Systems}, 2020.

\bibitem[Derezi{\'n}ski et~al.(2020{\natexlab{b}})Derezi{\'n}ski, Khanna, and
  Mahoney]{DKM20_TR}
M.~Derezi{\'n}ski, R.~Khanna, and M.~W. Mahoney.
\newblock Improved guarantees and a multiple-descent curve for {C}olumn
  {S}ubset {S}election and the {N}ystrom method.
\newblock Technical Report Preprint: arXiv:2002.09073, 2020{\natexlab{b}}.

\bibitem[Yang et~al.(2020)Yang, Yu, You, Steinhardt, and
  Ma]{yang2020rethinking}
Zitong Yang, Yaodong Yu, Chong You, Jacob Steinhardt, and Yi~Ma.
\newblock Rethinking bias-variance trade-off for generalization of neural
  networks.
\newblock In \emph{International Conference on Machine Learning}, pages
  10767--10777, 2020.

\bibitem[Netzer et~al.(2011)Netzer, Wang, Coates, Bissacco, Wu, and
  Ng]{netzer2011reading}
Yuval Netzer, Tao Wang, Adam Coates, Alessandro Bissacco, Bo~Wu, and Andrew~Y
  Ng.
\newblock Reading digits in natural images with unsupervised feature learning.
\newblock 2011.

\bibitem[Cettolo et~al.(2012)Cettolo, Girardi, and Federico]{cettolo2012wit3}
Mauro Cettolo, Christian Girardi, and Marcello Federico.
\newblock Wit3: Web inventory of transcribed and translated talks.
\newblock In \emph{Conference of european association for machine translation},
  pages 261--268, 2012.

\bibitem[Vaswani et~al.(2017)Vaswani, Shazeer, Parmar, Uszkoreit, Jones, Gomez,
  Kaiser, and Polosukhin]{vaswani2017attention}
Ashish Vaswani, Noam Shazeer, Niki Parmar, Jakob Uszkoreit, Llion Jones,
  Aidan~N Gomez, {\L}ukasz Kaiser, and Illia Polosukhin.
\newblock Attention is all you need.
\newblock In \emph{Advances in neural information processing systems}, pages
  5998--6008, 2017.

\end{thebibliography}



\ifisarxiv
\appendix
\section{Formal definitions on connectivity and similarity}\label{sec:definitions} 

In this section, we more formally define what we mean by globally nice loss landscapes, characterized by connectivity and similarity. 
Recall that we consider the nominal empirical risk minimization formulation, defined in Eqn. \eqref{eqn:training_loss}, and solved using SGD iterations described in Eqn. \eqref{eq:sgd-def}. 
We denote the underlying distribution from which we draw training/test samples as $P$.

In many cases, such as for deep NNs, a single input-output mapping $f: \mathbb{R}^{d_\text{in}}\to \mathbb{R}^{d_\text{out}}$ can be realized by an infinite set of different weights $\theta$, where two weights $\theta$ and $\theta'$ give the same mapping if for any $\mathbf{x}\in \mathbb{R}^{d_\text{in}}$, $f_{\theta}(\mathbf{x}) = f_{\theta'}(\mathbf{x})$.
For example, for networks using the ReLU activation, if we multiply the weights of a particular layer by a constant and divide the weights of the next layer by the same constant, we get the same mapping.
Thus, we view the different weights $\theta$ and $\theta'$ that give the same input-output mapping as being essentially the same.
We use the definitions in the following section to ensure that the similarity between $f_{\theta}$ and $f_{\theta'}$ equals 1 (which is the largest similarity in our definition) if they give the same input-output mapping.

\subsection{Definition of similarity}
\label{sec:definitions_CKA}

In this subsection, we quantify the closeness between weights trained from different initializations. We use $s(U,V)$ to denote a similarity metric between two matrices $U$ and $V$, i.e., it equals $1$ if $U=V$ and is close to $0$ when $U$ and $V$ are dissimilar. 

\begin{myDef}[Similarity between weights]\label{def:distance}
Given the similarity metric $s(U,V)$ and the concatenation of the outputs of $f_{\theta}$ and $f_{\theta'}$ on $m$ i.i.d. samples $(\mathbf{x}_1,\mathbf{x}_2,...,\mathbf{x}_m)$ from a perturbed distribution $Q$, written as
\begin{equation}
    F_{\theta} = [f_{\theta}(\mathbf{x}_1), f_{\theta}(\mathbf{x}_2),...,f_{\theta}(\mathbf{x}_m)]^\top,
\end{equation}
and
\begin{equation}
    F_{\theta'} = [f_{\theta'}(\mathbf{x}_1), f_{\theta'}(\mathbf{x}_2),...,f_{\theta'}(\mathbf{x}_m)]^\top,
\end{equation}
we define the similarity between two weights $\theta$ and $\theta'$ as 
\begin{equation}\label{eqn:similarity_definition}
    {\bf Similarity}(\theta,\theta') = \mathbb{E}_{(\mathbf{x}_1,\mathbf{x}_2,...,\mathbf{x}_m)\widesim[2]{i.i.d.} Q}s(F_{\theta}, F_{\theta'}).
\end{equation}
\end{myDef}
We say that a loss landscape defined by the training data $S_\text{train}$ and the space $\mathcal{F}$ over which $\theta$ is optimized has a \emph{similarity level} $\mu$ for a particular random training scheme $\mathcal{T}$, if for two weights $\theta$ and $\theta'$, trained using $\mathcal{T}$ with different random initializations,
\begin{equation}
    \mathbb{E}[{\bf Similarity}(\theta,\theta')] = \mu,
\end{equation}
where the expectation is taken over the randomness of the training scheme $\mathcal{T}$, including the random initializations and the random shuffling of data during training.
A similarity level close to 1 means that the trained models concentrate around the same mapping in terms of making predictions on $P$.

{\bf Measuring similarity.} 
In this paper, we use the CKA similarity defined in \eqref{eq:CKA} to measure the similarity between two weights.
The definition in \eqref{eq:CKA} can be obtained from \eqref{eqn:similarity_definition} by setting
\begin{equation}
    s(U,V)=\frac{\text{Cov}(U, V)}{\sqrt{\text{Cov}(U, U)\text{Cov}(V, V)}},
\end{equation}
where $\text{Cov}(X,Y) = (m-1)^{-2}\text{tr}(XX^\top H_m YY^\top H_m)$, and $H_m = I_m - m^{-1}\mathbf{1}\mathbf{1}^\top$ is the centering matrix.
We define the perturbed distribution $Q$ as the distribution of interpolations between training samples. Thus, we can sample from $Q$ by linearly combining samples from the training data $S_\text{train}$ \citep{zhang2017mixup}.
To measure CKA similarity, we compute 640 interpolated samples from the training set, which are obtained using linear coefficients following the Beta(16,16) distribution. 
In Appendix~\ref{sec:CKA_ablation}, we test different configurations on the perturbed training set.

\begin{remark}
The main reason that we use perturbed samples instead of the original training data is that measuring similarity on the training data can lead to trivial similarity, especially if the training aims to completely fit the training data. 
To explore this, we provide ablation study on different perturbed distributions in Appendix~\ref{sec:CKA_ablation}. 
\end{remark}

\subsection{Definition of connectivity}
\label{sec:mode_connectivity}

In this subsection, we quantify the connectivity between trained models.

\begin{myDef}[A curve between weights]
Suppose $\theta$ and $\theta'$ are two weights. Then, a curve between the two weights is a continuous mapping $\gamma(t), t\in[0,1]$, such that $\gamma(0) = \theta$ and $\gamma(1) = \theta'$. We say that the curve $\gamma(t)$ is $\epsilon$-low-energy if
\begin{equation}
    \int_{t=0}^1\mathcal{L}(\gamma(t))dt=\epsilon.
\end{equation}
\end{myDef}

\noindent
In this paper, we use mode connectivity $\textsf{mc}(\theta,\theta')$, defined in \eqref{eq:MD}, to measure the connectivity between two~weights.

Similar to the definition of similarity level, we say that a loss landscape defined by the training data $S_\text{train}$ and the function class $\mathcal{F}$ has a \emph{connectivity level} $\beta$ for a particular random training scheme $\mathcal{T}$ if
\begin{equation}\label{eqn:mode_connectivity}
     \mathbb{E}[\textsf{mc}(\theta,\theta')] = \beta,
\end{equation}
where the expectation is taken over the randomness of the training scheme $\mathcal{T}$, including the random initializations, shuffling of data, and the random training schemes applied to find the curve $\gamma(t)$.

Intuitively speaking, if trained models from random initializations on a loss landscape have negative connectivity level, it means the loss landscape is hard for the purpose of exploration. 
However, a large positive connectivity level may simply suggest that the training loss achieved by the two fixed end points is high. 
Thus, for our evaluation, a ``good'' connectivity level is achieved when $\beta\approx 0$.

{\bf Measuring connectivity.}
To find a low-energy curve between two weights $\theta$ and $\theta'$, we follow the procedures in \citep{garipov2018loss} and use the Bezier curve given by
\begin{equation}
    \gamma_\phi(t) = \sum_{j=0}^{k} \binom{k}{j}(1-t)^{k-j}t^{j}\theta_j, t\in[0,1],
\end{equation}
where $\theta_0 = \theta, \theta_k = \theta'$, and $\phi = \{\theta_1,\dots,\theta_{k-1}\}$ are trainable parameters of additional models, defining ``bends'' on the curve $\gamma_\phi(t)$.
The minimization of the loss on the curve is realized through repeated sampling of $t$ and minimizing the loss with respect to the weights $\gamma(t)$.

For most experiments, we follow \citep{garipov2018loss} and use three bends (including the two fixed end points) to parameterize the trainable curve.
To train the curve, we use SGD with 50 epochs, initial learning rate 0.01, and decay the learning rate to 1\% of the initial value linearly from epoch 25 to epoch 45.
The learning rate schedule follows \citep{garipov2018loss}, and we provide additional ablation results in~Appendix~\ref{sec:MD_lr} to show that the mode connectivity results are not significantly affected by hyperparameter settings.
We evaluate five $t$ values on the curve when calculating the maximum point $t^*$ in \eqref{eq:MD}.

\subsection{Globally nice loss landscapes}

In this subsection, using the similarity level and the connectivity level, we can define what we mean by globally nice loss landscapes.

\begin{myDef}
For a learning problem with the training data $S_\text{train}$, a loss function $\mathcal{L}$, a randomized training scheme $\mathcal{T}$, and a perturbed distribution $Q$ from the ground-truth distribution $P$, we say that the problem has a ($\mu$, $\beta$)-nice loss landscape if the following conditions hold:
\begin{itemize}
    \item The similarity level is equal to $\mu$.
    \item The connectivity level is equal to $\beta$.
\end{itemize}
\end{myDef}

\noindent
If $\mu$ is close to 1 and $\beta$ is larger than 0, we say that the loss landscape is \emph{globally nice}. 
We note that the best connectivity is achieved when the connectivity level is close to 0.

\subsection{Ablation study on different metrics}

In this subsection, we study different configurations when measuring the CKA similarity and the mode connectivity.

\subsubsection{Ablation study on measuring CKA similarity}\label{sec:CKA_ablation}

Here, we provide the details of the perturbed distribution when measuring the CKA similarity.
We study two ways of creating the perturbed distribution. The first one samples each perturbed image from two randomly sampled training images, i.e., $\tilde{x}=\lambda x_1 + (1-\lambda) x_2$, where $\lambda$ follows the distribution Beta($\alpha$,$\alpha$) \citep{zhang2017mixup}. 
In Figure \ref{fig:CKA_mixup}, we reproduce Figure \ref{fig:ResNet18_CKA} with different $\alpha$ values.

The second way of creating the perturbed distribution samples each image by randomly adding uniformly distributed noise to each pixel value of a randomly sampled training image. 
In Figure \ref{fig:CKA_noisy}, we reproduce Figure \ref{fig:ResNet18_CKA} with different magnitude of pixel noise added to the training images.

From the ablation results, we have the following conclusions. 
First, for perturbed distributions generated with pixel noise, i.e., Figure~\ref{fig:CKA_noisy}, if the noise perturbation is too small, the CKA similarity becomes less informative especially for Phase III and Phase IV, i.e., the two phases that achieve an almost exactly zero training loss (see Figure \ref{fig:ResNet18_loss}).
This is exactly the reason we use a perturbed distribution instead of the original training data distribution.
Second, if we choose linear combinations, i.e, Figure~\ref{fig:CKA_mixup}, the CKA measurement becomes more informative and is insensitive to the specific choice of $\alpha$.
We note that Beta($\alpha$,$\alpha$) with a large $\beta$ value concentrates at $1/2$. 
Thus, we effectively create linearly combined samples close to $\tilde{x}=\frac{1}{2} x_1 + \frac{1}{2} x_2$.
In the main paper, we choose $\alpha=16.0$.

\begin{figure}
    \centering
    \includegraphics[width=\textwidth]{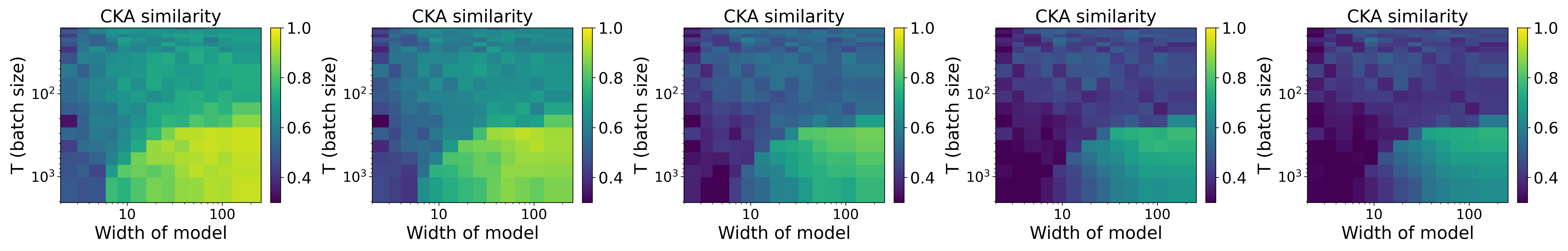}
    \caption{{\bf (Ablation study on CKA).} Ablation study on different Beta($\alpha$,$\alpha$) distributions when plotting the CKA results in Figure \ref{fig:ResNet18_CKA} using linear combinations of samples. From left to right: $\alpha=0.5$, $\alpha=1.0$, $\alpha=4.0$, $\alpha=16.0$, $\alpha=32.0$.}
    \label{fig:CKA_mixup}
\end{figure}

\begin{figure}
    \centering
    \includegraphics[width=\textwidth]{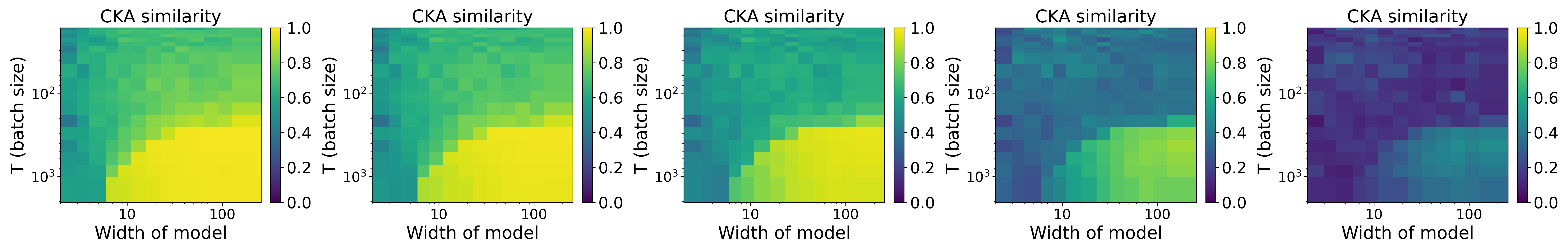}
    \caption{{\bf (Ablation study on CKA).} Ablation study on different magnitude of pixel noise when plotting the CKA results in Figure \ref{fig:ResNet18_CKA}. From left to right: noise magnitude = 5, 10, 20, 40, 80.
    }
    \label{fig:CKA_noisy}
\end{figure}

\subsubsection{Ablation study on measuring mode connectivity}\label{sec:MD_lr}

\begin{figure}
    \centering
    \includegraphics[width=.23\textwidth]{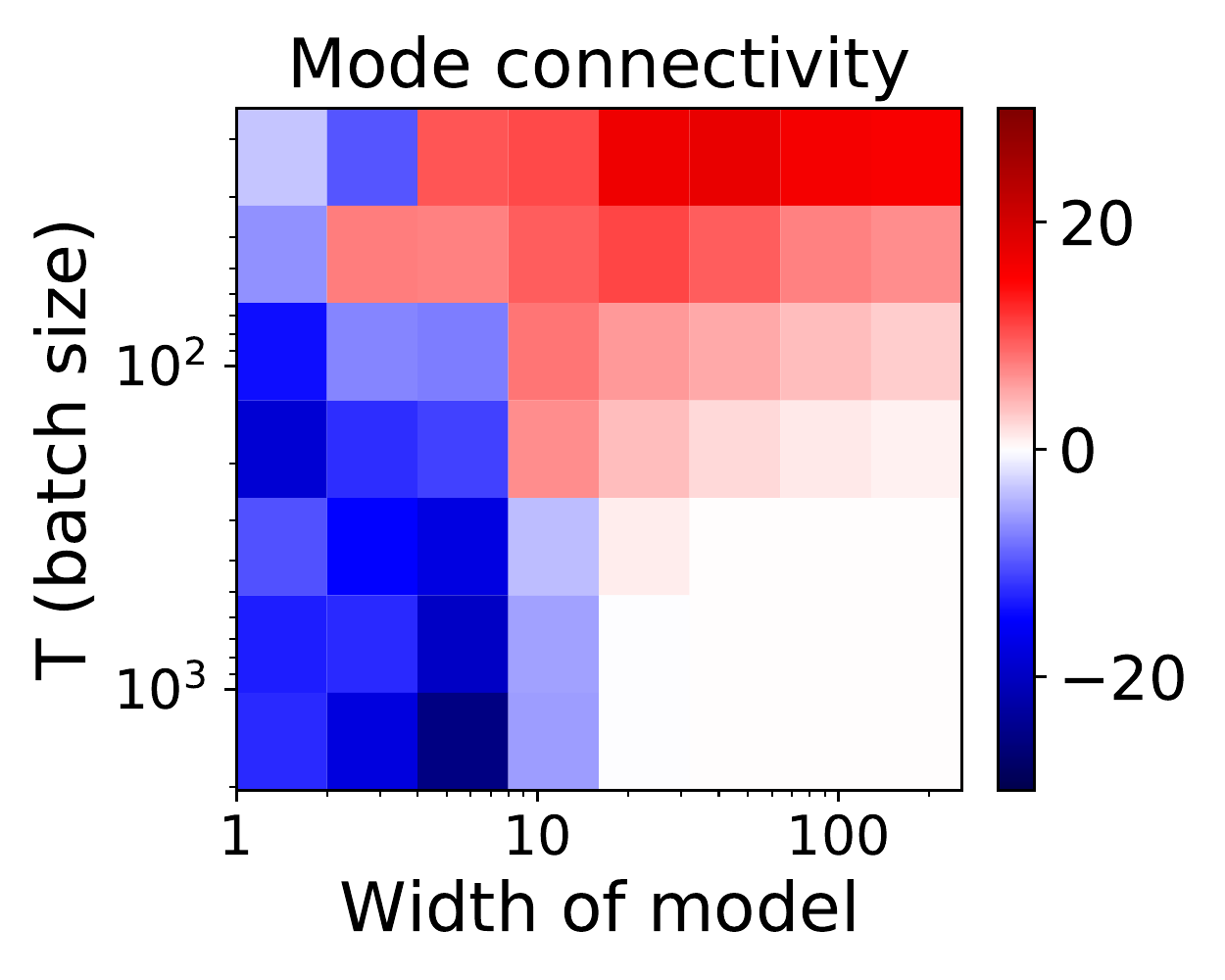}
    \includegraphics[width=.23\textwidth]{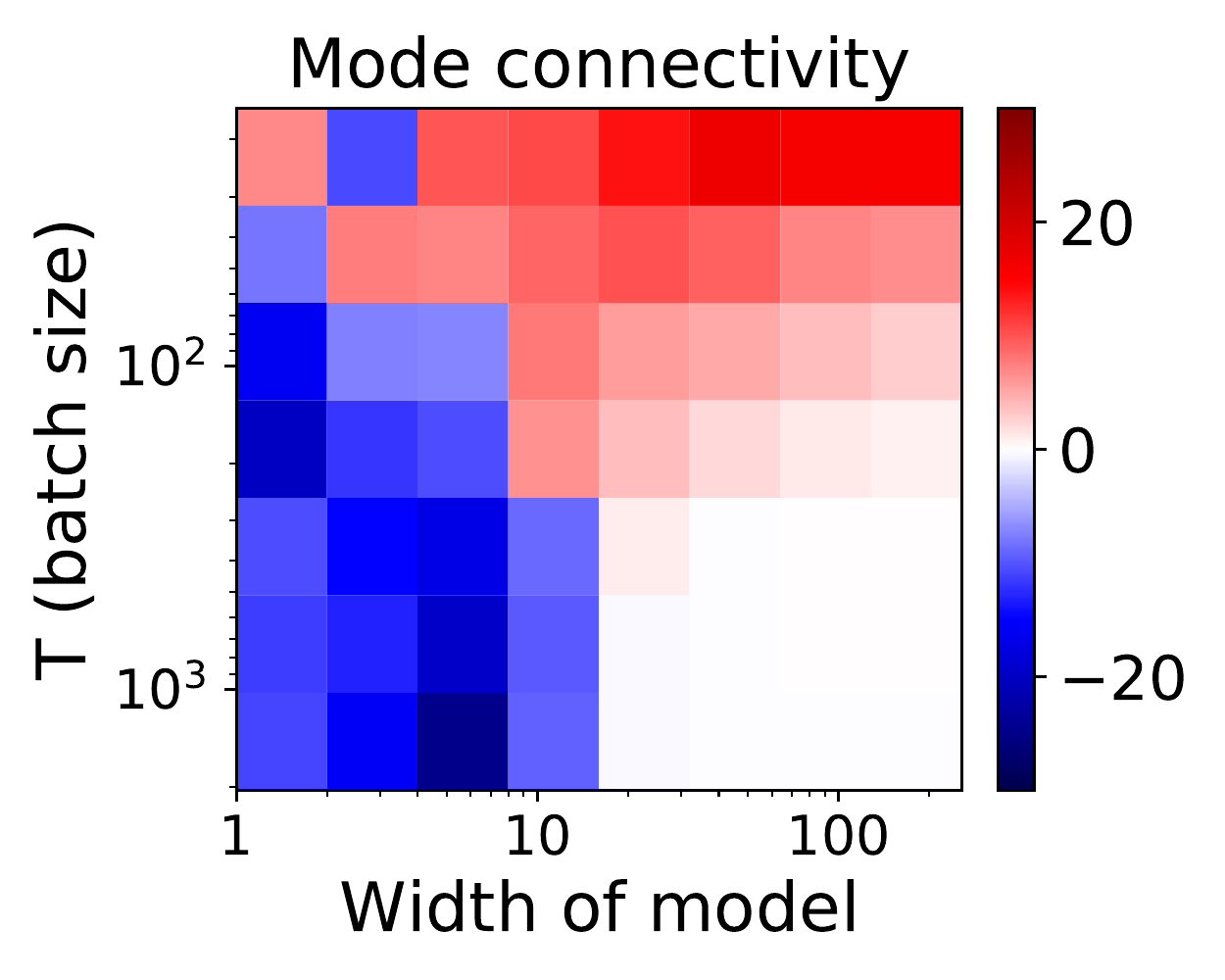}
    \includegraphics[width=.23\textwidth]{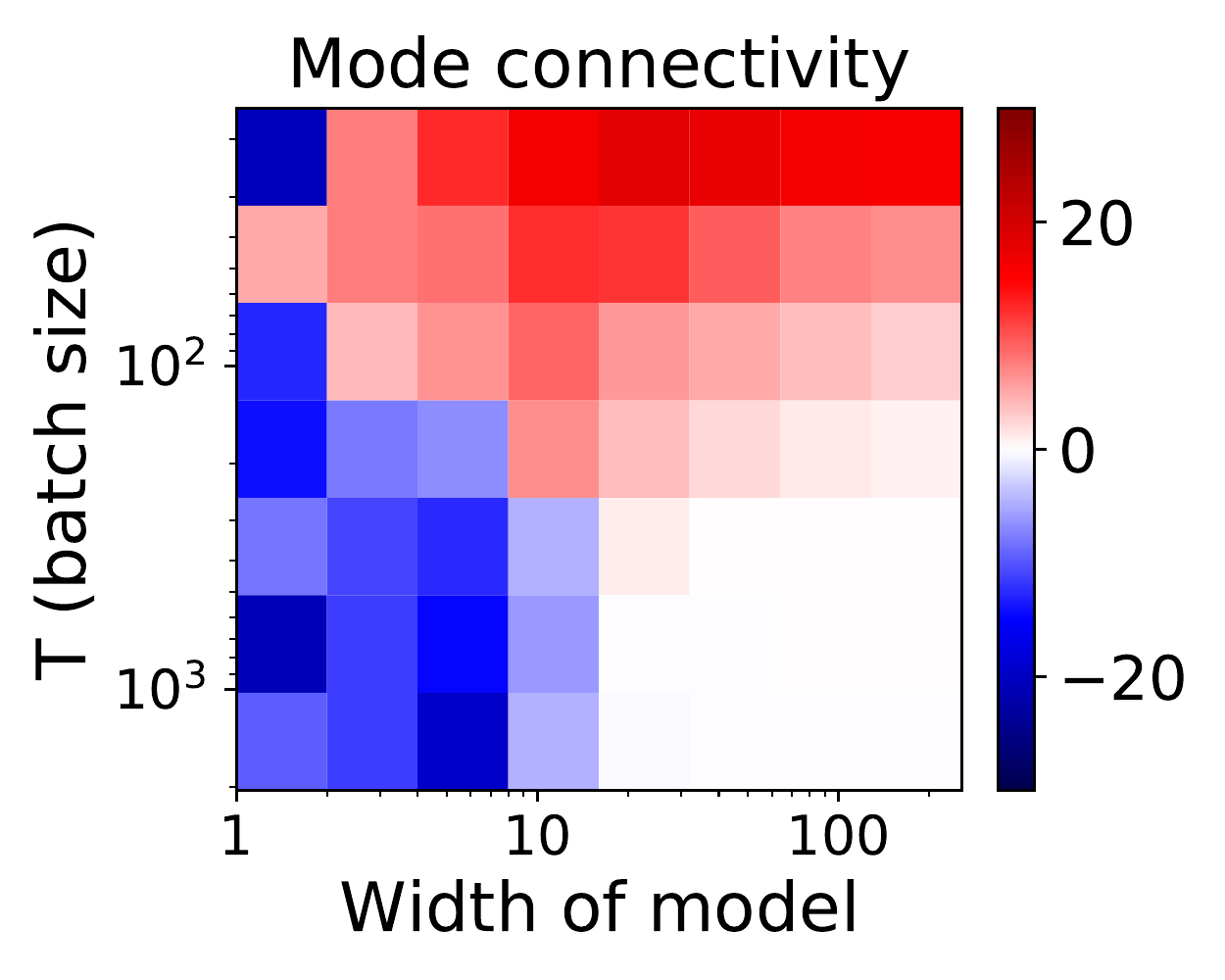}
    \includegraphics[width=.23\textwidth]{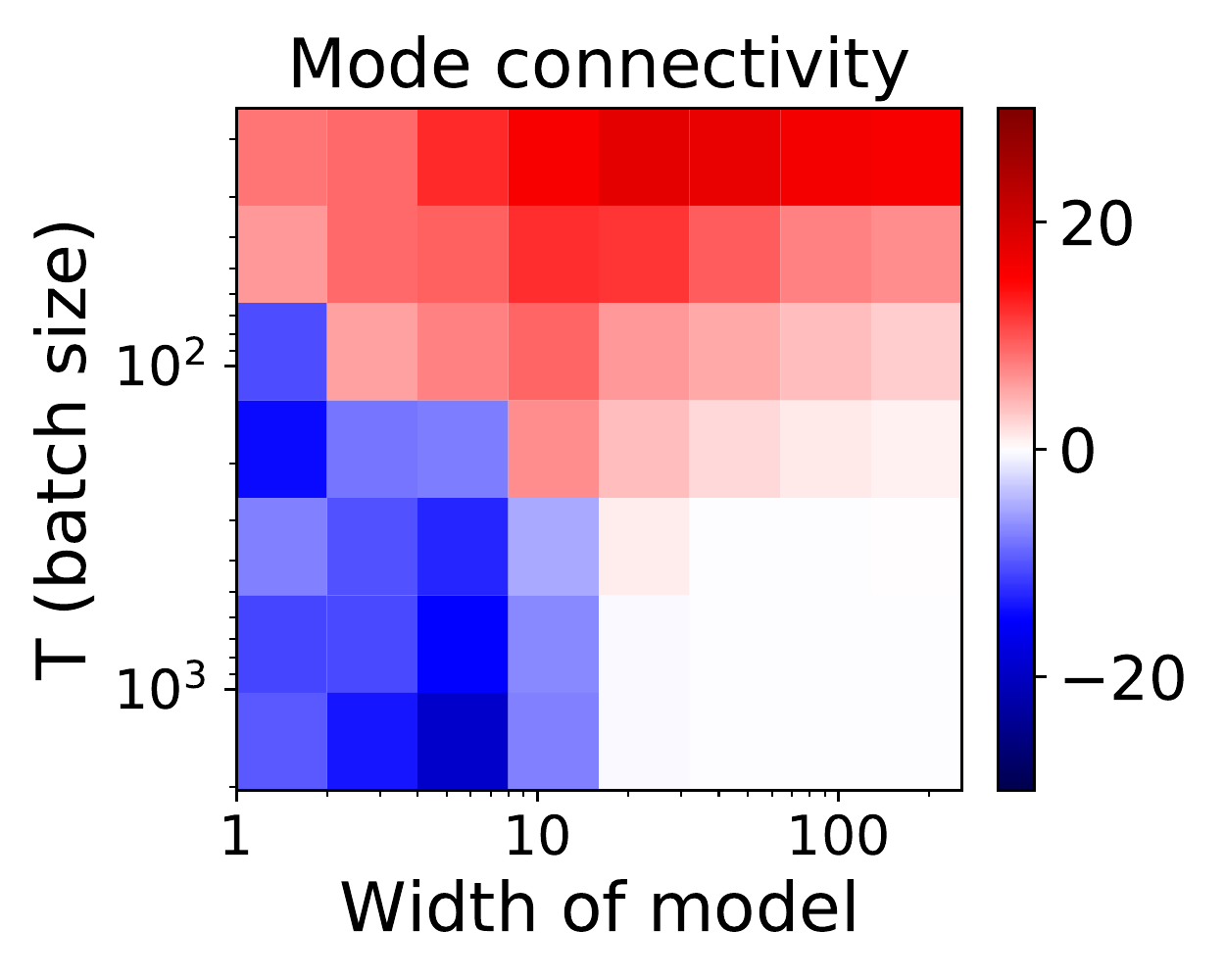}
    \caption{{\bf(Ablation study on mode connectivity).} Ablation study on different ways of measuring mode connectivity.
    From left to right: 
    {\bf i)} Small learning rate 0.003 with three bends;
    {\bf ii)} Medium learning rate 0.01 with three bends;
    {\bf iii)} Large learning rate 0.03 with three bends; and 
    {\bf iv)} Medium learning rate 0.01 with four bends.}
    \label{fig:mode_connectivity_ablation}
\end{figure}

Here, we study the best configuration to produce the mode connectivity plot shown in Figure \ref{fig:ResNet18_curve}.
We test four configurations. In the first three configurations, we use learning rate = 0.003, 0.01, and 0.03, respectively, and we use three bends (including the two end checkpoints) that form a quadratic Bezier curve.
In the fourth configuration, we use learning rate = 0.01, and we use four bends that form a cubic Bezier curve.
See Figure \ref{fig:mode_connectivity_ablation}.
Note that we scale the training time to match the learning rates.
When training with learning rate 0.01, we train for 50 epochs.
When training with learning rate 0.003, we train for 150 epochs.
When training with learning rate 0.03, we train for 40 epochs.
We use learning rate decay in all these experiments.
From Figure \ref{fig:ResNet18_curve}, the mode connectivity results are robust to the choice of specific configurations.
We note that training with a medium learning rate 0.01 and using three bends is the standard setting that we use in all the other~experiments.

\subsection{Visualizing individual low-energy curves in the mode connectivity plots}\label{sec:individual_curves}

In this subsection, we present the low-energy curves calculated from mode connectivity, which we have used to draw the mode connectivity plots such as Figure \ref{fig:ResNet18_curve}. The individual curves are shown in Figure \ref{fig:individual_curves}.
Each subfigure in Figure \ref{fig:individual_curves} represents the low-energy loss curve found using the curve searching algorithm described in Appendix~\ref{sec:mode_connectivity} for a specific (batch size, width) configuration.
The loss curve represents the loss evaluated at the interpolation points between two trained models for the specific (batch size, width) configuration. 
The two trained models are fixed during the curve searching process.
Then, we summarize each loss curve using the $\textsf{mc}$ value defined in \eqref{eq:MD}. 
When the low-energy curve is convex, e.g., in the top-right corner of the figure, $\textsf{mc}$ is positive. 
This shape of curve implies that the two end points (which are two same models trained with different random initialization) are not able to find a good local minima with a large temperature. 
When the low-energy curve has a high ``barrier'' at the middle point, e.g., in the left three columns of the figure, $\textsf{mc}$ is negative. 
This shape of curve implies that the loss landscape is poorly-connected, because it is hard to find a low-energy path between two trained models with different initializations.
When the entire curve remains at 0, e.g., in the bottom-right corner of the figure, we get zero $\textsf{mc}$ value, which corresponds to the white regions in the bottom-right corner of Figure \ref{fig:ResNet18_curve}.
This shape of curve implies that the loss landscape is well-connected.

\begin{figure}
    \centering
    \includegraphics[width=\textwidth]{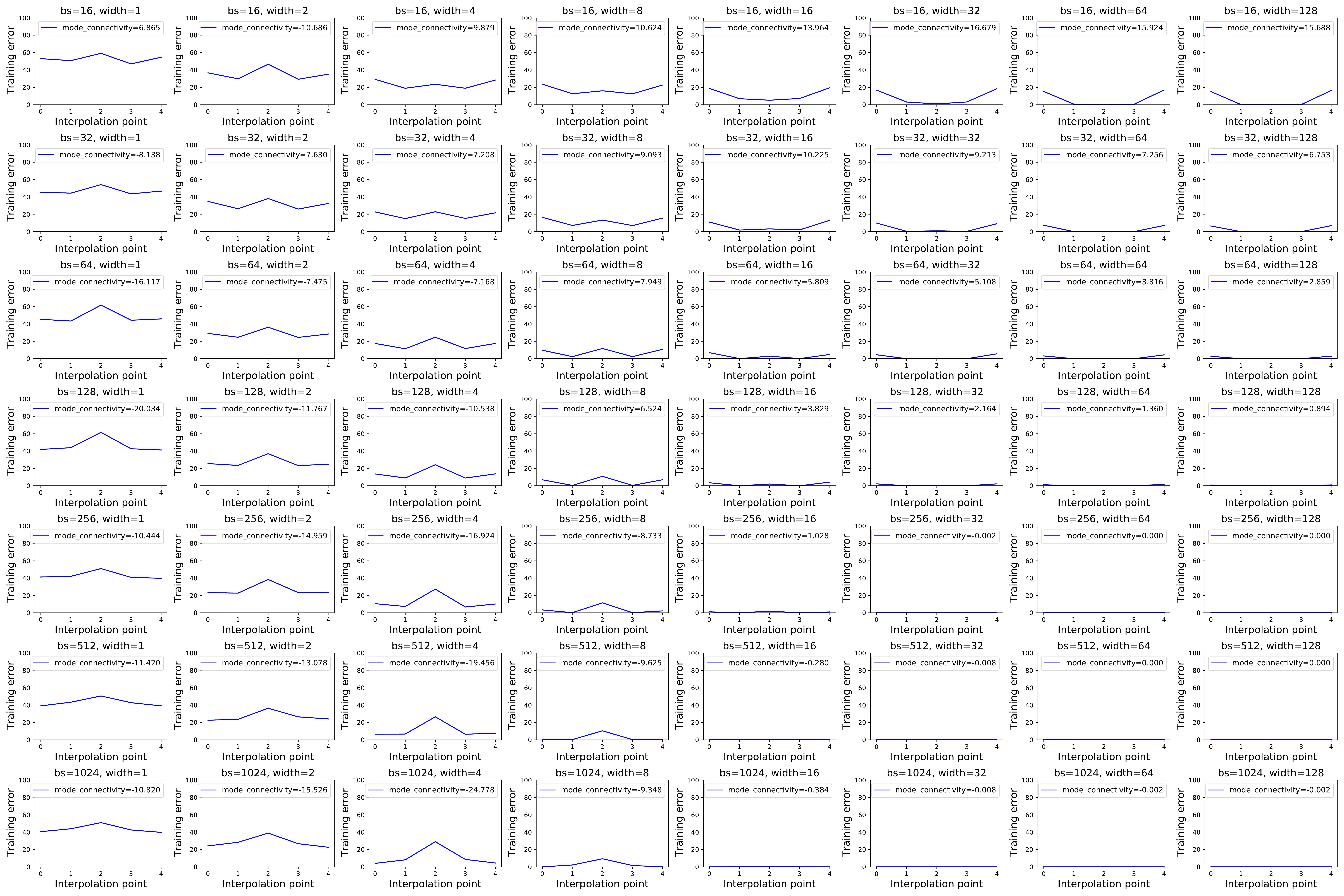}
    \caption{{\bf (Individual mode connectivity curves). } Plotting the individual low-energy curves found by mode connectivity, which are used to draw Figure \ref{fig:ResNet18_curve}.
    Subplots on the left three columns show barriers in the middle, which imply poorly-connected loss landscapes. 
    Subplots on the top-right corner show convex-like curves, which suggest the two models on the two ends are not sufficiently well-trained.
    Subplots on the bottom-right corner show flat curves with error close to 0, which suggest reasonably well-connected loss landscapes and reasonably well-trained models.
    }
    \label{fig:individual_curves}
\end{figure}

\section{Implementation details}
\label{sec:implementation}

In this section, we provide some implementation details for our empirical evaluations.

\subsection{Datasets}

For most of our experiments, we use CIFAR-10 \citep{krizhevsky2009learning}. 
We provide additional results on SVHN \citep{netzer2011reading} and CIFAR-100 in Appendix~\ref{sec:additional_dataset}.
For natural language processing, we use IWSLT16 German to English (De-En), which is a common machine translation dataset \citep{cettolo2012wit3}.
The results on IWSLT16 De-En are given in Appendix~\ref{sec:NLP}.
Following \citep{nakkiran2019deep}, we randomly sample 4K sentence pairs in our training.
In the experiments on training with random labels, e.g., Figure~\ref{fig:Noisy_label} and~\ref{fig:Scaling_noisy_label}, we randomly pick a certain percentage of training samples and change their labels to random target different from the original label.

\subsection{Architectures}

We use ResNets \citep{he2016deep} in the standard setting.
We scale the network width to change the size of the network.
For ResNet18 which contains four major blocks with channel width $\{k, 2k, 4k, 8k\}$, we select different values of $k$ to obtain ResNets with different widths. 
Similarly, we use VGG11 which contains blocks with channel width $\{k, 2k, 4k, 4k, 8k, 8k, 8k, 8k\}$, and we vary $k$.
See Appendix~\ref{sec:additional_architecture}.
For Transformers, following \citep{vaswani2017attention,nakkiran2019deep}, we use a six-layer Transformer with eight attention heads, and we vary the embedding dimension to change the model width. 
The experiments on Transformers are reported in Appendix~\ref{sec:NLP}.

\subsection{Training procedures}

In the standard setting, we purposely use constant batch size, learning rate, and weight decay throughout the training, to study the interactions between temperature-like parameters, load-like parameters, and the loss landscape.
We also provide results when training with learning rate decay in Figure \ref{fig:Lr_decay}.
Note that this also means that, when training with different batch sizes, we do not change learning rate accordingly \footnote{Tuning more than one hyperparameter confounds load and/or temperature. A primary purpose of this paper is to show how load and temperature-like parameters affect the loss landscape by putting them in a 2D grid. By keeping other hyperparameters constant, moving along one of the axes correlates directly to a change in load or temperature. However, if we tune, for example, learning rate and batch size altogether, we will not know how the load or temperature changes.}, which is different from the commonly used ``linear scaling rule'' \citep{goyal2017accurate}.
We provide additional results on tuning batch size with the linear scaling rule in Appendix~\ref{sec:LR_scaling}.

For training on CIFAR-10, following \citep{martin2018implicit_JMLRversion}, we use SGD and stop the training if the change of training loss is smaller than 0.0001 for 5 consecutive epochs.
If this cannot be satisfied, we train for 150 epochs to ensure that the training enter the steady stage, and we save the model with the best training loss. 
For the standard setting, we train with constant learning rate 0.05, batch size 128, and weight decay 5e-4.
We note that training with learning rate decay can potentially increase the test accuracy on CIFAR-10, and we provide the results with learning rate decay in Figure~\ref{fig:Lr_decay}.
For training on IWSLT16, following \citep{nakkiran2019deep}, we use Adam and train for 80K gradient updates, with 4K steps of linear warmup, 10\% label smoothing, and no drop-out. 
We repeat each experiment for five individual runs with random initialization and average the results.
We observe that using data augmentation (such as random flipping and cropping) makes training with noisy labels harder to converge. Thus, again to avoid confounding factors, we do not apply data augmentation in this paper.
However, it should be noted that data augmentation can improve test accuracy if it is used properly.

\subsection{Hyperparameters for different metrics}

To compute Hessian information,
we use the PyHessian software \citep{yao2020pyhessian} to measure the Hessian trace and leading eigenvalues. 
We find that using one batch of 200 random samples can already give stable results, and so we use that in all of our experiments.
PyHessian uses the power iteration method to measure the leading eigenvalues, and it uses the Hutchinson's method to measure the Hessian trace.
The maximum number of iterations used in these methods is set to be 100, while a relative tolerance level of 1e-3 is used to early stop the computation.

To compute CKA and mode connectivity, see Appendix~\ref{sec:definitions_CKA} and~\ref{sec:mode_connectivity}, respectively.

\subsection{Computing infrastructure} 

All experiments use NVIDIA GPU servers as computing nodes. Each server contains 8 Tesla V100 GPUs. The experiments are implemented in PyTorch. Each test accuracy plot in the main paper requires several days on one server to reproduce. The exact time depends on the granularity of the plot and the data/model~configurations.

\section{Learning rate decay helps the most when the loss landscape is close to being well-connected}\label{sec:lr_decay_helps}

In this section, we continue to study the improvement on test accuracy provided by learning rate decay.
If we find the optimal test accuracy for each width value in Figure \ref{fig:Lr_decay_accuracy} (i.e., finding the optimal test accuracy in each column slice in Figure \ref{fig:Lr_decay_accuracy}), and we compare it with the optimal test accuracy for each width value in Figure \ref{fig:ResNet18_accuracy}, we obtain the accuracy improvement purely from using learning rate decay.
We plot the improved accuracy shown in Figure \ref{fig:improved_accuracy}.
Observe that the peak of the improved accuracy happens almost exactly at the transition between globally well-connected and poorly-connected phases. 
We conjecture the following: 
i) this means that training with a large temperature at the beginning to \emph{explore} the loss landscape helps when the loss landscape is poorly-connected; and also
ii) this means that the exploration can help the most when the loss landscape is close to being well-connected. 
When the loss landscape is very well-connected, e.g., when the width is large, the improvement reduces.

\begin{figure}
    \centering
    \includegraphics[width=.40\textwidth]{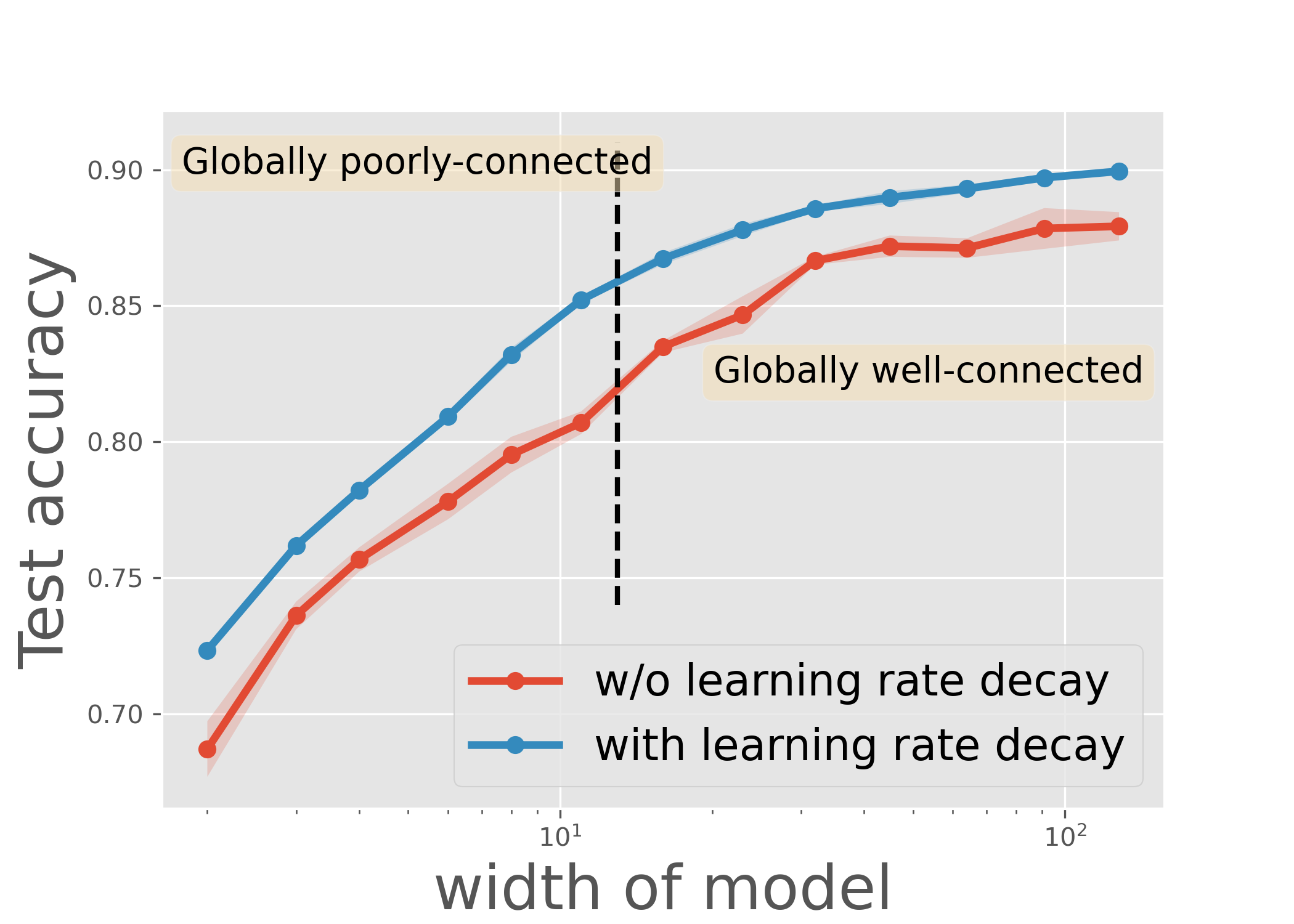}
    \quad
    \includegraphics[width=.40\textwidth]{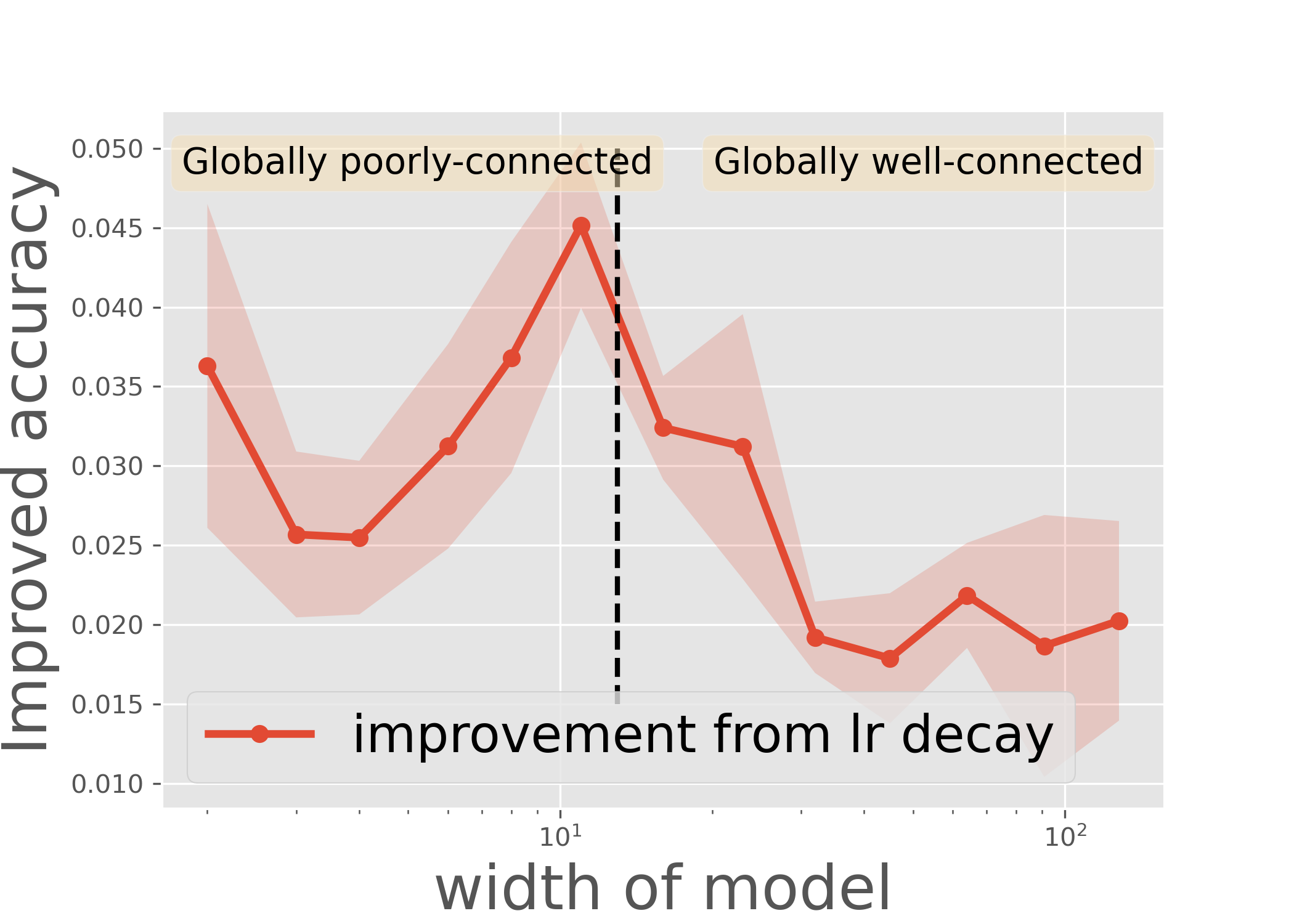}
    \caption{{\bf (Improvement from learning rate decay).} {\bf Left.} Best test accuracy with or without learning rate decay. {\bf Right.} Improved accuracy due to learning rate decay. Transition between globally well-connected and poorly-connected regions coincides with the peak of improved accuracy. }
    \label{fig:improved_accuracy}
\end{figure}

\section{Additional results}
\label{sec:additional_results}

In this section, we provide additional empirical results supporting those presented in the main paper.

\subsection{Additional datasets}\label{sec:additional_dataset}

In this subsection, we provide results on additional datasets.

First, we show the results on the SVHN dataset. 
See Figure~\ref{fig:SVHN}, which shows similar plots but to a lower resolution.
Again, we have the same conclusion that the best test accuracy is achieved when mode connectivity is close to zero, CKA similarity is large, and Hessian eigenvalue and trace are small.
Interestingly, for SVHN the smallest Hessian trace is achieved when the batch size is small. In this case, the test accuracy is not optimal. 
Thus, one has to use mode connectivity and CKA to find the optimal test accuracy.
Next, we show the results on CIFAR-100. 
See Figure~\ref{fig:Cifar100}. 
The results are quite similar to Figure \ref{fig:ResNet18}.

\def \figname {SVHN}
\begin{figure}[h]
  \centering
    \begin{subfigure}{0.23\textwidth}
      \includegraphics[width=\textwidth]{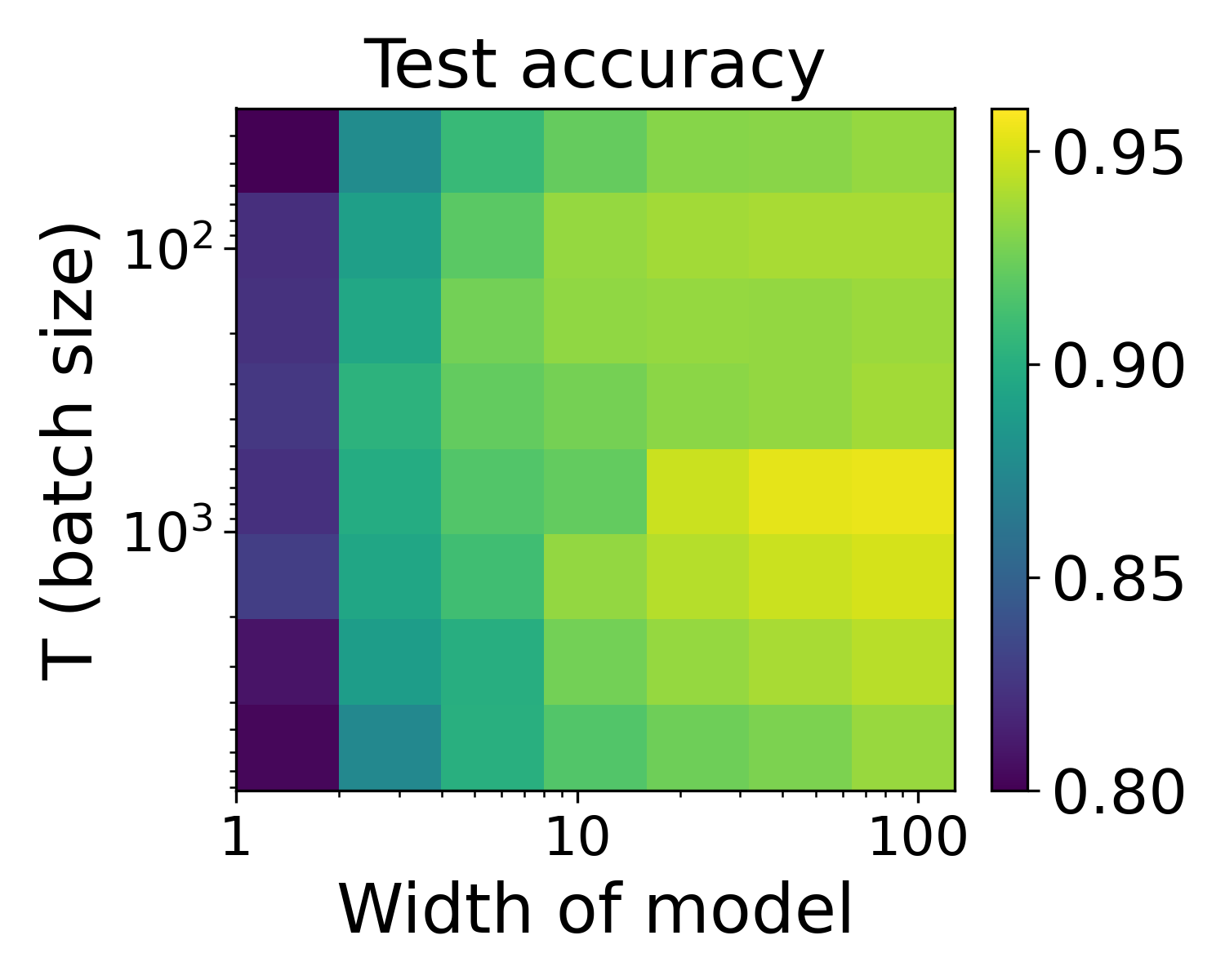}
      \caption{Test accuracy\label{fig:\figname_accuracy}}
    \end{subfigure}
    \begin{subfigure}[c]{0.23\textwidth}
      \includegraphics[width=\textwidth]{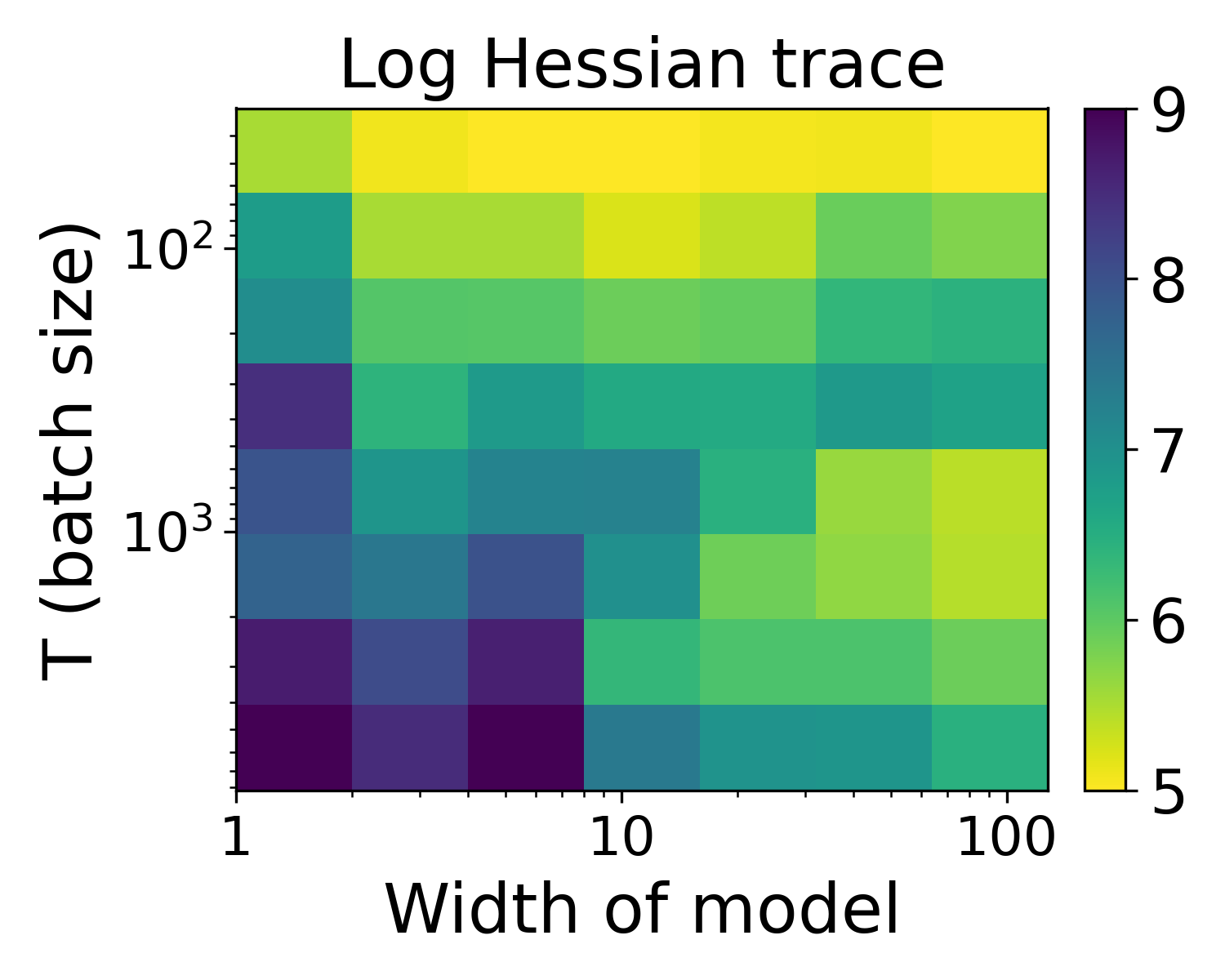}
      \caption{Hessian trace\label{fig:\figname_hessian_t}}
    \end{subfigure}
  \begin{subfigure}[c]{0.23\textwidth}
      \includegraphics[width=\textwidth]{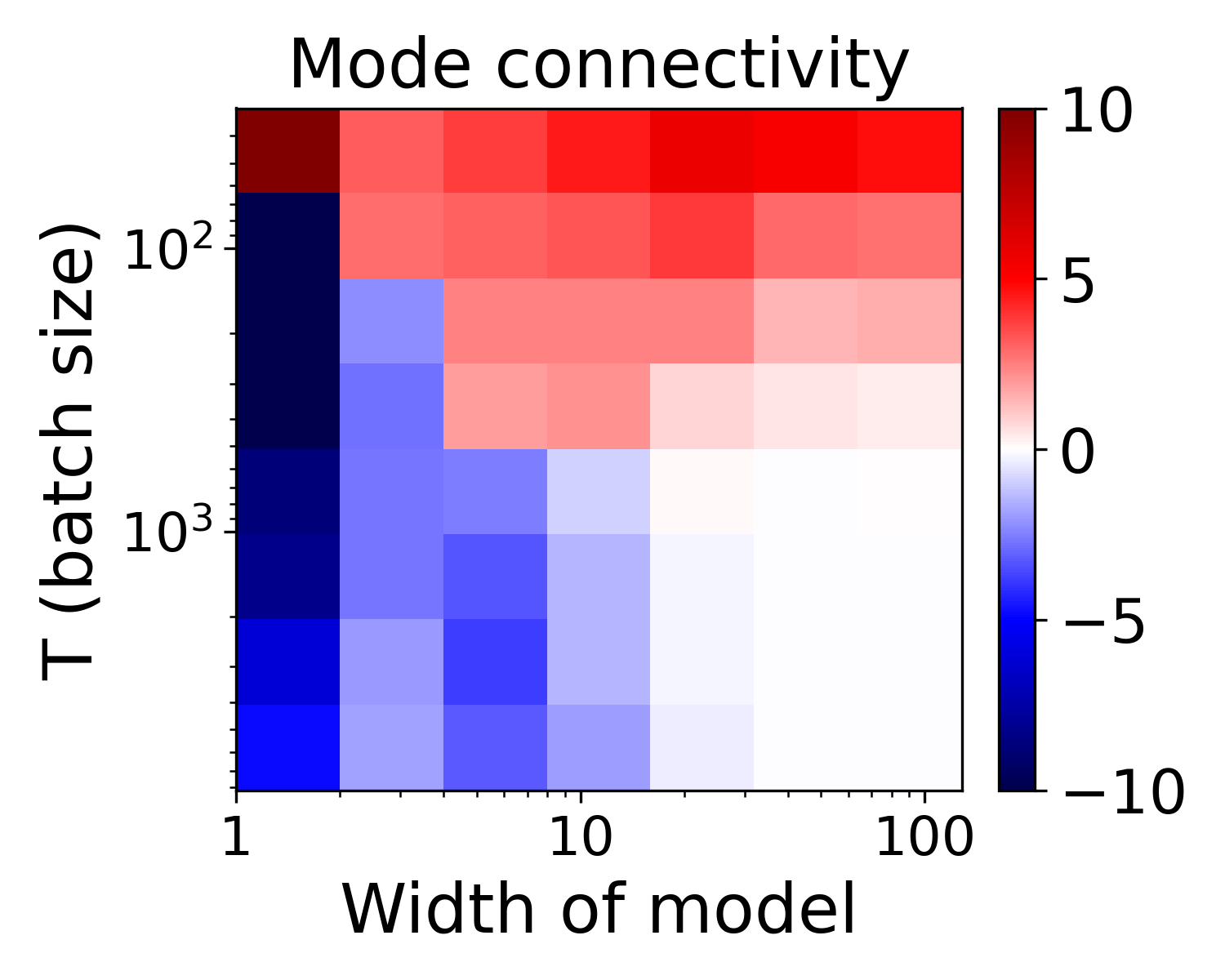}
      \caption{Mode connectivity\label{fig:\figname_curve}}
    \end{subfigure}
    \begin{subfigure}[c]{0.23\textwidth}
      \includegraphics[width=\textwidth]{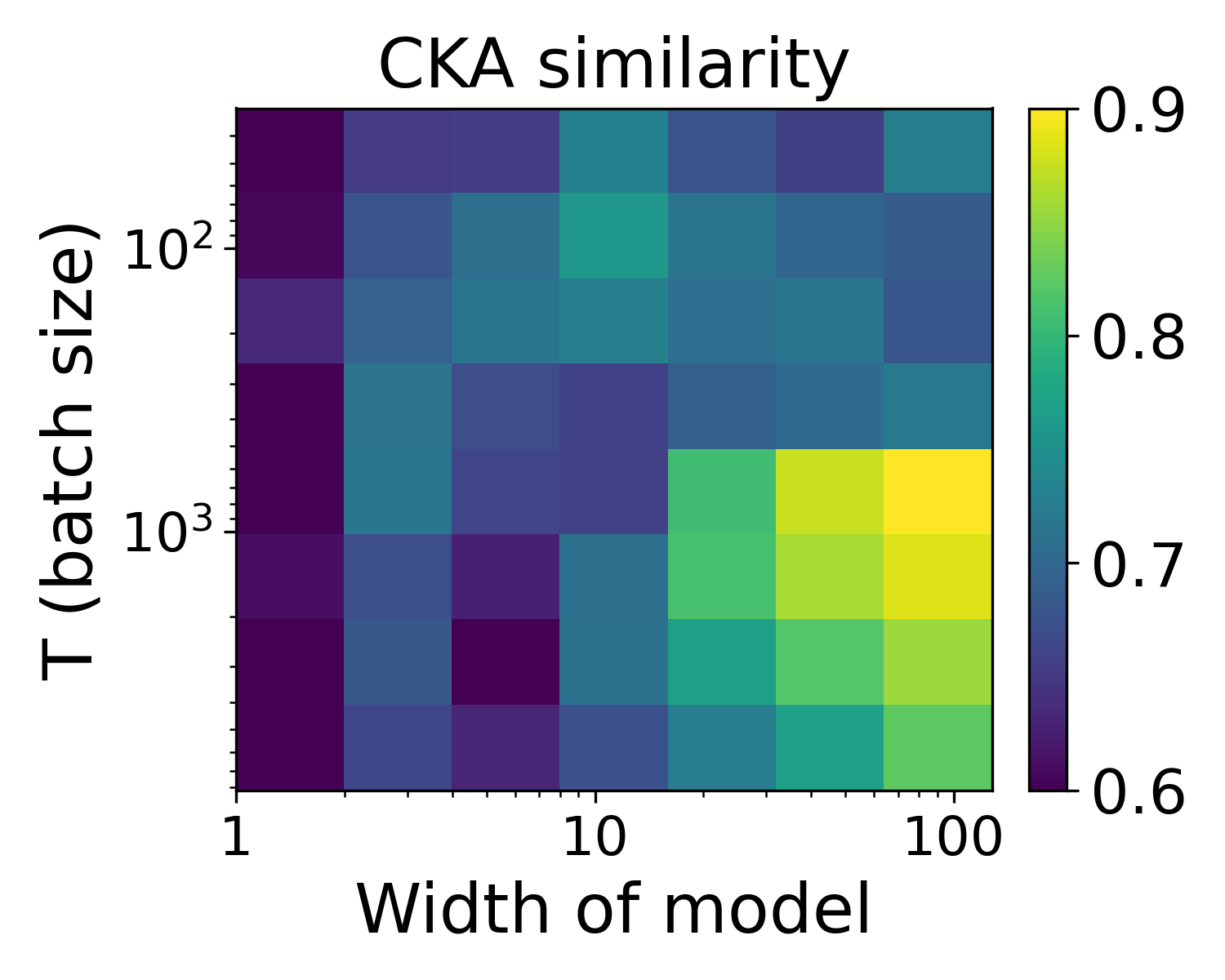}
      \caption{CKA similarity\label{fig:\figname_CKA}}
    \end{subfigure}
  \caption{{\bf (SVHN).} Partitioning the 2D load-like---temperature-like diagram into different phases of learning, using batch size as the temperature and varying model width to change load.  Models are trained with ResNet18 on SVHN. All plots are on the same set of axes. }
  \label{fig:SVHN}
\end{figure}

\def \figname {cifar100}
\begin{figure}[h]
  \centering
    \begin{subfigure}{0.23\textwidth}
      \includegraphics[width=\textwidth]{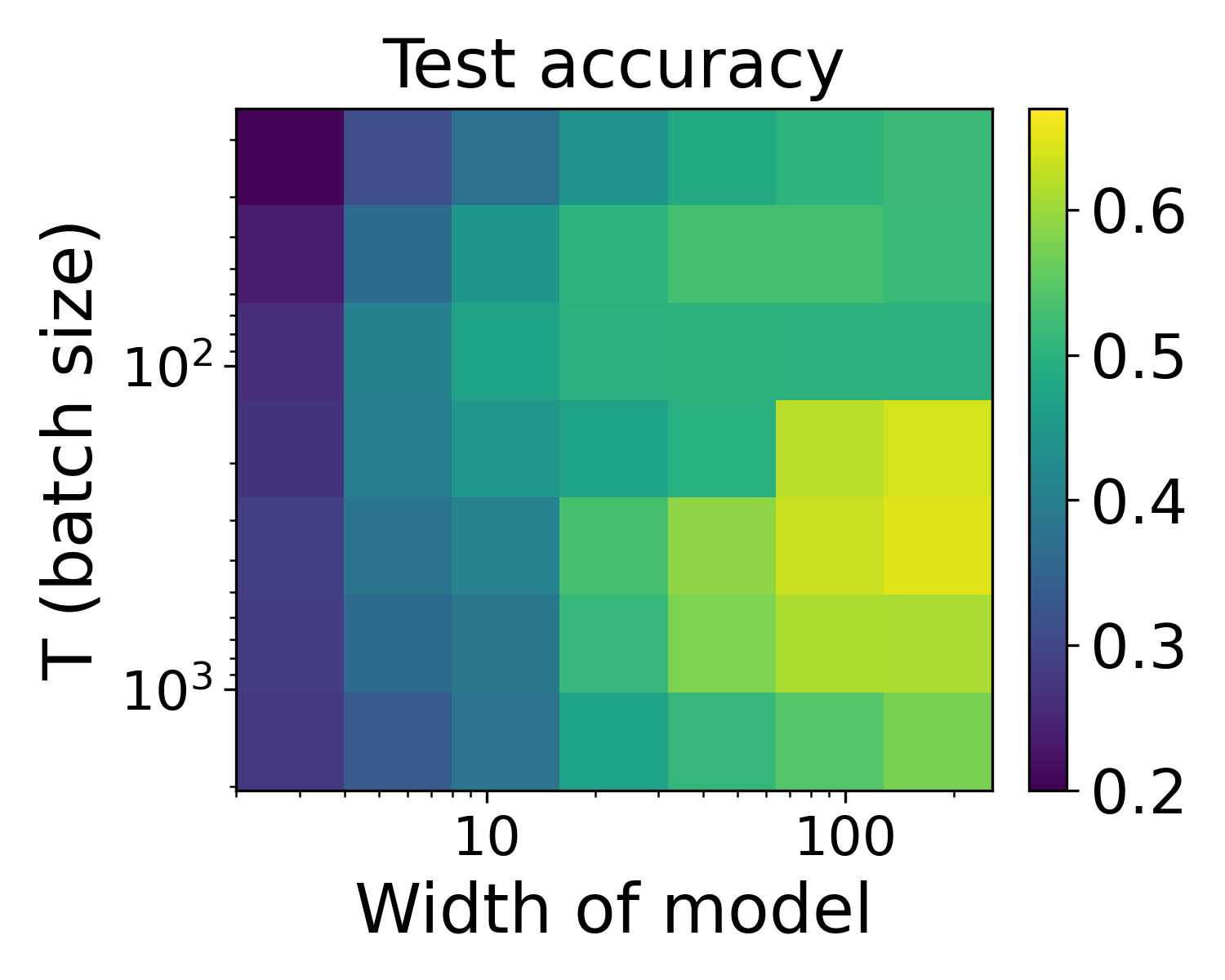}
      \caption{Test accuracy\label{fig:\figname_accuracy}}
    \end{subfigure}
    \begin{subfigure}[c]{0.23\textwidth}
      \includegraphics[width=\textwidth]{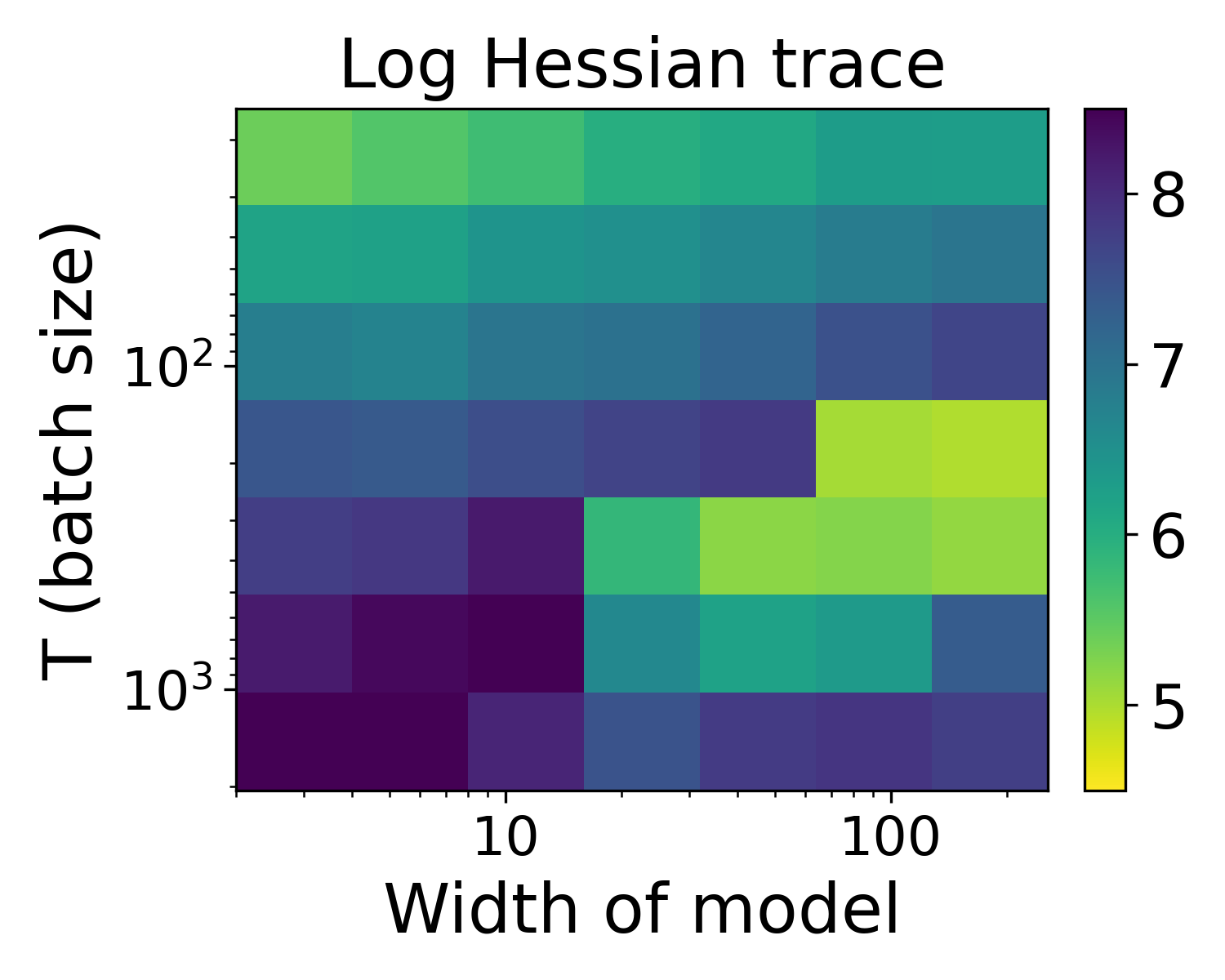}
      \caption{Hessian trace\label{fig:\figname_hessian_t}}
    \end{subfigure}
  \begin{subfigure}[c]{0.23\textwidth}
      \includegraphics[width=\textwidth]{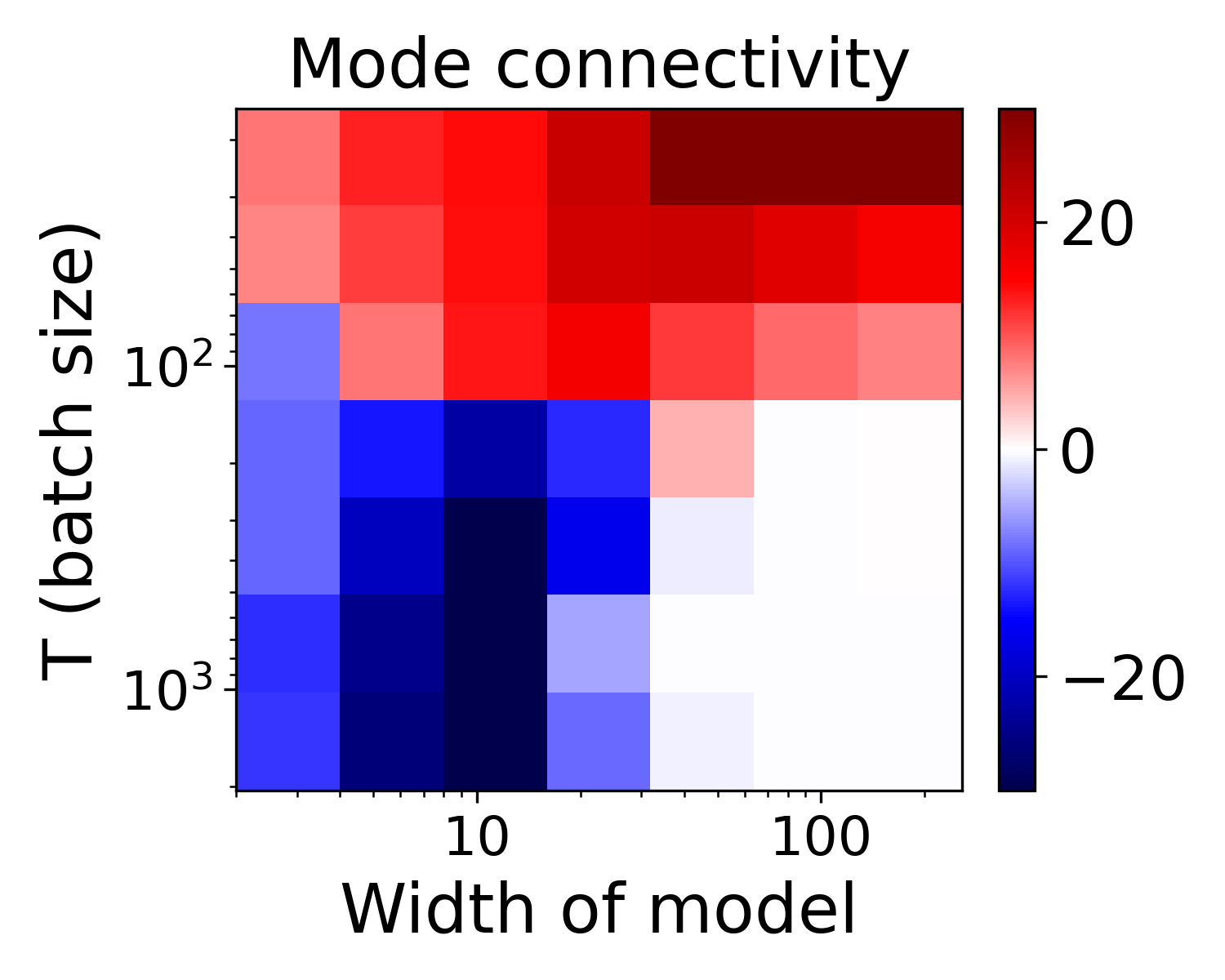}
      \caption{Mode connectivity\label{fig:\figname_curve}}
    \end{subfigure}
    \begin{subfigure}[c]{0.23\textwidth}
      \includegraphics[width=\textwidth]{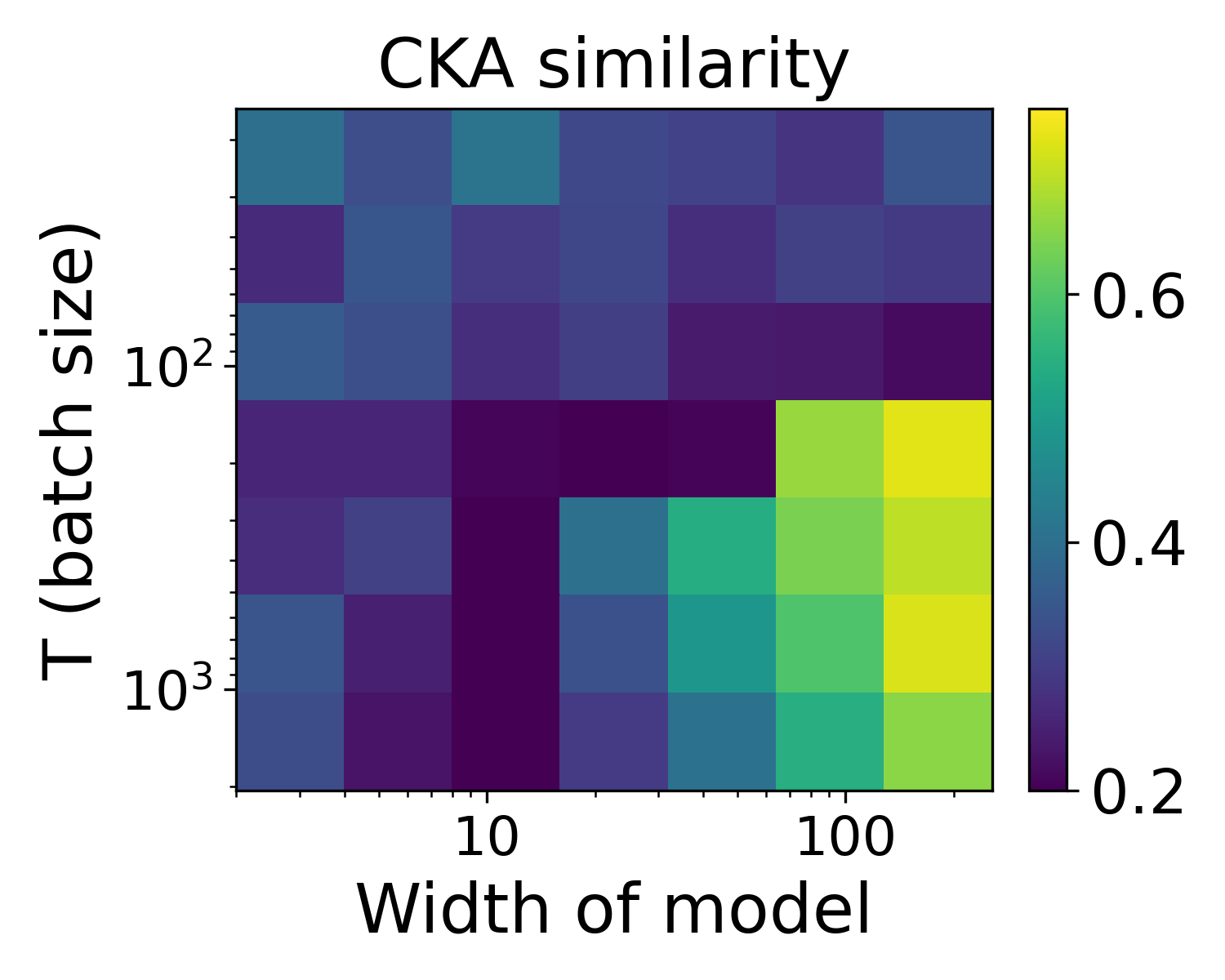}
      \caption{CKA similarity\label{fig:\figname_CKA}}
    \end{subfigure}
  \caption{{\bf (CIFAR-100).} Partitioning the 2D load-like---temperature-like diagram into different phases of learning, using batch size as the temperature and varying model width to change load.  Models are trained with ResNet18 on CIFAR-100. All plots are on the same set of axes. }
  \label{fig:Cifar100}
\end{figure}

\subsection{Additional network architectures}\label{sec:additional_architecture}

In this subsection, we show the results on the VGG networks in addition to ResNets studied in the main paper. See the results in Figure~\ref{fig:VGG}.

\def \figname {VGG}
\begin{figure}[h]
  \centering
    \begin{subfigure}{0.23\textwidth}
      \includegraphics[width=\textwidth]{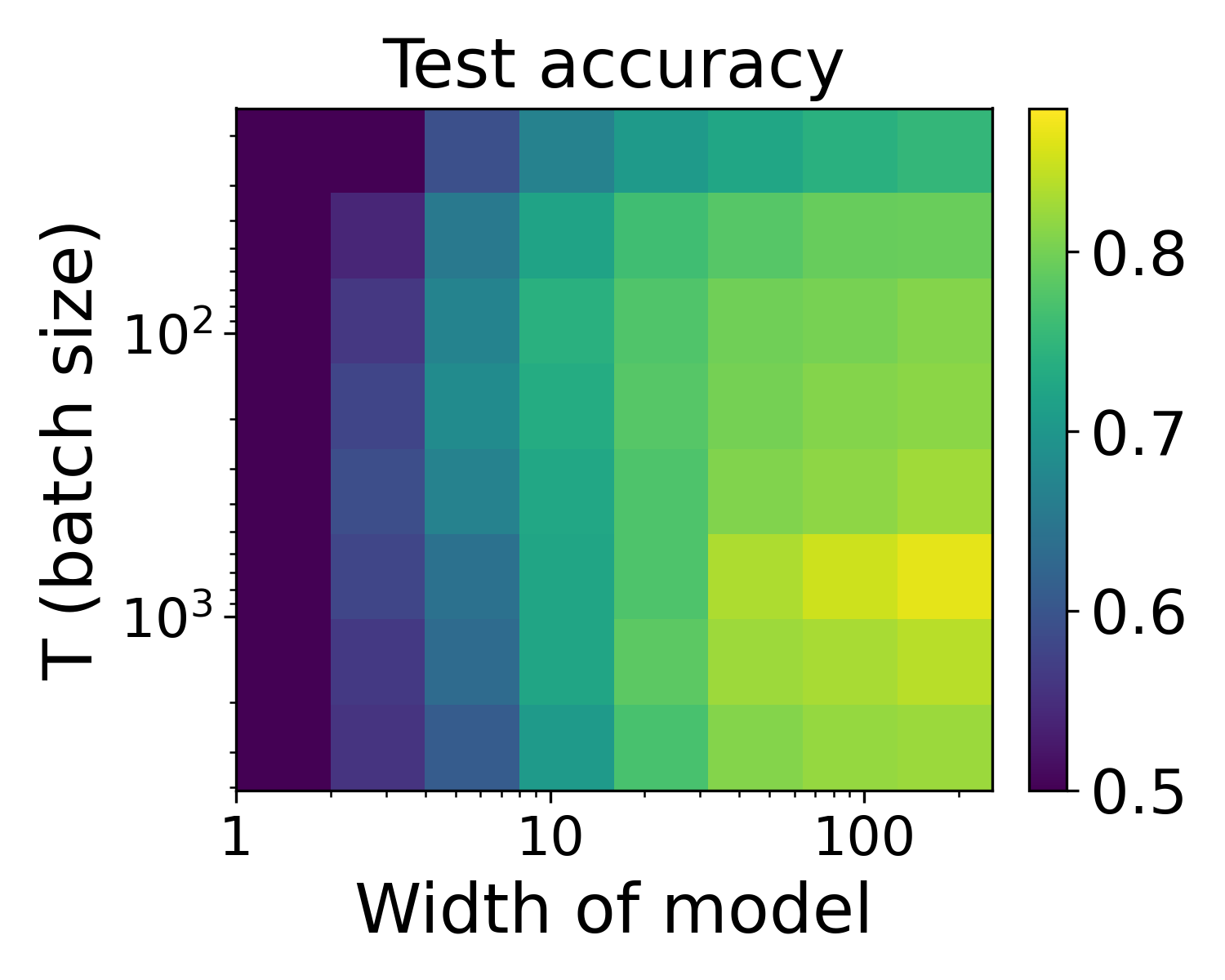}
      \caption{Test accuracy\label{fig:\figname_accuracy}}
    \end{subfigure}
    \begin{subfigure}[c]{0.23\textwidth}
      \includegraphics[width=\textwidth]{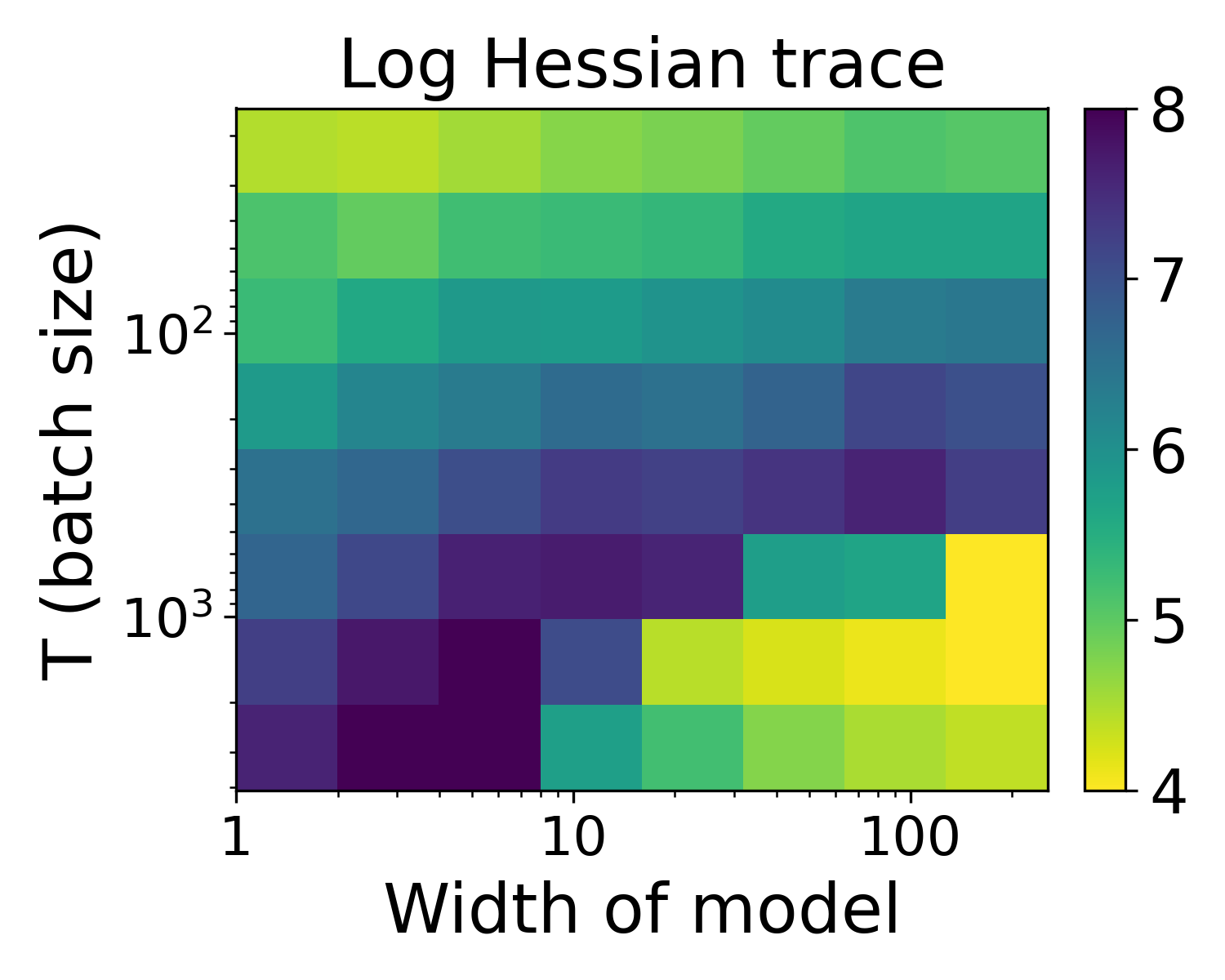}
      \caption{Hessian trace\label{fig:\figname_hessian_t}}
    \end{subfigure}
  \begin{subfigure}[c]{0.23\textwidth}
      \includegraphics[width=\textwidth]{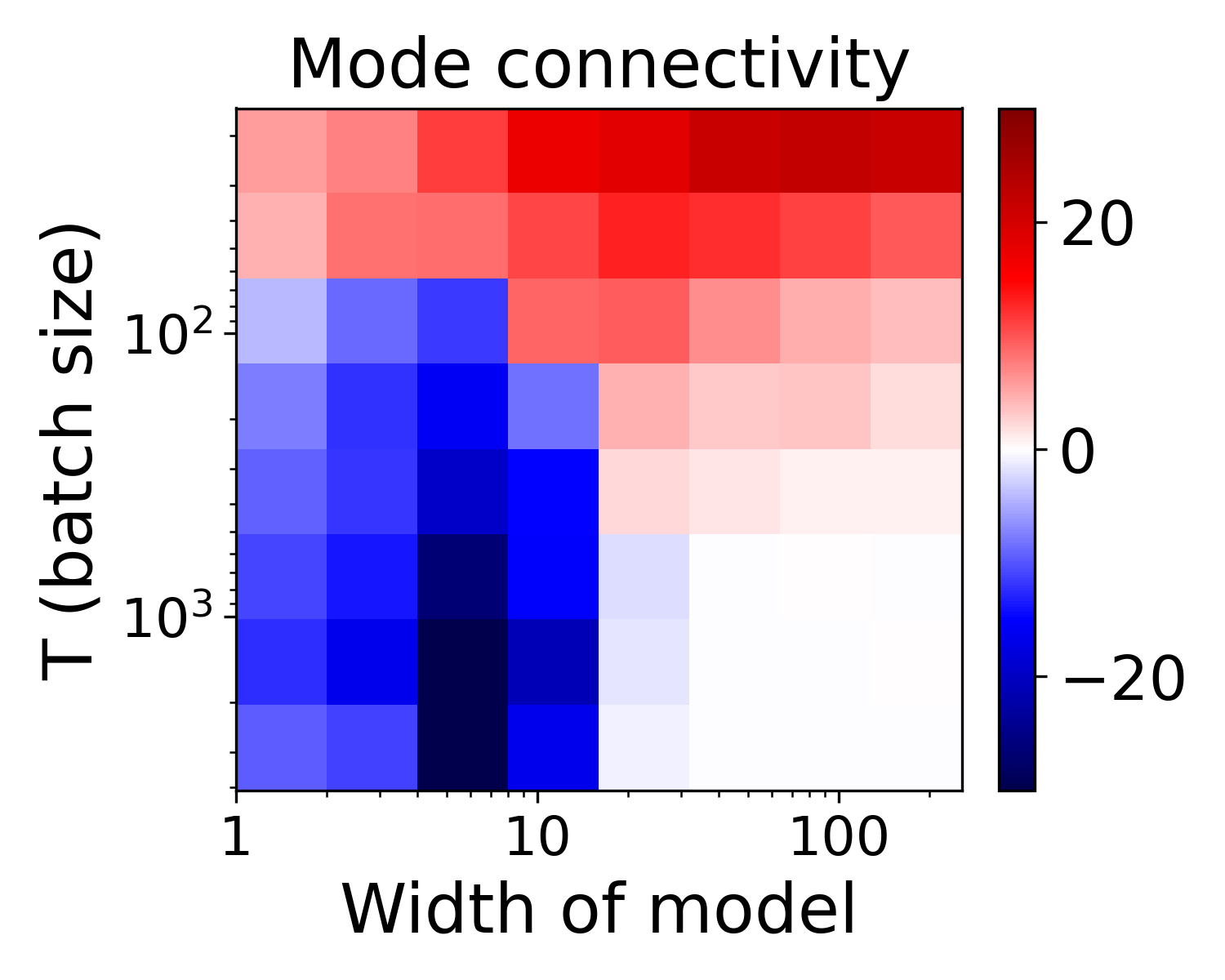}
      \caption{Mode connectivity\label{fig:\figname_curve}}
    \end{subfigure}
    \begin{subfigure}[c]{0.23\textwidth}
      \includegraphics[width=\textwidth]{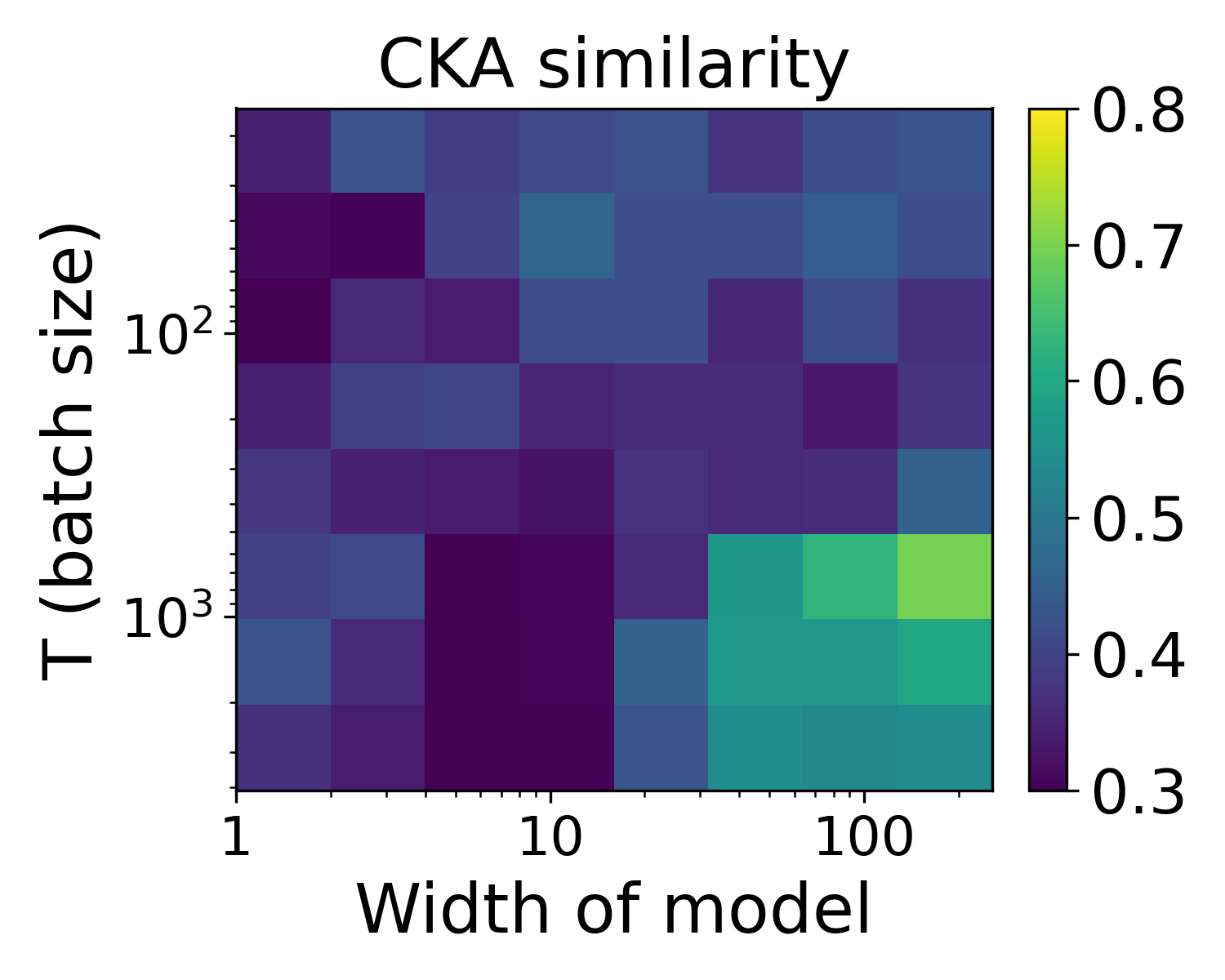}
      \caption{CKA similarity\label{fig:\figname_CKA}}
    \end{subfigure}
  \caption{{\bf (VGG11).} Partitioning the 2D load-like---temperature-like diagram into different phases of learning, using batch size as the temperature and varying model width to change load.  Models are trained with VGG11 on CIFAR-10. All plots are on the same set of axes.}
  \label{fig:VGG}
\end{figure}

\subsection{Additional results on double descent and noisy labels}\label{sec:double_descent}

In this subsection, we provide an additional experiment on training with noisy labels and the double descent phenomenon. 
In Figure~\ref{fig:Noisy_label_lr}, we show an analogous result to Figure \ref{fig:Noisy_label} but with learning rate as the temperature. 
The results in Figure~\ref{fig:Noisy_label_lr} are almost identical to Figure~\ref{fig:Noisy_label}. 
In particular, the optimal test accuracy for a column slice on the left part could be achieved in Phase I instead of Phase III.

From the results shown in Figure~\ref{fig:Noisy_label_lr} and Figure~\ref{fig:Noisy_label}, an operational way to decide whether we should train to zero loss is by using the mode connectivity.
From the mode connectivity plot in Figure \ref{fig:Noisy_label_curve}, for a specific width value, we first train with low temperature and see if the mode connectivity $\textsf{mc}$ (defined in Eqn.\eqref{eq:MD}) is close to zero, i.e., if it falls into the bottom-right white region. 
If $\textsf{mc}$ is indeed close to zero (when trained with low temperature), we can safely train to zero training loss. 
However, if $\textsf{mc}$ has a large negative value, it means that the loss landscape is still relatively poorly-connected, and training to zero loss may harm test accuracy.
In this case, one should try to explore the loss landscape to find the best minima before reducing the training loss to zero.
Note that an incorrect interpretation of this observation is to say that we should always passively avoid reducing the training loss to zero as long as the mode connectivity $\textsf{mc}$ is negative (which means that the loss landscape is poorly-connected).
This is because one can either increase the model width to improve mode connectivity or design exploration schemes before training to zero loss.

\def \figname {Noisy_label_lr}
\begin{figure}[h]
  \begin{tabular}[c]{cccc}
  \hspace{-5mm}
    \multirow{2}{*}{
    \begin{subfigure}{0.30\textwidth}
        \vspace{-5mm}
      \includegraphics[width=\textwidth]{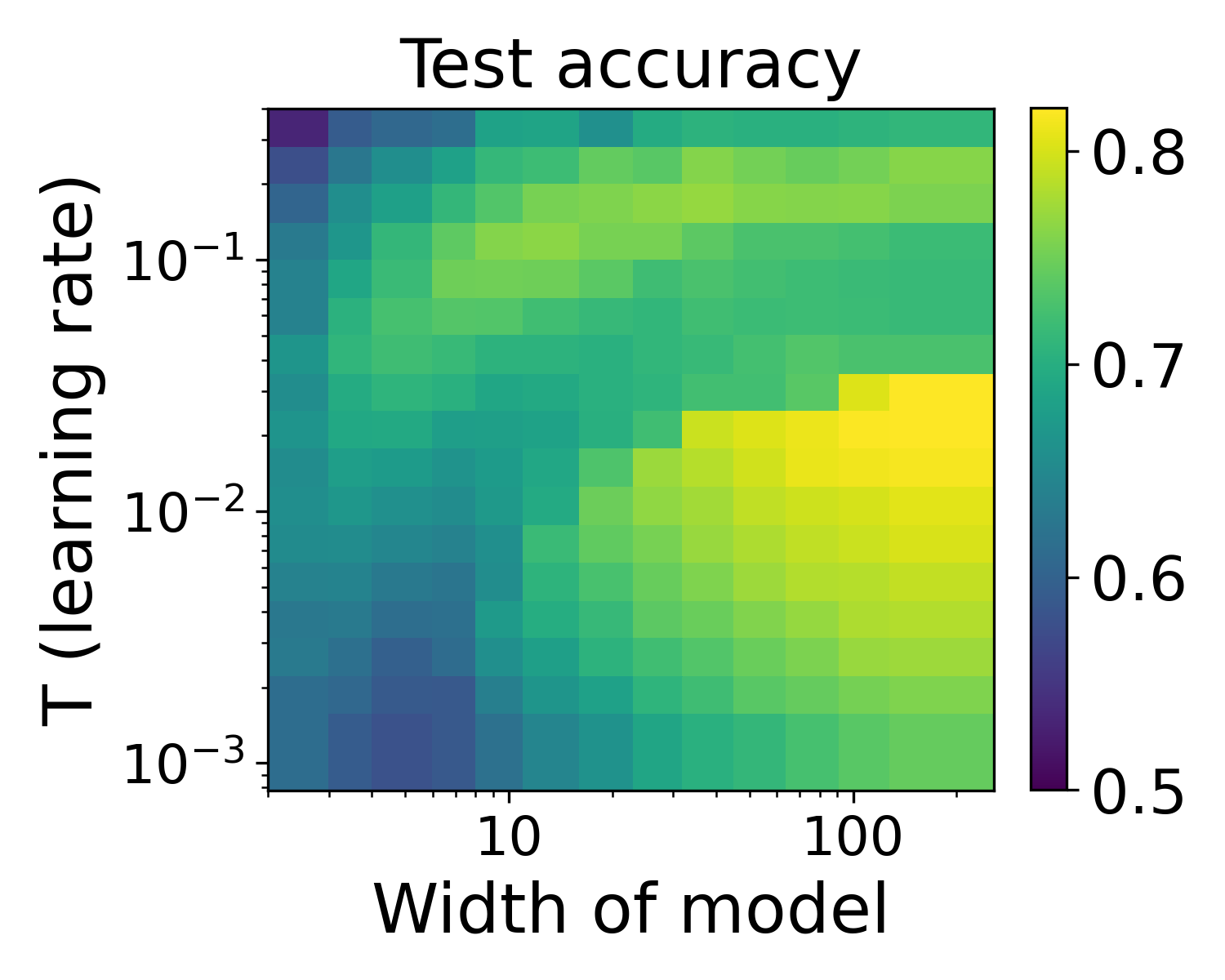}
      \caption{Test accuracy\label{fig:\figname_accuracy}}
    \end{subfigure}
} 
& \begin{subfigure}[c]{0.21\textwidth}
      \includegraphics[width=\textwidth]{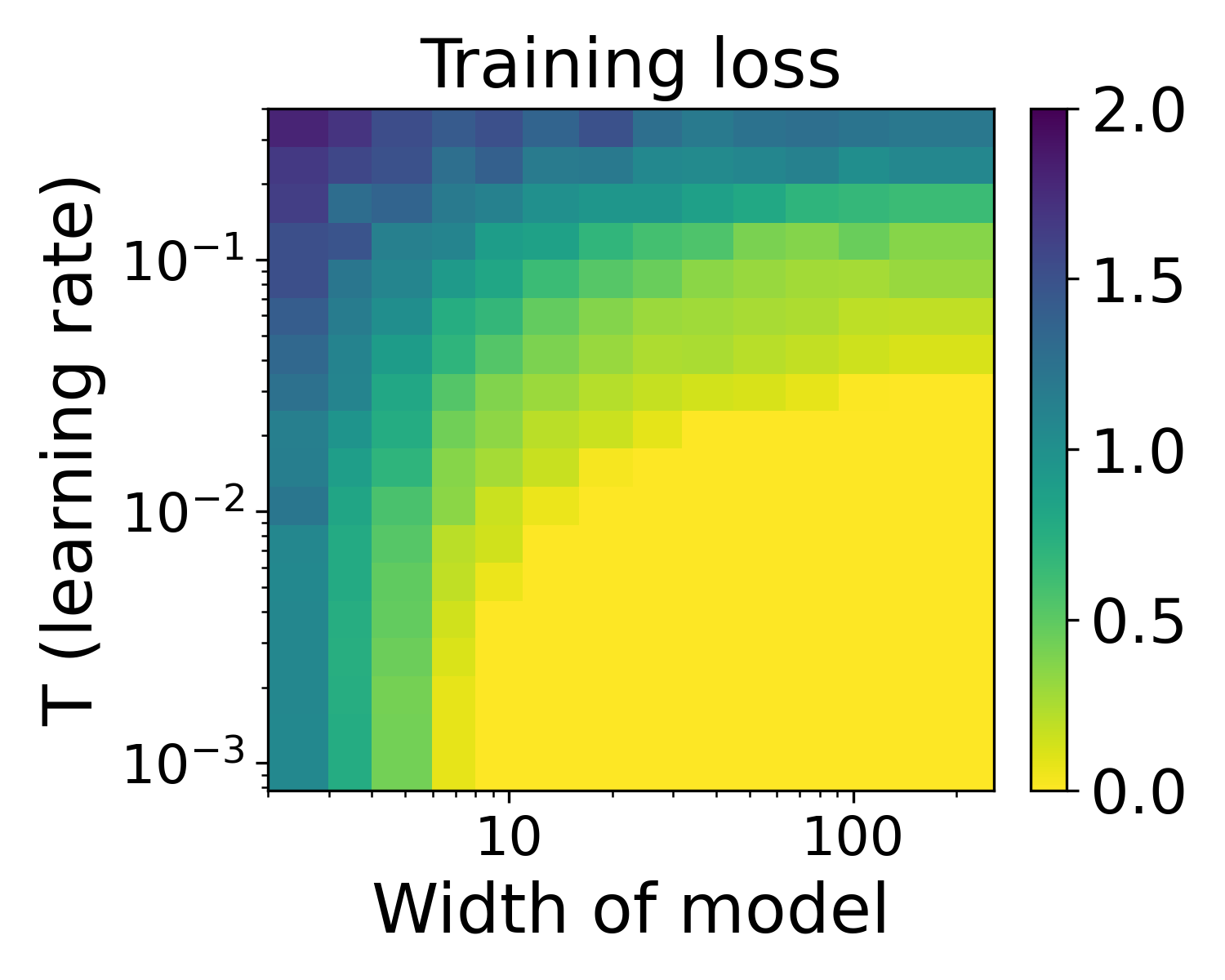}
      \caption{Training loss\label{fig:\figname_loss}}
    \end{subfigure}&
    \begin{subfigure}[c]{0.21\textwidth}
      \includegraphics[width=\textwidth]{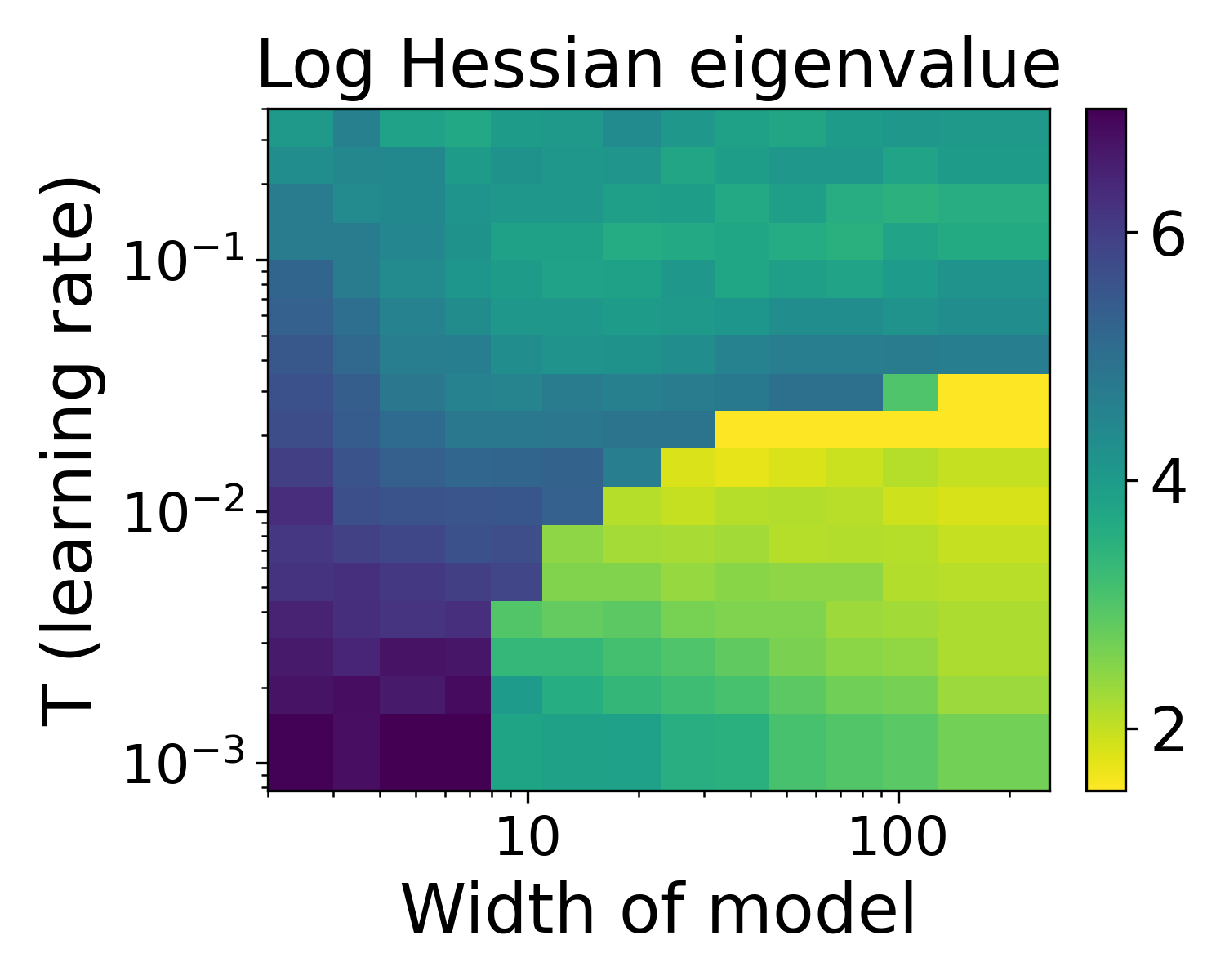}
      \caption{Hessian eigenvalue\label{fig:\figname_hessian_e}}
    \end{subfigure}&
    \begin{subfigure}[c]{0.21\textwidth}
      \includegraphics[width=\textwidth]{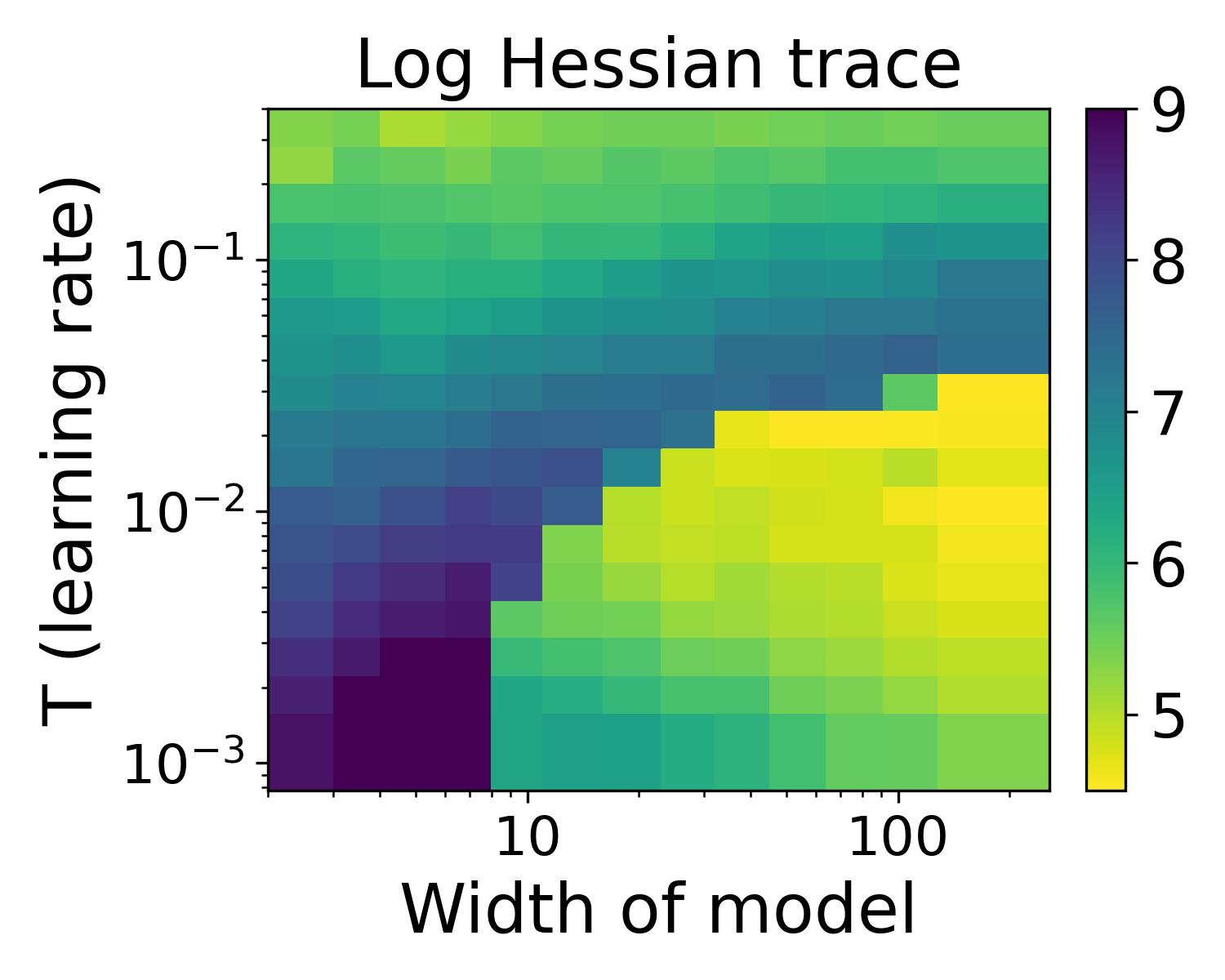}
      \caption{Hessian trace\label{fig:\figname_hessian_t}}
    \end{subfigure}\\
& \begin{subfigure}[c]{0.21\textwidth}
      \includegraphics[width=\textwidth]{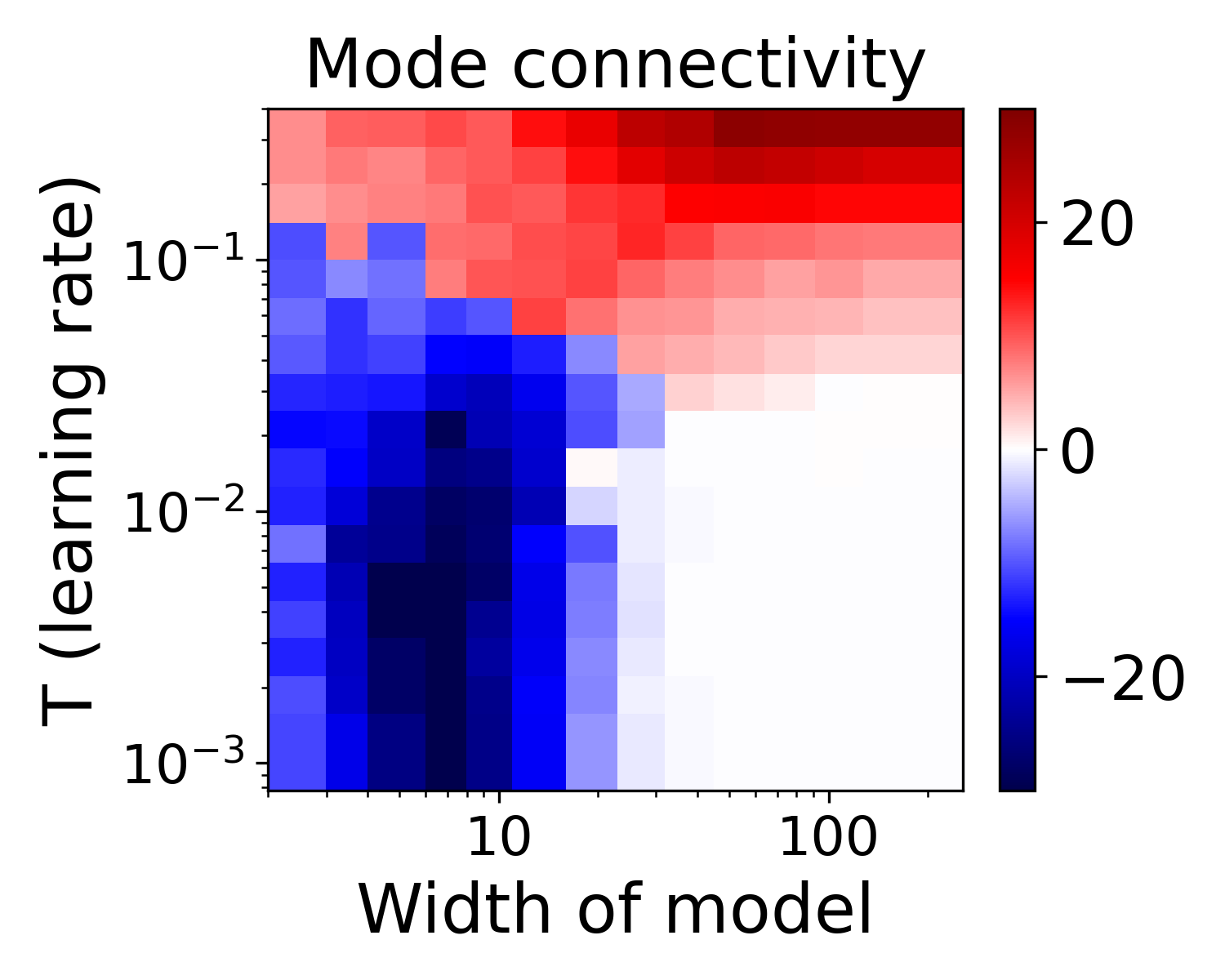}
      \caption{Mode connectivity\label{fig:\figname_curve}}
    \end{subfigure}&
    \begin{subfigure}[c]{0.21\textwidth}
      \includegraphics[width=\textwidth]{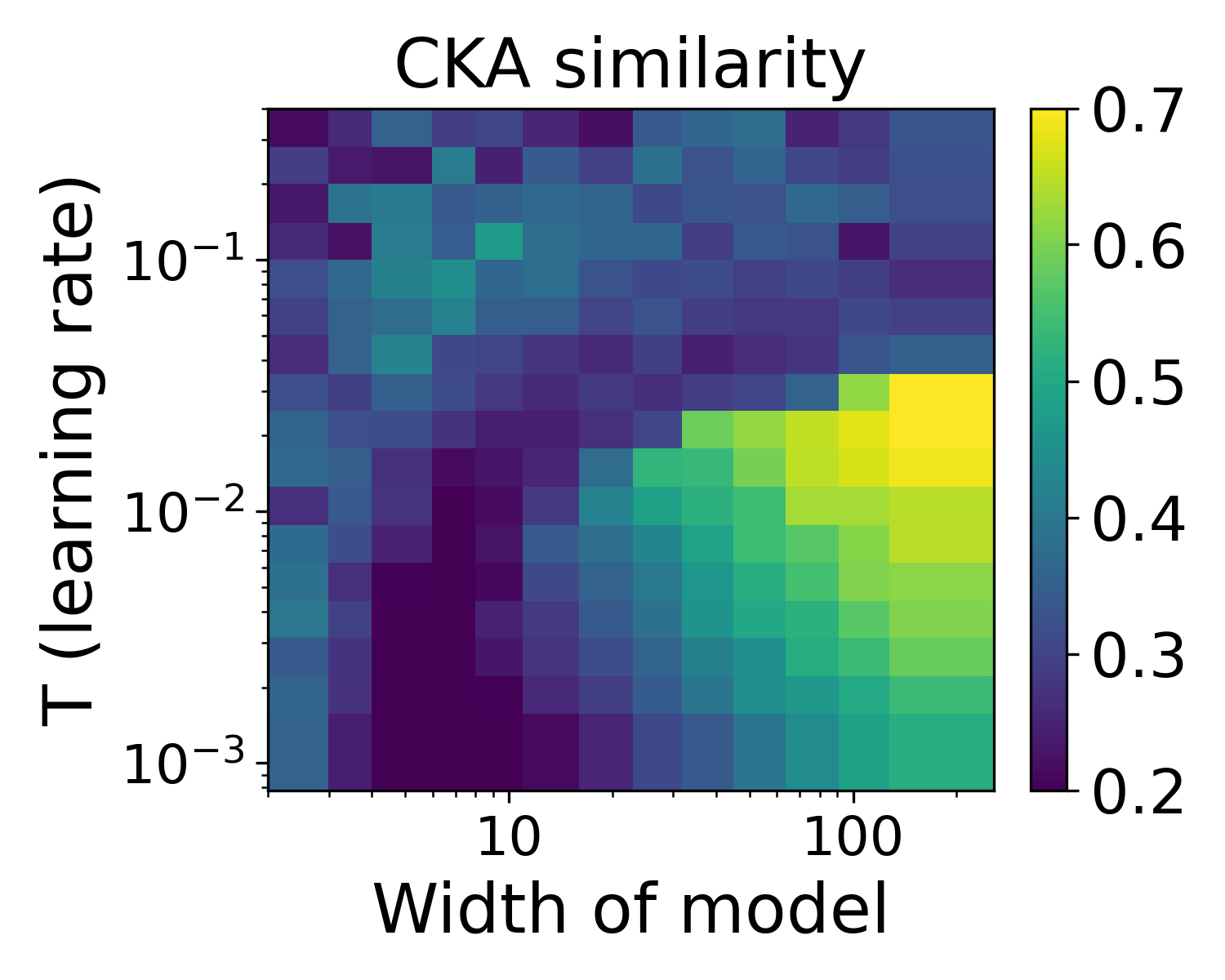}
      \caption{CKA similarity\label{fig:\figname_CKA}}
    \end{subfigure}&
    \begin{subfigure}[c]{0.21\textwidth}
      \includegraphics[width=\textwidth]{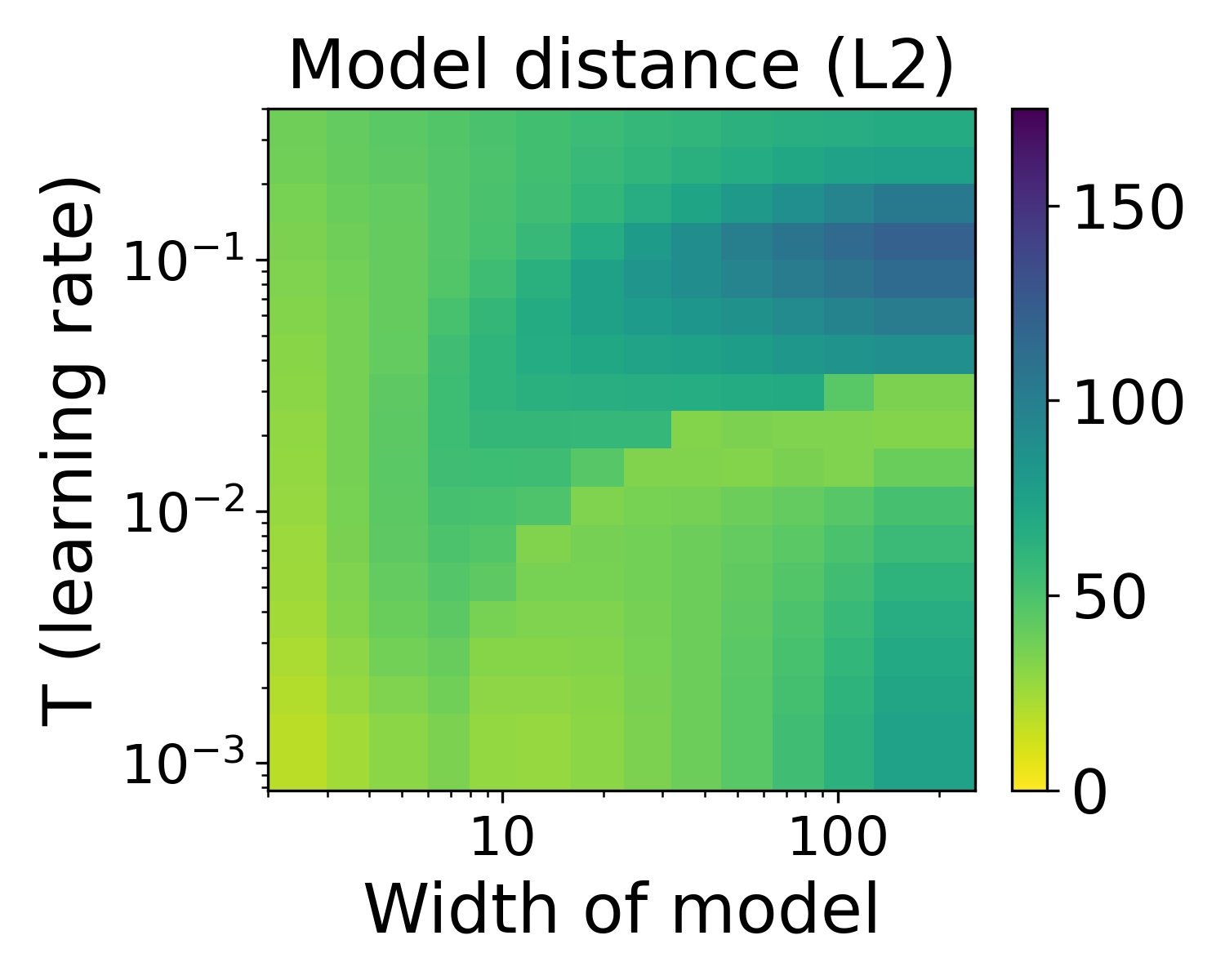}
      \caption{$\ell_2$ distance\label{fig:\figname_dist}}
    \end{subfigure}\\
  \end{tabular}    
  \caption{{\bf (Training to noisy labels, with learning rate as temperature).} Partitioning the 2D load-like---temperature-like diagram into different phases of learning, using learning rate as the temperature and varying model width to change load. 
  10\% of labels are randomized, and double descent is observed between phases.
  Models are trained with ResNet18 on CIFAR-10. All plots are on the same set of~axes.}
  \label{fig:Noisy_label_lr}
\end{figure}

A side note is that, apart from distinguishing the two subcategories in Phase IV, CKA can also help categorize the other phases. 
In particular, for Figure \ref{fig:ResNet18}, CKA(Phase I and II) $<$ CKA(Phase III) $<$ CKA(Phase IV). 
However, as we will see in Figure \ref{fig:Noisy_label}, the CKA in Phase III (globally poorly-connected and locally flat) can become worse than Phase I and II, suggesting that both global structural deficiency and local structural deficiency can dictate the large dissimilarity between models. 
(Clearly, this suggests the need for improved global metrics to compare models.)
What remains a consistent trend, however, is that Phase IV-B always has the largest CKA similarity.

\subsection{Results on machine translation}\label{sec:NLP}

In this subsection, we show the results on the neural machine translation task on IWSLT 2016 De-En. 
In this experiment, we still define load as the width of the Transformer model, which is the dimension of embedding vectors.
See the results in Figure \ref{fig:NMT_constant_lr}. 
Similar to Figure~\ref{fig:Noisy_label}, the results exhibit both width-wise and temperature-wise double descent.
The first type of double descent matches previous work~\citep{nakkiran2019deep}.

More importantly, the mode connectivity shown in Figure \ref{fig:NMT_constant_lr_curve} remains poor even if we increase the width of the model to a large value. 
We also note that we only use 4K samples, which, according to the results on subsampled CIFAR-10 shown in Figure~\ref{fig:different_data}, means that mode connectivity should be easier to become closer to zero than if we used the whole dataset.
Thus, the mode connectivity result here suggests that the loss landscape in this machine translation task is significantly worse than that of image classification on CIFAR-10 (shown in Figure~\ref{fig:ResNet18} and Figure~\ref{fig:Noisy_label}).
This suggests that, even for the large-width Transformers, we still have not transitioned to globally well-connected loss landscapes---in particular, this means that the entire Figure \ref{fig:NMT_constant_lr_accuracy} only covers the ``top-left corner'' of the 2D phase diagram shown in Figure \ref{fig:ResNet18}, i.e., Phase I. 
From the discussion in Appendix~\ref{sec:double_descent}, one should be careful about training to zero loss in globally poorly-connected loss landscapes. 
Indeed, we show in the next paragraph that early stopping can significantly reduce the test cross entropy loss.
It is worth noting that the training on the top-right corner of each subplot in Figure~\ref{fig:NMT_constant_lr} is difficult to converge, due to large temperature and large size of the Transformer, so we explicitly mark that region with ``NC'' (meaning ``not converged'').

\def \figname {NMT_constant_lr}
\begin{figure}
    \centering
    \begin{subfigure}{0.23\textwidth}
      \includegraphics[width=\textwidth]{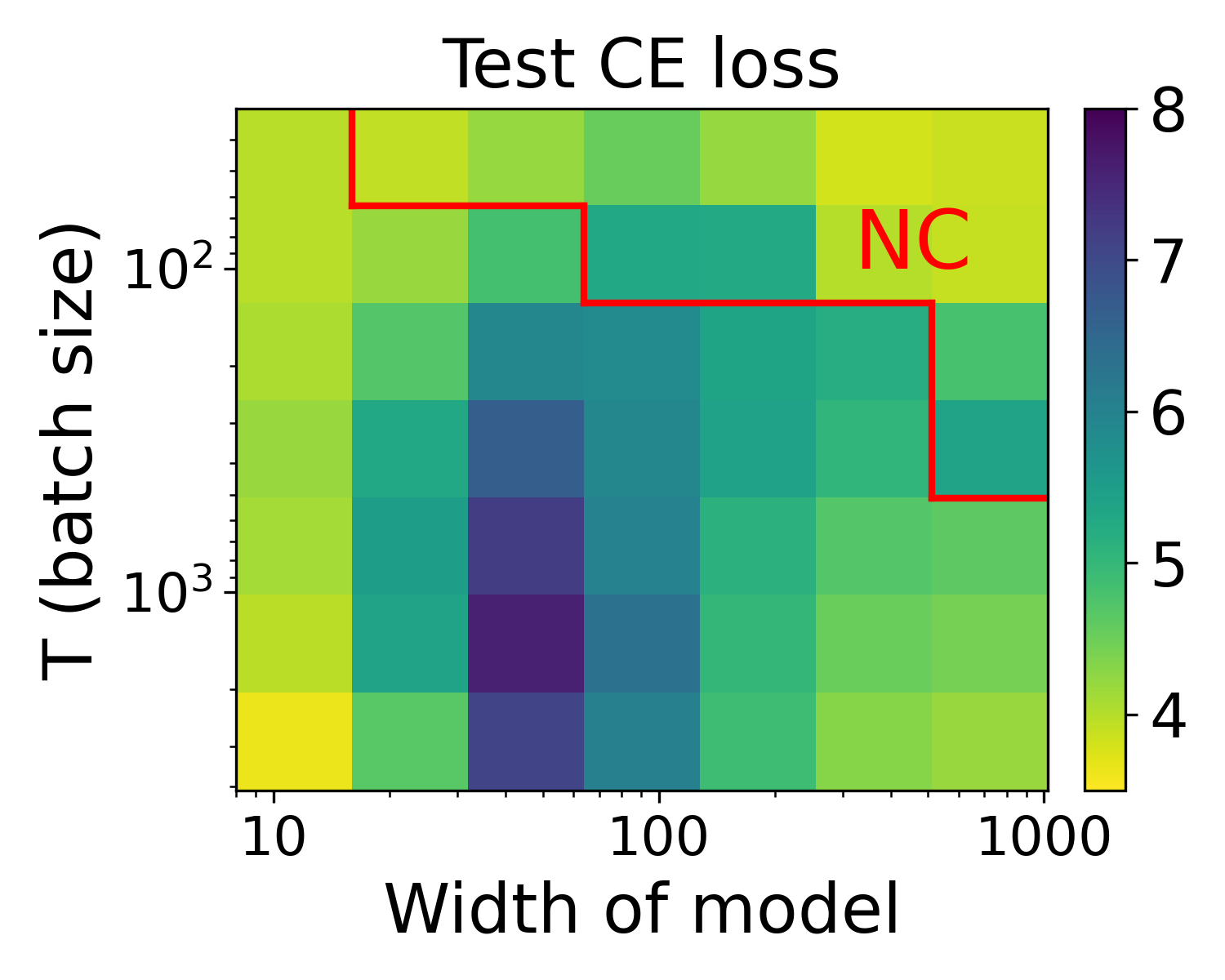}
      \caption{Test CE loss\label{fig:\figname_accuracy}}
    \end{subfigure} 
    \begin{subfigure}{0.23\textwidth}
      \includegraphics[width=\textwidth]{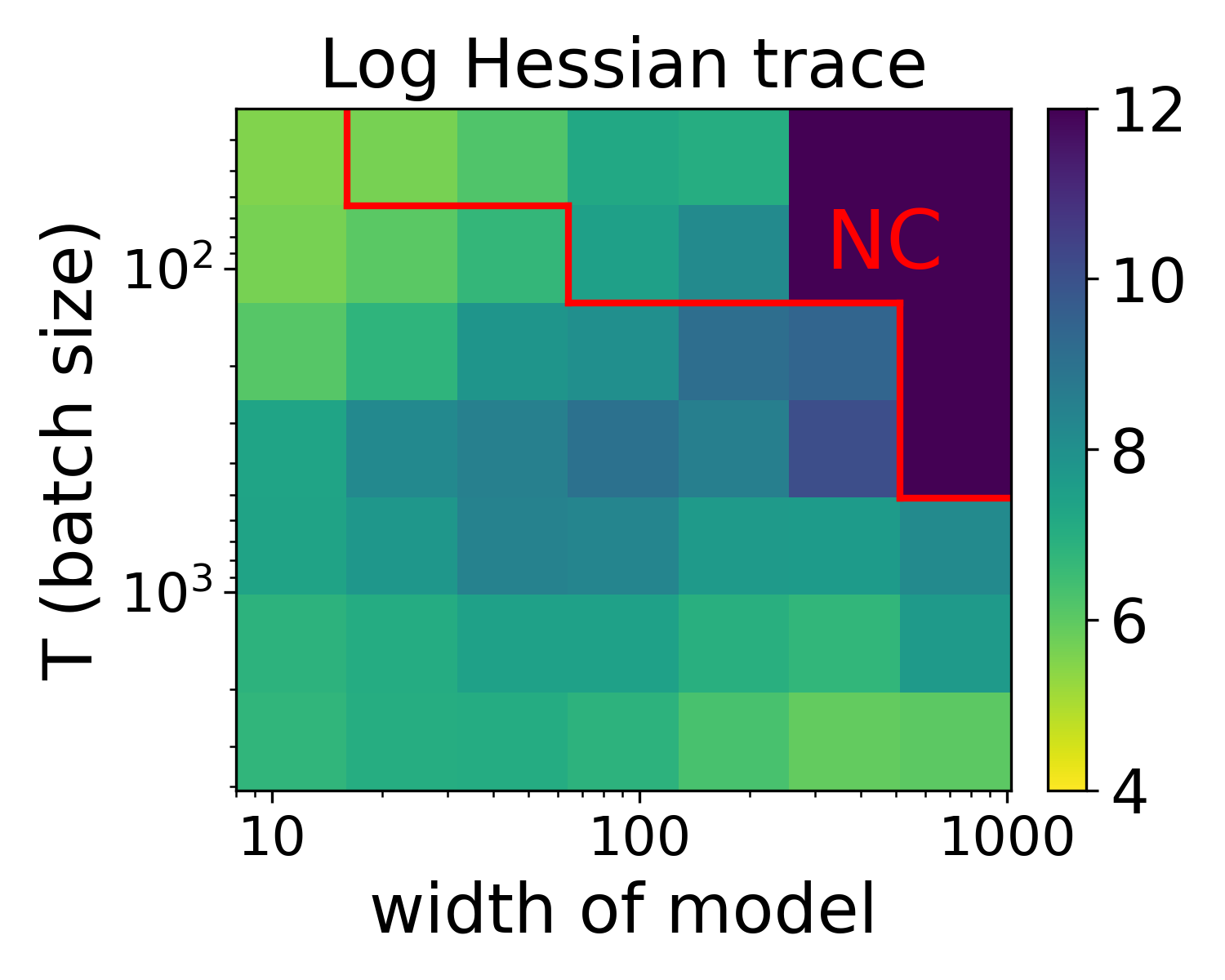}
      \caption{Hessian trace\label{fig:\figname_hessian_t}}
    \end{subfigure}
    \begin{subfigure}{0.23\textwidth}
      \includegraphics[width=\textwidth]{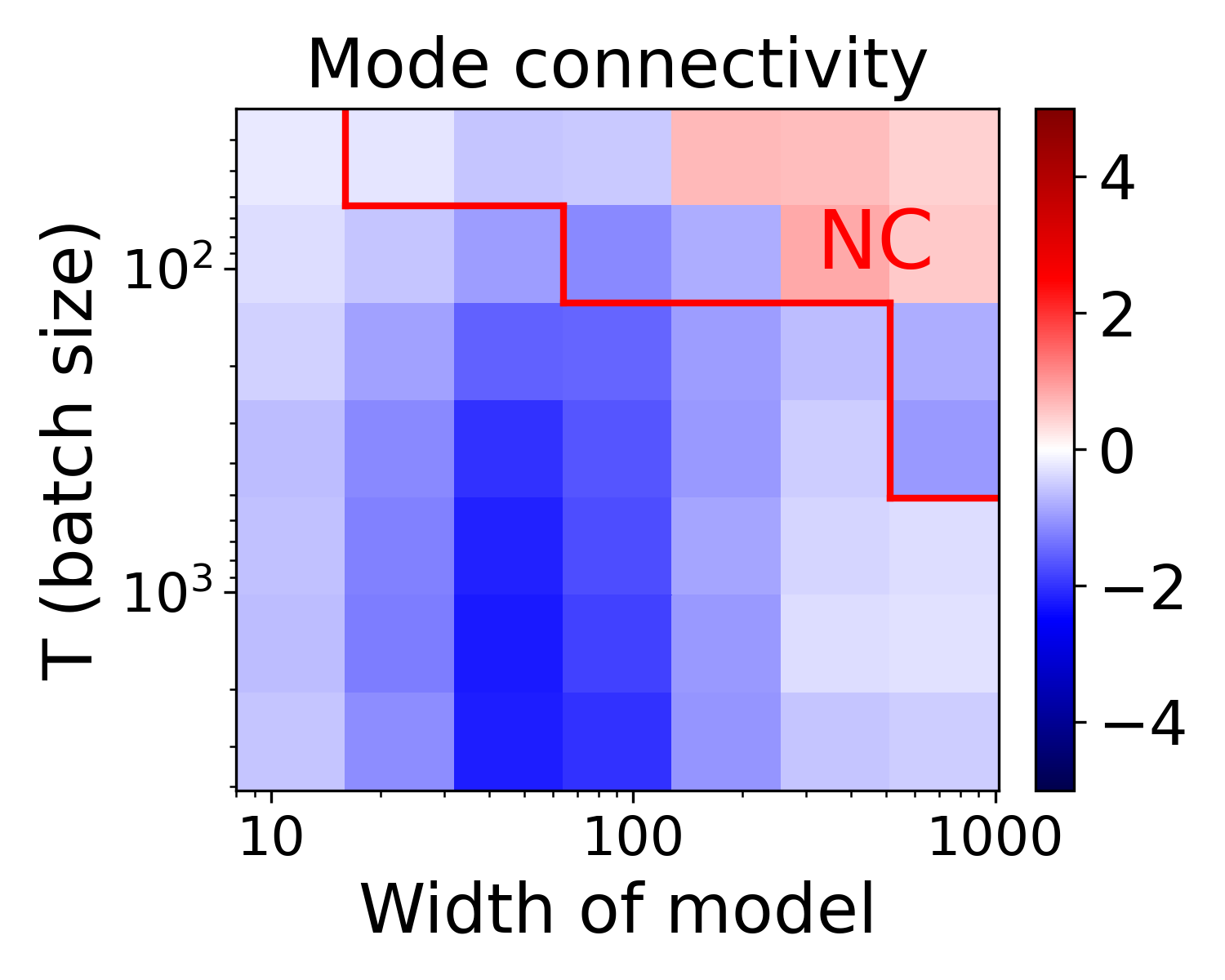}
      \caption{Mode connectivity\label{fig:\figname_curve}}
    \end{subfigure}
    \begin{subfigure}{0.23\textwidth}
      \includegraphics[width=\textwidth]{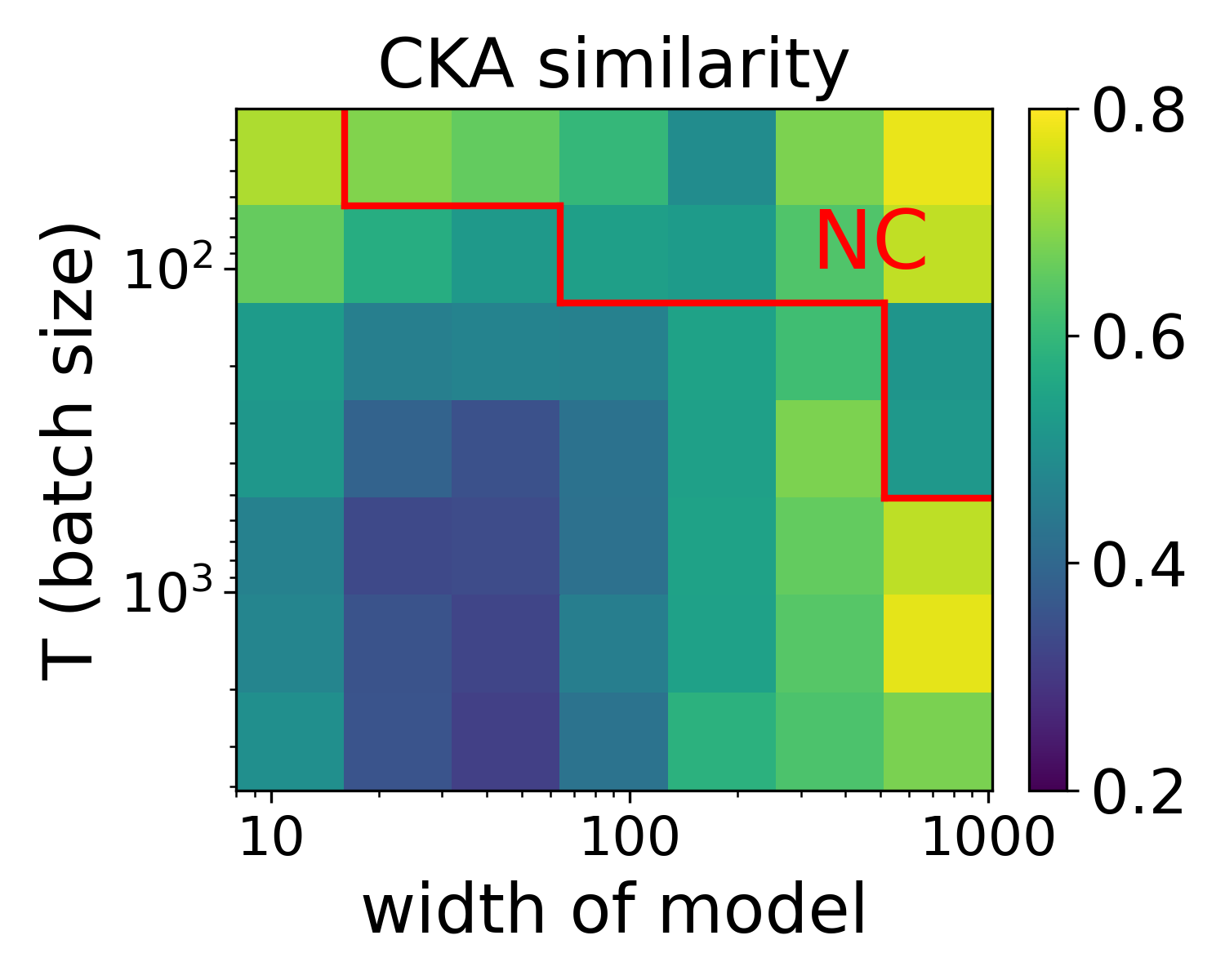}
      \caption{CKA similarity\label{fig:\figname_CKA}}
    \end{subfigure}
    \caption{{\bf (Machine translation).} Partitioning the 2D load-like---temperature-like diagram into different phases of learning, using batch size as the temperature and varying model width (token embedding dimension) to change load.
    Models are trained with Transformers on IWSLT 2016 De-En with 4K subsamples. All plots are on the same set of axes. Mode connectivity shows that the loss landscape is poorly-connected even for a large embedding dimension.
    We find the training on the upper-right corner of each subplot hard to converge.
    }
    \label{fig:NMT_constant_lr}
\end{figure}

{\bf Optimal early stopping helps when the global connectivity is low}
Now, we provide more results to the machine translation task.
In particular, we report the results on training with optimal early stopping and training with an inverse square-root learning rate.
Training with optimal early stopping means that we choose the best test accuracy during the entire training process \citep{nakkiran2019deep}, which is used to show the theoretically optimal accuracy improvement from using early stopping.
See Figure \ref{fig:NMT_inverse_sqrt}.

\begin{figure}
    \begin{tabular}{cc}
        \begin{tabular}{c}
        \centering
             \begin{subfigure}[t]{0.40\textwidth}
      \includegraphics[width=\linewidth]{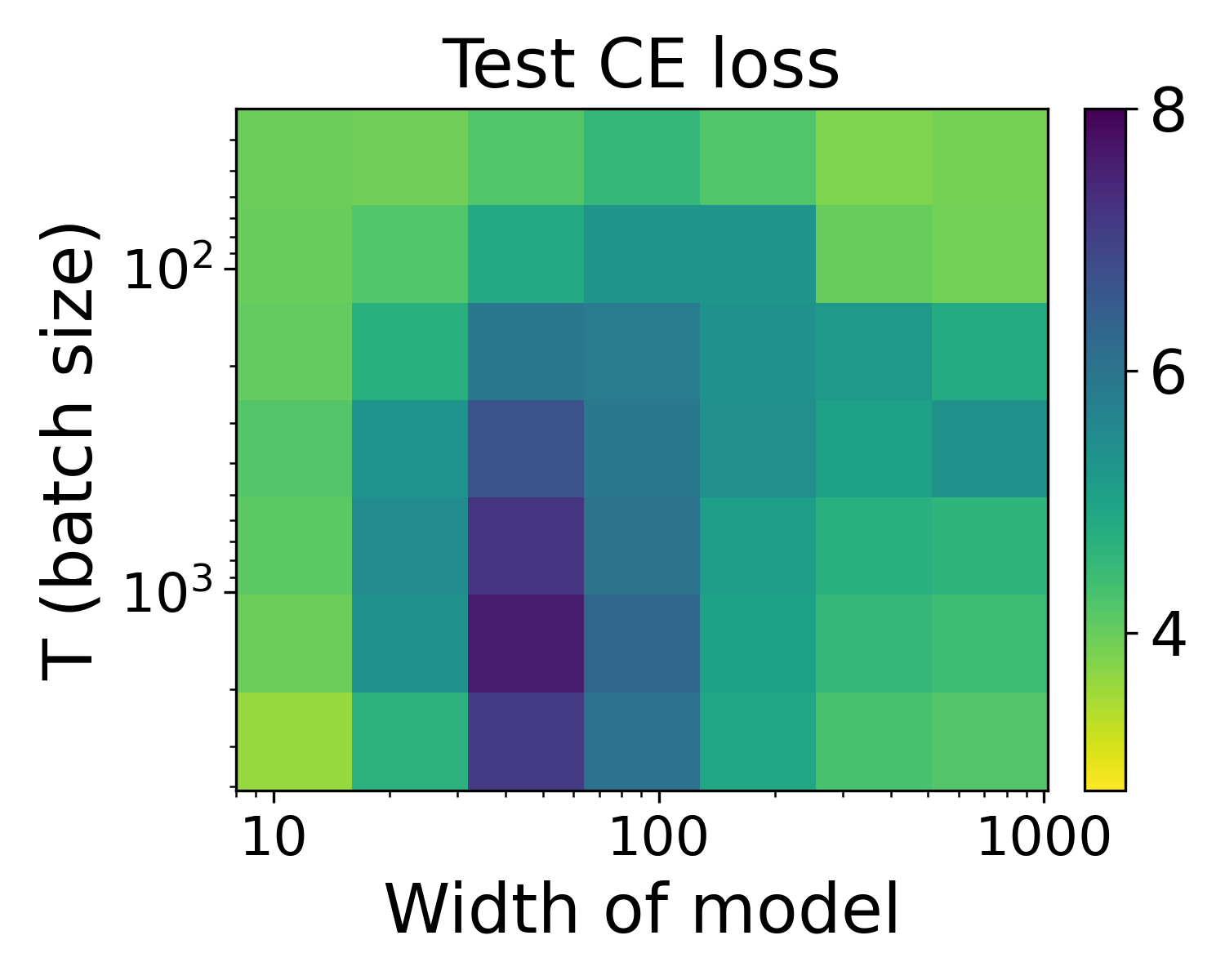}
      \caption{Training with a constant learning rate and 80K gradient updates}
    \end{subfigure} 
        \end{tabular} & \begin{tabular}{c}
             \begin{subfigure}[t]{0.40\textwidth}
      \includegraphics[width=\linewidth]{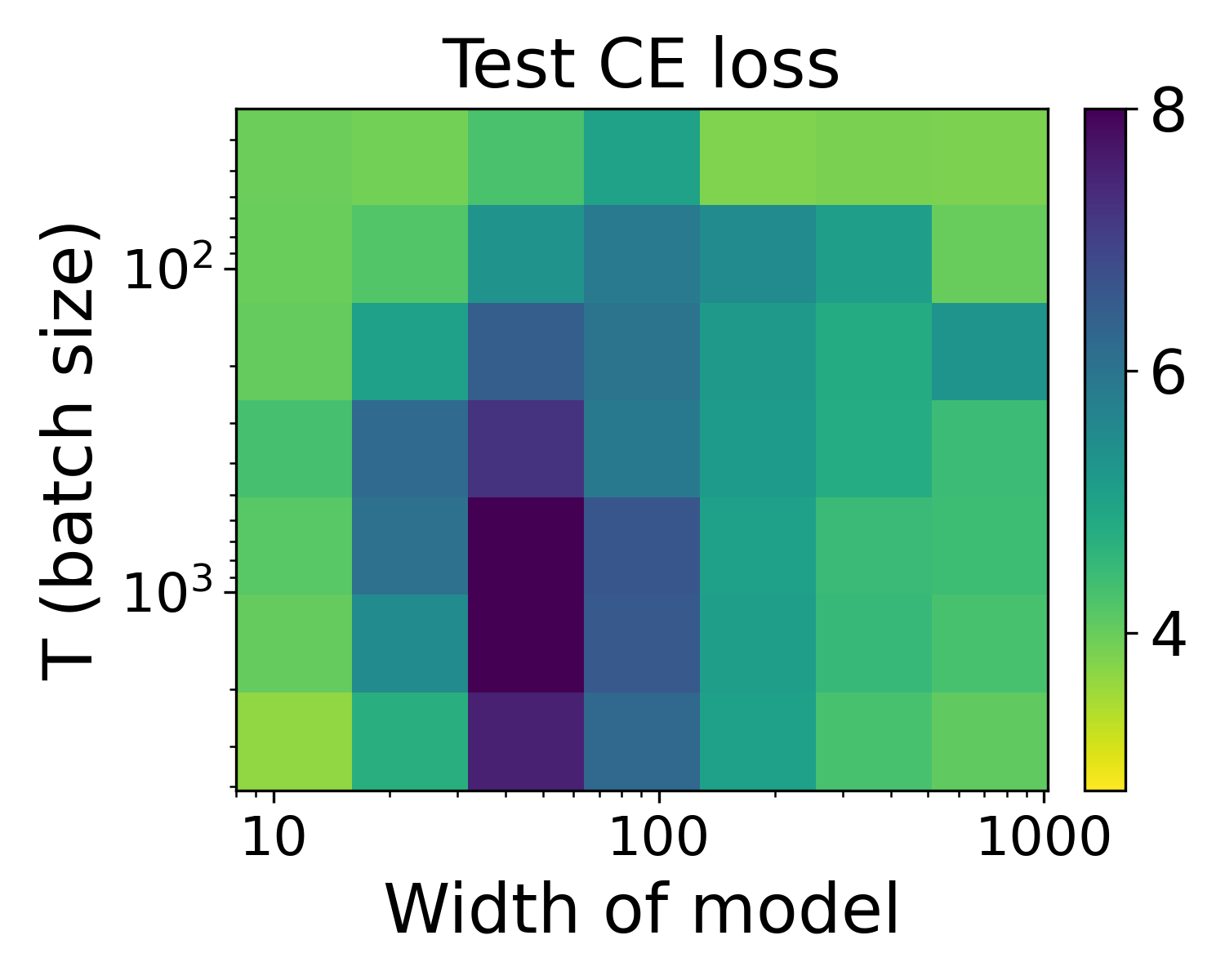}
      \caption{Training with an inverse square-root learning rate and 80K gradient updates}
    \end{subfigure}
        \end{tabular} \\
        \begin{tabular}{c}
             \begin{subfigure}[t]{0.40\textwidth}
      \includegraphics[width=\linewidth]{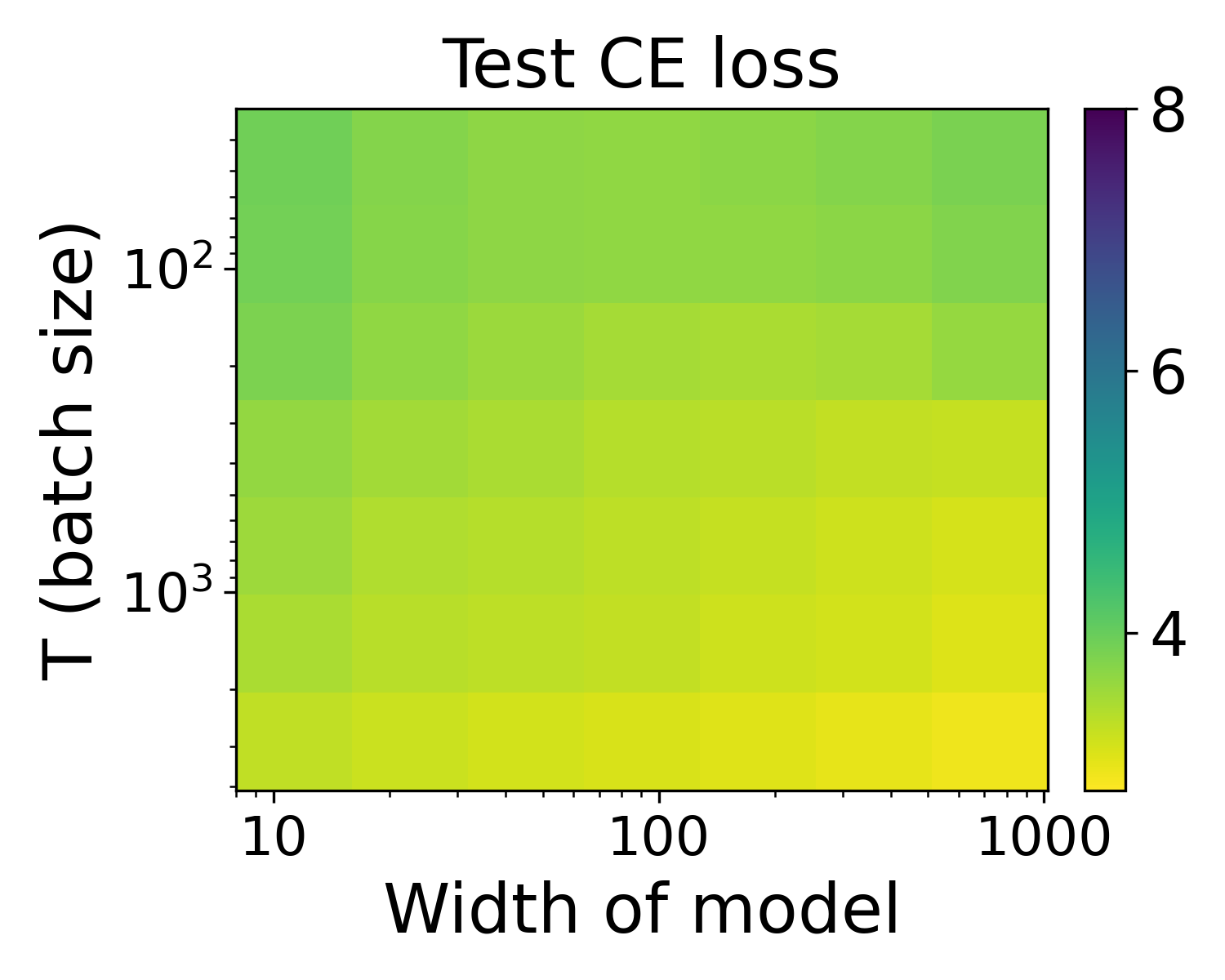}
      \caption{Training with a constant learning rate and optimal early stopping}
    \end{subfigure} 
        \end{tabular} & \begin{tabular}{c}
             \begin{subfigure}[t]{0.40\textwidth}
      \includegraphics[width=\linewidth]{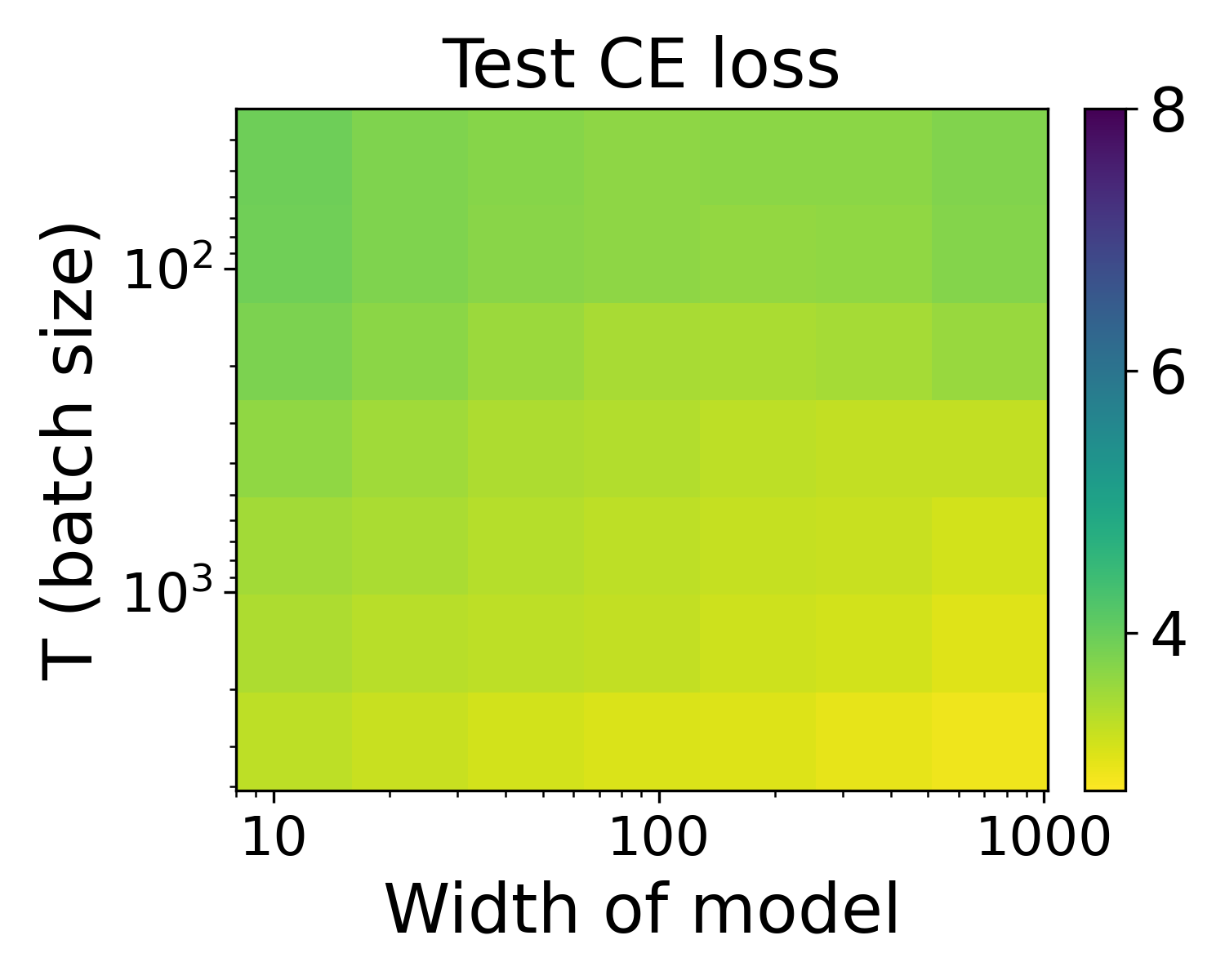}
      \caption{Training with an inverse square-root learning rate and optimal early stopping}
    \end{subfigure}
        \end{tabular}
    \end{tabular}
\caption{{\bf (Early stopping).} {\bf (First row).} Training with 80K gradient updates. {\bf (Second row).} Training with optimal early stopping. {\bf (First column).} Training with a constant learning rate. {\bf (Second column).} Training with an inverse square-root learning rate. 
The main conclusion is that training with optimal early stopping significantly improves test accuracy in this case. }
    \label{fig:NMT_inverse_sqrt}
\end{figure}

Comparing the left and the right column in Figure~\ref{fig:NMT_inverse_sqrt}, we see that the inverse square-root learning rate does not significantly change the results.

However, comparing the first and the second row in Figure~\ref{fig:NMT_inverse_sqrt}, we see that optimal early stopping significantly improves the test accuracy. 
We note that this is expected and has been observed in \citep{nakkiran2019deep} that optimal early stopping can significantly mitigate double descent.
Since the global connectivity in this task is low even for large width (shown in Figure \ref{fig:NMT_constant_lr_curve}), the observation here further supports our conclusion in Appendix~\ref{sec:double_descent}, which is that one should not completely fit the training data when the global connectivity is~low.

\subsection{Additional temperature parameters}\label{sec:temperature_LR}

In this subsection, we reproduce the results in Figure \ref{fig:ResNet18}, but we change the temperature parameter from batch size to learning rate. 
See the results shown in  Figure \ref{fig:LR}.
We see that the results are very similar to Figure \ref{fig:ResNet18}.

\def \figname {LR}
\begin{figure}[h]
  \centering
    \begin{subfigure}{0.23\textwidth}
      \includegraphics[width=\textwidth]{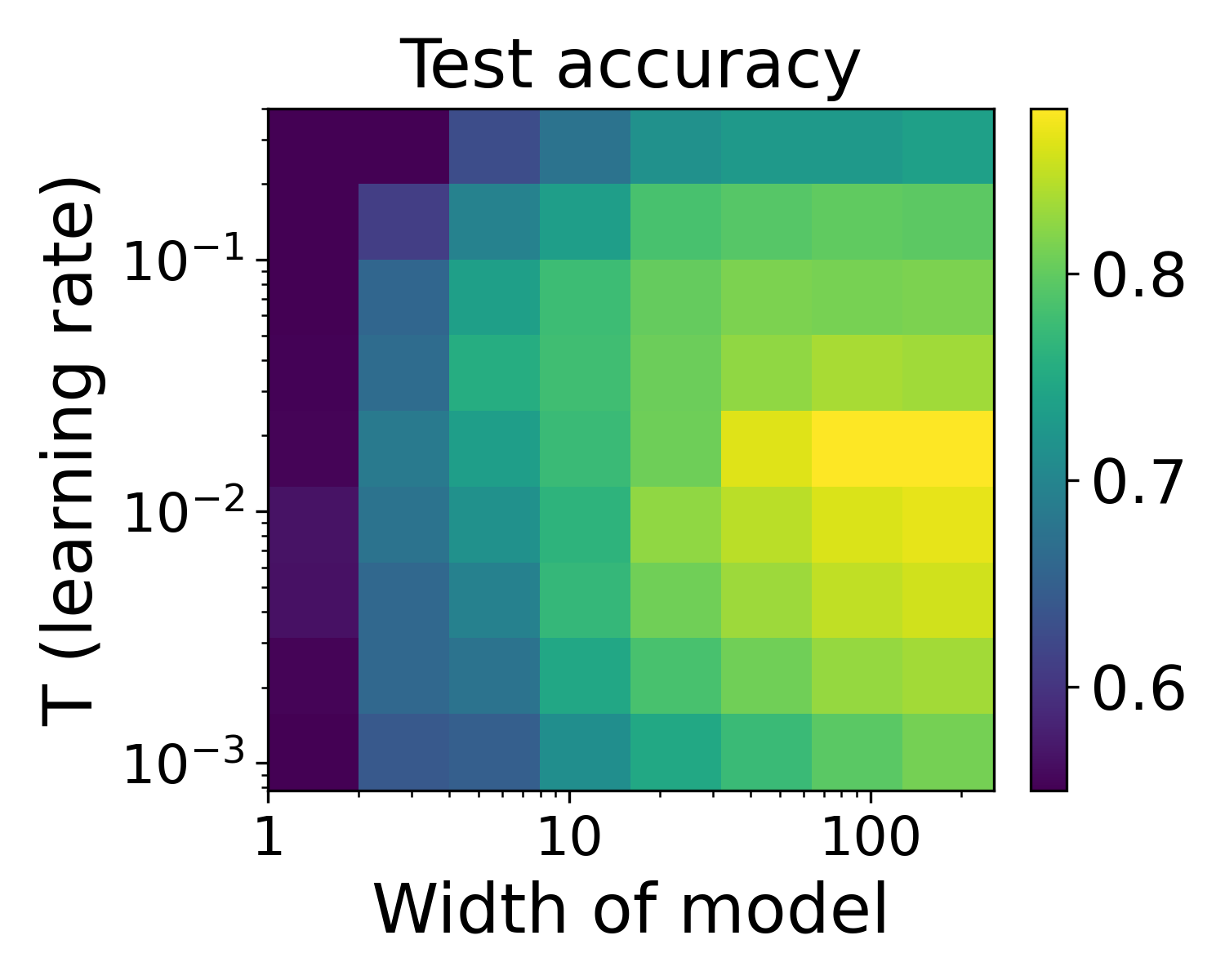}
      \caption{Test accuracy\label{fig:\figname_accuracy}}
    \end{subfigure}
    \begin{subfigure}[c]{0.23\textwidth}
      \includegraphics[width=\textwidth]{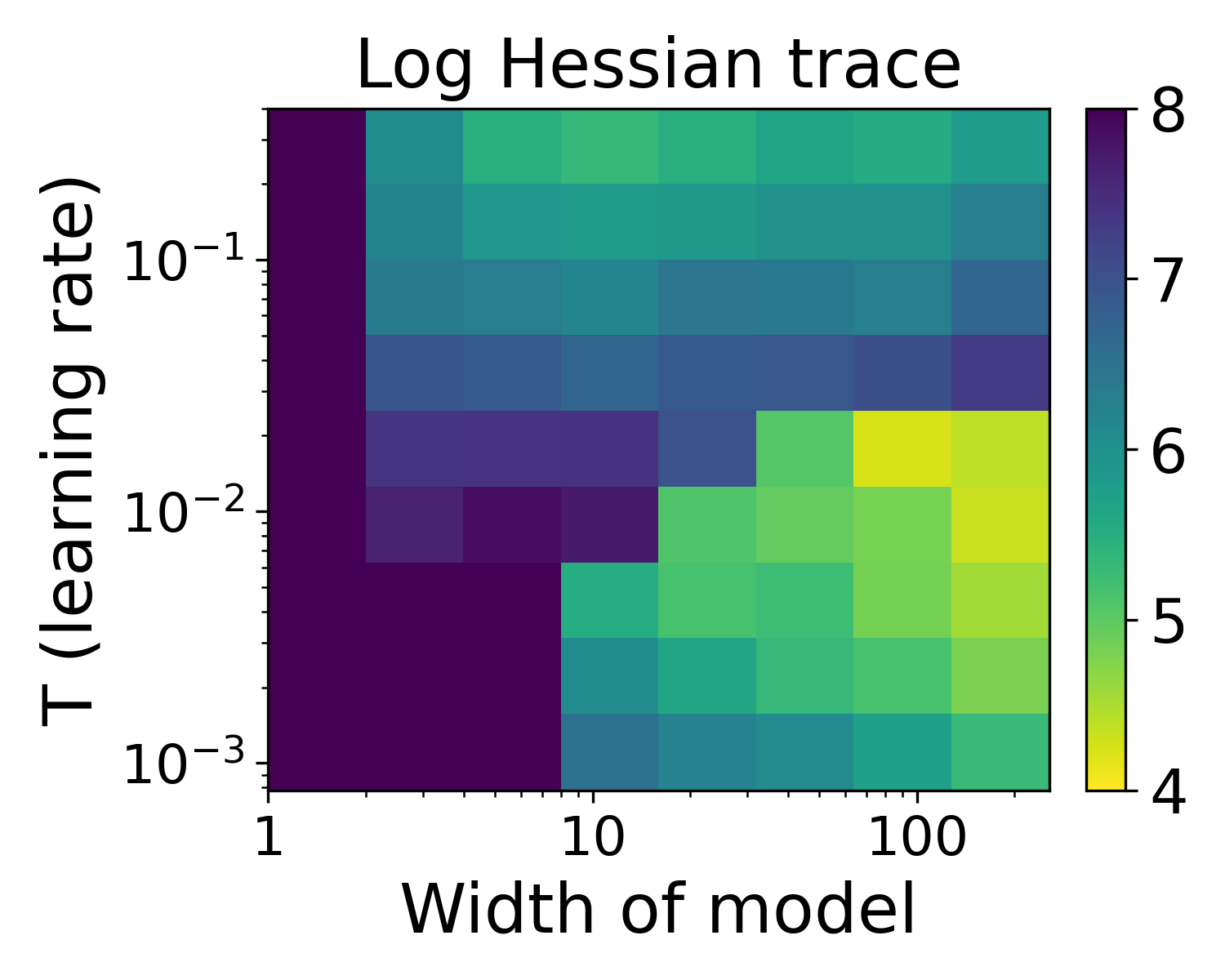}
      \caption{Hessian trace\label{fig:\figname_hessian_t}}
    \end{subfigure}
  \begin{subfigure}[c]{0.23\textwidth}
      \includegraphics[width=\textwidth]{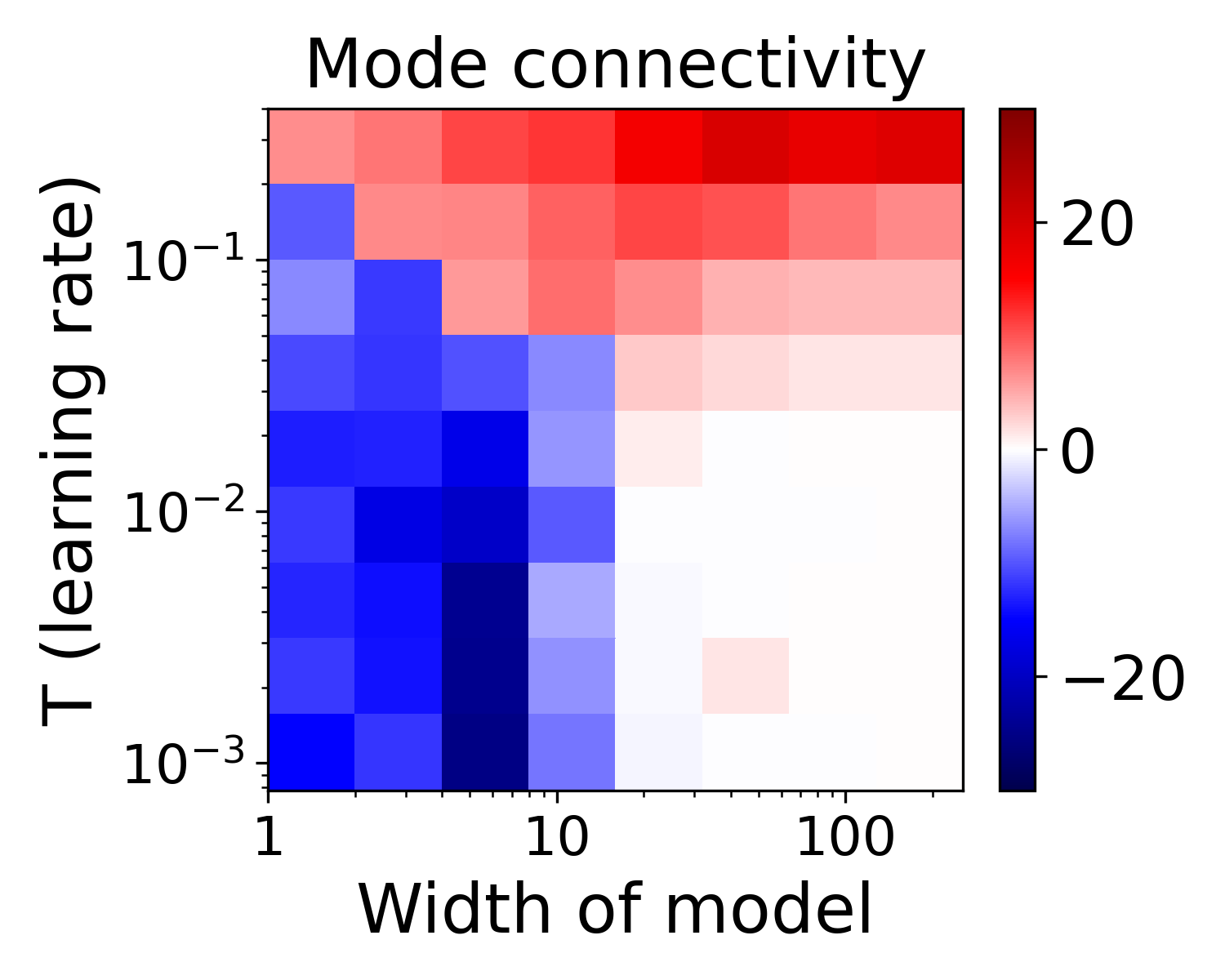}
      \caption{Mode connectivity\label{fig:\figname_curve}}
    \end{subfigure}
    \begin{subfigure}[c]{0.23\textwidth}
      \includegraphics[width=\textwidth]{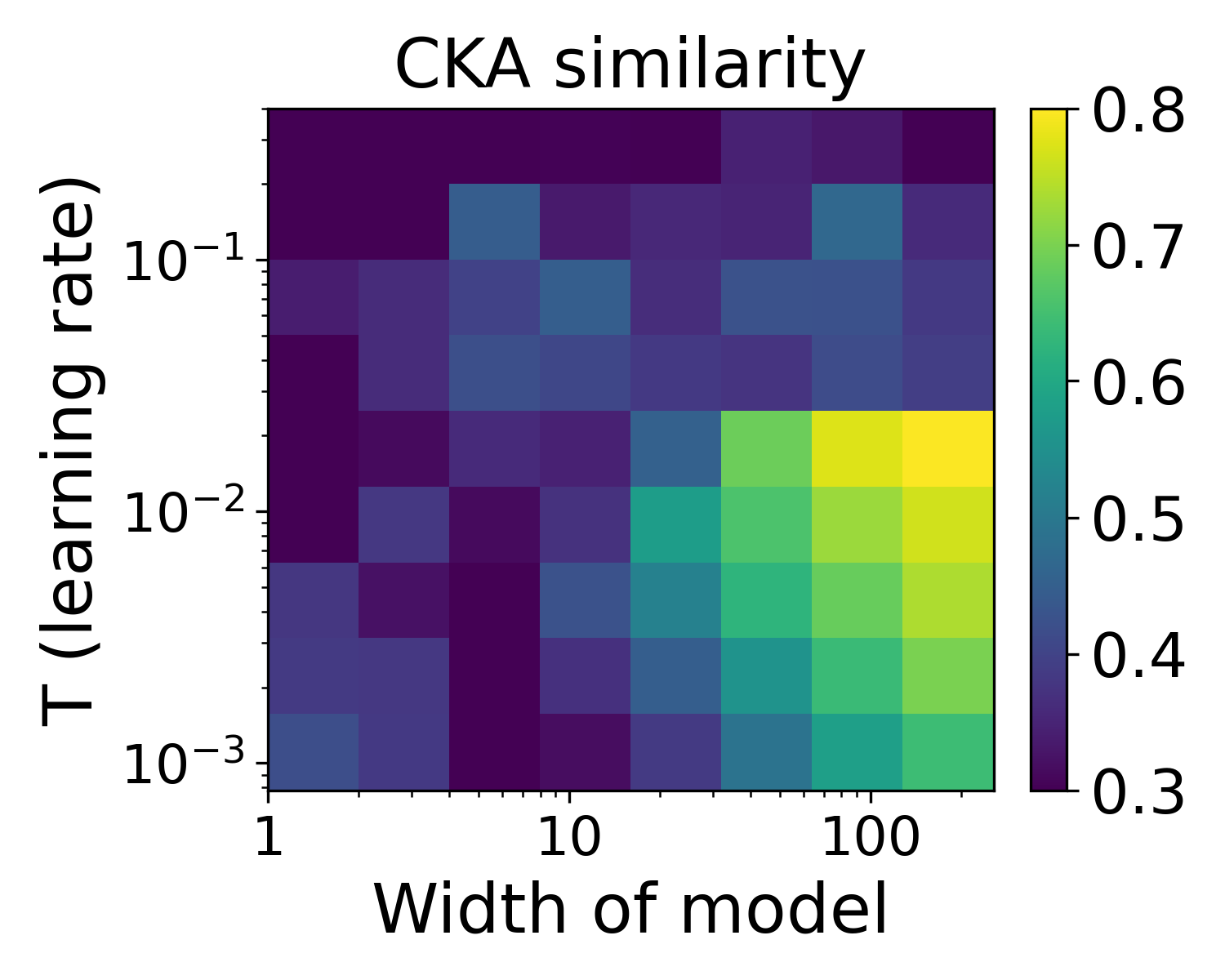}
      \caption{CKA similarity\label{fig:\figname_CKA}}
    \end{subfigure}
  \caption{{\bf (Learning rate as temperature).}
  Partitioning the 2D load-like---temperature-like diagram into different phases of learning, using learning rate as the temperature and varying model width to change load.
  Models are trained with ResNet18 on CIFAR-10. All plots are on the same set of axes. }
  \label{fig:LR}
\end{figure}

\subsection{Additional ways to change load}

In this subsection, we present results analogous to Figure~\ref{fig:Scaling_noisy_label}, but we change the load parameter by changing the amount of additive noise to each image, instead of the amount of randomized labels. 
More specifically, to change the amount of noise added to each image sample, we put a random additive noise with uniform distribution in $[0,u]$ to each pixel in the image, and we vary $u$.
See the results shown in Figure~\ref{fig:Noisy_image}.

When we compare Figure~\ref{fig:Noisy_image} to Figure~\ref{fig:Scaling_noisy_label}, we find that mode connectivity in Figure~\ref{fig:Noisy_image_curve} exhibits well-connected loss landscape despite changing amount of noise. 
We conjecture that this is because the amount of noise added to each pixel does not degrade the quality of data as substantially, and that this type of noise certainly is not as strong as changing the label.
However, CKA still captures the degradation of data quality -- if we move from the right to the left (meaning adding more noise) at the bottom of Figure~\ref{fig:Noisy_image_CKA}, we see gradually darker color.

\def \figname {Noisy_image}
\begin{figure}
  \centering
    \begin{subfigure}{0.23\textwidth}
      \includegraphics[width=\textwidth]{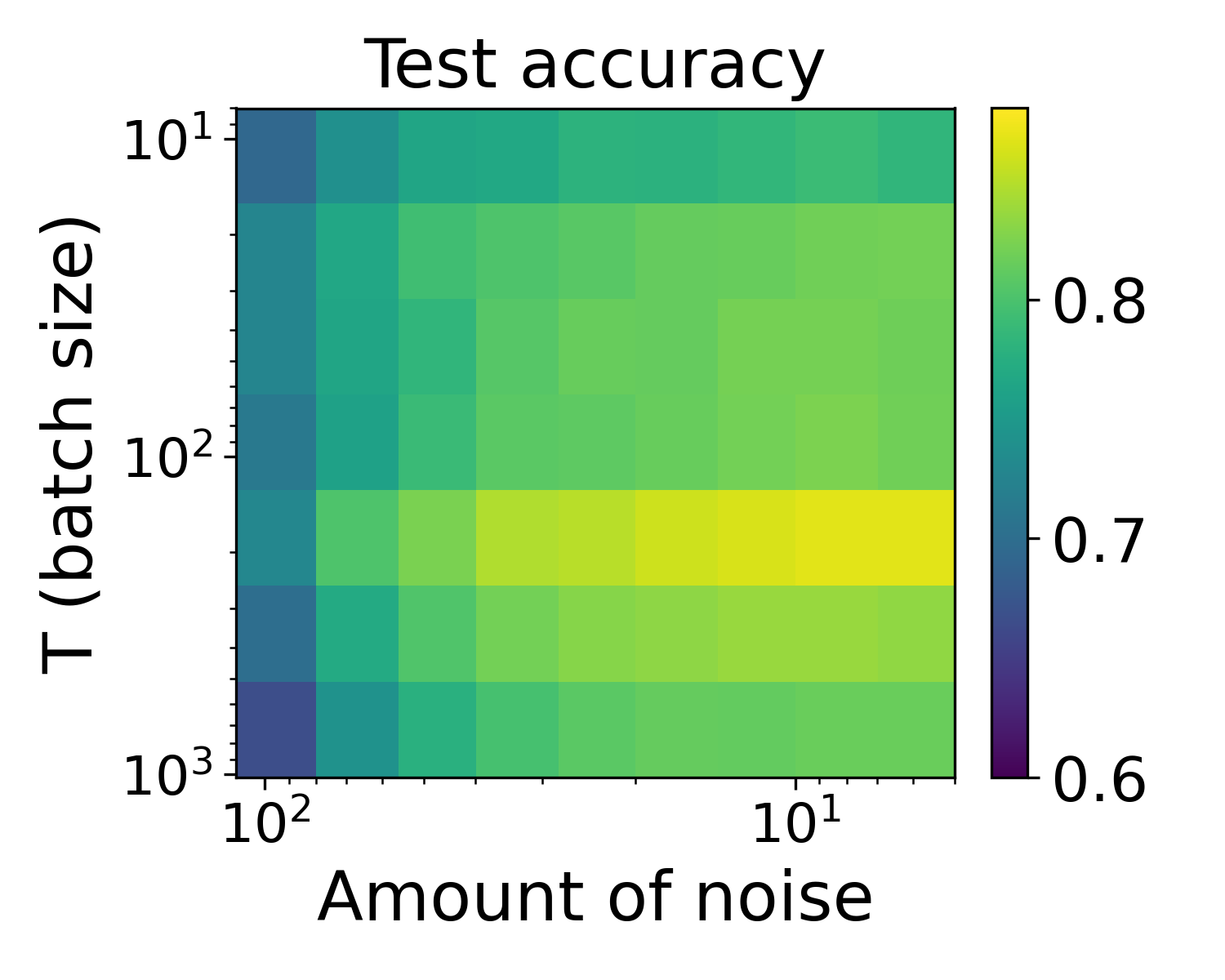}
      \vspace{-6mm}
      \caption{Test accuracy\label{fig:\figname_accuracy}}
    \end{subfigure}
    \begin{subfigure}[c]{0.23\textwidth}
      \includegraphics[width=\textwidth]{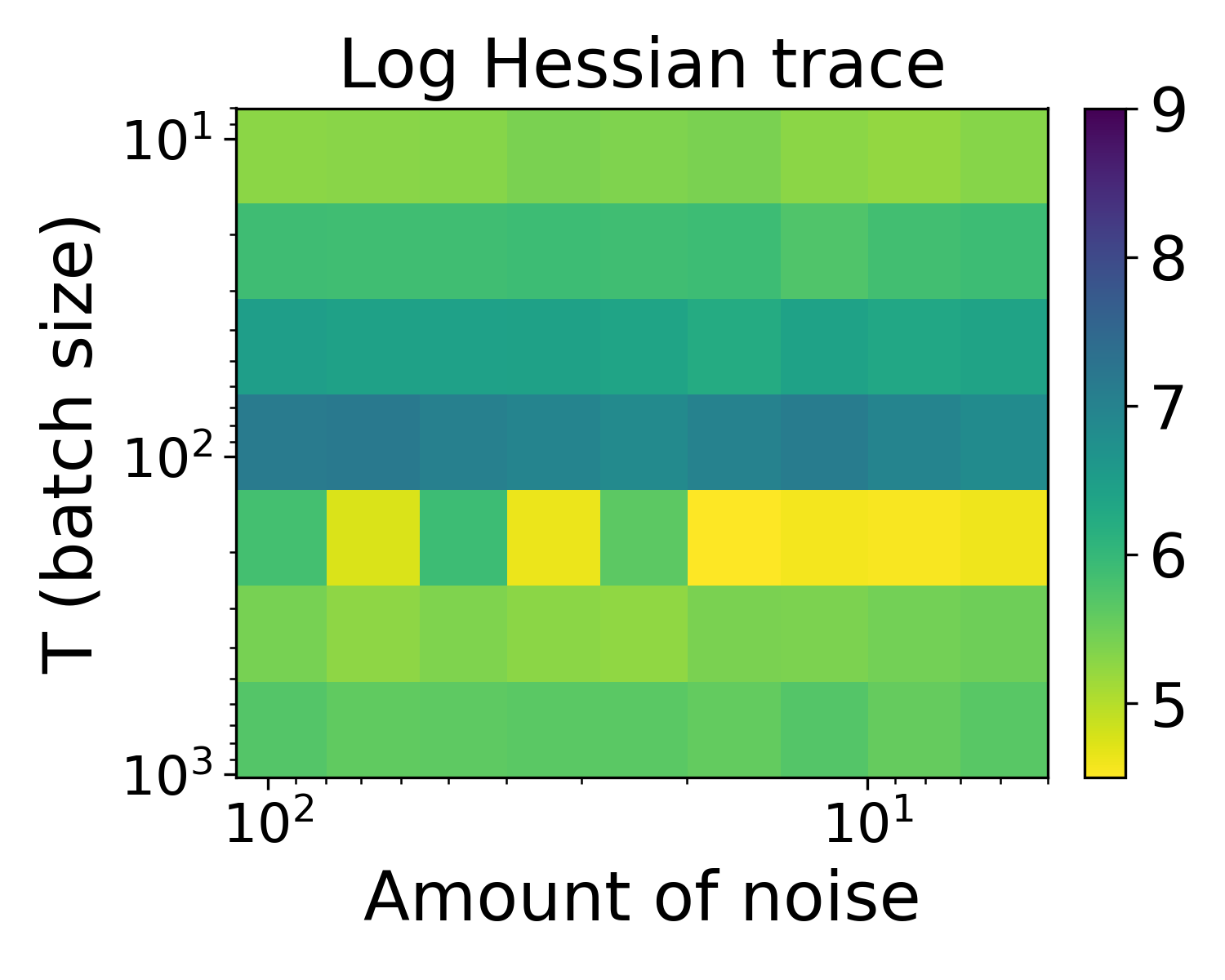}
      \vspace{-6mm}
      \caption{Hessian trace\label{fig:\figname_hessian_t}}
    \end{subfigure}
  \begin{subfigure}[c]{0.23\textwidth}
      \includegraphics[width=\textwidth]{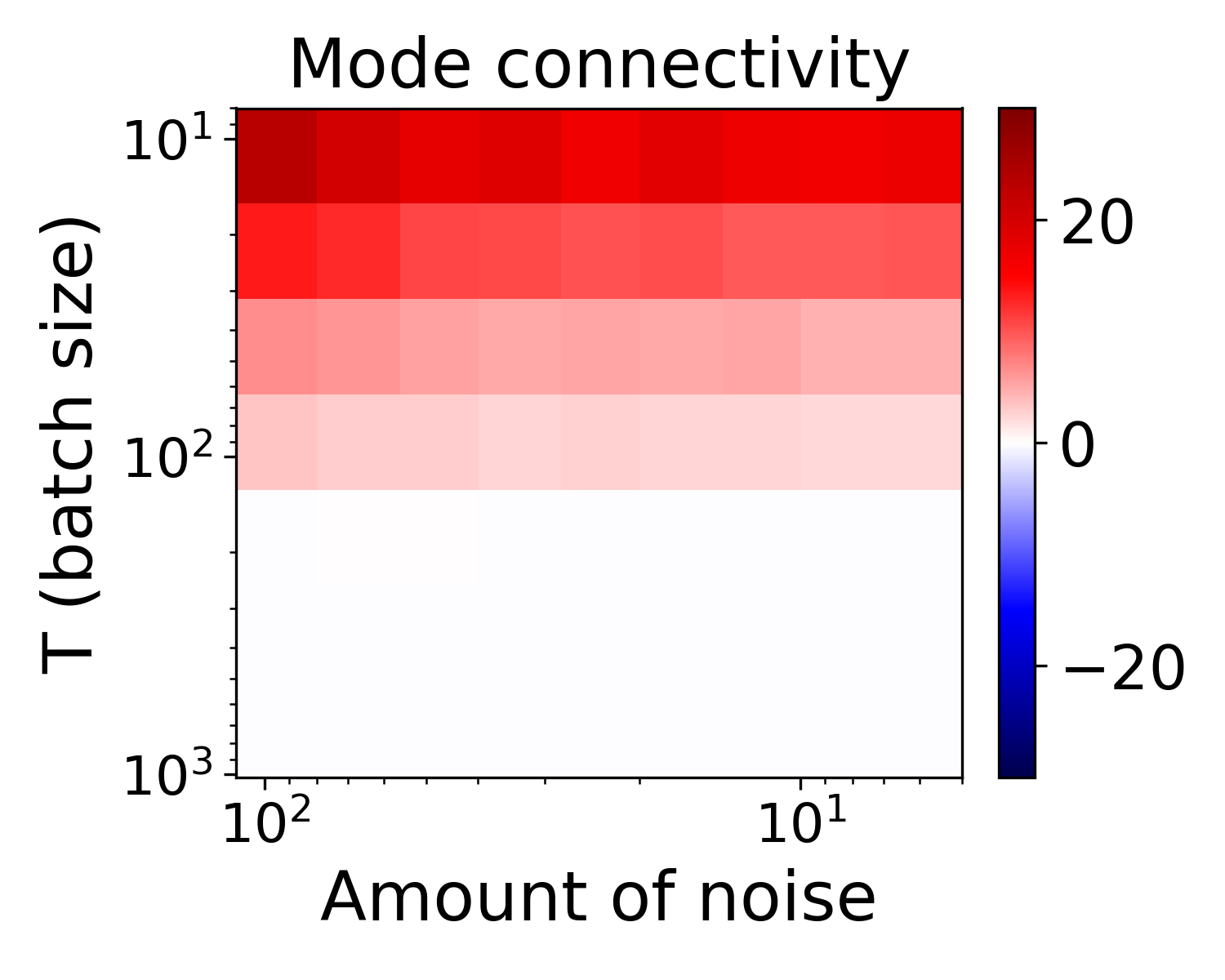}
      \vspace{-6mm}
      \caption{Mode connectivity\label{fig:\figname_curve}}
    \end{subfigure}
    \begin{subfigure}[c]{0.23\textwidth}
      \includegraphics[width=\textwidth]{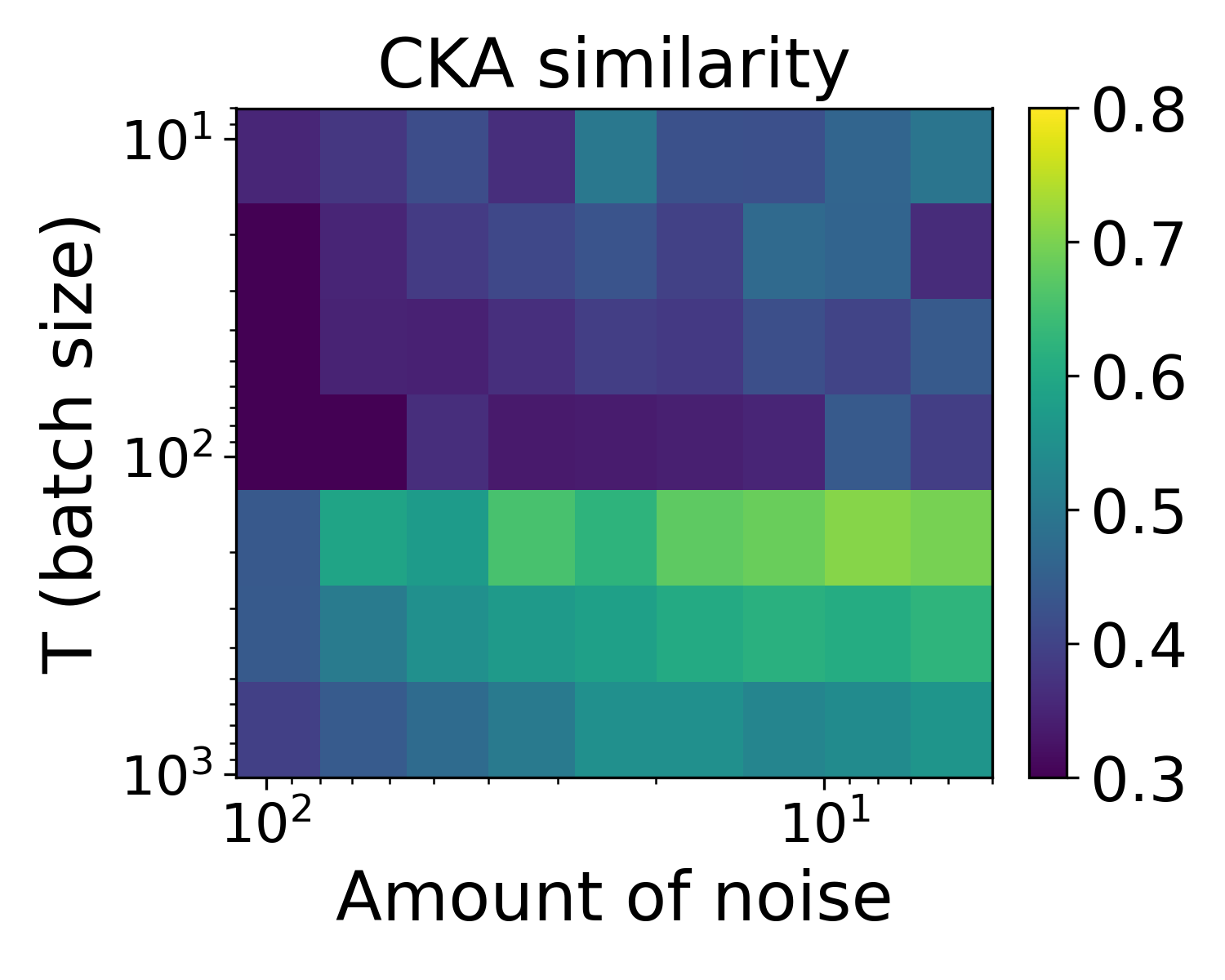}
      \vspace{-6mm}
      \caption{CKA similarity\label{fig:\figname_CKA}}
    \end{subfigure}
  \caption{{\bf (Varying additive noise on images to change load).}
  Partitioning the 2D load-like---temperature-like diagram into different phases of learning, using batch size as the temperature and varying amount of additive noise to each image to change load.
  Models are trained with ResNet18 on CIFAR-10. Mode connectivity remains close to zero even for a large amount of additive noise on each image. All plots are on the same set of axes. 
  }
  \label{fig:Noisy_image}
\end{figure}

\subsection{Additional training schemes}

In this subsection, we provide additional results on additional training schemes.

\subsubsection{Large-batch training}\label{sec:large_batch_size}

\def \figname {ResNet18_lbs}
\begin{figure}[h]
  \centering
    \begin{subfigure}{0.23\textwidth}
      \includegraphics[width=\textwidth]{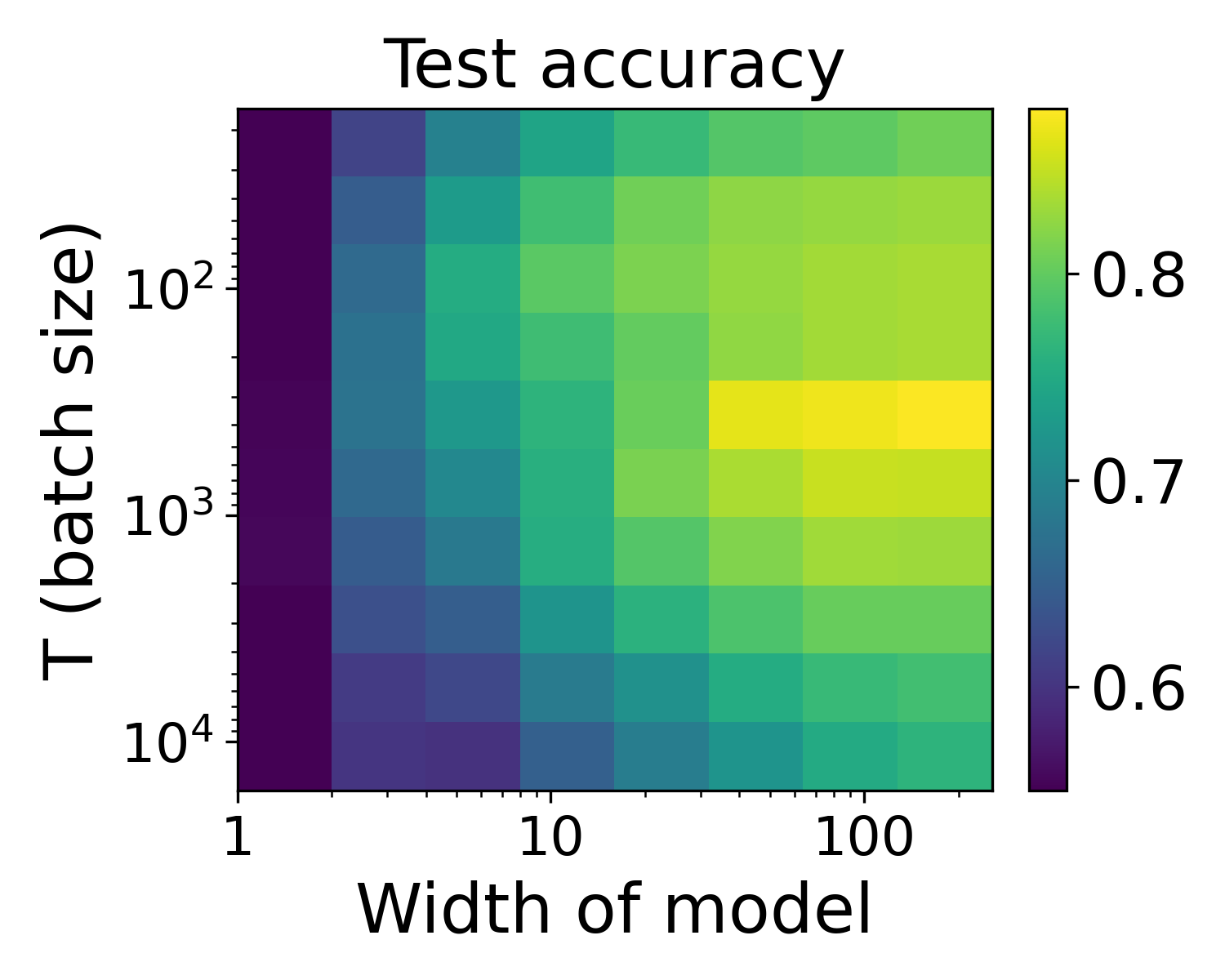}
      \caption{Test accuracy\label{fig:\figname_accuracy}}
    \end{subfigure}
    \begin{subfigure}[c]{0.23\textwidth}
      \includegraphics[width=\textwidth]{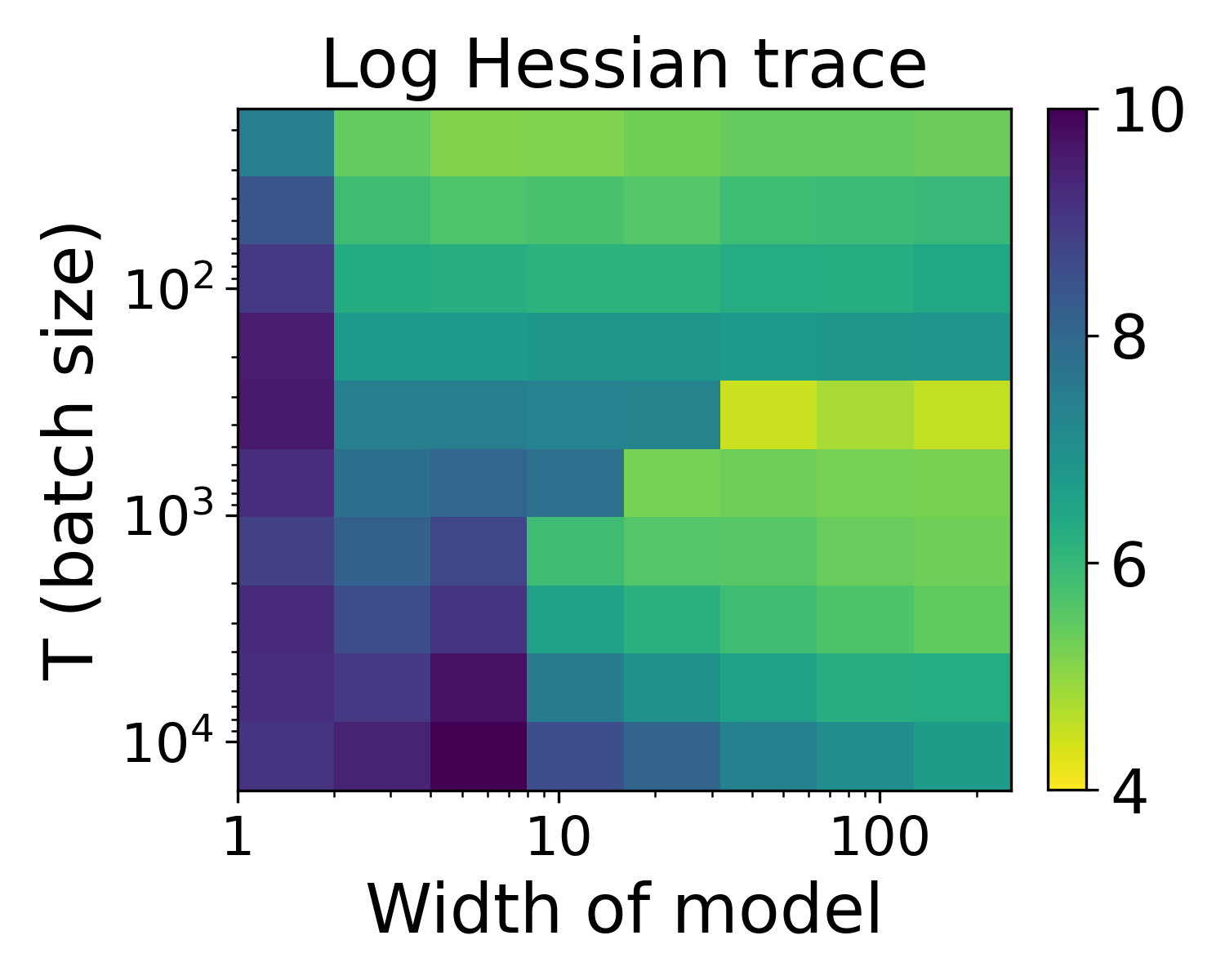}
      \caption{Hessian trace\label{fig:\figname_hessian_t}}
    \end{subfigure}
  \begin{subfigure}[c]{0.23\textwidth}
      \includegraphics[width=\textwidth]{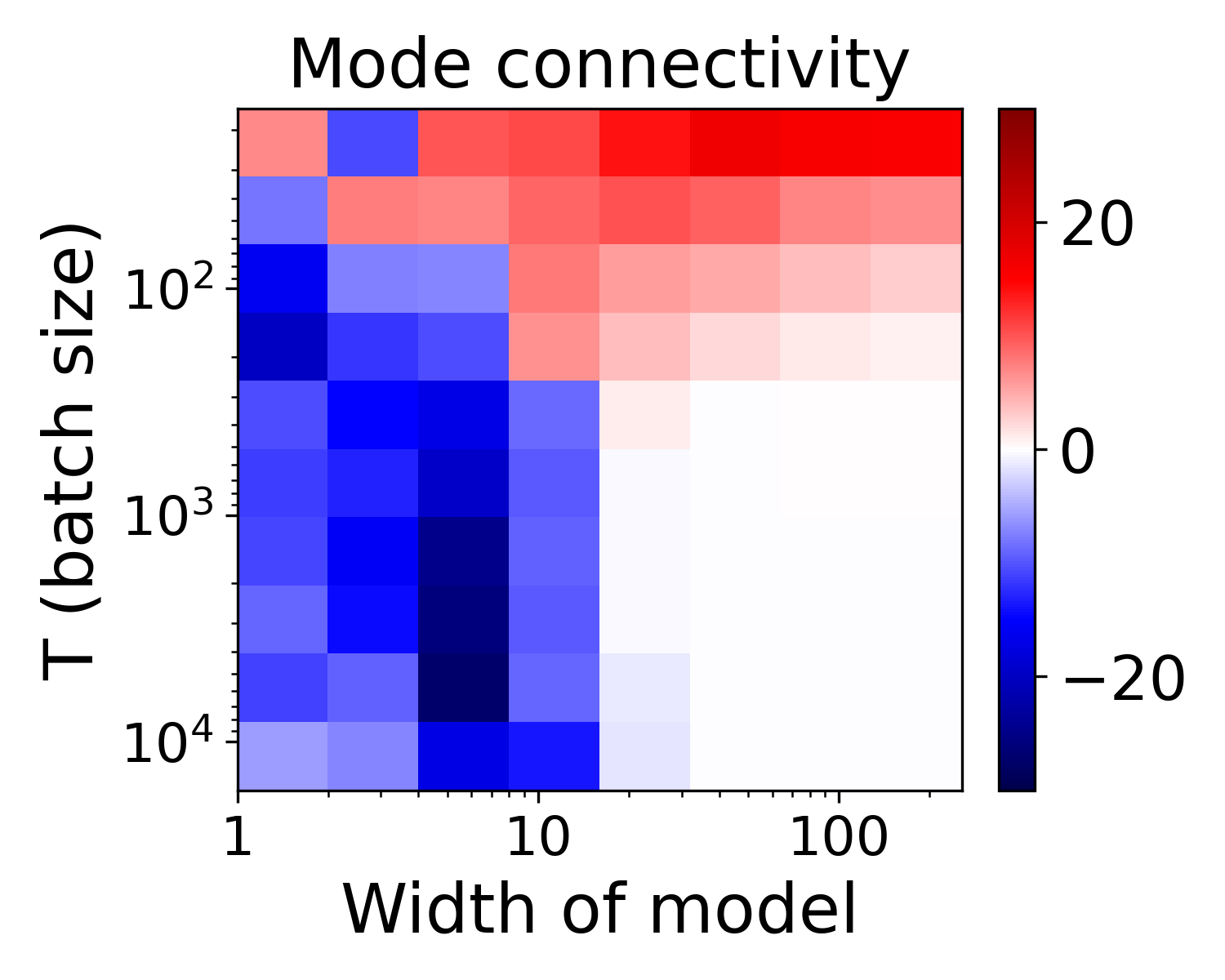}
      \caption{Mode connectivity\label{fig:\figname_curve}}
    \end{subfigure}
    \begin{subfigure}[c]{0.23\textwidth}
      \includegraphics[width=\textwidth]{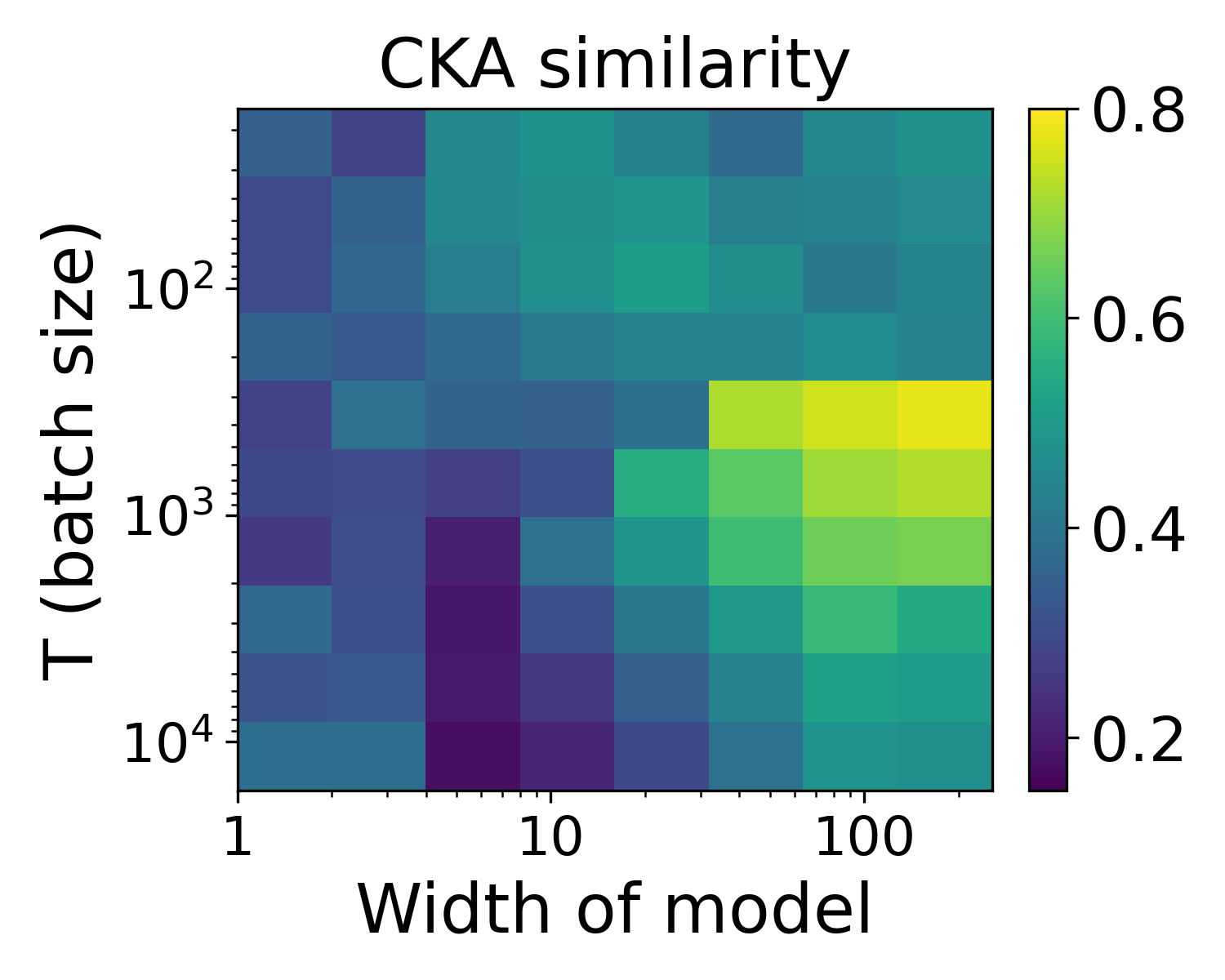}
      \caption{CKA similarity\label{fig:\figname_CKA}}
    \end{subfigure}
  \caption{{\bf (Large-batch training).} Partitioning the 2D load-like---temperature-like diagram into different phases of learning, using batch size as the temperature and varying model width to change load. 
  Large-batch training is used.
  Models are trained with ResNet18 on CIFAR-10. All plots are on the same set of axes. }
  \label{fig:ResNet18_lbs}
\end{figure}

Here, we present results analogous to our main results in Figure~\ref{fig:ResNet18}, but with a large batch size. We increase the maximum batch size to 8192 in this subsection. See results in Figure \ref{fig:ResNet18_lbs}.

The results show a similar trend to Figure \ref{fig:ResNet18}, in that the best accuracy is achieved when Hessian is small, mode connectivity is close to zero, and CKA similarity is large.
The only slight difference is that the Hessian becomes much larger than in Figure~\ref{fig:ResNet18} when trained with a large batch size.
We note that this observation matches the common belief that local minima becomes sharper in large-batch training \citep{keskar2016large}.

\subsubsection{Training with the linear scaling rule}\label{sec:LR_scaling}

Here, we present analogous results to Figure~\ref{fig:ResNet18}, but train with the linear scaling rule, i.e., when the batch size is multiplied by $k$, the learning rate is multiplied by the same constant $k$ \citep{goyal2017accurate}.
We choose the ``standard setting'' to be training with learning rate 0.05 and batch size 128, which matches the settings in other experiments, and we change learning rate to 0.05$\cdot$k when we change the batch size to 128$\cdot$k.
The results are shown in Figure~\ref{fig:Lr_scaling}. 

First, if we compare the test accuracy in Figure \ref{fig:Lr_scaling_accuracy} to that without using the linear scaling rule, shown in Figure \ref{fig:ResNet18_accuracy}, we see that the test accuracy changes much more slowly with batch size (along the Y-axis) in the first case. 
Moreover, the mode connectivity and Hessian almost do not change at all along the $Y$-axis. 
This phenomenon is expected because the linear scaling rule aims to maintain a constant noise variance when scaling up the batch size.

\def \figname {Lr_scaling}
\begin{figure}[h]
  \centering
    \begin{subfigure}{0.23\textwidth}
      \includegraphics[width=\textwidth]{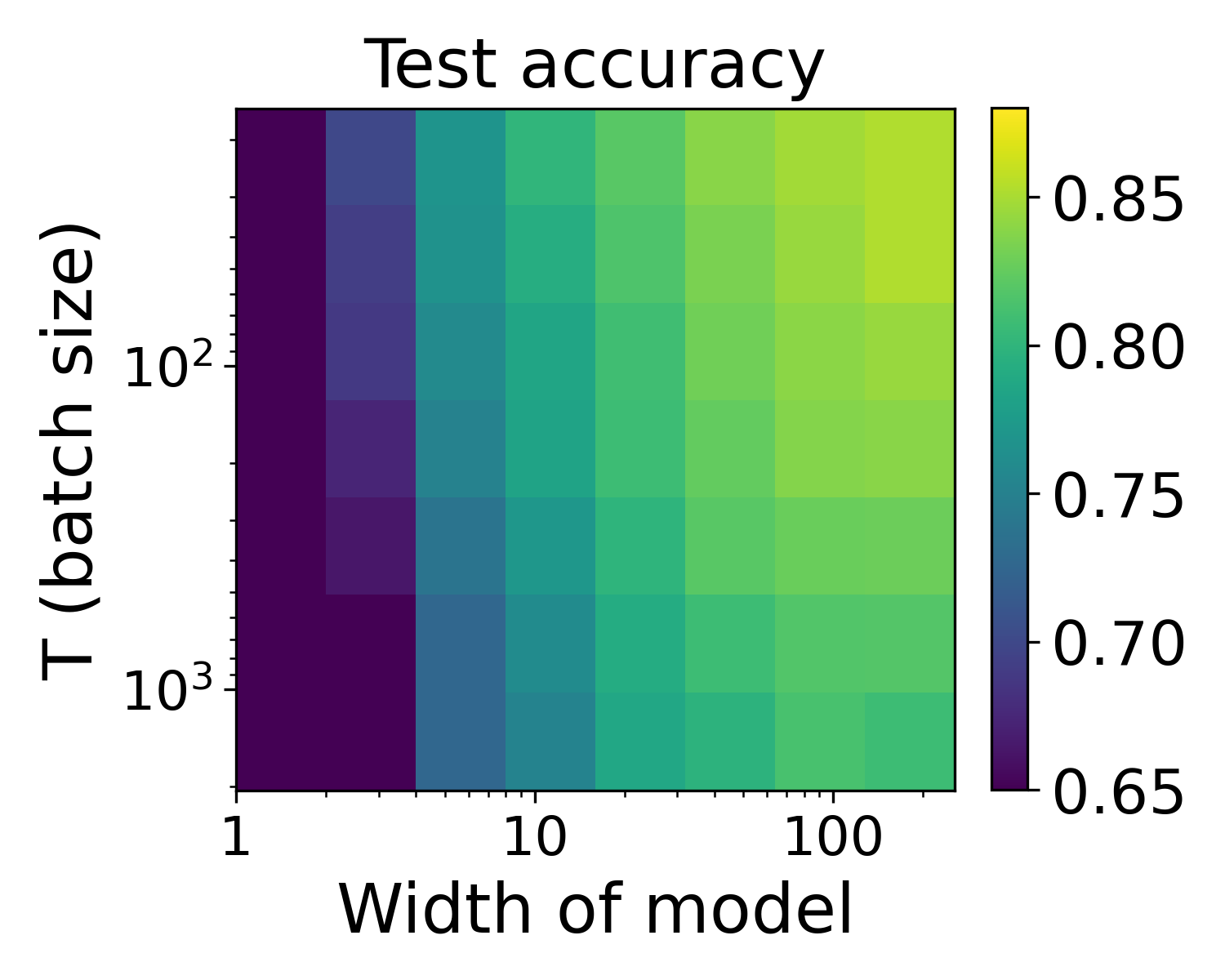}
      \caption{Test accuracy\label{fig:\figname_accuracy}}
    \end{subfigure}
    \begin{subfigure}[c]{0.23\textwidth}
      \includegraphics[width=\textwidth]{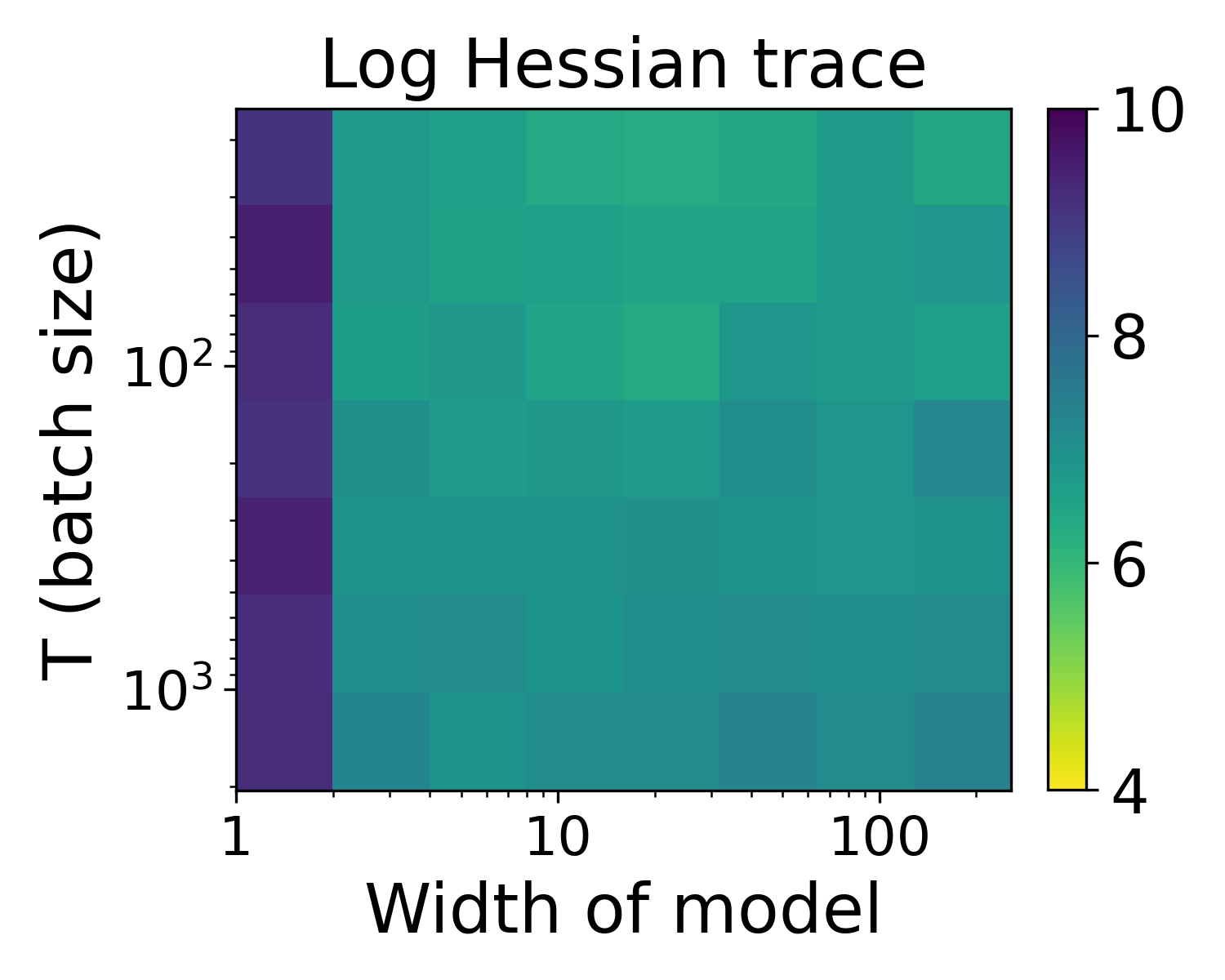}
      \caption{Hessian trace\label{fig:\figname_hessian_t}}
    \end{subfigure}
  \begin{subfigure}[c]{0.23\textwidth}
      \includegraphics[width=\textwidth]{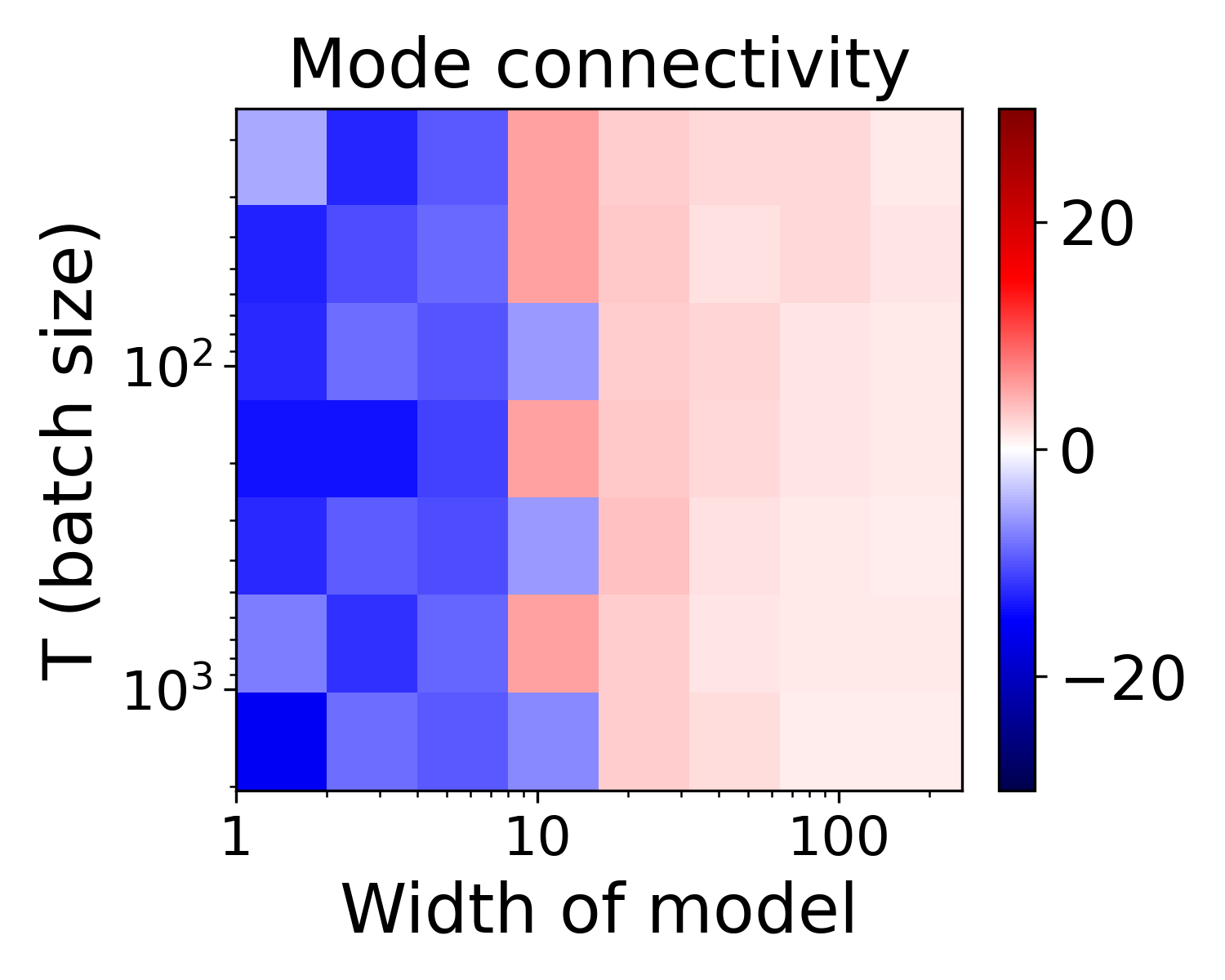}
      \caption{Mode connectivity\label{fig:\figname_curve}}
    \end{subfigure}
    \begin{subfigure}[c]{0.23\textwidth}
      \includegraphics[width=\textwidth]{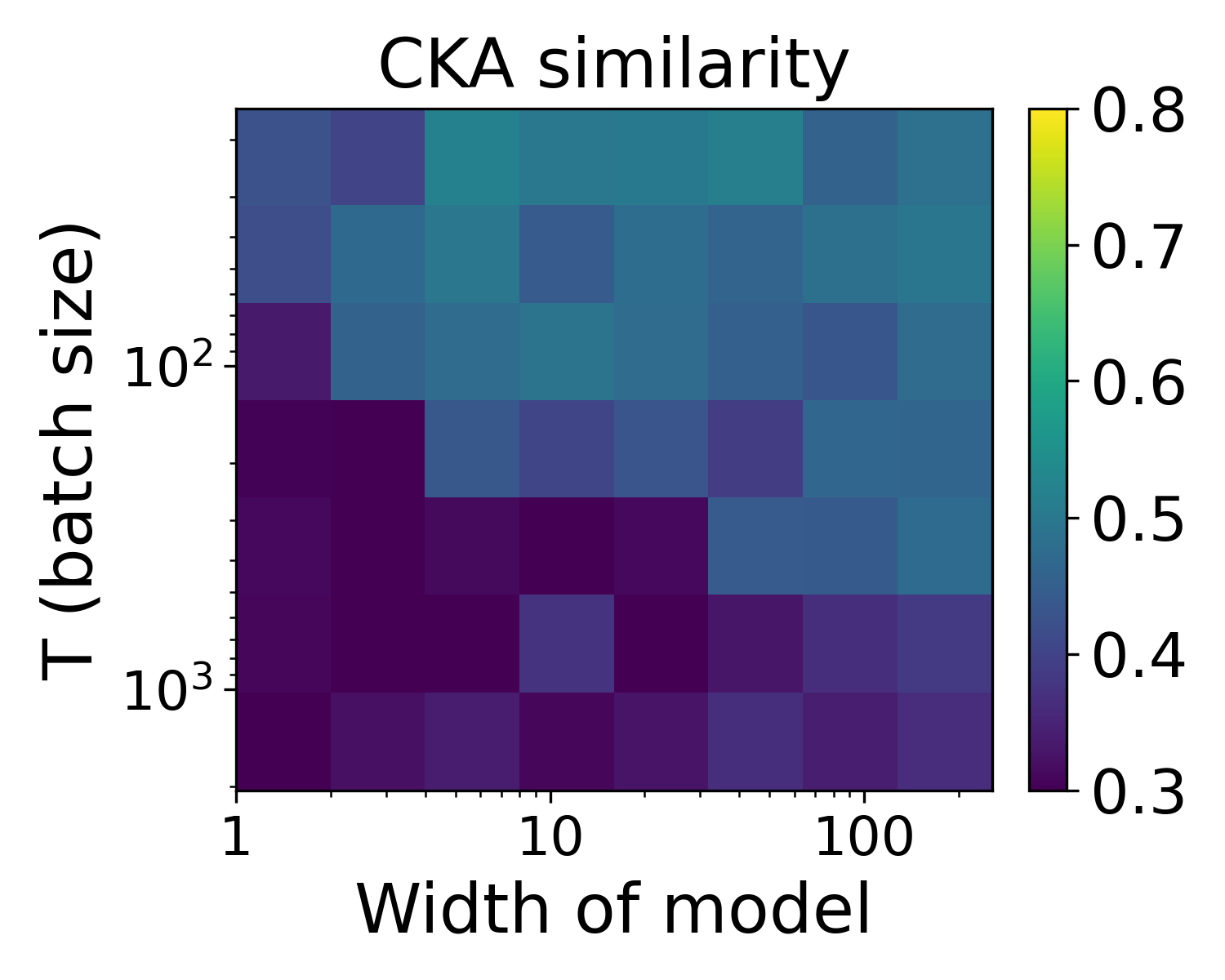}
      \caption{CKA similarity\label{fig:\figname_CKA}}
    \end{subfigure}
  \caption{{\bf (Tuning learning rate with batch size).}
  Partitioning the 2D load-like---temperature-like diagram into different phases of learning, using batch size as the temperature and varying model width to change load. 
  Linear scaling rule is used to tune learning rate with batch size.
  Models are trained with ResNet18 on CIFAR-10. All plots are on the same set of axes.}
  \label{fig:Lr_scaling}
\end{figure}

\fi
\end{document}